\documentclass[10pt,twocolumn,letterpaper]{article}

\PassOptionsToPackage{table}{xcolor}

\usepackage[pagenumbers]{wacv} 

\input{preamble}

\definecolor{wacvblue}{rgb}{0.21,0.49,0.74}
\usepackage[pagebackref,breaklinks,colorlinks,allcolors=wacvblue]{hyperref}

\def\wacvPaperID{1764} 
\def\confName{WACV}
\def\confYear{2027}

\title{OpenVAM: Open-World Visual Attention Modeling with VLMs}

\author{
	Kiana Hooshanfar\footnotemark[1]$^{\;\, ,1}$ \quad
	Amirhossein Kazerouni\footnotemark[1]$^{\;\,, 2, 3, 4}$ \quad
	Alireza Hosseini\footnotemark[1]$^{\;\, ,1}$ \\
	Michael Brudno$^{\;2, 3, 4}$ \quad
	Babak Taati$^{\;2, 3, 4}$ \\
	${^1}$ University of Tehran
	${^2}$ University of Toronto 
	${^3}$ Vector Institute \\
	${^4}$ University Health Network \\
	{\tt\small \{k.hooshanfar , arhosseini77\}@ut.ac.ir, \{amirhossein, brudno\}@cs.toronto.edu} \\
	{\tt\small babak.taati@uhn.ca} \\
	{\tt\small \url{https://k-hooshanfar.github.io/openvam/}}
}

\begin{document}
\maketitle
\begingroup
\renewcommand{\thefootnote}{\fnsymbol{footnote}}
\footnotetext[1]{Equal contribution}
\endgroup
\addtocontents{toc}{\protect\setcounter{tocdepth}{-1}}

 \begin{abstract}
Predicting human gaze is a core capability for applications ranging from web/UI design analysis to robotics and human-computer interaction. Yet, most visual attention modeling methods output only a dense saliency map, which is often insufficient for action: practitioners need to connect attention peaks to discrete elements in the scene (\emph{what}) and understand the drivers of those peaks in context (\emph{why}), while remaining robust to domain shift across natural images, commercial content, and UI/web layouts. We, therefore, introduce \textbf{OpenVAM} (\textbf{Open}-world \textbf{V}isual \textbf{A}ttention \textbf{M}odeling with VLMs), a unified framework that jointly addresses universality and explainability across heterogeneous domains (natural scenes, commercial imagery, and UI/web layouts) and supervision modalities. OpenVAM adopts a \emph{decoupled-but-aligned} design: a dedicated dense visual pathway provides stable, spatially precise localization, while an instruction-following vision--language semantic head generates grounded \emph{what/why} explanations conditioned on the same image and data-type prompts. A three-stage training strategy preserves strong localization priors while progressively introducing language grounding and improving explanation alignment via parameter-efficient adaptation without perturbing the saliency branch. We further propose a scalable pipeline to generate multi-domain saliency-reason annotations for training and systematic evaluation. Experiments across diverse datasets show that OpenVAM improves robustness under domain shift while producing image-grounded explanations that make saliency predictions more interpretable.
\end{abstract}
\vspace{-6pt}

\vspace{-1em}
\section{Introduction}
\vspace{-0.25em}
Visual attention selectively prioritizes processing toward particular locations, features, or objects in a scene~\cite{grent2007timing,kastner2000texture}. Modeling this process aims to predict this selection from visual input, typically by estimating the spatial distribution of human gaze over an image~\cite{judd2009learning}. This is commonly formulated as saliency prediction: given an image, produce a dense saliency map whose values approximate the probability (or density) of human fixations at each location~\cite{jiang2015salicon}.
Saliency maps provide compact and interpretable cues about where people are likely to look, supporting applications in marketing and e-commerce~\cite{hosseini2024brand}, UI design~\cite{jiang2023ueyes}, media understanding and compression~\cite{mishra2021multi,patel2021saliency}, and robotics or HCI systems that allocate attention to informative regions~\cite{samani2023eye,jacob2003eye,hooshanfar2025dtfsal}.

Despite the utility of saliency maps, a heatmap alone is often insufficient for downstream decisions~\cite{chen2025explainable}. In practice, users must (i) link attention peaks to discrete elements in the scene, such as a headline, logo, face, product image, or call-to-action button, and (ii) identify the visual cues associated with those peaks in context, such as semantic relevance, size and position, color/contrast, typography, or layout conventions. Accordingly, we frame practical attention modeling around three complementary questions: \textbf{(1)} \textit{where observers look}, \textbf{(2)} \textit{what elements account for the predicted attention}, and \textbf{(3)} \textit{why these elements draw attention in context.} For instance, when a landing page exhibits a strong attention peak in the top-left region, improving the design depends on whether that peak is caused by a prominent headline, a bright call-to-action button, a distracting icon, or an unintended contrast imbalance~\cite{jiang2023ueyes}. Our scope here is static visual attention: spatial saliency and semantic interpretation, rather than temporal fixation sequences.

Most saliency predictors still focus on the \emph{where} question. Most models commonly adopt an encoder-decoder architecture~\cite{han2022survey}, and with strong backbones and large-scale training can perform well on in-domain benchmarks~\cite{lou2022transalnet,aydemir2023tempsal}. However, deployment is limited by the inherently multi-domain nature of saliency: attention cues vary across natural scenes, commercial imagery, and UI/web layouts, and supervision signals exhibit distinct biases (e.g., data acquisitions using eye vs.\ mouse tracking)~\cite{tavakoli2017saliency}. As a result, models trained on a single dataset or visual context often degrade under domain shift, and it is unclear which specialized model to apply to previously unseen images. Recent work has trained unified saliency models across datasets and modalities~\cite{li2024uniar,droste2020unified}. SUM~\cite{hosseini2025sum} conditions the predictor on the input domain (e.g., natural images, commercial imagery, or UI/web layouts) and the data acquisition type to support a single model across heterogeneous visual contexts. However, these approaches typically still output only a saliency map and may rely on explicit domain identifiers at inference, which can be unreliable for mixed, ambiguous, or open-world inputs. Progress toward universal, explainable attention modeling is also limited by data. Standard saliency benchmarks provide fixation points or fixation-density maps, but rarely include aligned natural-language rationales or explicit descriptions of attended elements~\cite{jiang2015salicon,borji2015cat2000,jiang2023ueyes,jiang2022does}. Scaling such annotations is difficult: explanations are costly to collect and inherently ambiguous because attention may reflect low-level cues (e.g., contrast or edges), high-level semantics (e.g., faces or text), and domain-specific conventions (e.g., UI layout patterns or brand placement). So, there is no widely adopted multi-domain benchmark that pairs saliency supervision with free-form, image-grounded explanations.

In this work, we introduce \textbf{OpenVAM}: \textbf{Open}-\textbf{W}orld \textbf{V}isual \textbf{A}ttention \textbf{M}odeling with VLMs, a unified framework designed to address \emph{universality} and \emph{explainability} jointly. OpenVAM couples a dedicated encoder-decoder saliency predictor for accurate dense estimation with a VLM-based semantic head that produces grounded, concise explanations. Crucially, OpenVAM uses a \emph{decoupled-but-aligned} design: dense localization is learned in a stable visual pathway, while language is trained to explain the prediction without destabilizing spatial precision. This enables a single system to output a saliency map (\emph{where}) together with open-vocabulary descriptions of salient elements (\emph{what}) and short natural-language rationales (\emph{why}) within a unified framework. We further release a new multi-domain dataset that augments existing saliency benchmarks with saliency reason annotations, image-grounded explanations describing the salient elements and the cues that drive attention, spanning natural scenes, commercial imagery, and UI/web layouts. This dataset enables systematic training and evaluation of models that must generalize across domains while producing actionable \emph{where, what,} and \emph{why} outputs. Our contributions are as follows:
\begin{itemize}
    \item We propose \textbf{OpenVAM}, a unified framework for open-world visual attention modeling that jointly produces dense saliency (\emph{where}) and grounded natural-language outputs describing salient elements and cues (\emph{what/why}).
    \item We introduce a \emph{decoupled-but-aligned} architecture that separates localization and language generation while aligning them through the same image and data context.
    \item We present a multi-domain dataset that augments saliency datasets with saliency reason annotations, enabling training and evaluation of explainable saliency across heterogeneous visual domains.
\end{itemize}
\section{Related Works}
\noindent\textbf{Saliency Prediction.}
Early saliency models focused on bottom-up cues such as contrast and contextual distinctiveness~\cite{grent2007timing,kastner2000texture,jiang2015image,rajashekar2008gaffe,goferman2011context}. With large-scale gaze datasets~\cite{jiang2015salicon,borji2015cat2000,jiang2021deepvs2}, learning-based approaches became dominant, using pretrained backbones and encoder-decoder or attention-based architectures to predict fixation-density maps~\cite{vig2014large,kummerer2014deep,kruthiventi2017deepfix,cornia2018predicting,han2022survey,vaswani2017attention,lou2022transalnet}. Saliency modeling has since expanded beyond natural images to ads, e-commerce, UI/web layouts, information visualizations, and omnidirectional content~\cite{jiang2022does,hosseini2024brand,jiang2023ueyes,tavakoli2017saliency,matzen2017data,gutierrez2018toolbox}. In particular, DVS~\cite{matzen2017data} targets saliency prediction for abstract data visualizations, while Salient360!~\cite{gutierrez2018toolbox} supports visual-attention modeling for 360-degree images. However, most existing approaches are not universal: they are typically developed and evaluated within a single domain, and their performance often degrades when applied to new content distributions.

\noindent\textbf{Unified Models.}
Unified saliency modeling aims to train a single predictor across heterogeneous datasets and visual contexts, reducing the need to maintain specialized models per domain. Prior work, such as UNISAL, leverages domain adaptation to combine multiple saliency domains within one framework~\cite{droste2020unified}, while UniAR explores large-scale unified training with multimodal transformers to capture diverse attention behaviors across tasks and content~\cite{li2024uniar}. SUM explicitly targets cross-domain saliency by integrating efficient long-range modeling with a U-Net-style predictor and conditioning the network on the input domain (e.g., natural, UI/web, commercial) to adapt its behavior within one model~\cite{hosseini2025sum}. Although effective, these unified approaches remain focused on predicting saliency heatmaps, and they do not provide element-level grounding or explanations that shows what is being attended and why.

\begin{figure*}[t]
    \centering
    \resizebox{\linewidth}{!}{%
    \includegraphics[]{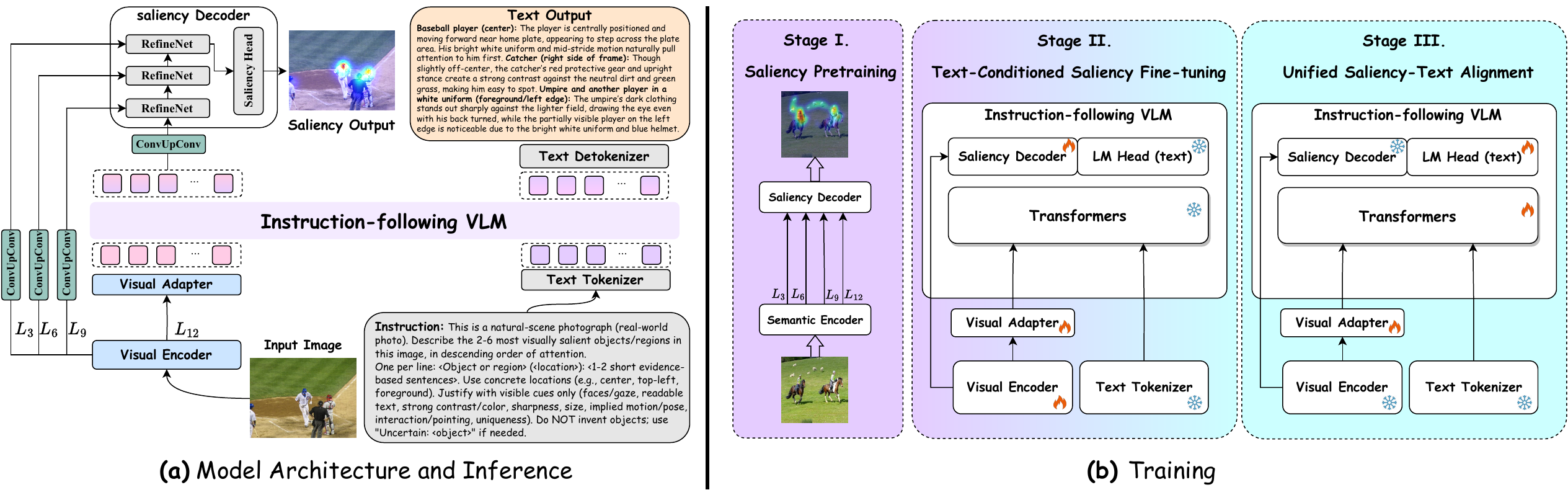}
    }
    \vspace{-1em}
\caption{\textbf{(a) OpenVAM architecture.}
An input image is encoded into multi-level visual features. ConvUpConv projections build a feature pyramid for the saliency decoder to predict dense saliency, while the deepest feature ($L_{12}$) is adapted into an instruction-following VLM to generate grounded explanations of salient regions.
\textbf{(b) Model training.}
\emph{Stage I:} the visual encoder and saliency decoder learn dense attention localization from multi-scale features ($L_3,L_6,L_9,L_{12}$).
\emph{Stage II:} the VLM is attached, and the visual encoder, adapter, and saliency decoder are trained while the tokenizer, transformer, and LM head remain frozen.
\emph{Stage III:} the encoder and saliency decoder are frozen, while the visual adapter and the language side are tuned to align explanations with saliency.}
    \vspace{-1.5em}
    \label{fig:model_arch}
\end{figure*}

\noindent\textbf{Scanpath Prediction.}
Scanpath models explicitly predict the temporal sequence of human fixations. SALYPATH~\cite{kerkouri2021salypath} jointly models saliency and scanpaths by deriving fixation trajectories from learned saliency representations, while UMSS~\cite{wang2023scanpath} predicts both saliency maps and fixation sequences for information visualizations. OAT~\cite{fang2024oat} models attention at the object level and predicts sequences of attended objects during visual search. GazeXplain~\cite{chen2024gazexplain} combines scanpath prediction with fixation-level natural-language explanations, bridging temporal gaze modeling and language-based interpretation. Relatedly, recent CLIP-based work~\cite{zanca2025contrastive} demonstrates that language-guided vision representations can serve as zero-shot human scanpath predictors. These approaches explicitly model temporal gaze dynamics and fixation order. In contrast, OpenVAM targets \emph{static, dense saliency prediction} together with grounded \emph{what/why} rationales, and does not model scanpaths or the temporal ordering of human fixations.

\noindent\textbf{VLMs in Visual Attention Modeling.}
Recent work uses VLMs to incorporate semantic or language cues into visual attention modeling. SalChartQA~\cite{wang2024salchartqa} predicts query-conditioned saliency for information visualizations, while XSal~\cite{chen2025explainable} uses a general-purpose VLM to generate semantic proposals that are mapped to image regions and converted into saliency predictions. GazeVLM~\cite{chen2025eye} instead uses human gaze to select informative visual tokens for more efficient VLM inference, assuming gaze is available at test time. In contrast, OpenVAM infers attention directly from the image and couples a dedicated dense saliency pathway with an instruction-following VLM semantic head to jointly provide image-wide saliency (\emph{where}) and grounded descriptions of salient elements and visual cues (\emph{what/why}) across natural images, e-commerce, and UI/web layouts.
\section{OpenVAM}
\label{sec:method}
Human visual attention is inherently \emph{dense} and \emph{spatial}, whereas language supervision is \emph{sparse} and \emph{semantic}. Directly fine-tuning a VLM end-to-end to satisfy both objectives leads to undesirable coupling: the dense prediction quality becomes sensitive to prompt phrasing and language-head training dynamics, and the model may sacrifice spatial precision to optimize token likelihood (see~\autoref{tab:openvam_stage_ablation_main}). \textbf{OpenVAM} addresses this mismatch with a \emph{decoupled-but-aligned} design that separates \emph{where} attention is (dense saliency) from \emph{what/why} it is (text explanation) while conditioning both on the same image and data context. Specifically, OpenVAM combines a domain-agnostic saliency backbone with a dedicated dense decoder, a VLM-based semantic head for grounded explanations, and a staged training strategy that preserves saliency priors while progressively introducing language supervision.

\subsection{Model Architecture}
As demonstrated in~\autoref{fig:model_arch}(a), given an image $I\in\mathbb{R}^{H\times W\times 3}$, OpenVAM outputs a saliency map $\hat{S}\in\mathbb{R}^{H\times W}$ and an explanation sequence $\hat{Y}=\{y_t\}_{t=1}^{T}$, where $T$ denotes the generated token length. OpenVAM consists of a DINOv3-based~\cite{simeoni2025dinov3} visual encoder, a coarse-to-fine saliency decoder, and an instruction-following vision-language module for attention grounding and explanation. The central design principle is \emph{decoupled but aligned learning}: dense attention localization is learned with a specialized visual pathway, while semantic grounding is learned with a language pathway that conditions on the same image and a data-type prompt, and the two are optimized jointly.

\noindent\textbf{Visual Encoder.}
\label{sec:visual_encoder}
We decouple dense saliency prediction from the VLM's native vision tower and extract multi-level features from a DINOv3~\cite{simeoni2025dinov3} ViT at layers \(\ell\in\{3,6,9,12\}\). Earlier blocks preserve fine spatial detail, while deeper blocks encode higher-level semantics and broader context. Let \(\mathbf{F}^{(\ell)}=\phi_{\text{DINO}}^{(\ell)}(I)\in\mathbb{R}^{N\times d}\) denote the token features at layer \(\ell\). The first three pyramid levels are obtained directly from \(L_3,L_6,L_9\), while the deepest feature \(L_{12}\) is projected through the visual adapter and VLM transformer for cross-modal alignment:
\begin{equation}
\footnotesize
\mathbf{f}^{(k)} = g_k\!\left(\mathbf{F}^{(\ell_k)}\right),\; k=1,2,3,
\qquad
\mathbf{f}^{(4)} = g_4\!\left(\widetilde{\mathbf{F}}^{(12)}\right),
\end{equation}
where \(\ell_k\in\{3,6,9\}\), \(\widetilde{\mathbf{F}}^{(12)}\) denotes the VLM-processed visual representation of \(L_{12}\), and
\(g_k(\cdot)=\mathrm{Conv}_{3\times3}\!\big(\uparrow(\mathrm{Conv}_{1\times1}(\cdot))\big)\).
The resulting pyramid \(\{\mathbf{f}^{(k)}\}_{k=1}^{4}\) combines fine spatial cues with high-level semantic representations for dense saliency prediction.

\noindent\textbf{Saliency Decoder.}
\label{sec:sal_decoder}
Given the four-level feature pyramid \(\{\mathbf{f}^{(k)}\}_{k=1}^{4}\), inspired by DPT~\cite{ranftl2021vision}, we predict saliency with a simple coarse-to-fine refinement decoder that progressively fuses features via skip connections:
\begin{equation}
\footnotesize
\mathbf{z}^{(4)} = r_4(\mathbf{f}^{(4)}), \;
\mathbf{z}^{(k)} = r_k\!\left(\mathbf{f}^{(k)},\, \uparrow(\mathbf{z}^{(k+1)})\right), \; k=3,2,1,
\end{equation}
where \(r_k\) denotes a RefineNet~\cite{lin2017refinenet} block that merges the upsampled coarse representation with the corresponding higher-resolution features to recover fine spatial detail. A lightweight output head then produces the final prediction:
\begin{equation}
\footnotesize
\hat{S}
=
\sigma\!\left(
\phi_{1}\!\left(
\rho\!\left(
\phi_{3}^{(2)}\!\left(
\uparrow\, \rho\!\left(\phi_{3}^{(1)}(\mathbf{z}^{(1)})\right)
\right)\right)\right)\right),
\quad
\hat{S}\in\mathbb{R}^{H\times W}.
\end{equation}
Here, \(\phi_{3}^{(1)}\) and \(\phi_{3}^{(2)}\) denote \(3\times3\) convolution layers, \(\phi_{1}\) denotes a \(1\times1\) convolution layer, \(\rho(\cdot)\) is ReLU, \(\uparrow\) denotes upsampling, and \(\sigma(\cdot)\) is the sigmoid function.

\noindent\textbf{Vision-Language Semantic Head.}
\label{sec:vlm}
OpenVAM uses an instruction-following Qwen-VL~\cite{bai2023qwen,wu2025qwen} model to generate a concise explanation conditioned on the same image. Visual tokens are incorporated into the language context via a lightweight adapter. The language model generates:~$\hat{Y} \sim p_{\theta}\!\left(Y \,\middle|\, I, \pi(\text{Data-type Instruction})\right),$ where $\pi(\cdot)$ is a data-type instruction template (e.g., natural scene, web/UI, document). This explicit data-type conditioning encourages consistent explanation style and grounding across domains. Importantly, the VLM is \emph{not} the primary source of dense spatial features; it functions as an auxiliary semantic head that improves interpretability and promotes domain-robust representations without destabilizing saliency learning.

Both heads operate on the same image and are optimized jointly. The saliency head is trained with dense saliency supervision to localize attention accurately, while the language head is trained to produce grounded descriptions of salient regions. This joint training couples \emph{spatial} and \emph{semantic} supervision without forcing the dense predictor to rely on VLM-native visual features, mitigating prompt sensitivity and stabilizing learning across heterogeneous data.
\subsection{Training Procedure}
As illustrated in~\autoref{fig:model_arch}(b), we train OpenVAM with three stages designed to preserve a strong saliency prior while progressively introducing language conditioning.

\noindent\textbf{Stage I.}
We first learn a strong \emph{saliency localization} model using saliency supervision only. Specifically, we train the visual encoder with the coarse-to-fine saliency decoder to predict fixation-derived saliency maps. This stage is crucial to capture robust, spatially precise attention priors without any language modeling component, and serves as the initialization for subsequent stages.

\noindent\textbf{Stage II.}
Starting from the Stage~I model, we integrate an instruction-following Qwen-VL module into the deepest visual pathway and provide a data-type instruction to the semantic head. The final DINOv3 representation is mapped through the vision-to-language adapter and the frozen VLM transformer, whose visual hidden states form the coarsest saliency feature. We optimize the DINOv3 encoder, multi-scale dense decoder, and visual adapter using saliency supervision while keeping the language backbone frozen. This stage aligns the dense visual representation with the pretrained vision-language feature space without introducing token-level language supervision.

\noindent\textbf{Stage III.}
To improve explanation and consistency without perturbing localization, we freeze the visual encoder and dense decoder while keeping the visual adapter trainable, and apply LoRA~\cite{hu2022lora} to the language transformer, the LM head, and required norms. Since the adapter still feeds the saliency pathway, we retain $\mathcal{L}_{\text{sal}}$ so that adapter updates do not degrade dense prediction. This parameter-efficient adaptation sharpens grounded descriptions while preserving the localization behavior learned in Stages~I--II.


\subsection{Loss Functions}
We train OpenVAM with a composite saliency objective inspired by prior saliency works~\cite{hosseini2025sum,lou2022transalnet}. The objective combines complementary terms that capture both distributional agreement and structural consistency between the predicted saliency map and human attention signals. Let $S^{g}$ denote the ground-truth saliency map, $F^{g}$ the ground-truth fixation map, and $\hat{S}$ the predicted saliency map. Our saliency loss is
\begin{equation}
\footnotesize
\begin{aligned}
\mathcal{L}_{\text{sal}} &=
\lambda_{1}\,\mathcal{L}_{\text{KL}}(S^{g},\hat{S})
-\lambda_{2}\,\mathcal{L}_{\text{CC}}(S^{g},\hat{S})
-\lambda_{3}\,\mathcal{L}_{\text{SIM}}(S^{g},\hat{S}) \\
&\quad-\lambda_{4}\,\mathcal{L}_{\text{NSS}}(F^{g},\hat{S})
+\lambda_{5}\,\mathcal{L}_{\text{MSE}}(S^{g},\hat{S}) \, ,
\end{aligned}
\label{eq:sal_loss}
\end{equation}
where $\lambda_i$ is a scaling factor. Following common practice, we minimize dissimilarity terms (KL, MSE) and maximize similarity terms (CC, SIM, NSS). We summarize the saliency loss terms, reporting the formulation of each term, what it measures, and its role in optimizing OpenVAM in Supp.~\suppref{sec:supp_loss}. \textbf{Stage I} learns a strong localization prior by optimizing the saliency pathway only: $\mathcal{L}^{\text{(I)}}=\mathcal{L}_{\text{sal}}$. \textbf{Stage II} continues saliency training under the VLM representation space; the LM backbone remains frozen, and we update the visual pathway: $\mathcal{L}^{\text{(II)}}=\mathcal{L}_{\text{sal}}$. \textbf{Stage III} optimizes the language objective for the explanation head, while keeping $\mathcal{L}_{\text{sal}}$ as a fixed localization constraint: $\mathcal{L}^{\text{(III)}}=\alpha\,\mathcal{L}_{\text{sal}}+\beta\,\mathcal{L}_{\text{text}}$. $\mathcal{L}_{\text{text}}$ is the standard autoregressive token-level cross-entropy between the generated explanation and the ground-truth text.

\section{Experiments}
\begin{figure}[H]
\centering
\includegraphics[width=0.72\linewidth]{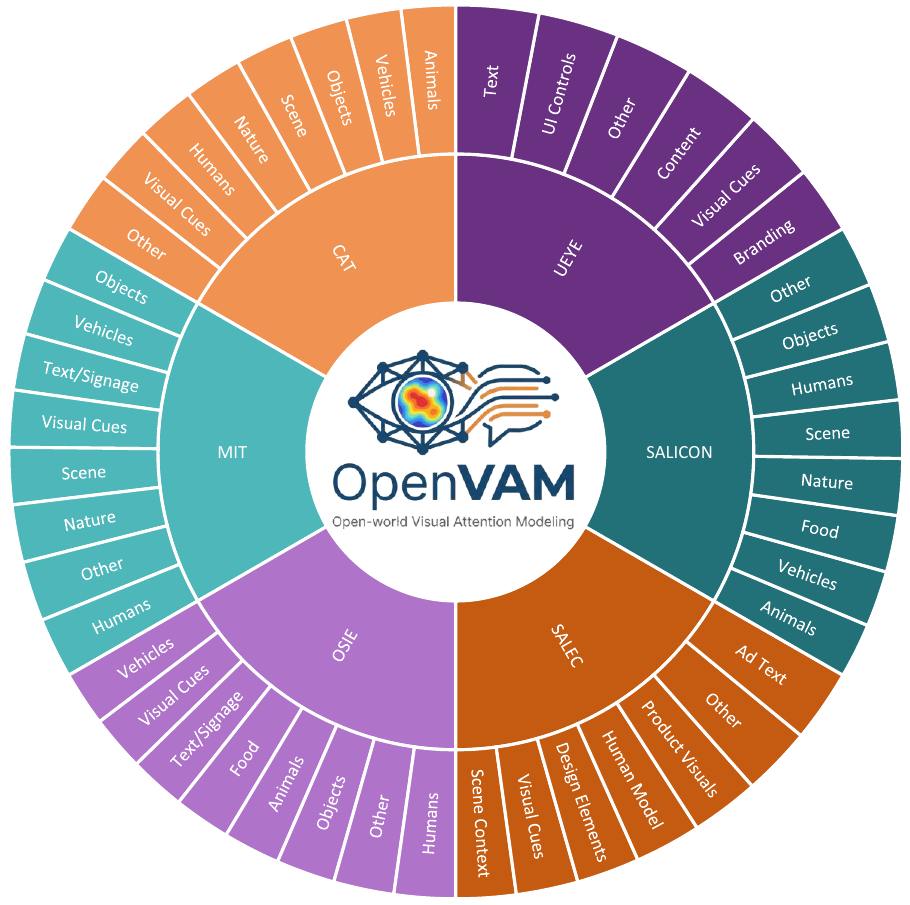}
\vspace{-0.5em}
\caption{Overview of the proposed dataset.}
\label{fig:dataset_main}
\end{figure}

\noindent\textbf{Training and Testing Datasets.} We build a unified multi-domain \emph{saliency and reason} corpus by augmenting six established saliency benchmarks with image-grounded textual rationales. Each image is paired with a concise explanation in the form \texttt{Object (location): reason}, complementing dense saliency supervision (\emph{where}) with aligned \emph{what} and \emph{why} descriptions. As shown in \autoref{fig:dataset_main}, our corpus spans natural images (SALICON~\cite{huang2015salicon}, MIT1003~\cite{judd2009learning}, CAT2000~\cite{borji2015cat2000}, OSIE~\cite{xu2014predicting}), e-commerce (SalECI~\cite{jiang2022does}), and web/UI layouts (U-EYE~\cite{jiang2023ueyes}), while covering both eye-tracking and mouse-tracking supervision. We generate rationales at scale using Gemini~2.5 Flash~\cite{comanici2025gemini}, conditioned on the stimulus image, ground-truth saliency map, and an instruction prompt (see Supp.~\suppref{sec:dataset_analysis} for details). The model identifies salient regions and describes visual cues associated with their saliency. To reduce hallucinations and mislocalization, an expert annotator verifies object visibility and location consistency with the saliency map, correcting failed samples.

\begin{figure*}[!]
    \centering
    \resizebox{\textwidth}{!}{
    \begin{tabular}{@{} *{8}c @{}}

    \includegraphics[width=0.155\textwidth]{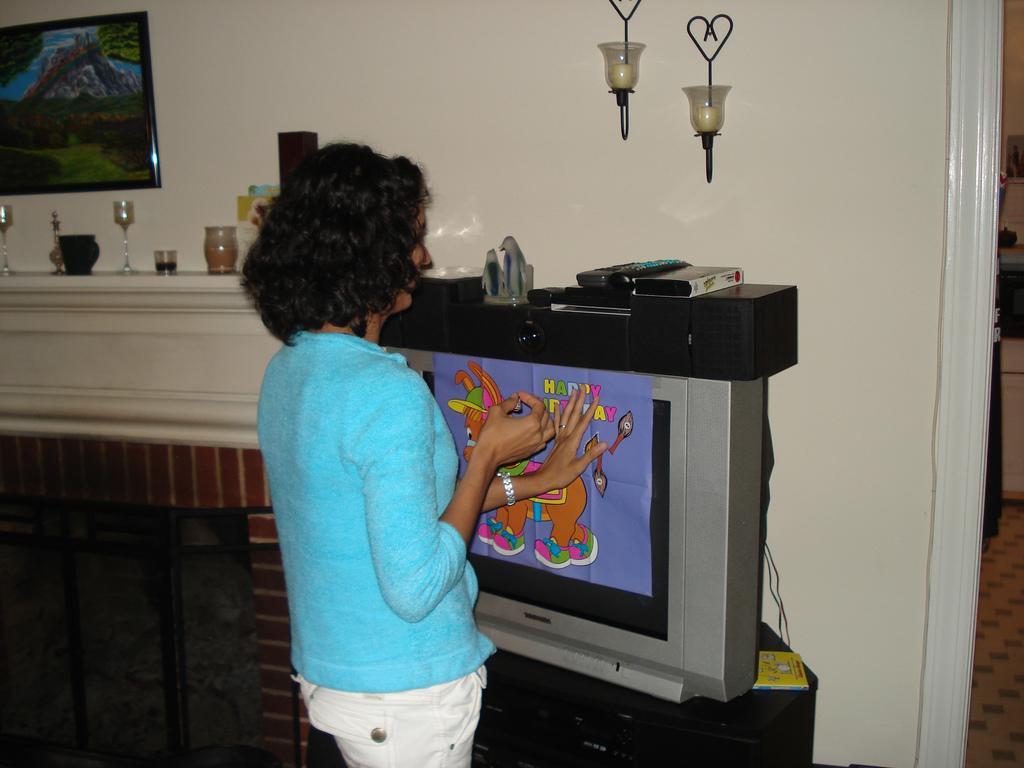} &
    \includegraphics[width=0.155\textwidth]{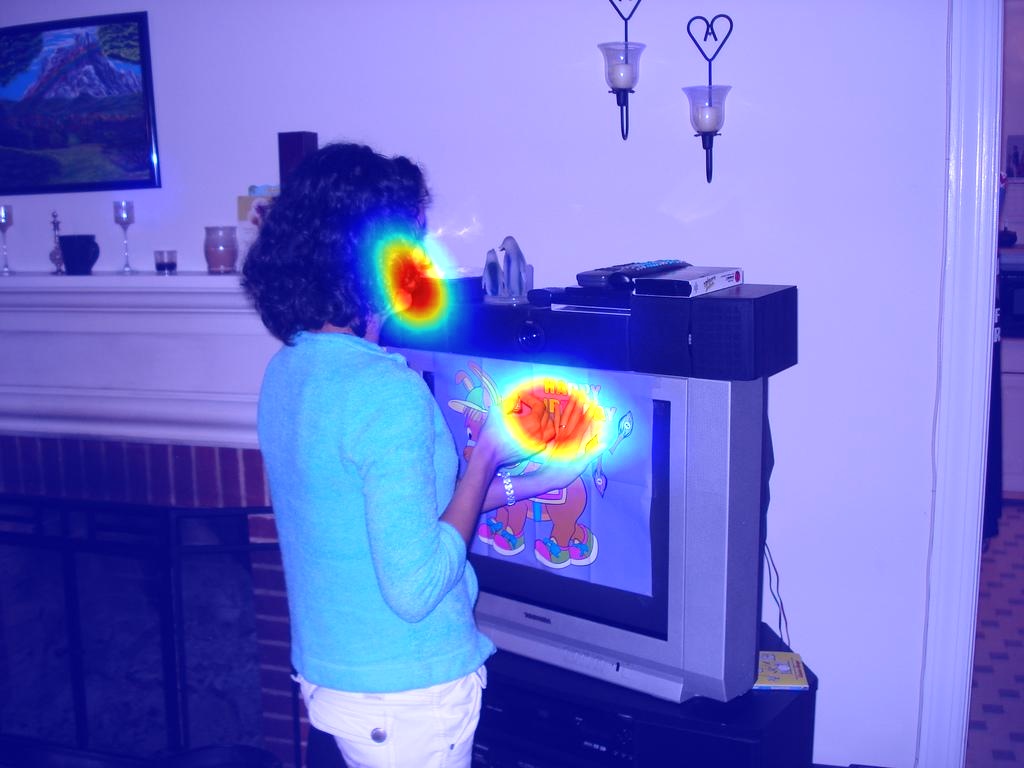} &
    \includegraphics[width=0.155\textwidth]{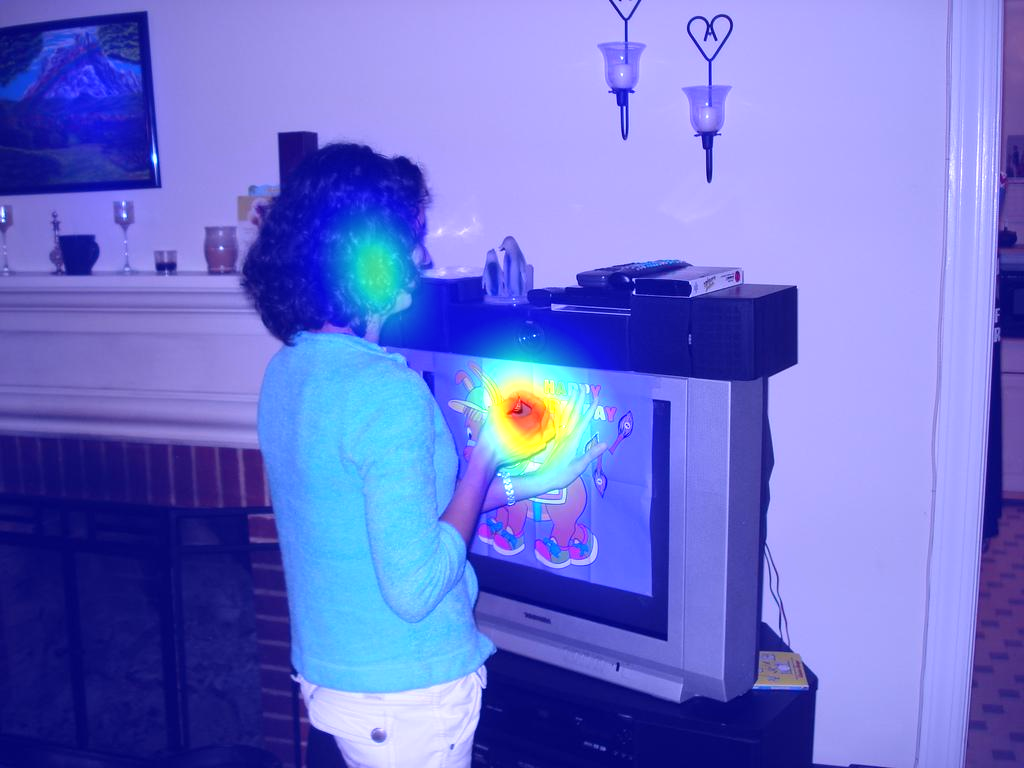} &
    \includegraphics[width=0.155\textwidth]{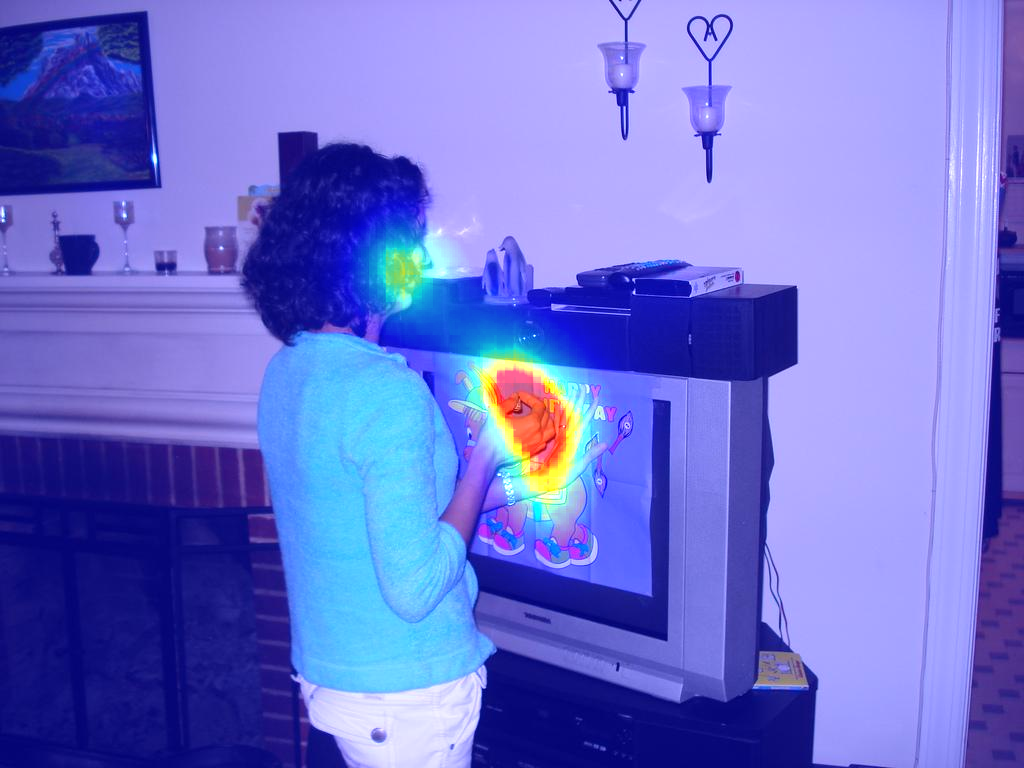} &
    \includegraphics[width=0.155\textwidth]{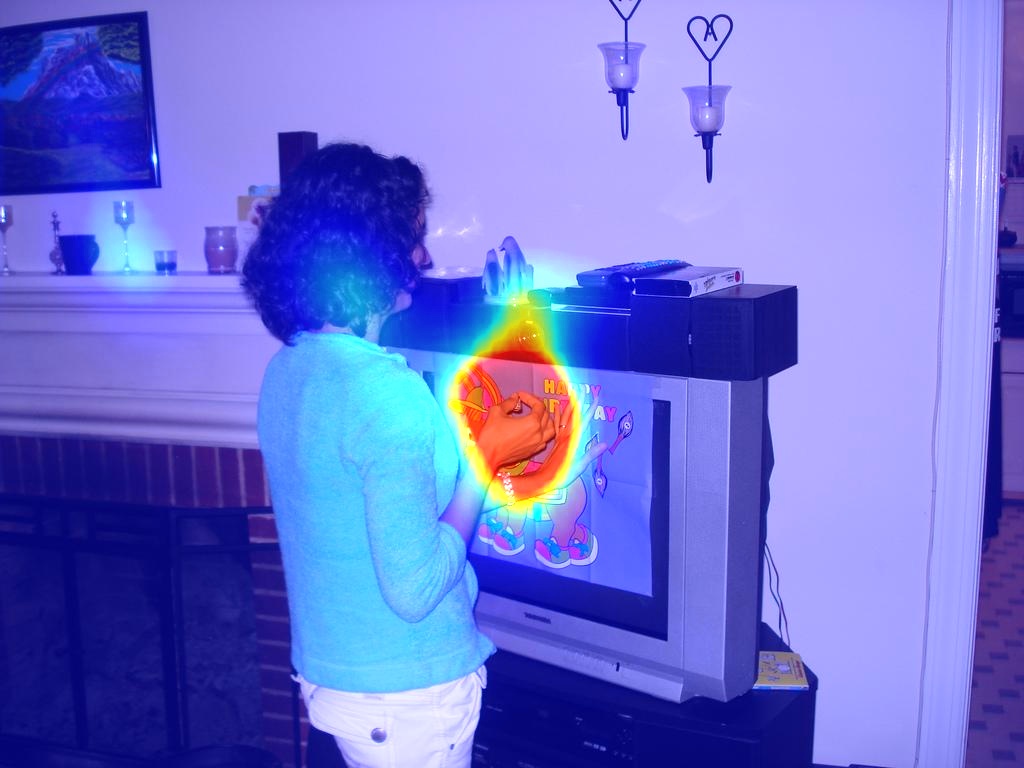} &
    \includegraphics[width=0.155\textwidth]{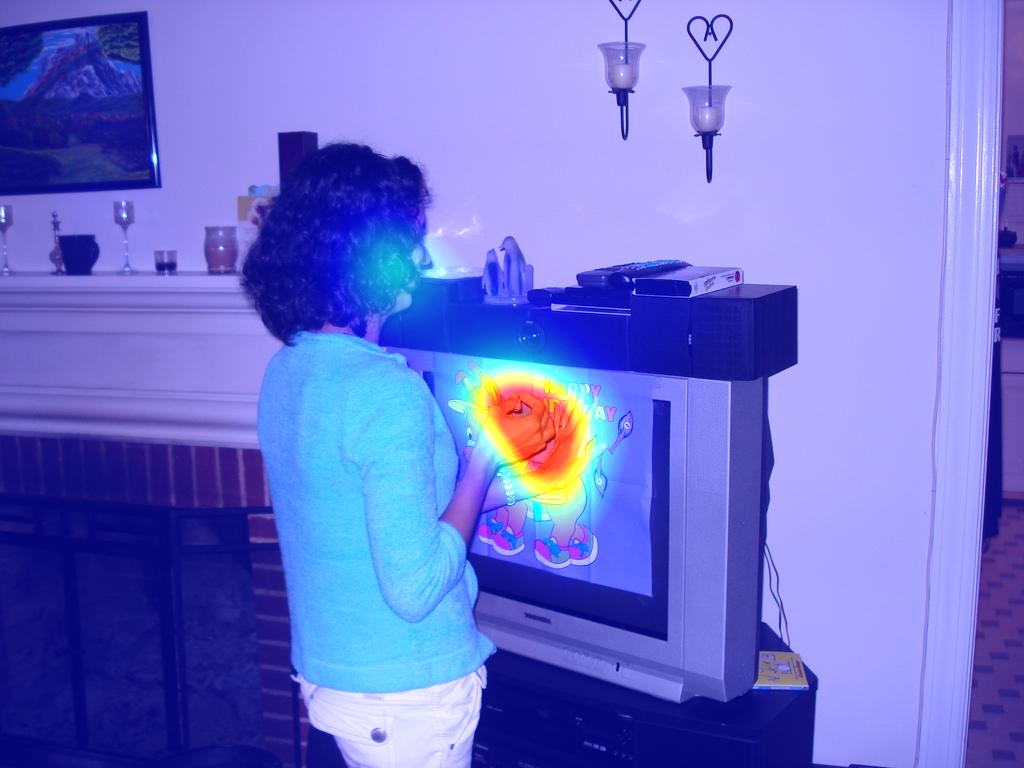} &
    \includegraphics[width=0.155\textwidth]{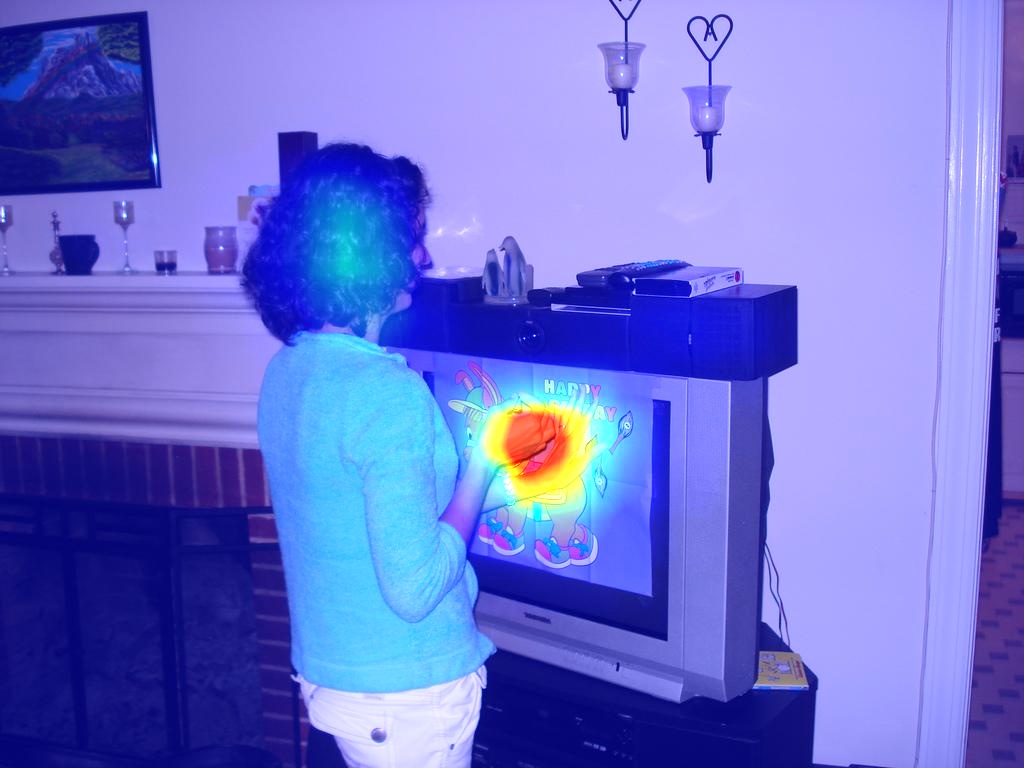} &
    \includegraphics[width=0.155\textwidth]{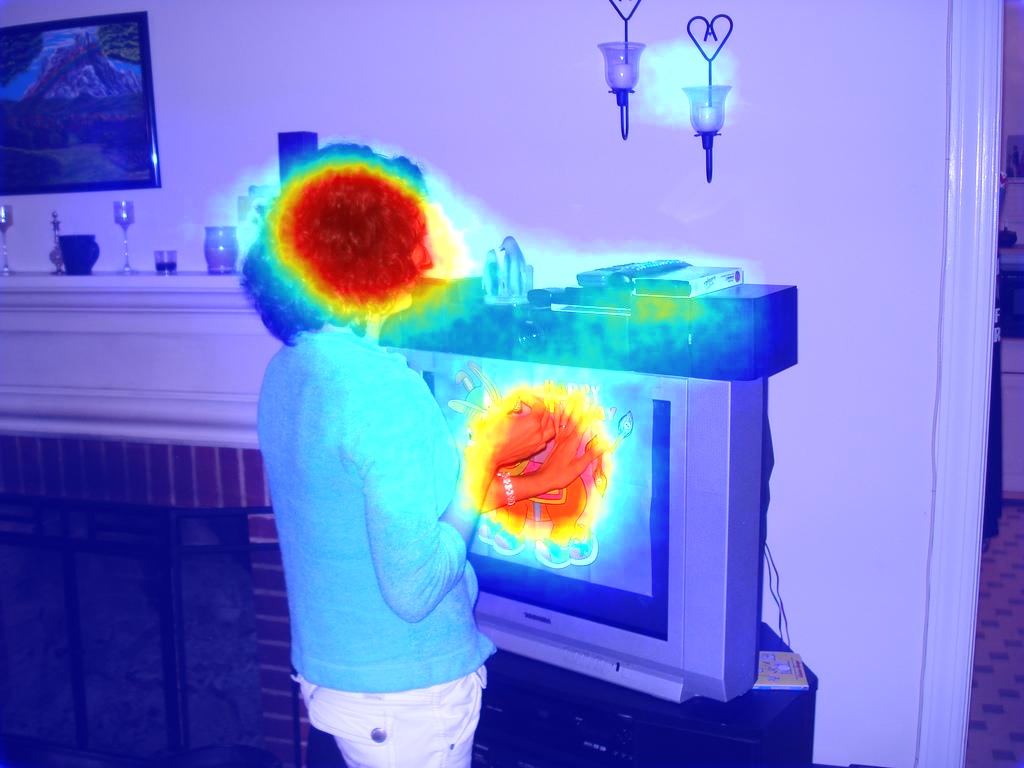}
    \\
    {\footnotesize Input Image} & {\footnotesize Ground Truth} &
    {\footnotesize \textbf{OpenVAM}} &
    {\footnotesize SUM~\cite{hosseini2025sum}} &
    {\footnotesize Transalnet~\cite{lou2022transalnet}} &
    {\footnotesize UNISAL~\cite{droste2020unified}} &
    {\footnotesize EML-NET~\cite{jia2020eml}} &
    {\footnotesize FastSal~\cite{hu2021fastsal}} \\


    \includegraphics[width=0.155\textwidth]{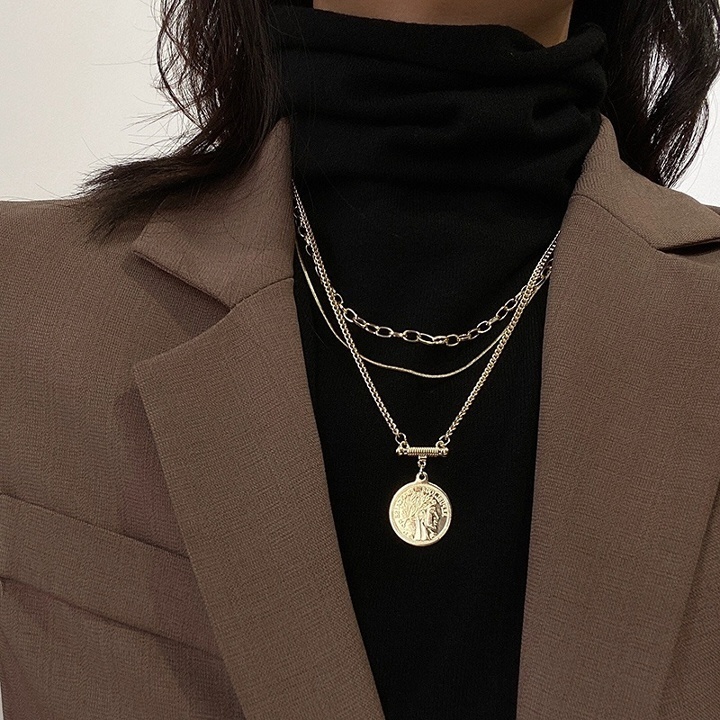} &
    \includegraphics[width=0.155\textwidth]{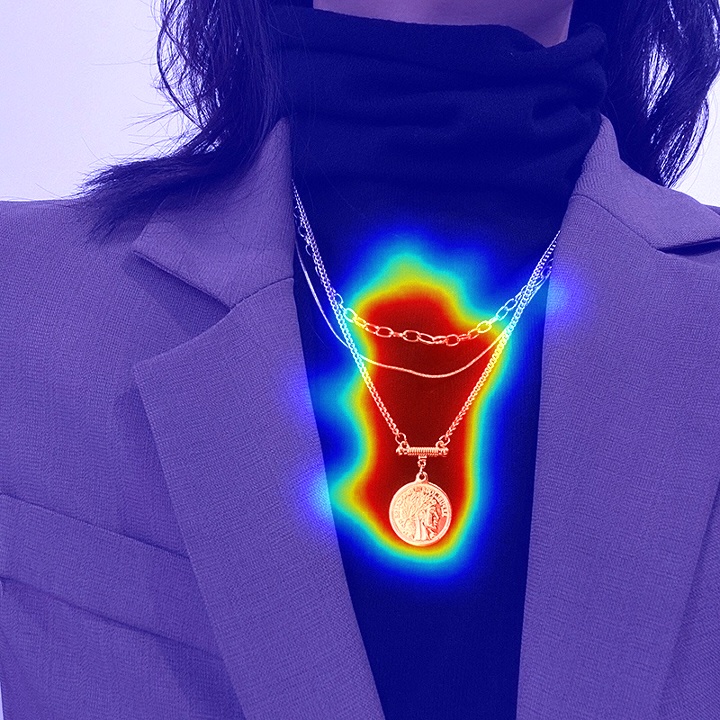} &
    \includegraphics[width=0.155\textwidth]{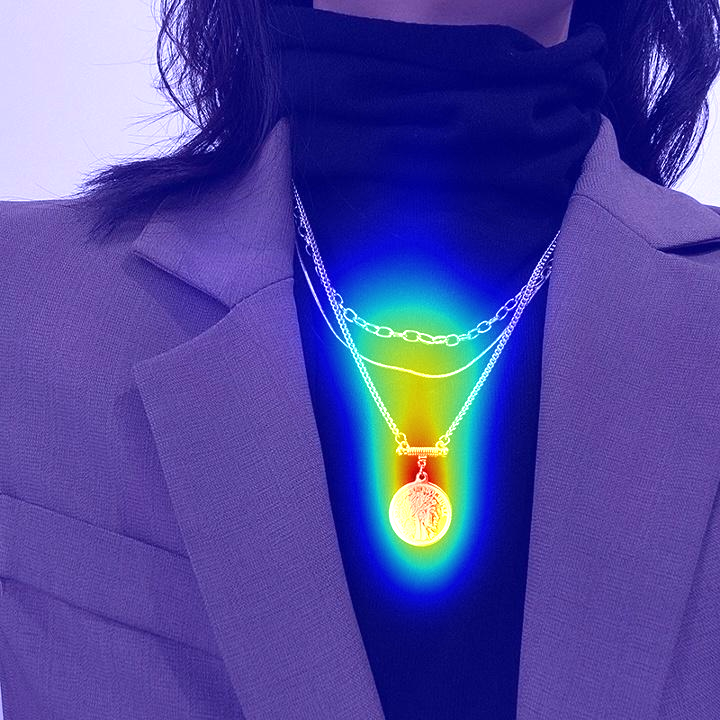} &
    \includegraphics[width=0.155\textwidth]{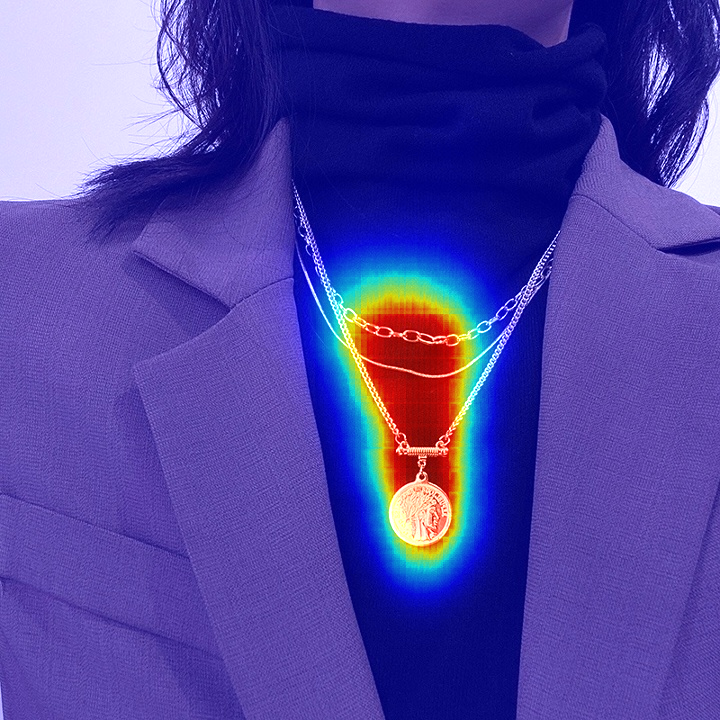} &
    \includegraphics[width=0.155\textwidth]{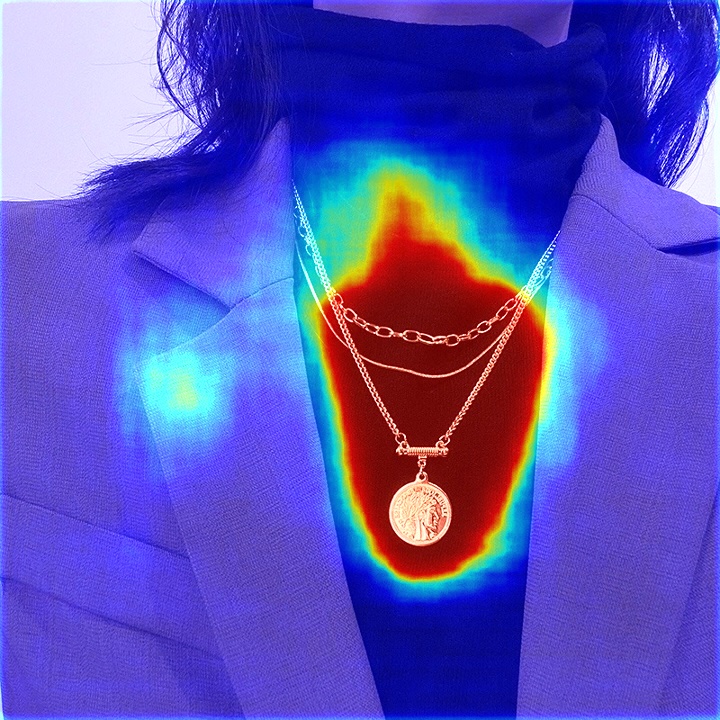} &
    \includegraphics[width=0.155\textwidth]{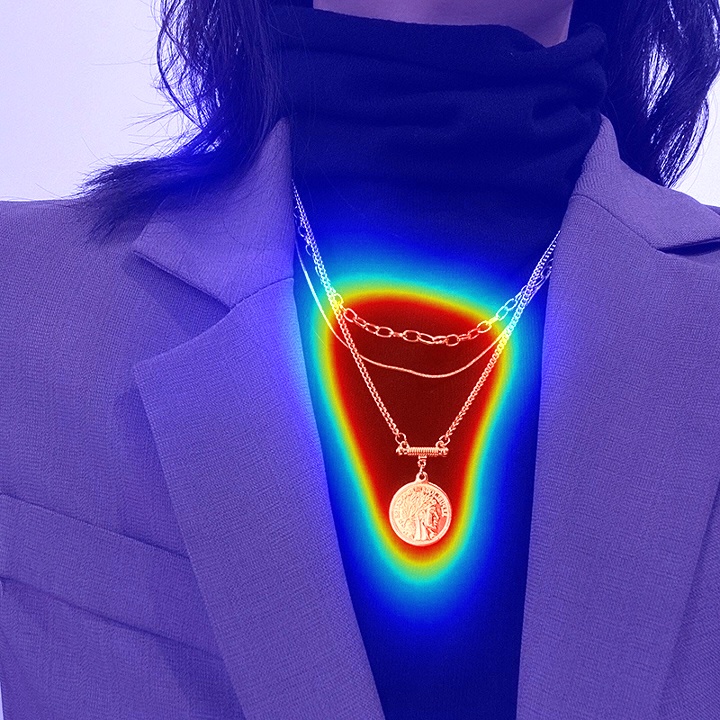} &
    \includegraphics[width=0.155\textwidth]{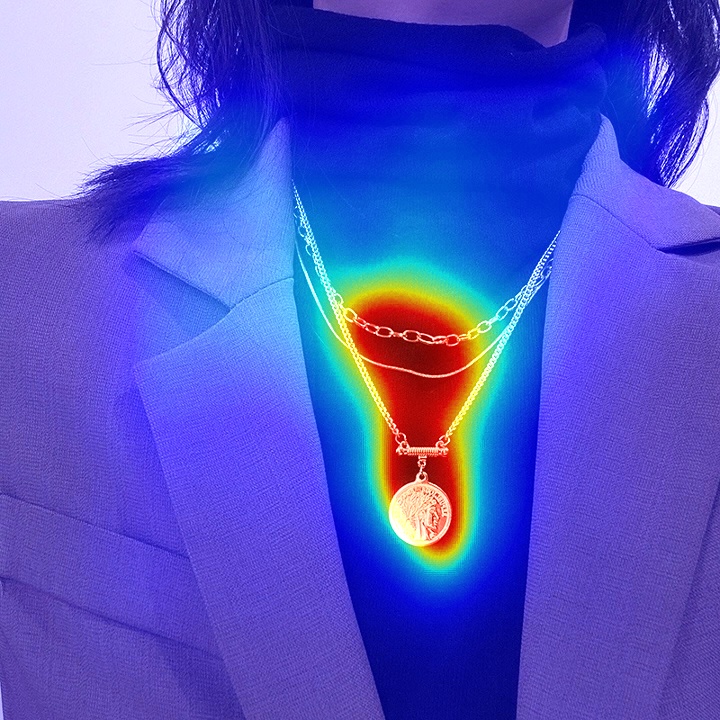} &
    \includegraphics[width=0.155\textwidth]{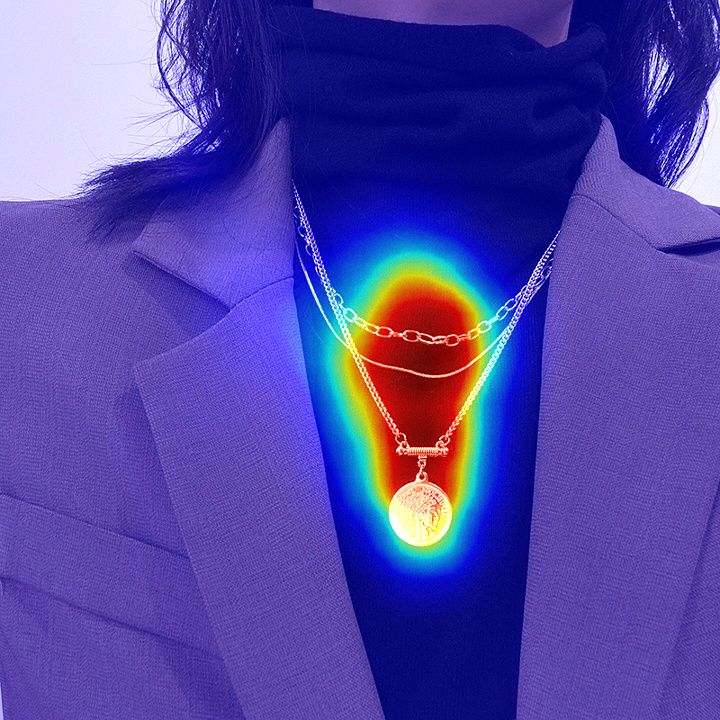}
    \\
    {\footnotesize Input Image} & {\footnotesize Ground Truth} &
    {\footnotesize \textbf{OpenVAM}} &
    {\footnotesize SUM~\cite{hosseini2025sum}} &
    {\footnotesize Transalnet~\cite{lou2022transalnet}} &
    {\footnotesize Temp-Sal~\cite{aydemir2023tempsal}} &
    {\footnotesize DeepGaze~\cite{linardos2021deepgaze}} &
    {\footnotesize BrandAttn~\cite{hosseini2024brand}} \\

    \includegraphics[width=0.155\textwidth]{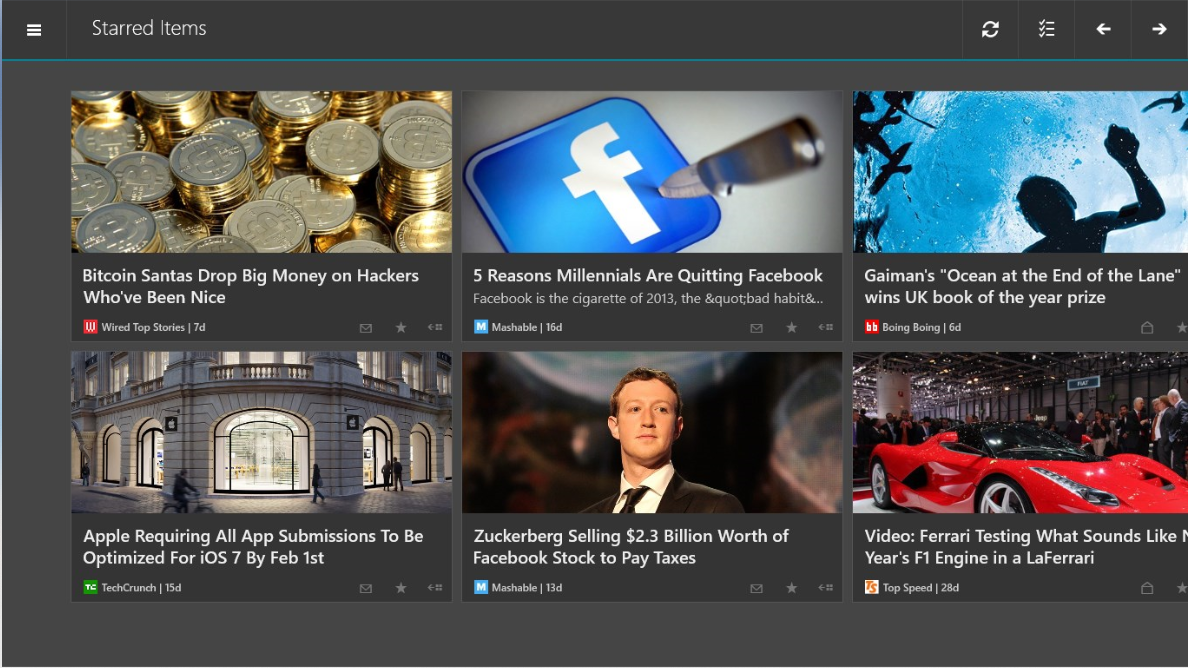} &
    \includegraphics[width=0.155\textwidth]{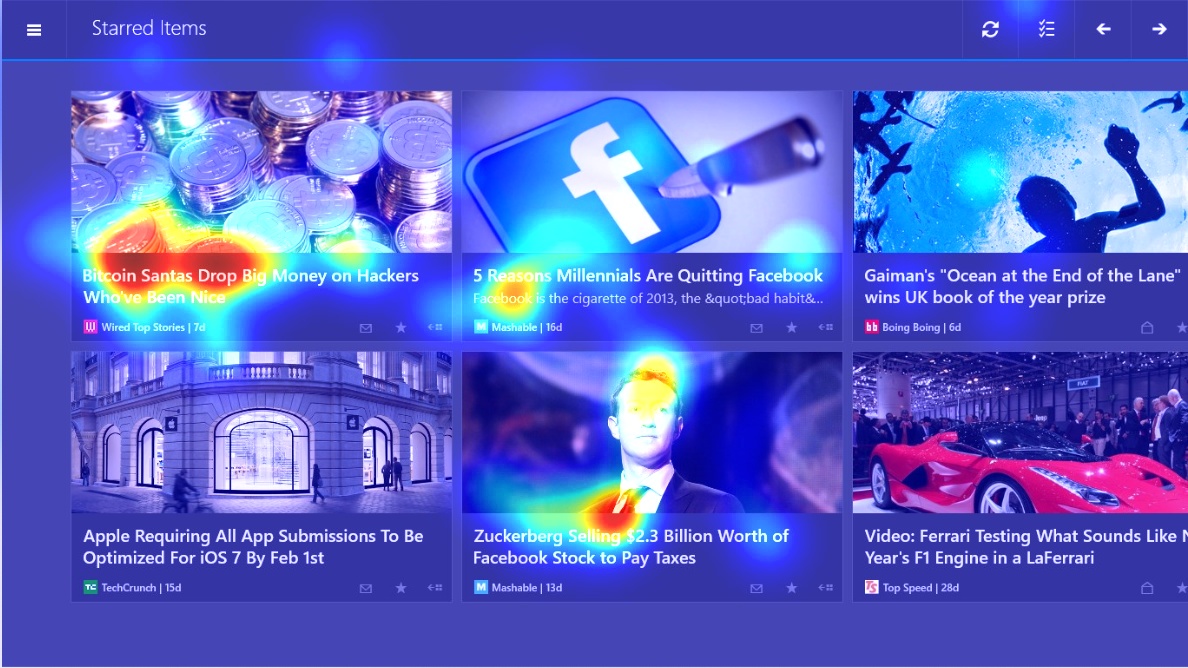} &
    \includegraphics[width=0.155\textwidth]{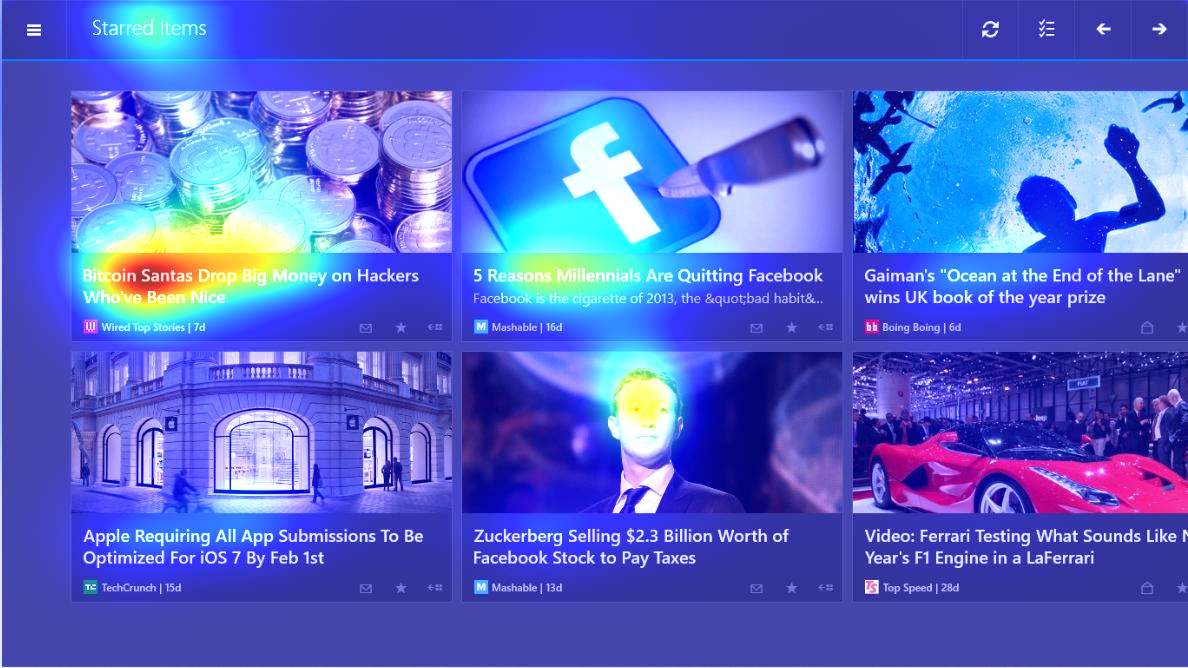} &
    \includegraphics[width=0.155\textwidth]{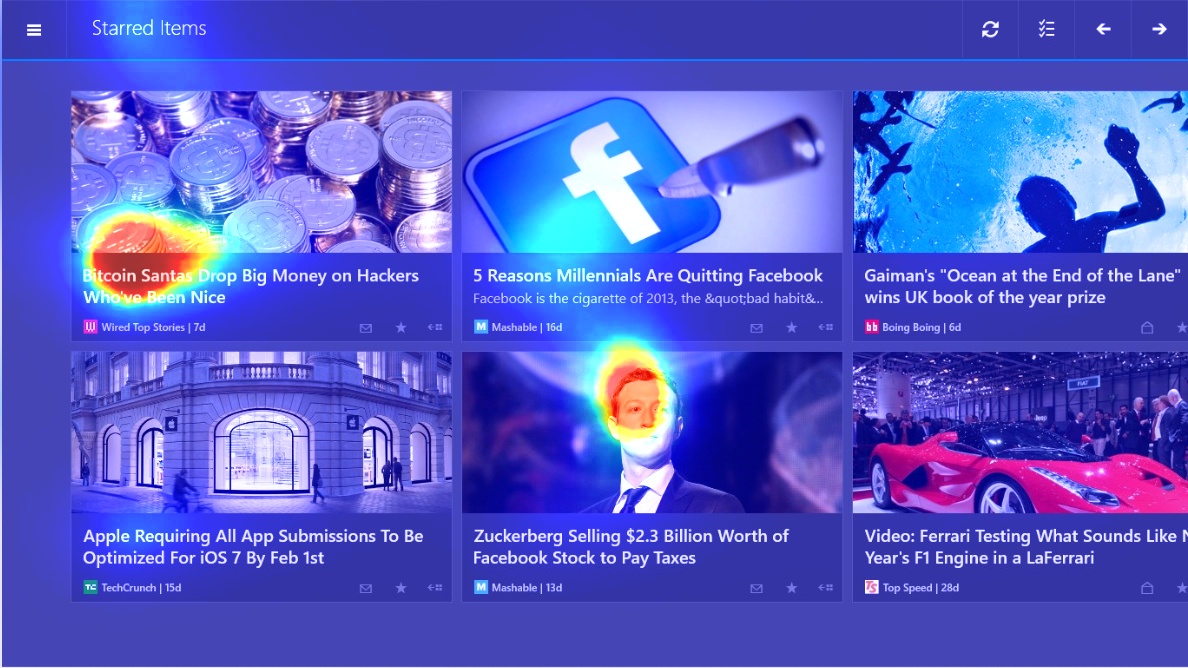} &
    \includegraphics[width=0.155\textwidth]{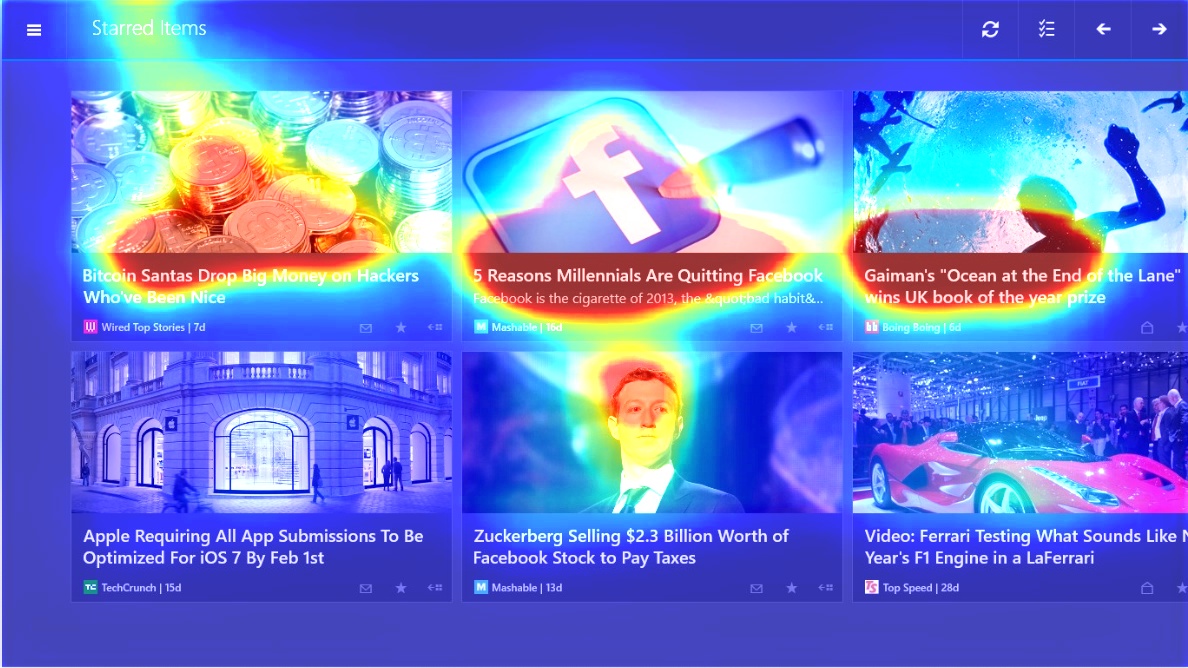} &
    \includegraphics[width=0.155\textwidth]{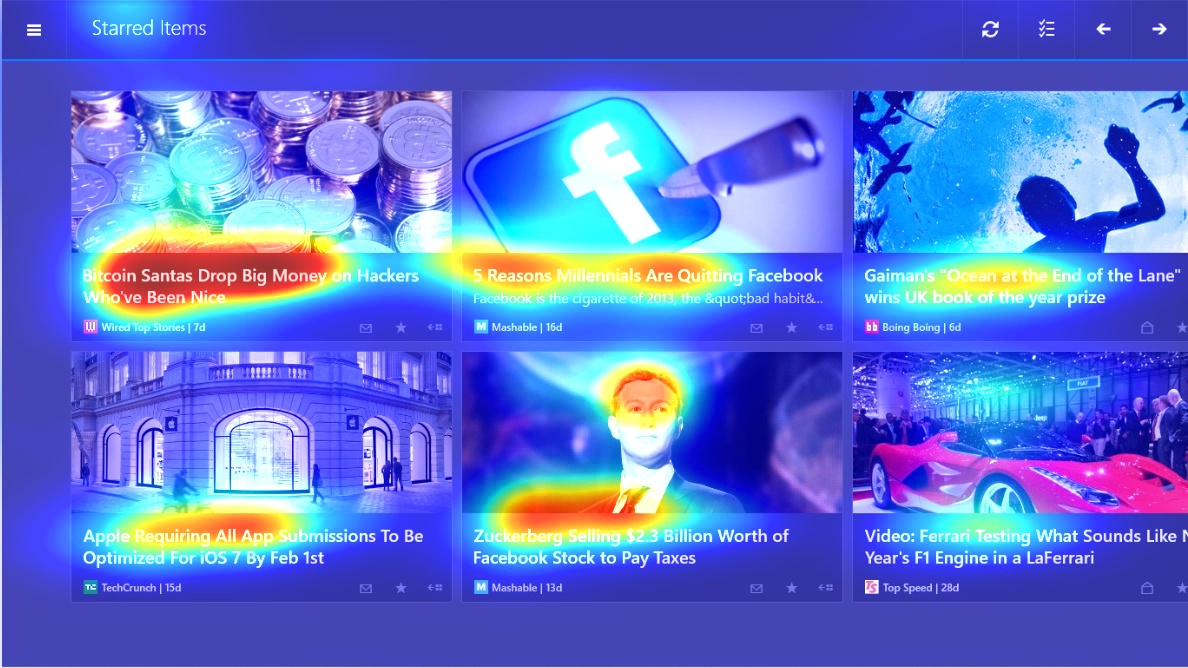} &
    \includegraphics[width=0.155\textwidth]{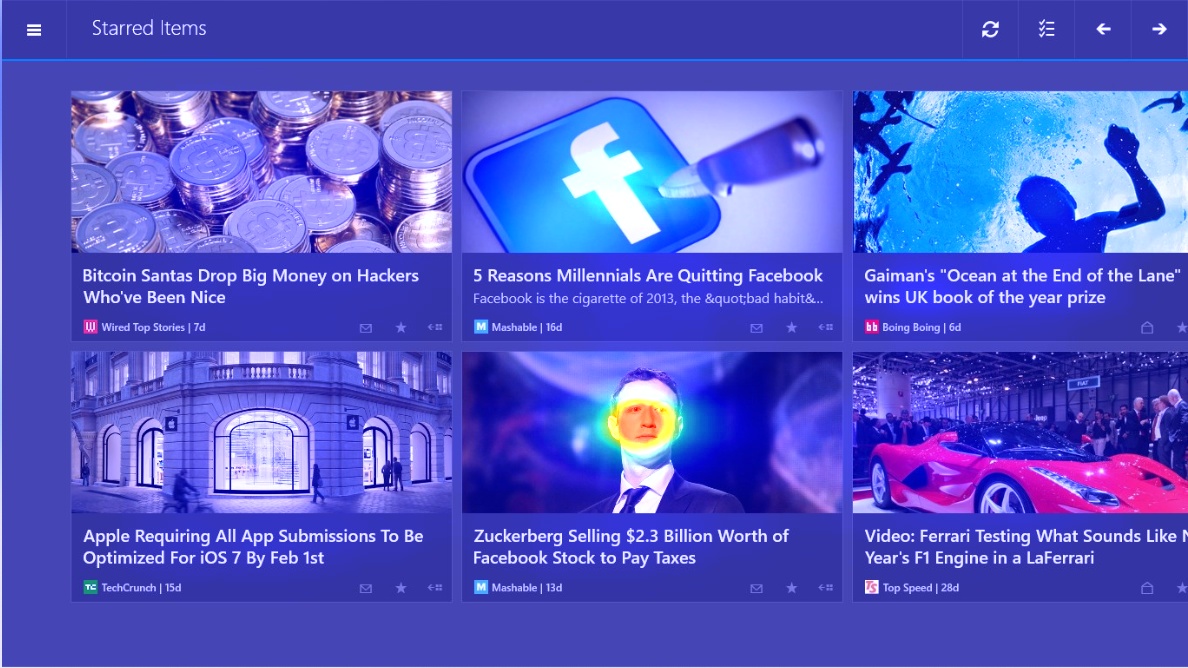} &
    \includegraphics[width=0.155\textwidth]{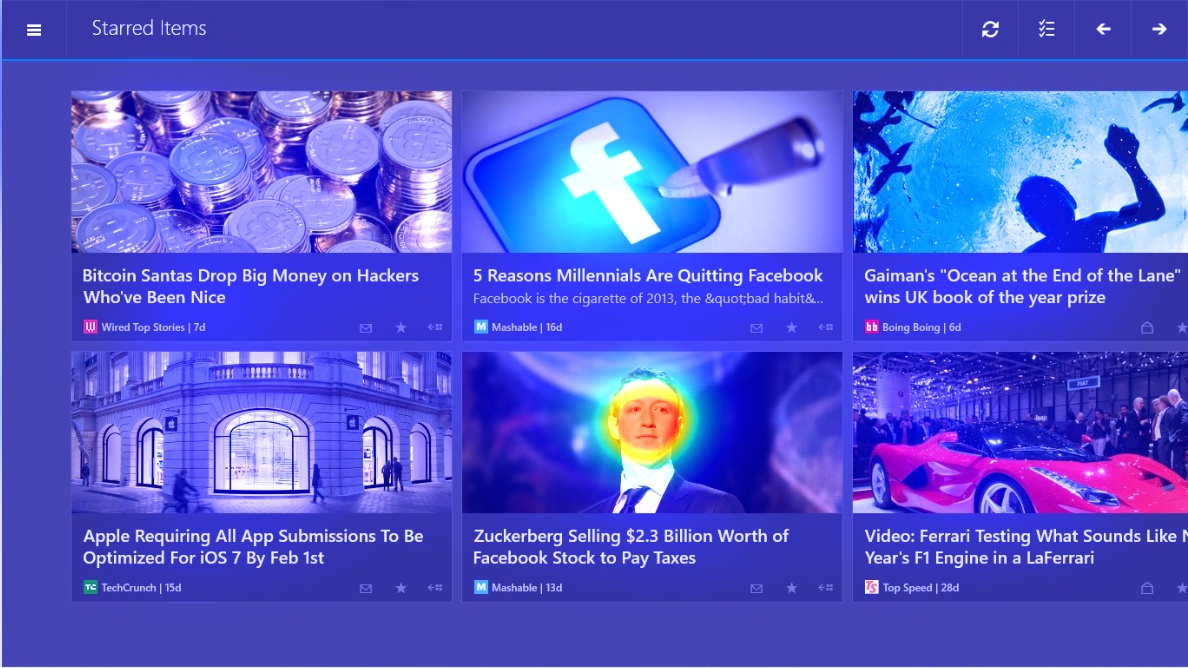}
    \\
    {\footnotesize Input Image} & {\footnotesize Ground Truth} &
    {\footnotesize \textbf{OpenVAM}} &
    {\footnotesize SUM~\cite{hosseini2025sum}} &
    {\footnotesize Transalnet~\cite{lou2022transalnet}} &
    {\footnotesize UMSI++~\cite{jiang2023ueyes}} &
    {\footnotesize SAM++~\cite{jiang2023ueyes}} &
    {\footnotesize UMSI~\cite{fosco2020predicting}} \\

    \includegraphics[width=0.155\textwidth]{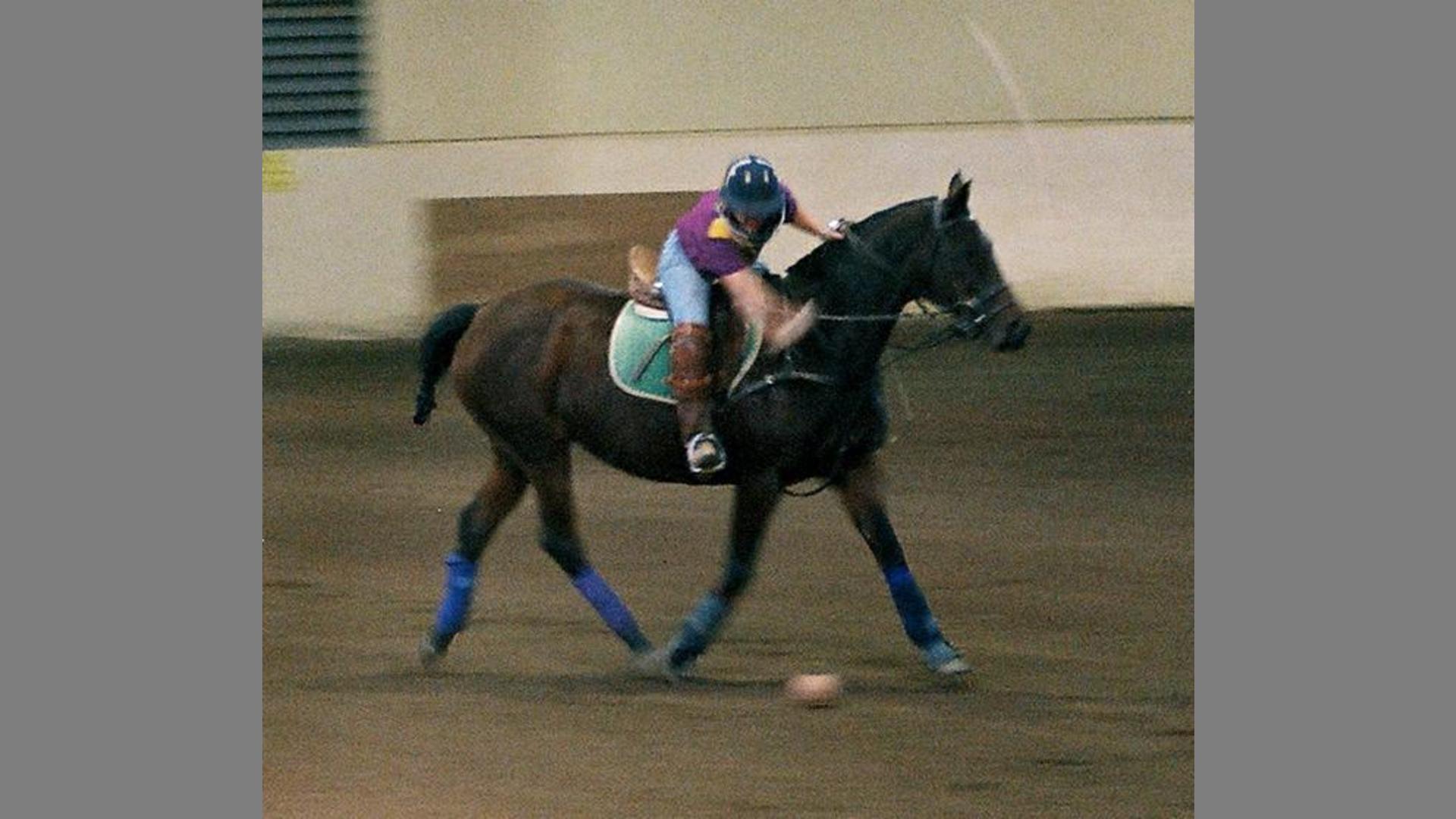} &
    \includegraphics[width=0.155\textwidth]{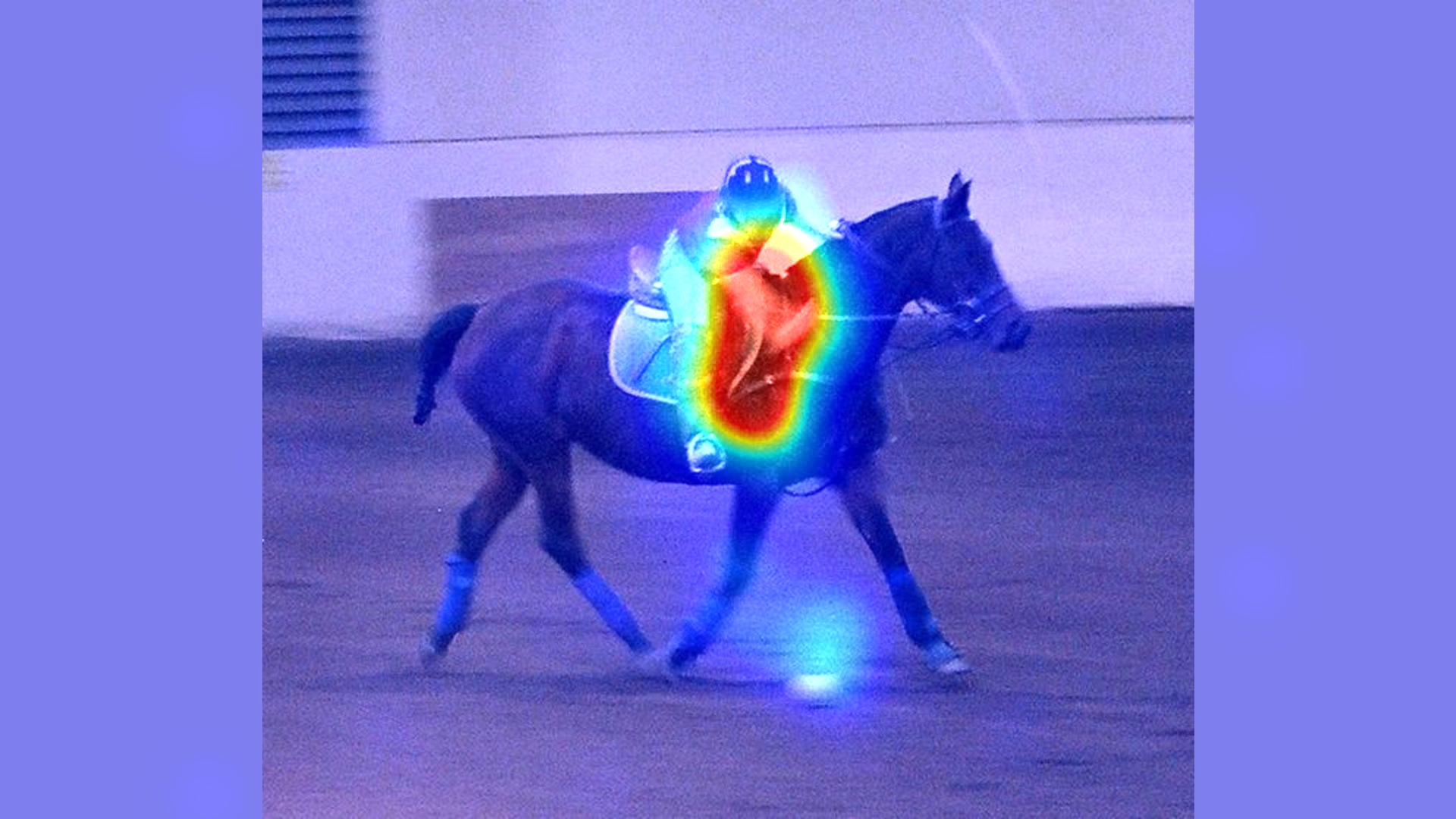} &
    \includegraphics[width=0.155\textwidth]{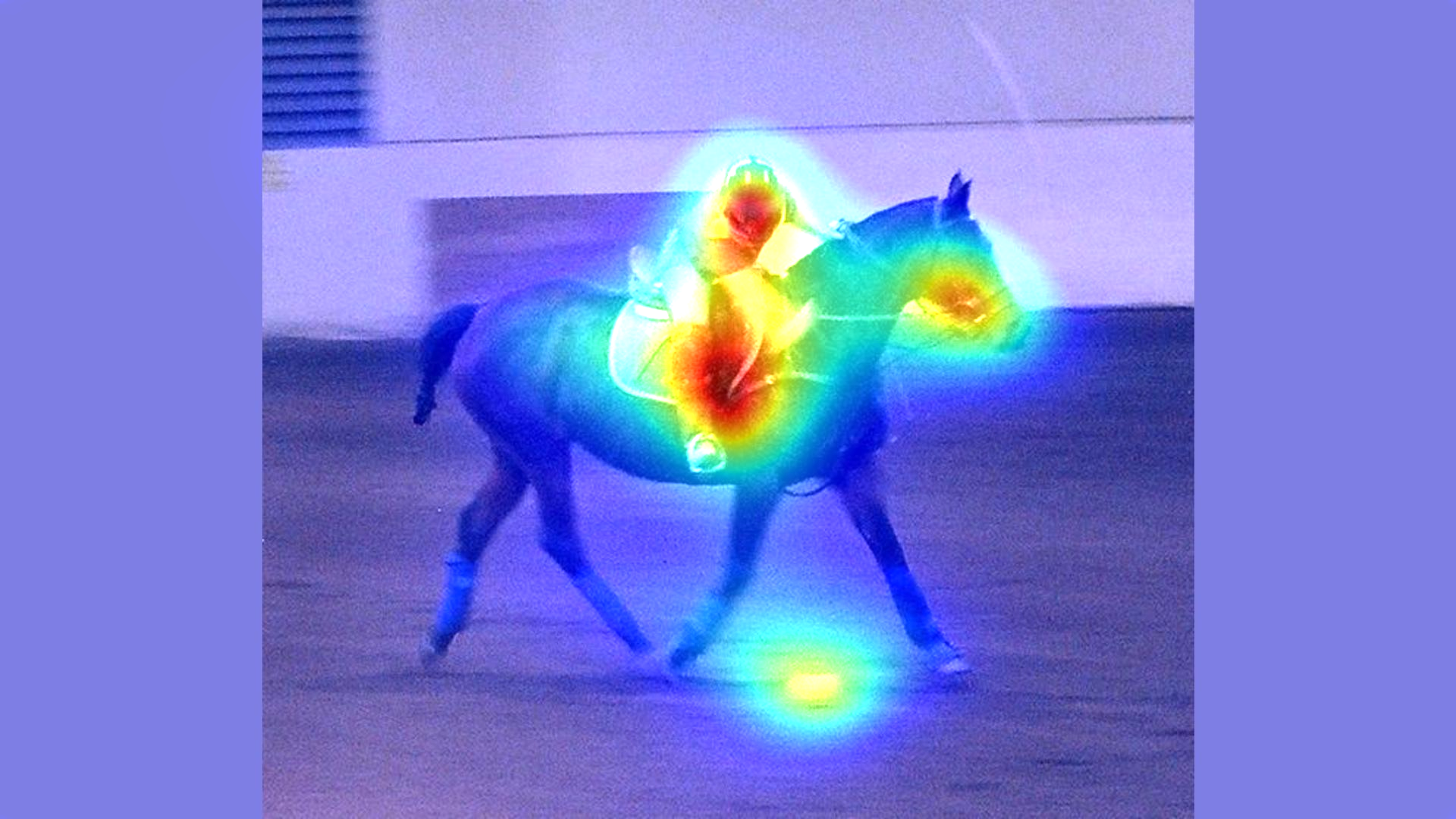} &
    \includegraphics[width=0.155\textwidth]{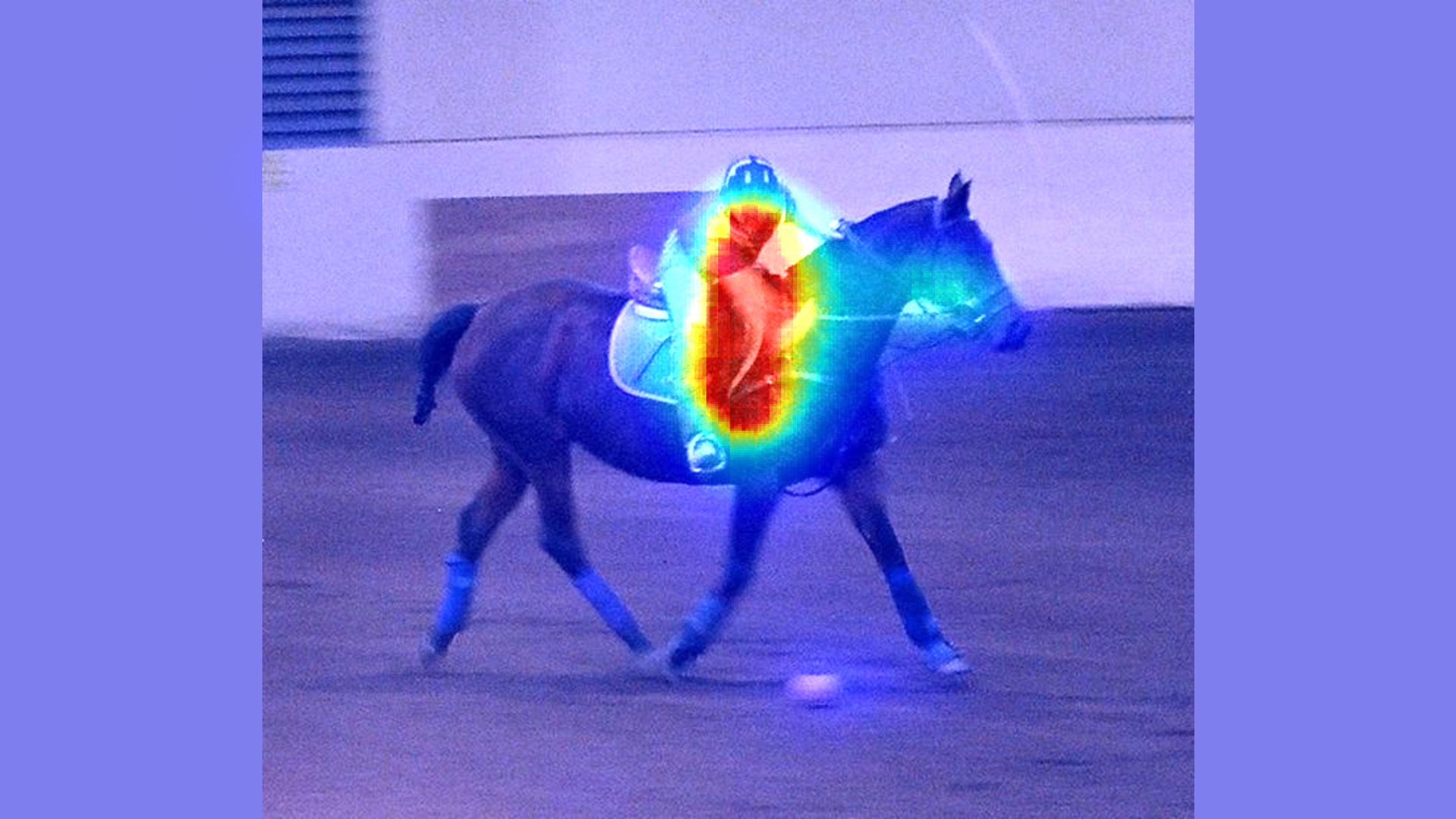} &
    \includegraphics[width=0.155\textwidth]{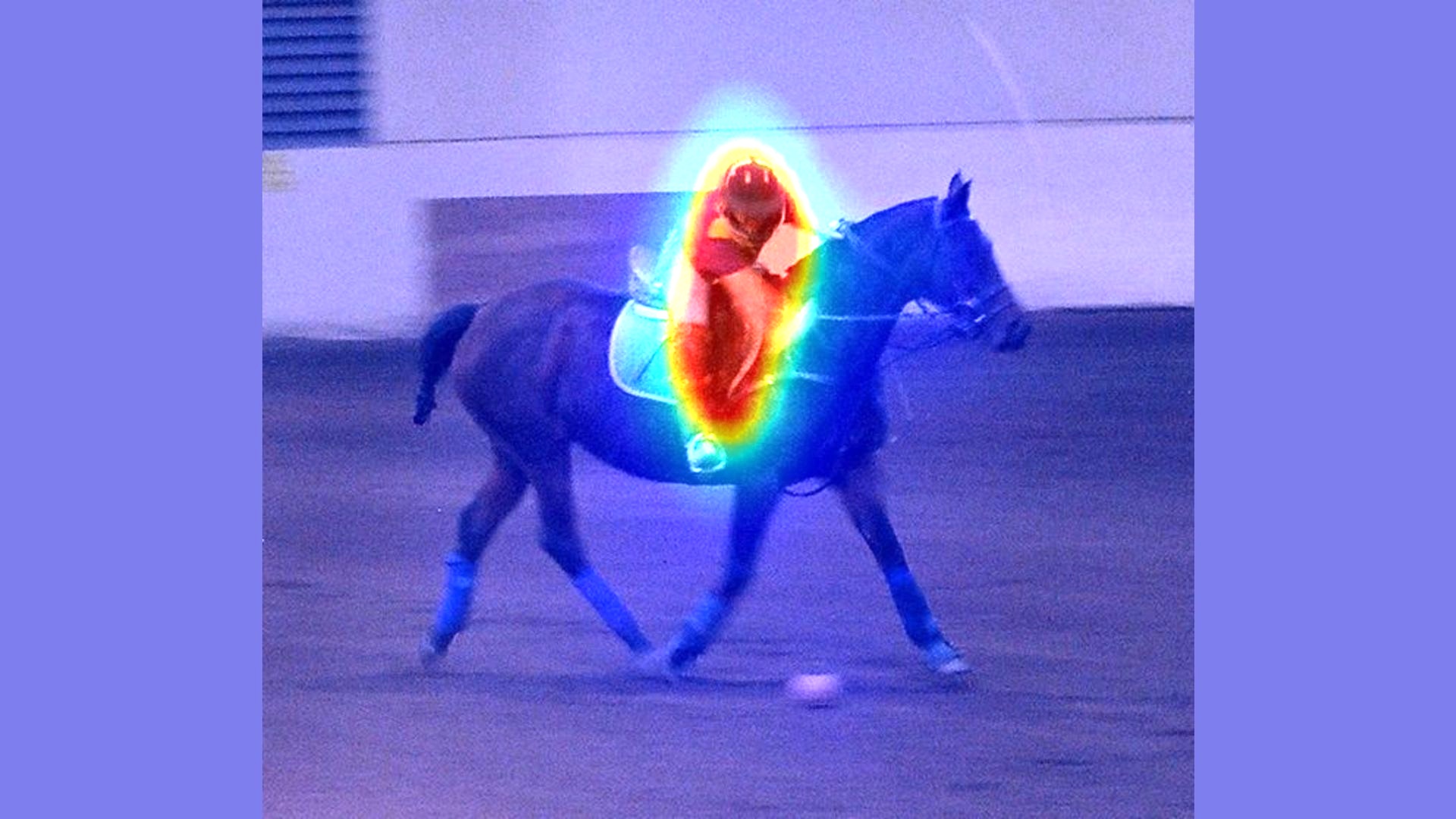} &
    \includegraphics[width=0.155\textwidth]{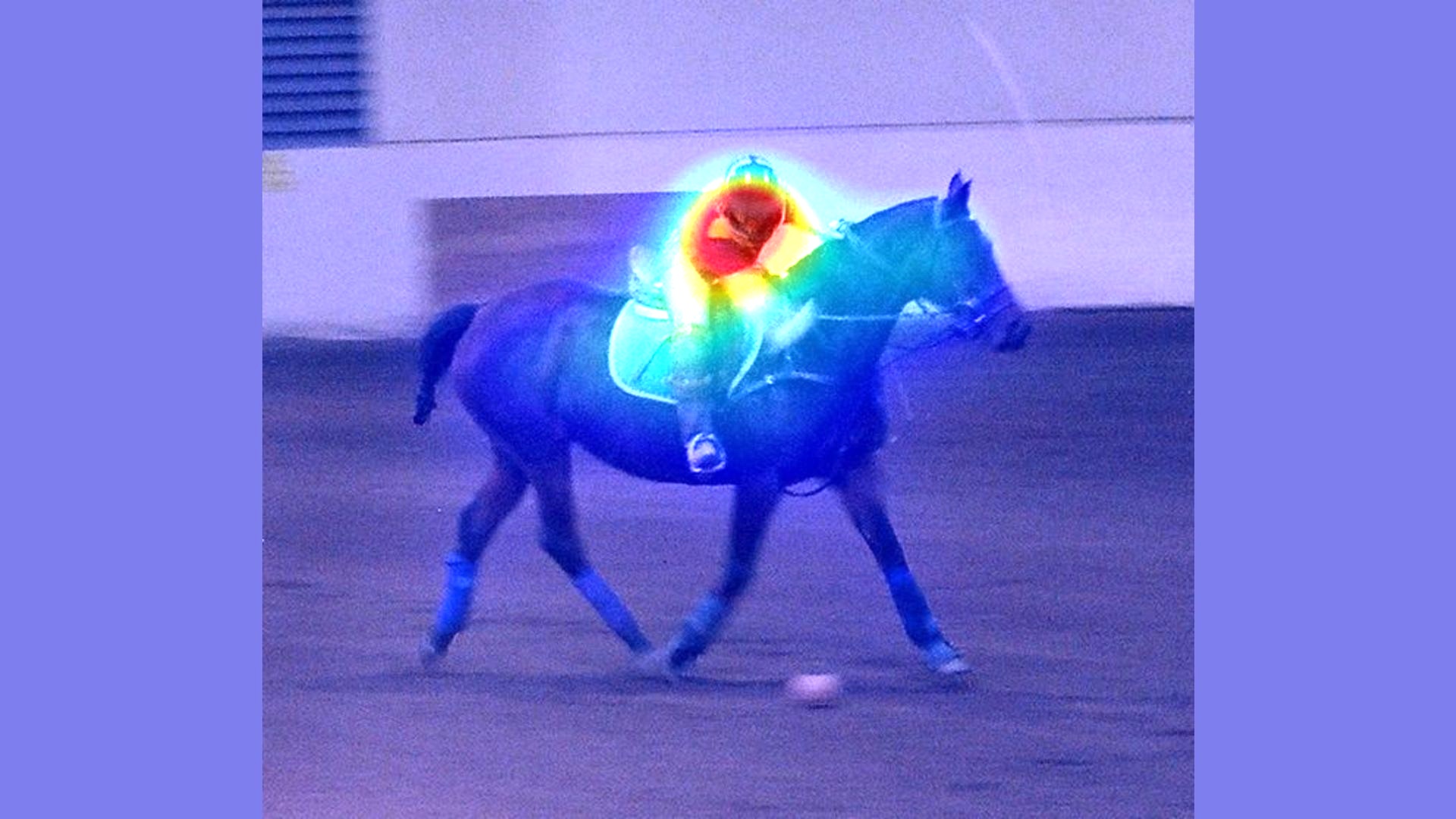} &
    \includegraphics[width=0.155\textwidth]{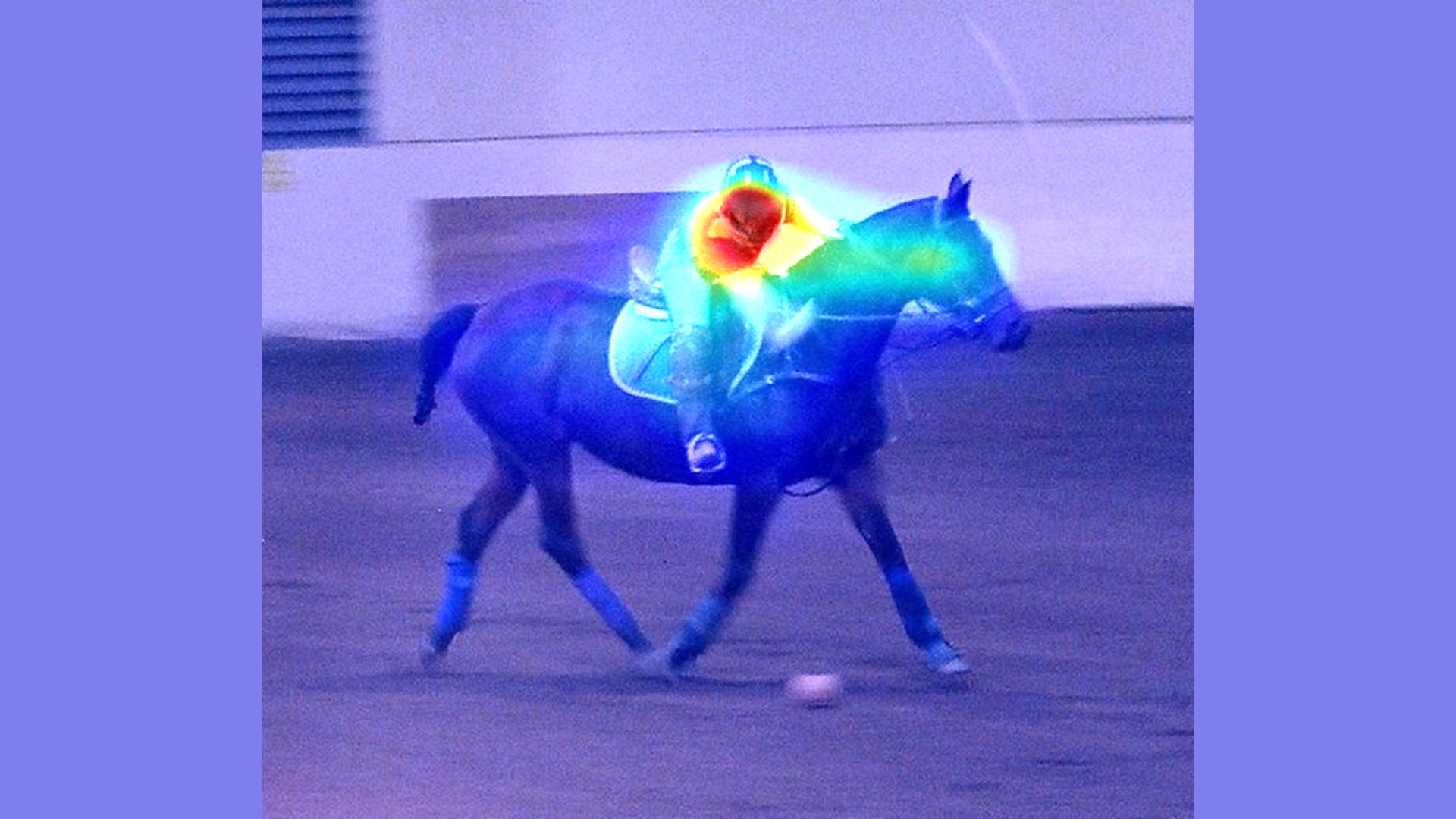} &
    \includegraphics[width=0.155\textwidth]{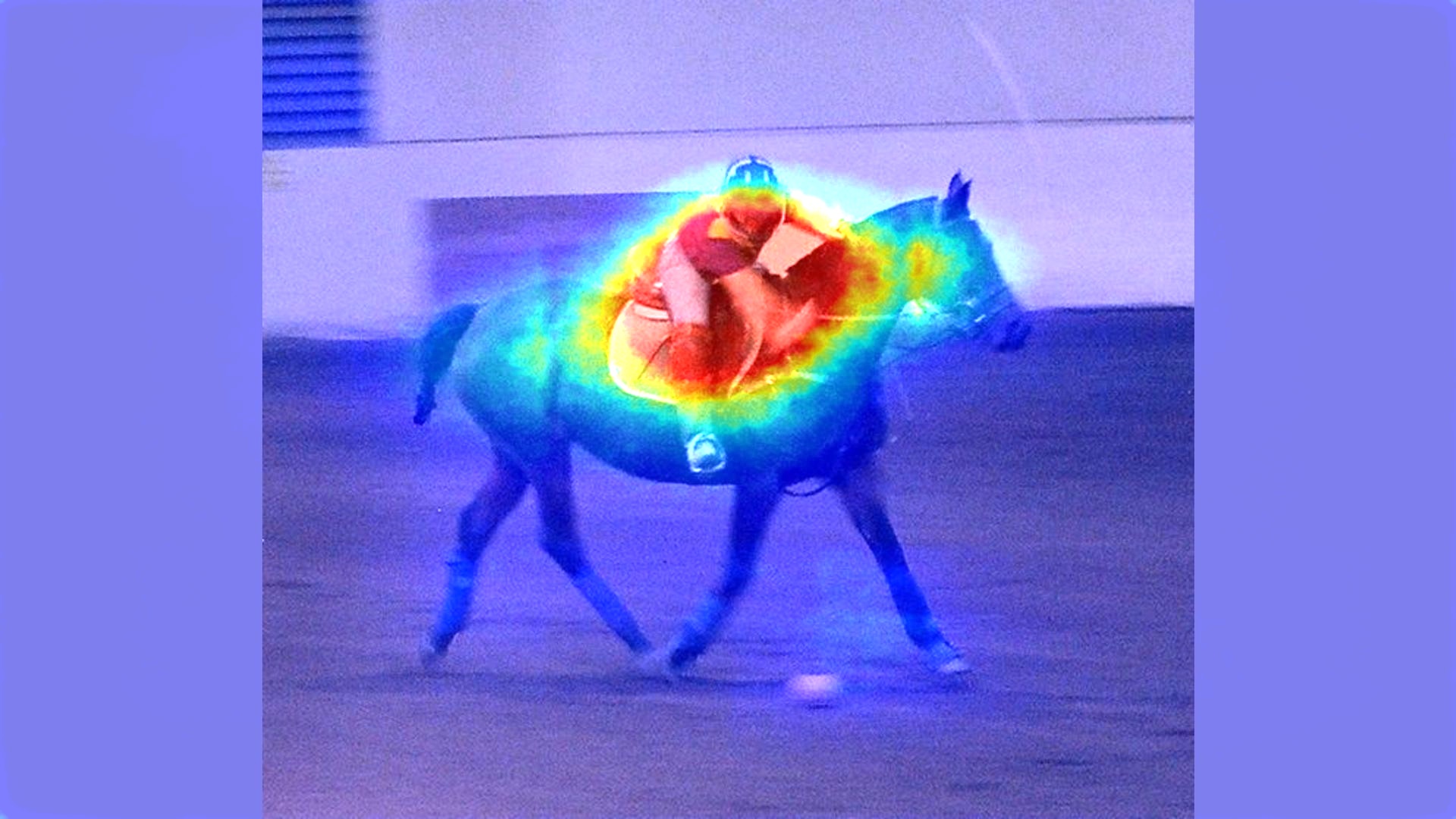}
    \\
    {\footnotesize Input Image} & {\footnotesize Ground Truth} &
    {\footnotesize \textbf{OpenVAM}} &
    {\footnotesize SUM~\cite{hosseini2025sum}} &
    {\footnotesize Transalnet~\cite{lou2022transalnet}} &
    {\footnotesize UNISAL~\cite{droste2020unified}} &
    {\footnotesize EML-NET~\cite{jia2020eml}} &
    {\footnotesize FastSal~\cite{hu2021fastsal}} \\
    \end{tabular}
    }
    \vspace{-0.8em}
    \caption{\textbf{Qualitative comparison of saliency predictions across domains.}
    Rows~1 and~4 show natural-scene images, Row~2 shows an e-commerce/commercial image, and Row~3 shows a UI/web layout.}
    \label{fig:visual_sal}
    \vspace{-0.75em}
\end{figure*}
\begin{table*}[!]
    \centering
    \caption{\textbf{Qualitative text comparison.} OpenVAM explanations versus ground-truth text for Samples~1--4, respectively.}
    \vspace{-1em}
    \label{tab:openvam_text_samples}
    \resizebox{\textwidth}{!}{%
    \setlength{\tabcolsep}{5pt}%
    \fontsize{6.5pt}{7.2pt}\selectfont
    \setlength{\parskip}{0pt}
    \renewcommand{\arraystretch}{0.4}%
    \begin{tabular}{c|p{0.31\textwidth}|p{0.31\textwidth}|p{0.31\textwidth}}
    \toprule
    \textbf{Sample} & \textbf{OpenVAM-3B} & \textbf{OpenVAM-7B} & \textbf{Ground Truth} \\
    \midrule
    \rowcolor{blue!6}
    Sample 1 &
    \textbf{The girl's face (center-left):} Faces are inherently attention-grabbing, and her direct gaze towards the television makes it a strong focal point. \textbf{The television screen content (right-center):} The image of an animal on the TV display creates high contrast against its dark frame. Its central placement within that part of this visual field also draws attention. \textbf{The girl's hands and arms (center-right):} Her active pose with raised hand suggests motion or interaction, drawing the eye to where she is engaged in an activity. &
    \textbf{The woman (center-left):} She is the largest subject, and her bright blue sweater creates strong color contrast against the more muted tones of the room. Her active pose and profile view direct attention toward the television.
\textbf{The television screen (center):} The woman's gaze and hands point directly to the screen, which is a focal point of interaction. The screen emits light and displays a distinct image, adding saliency. \textbf{The cat image on TV (center):} The animal picture on the screen is visually prominent and high-contrast compared to the rest. &
    \textbf{Woman's face (mid-left):} It is a human face, which naturally draws significant attention, and her head is angled, indicating focus. \textbf{Woman's arms (mid-right):} Positioned in an animated way, suggesting action and interaction with the TV. \textbf{TV screen (mid-right):} It's bright and displays an image of a cat, which attracts attention due to contrast and content. \\
    
    \midrule

    \rowcolor{orange!6}
    Sample 2 &
    \textbf{Pendant (bottom-center):} The coin-like pendant is a distinct and detailed feature of the necklace, drawing the eye with its circular shape and shiny texture at the lowest point of the jewelry. \textbf{Blazer and turtleneck (surrounding center):} The combination of a brown blazer over a black turtleneck provides a strong color contrast and frames the necklace, helping it stand out. \textbf{Necklace chain (upper center):} The chain is visible against the dark turtleneck and leads the eye down to the pendant. &
    \textbf{Pendant (bottom-center):} The coin-like pendant is a distinct and detailed feature of the necklace, drawing the eye with its circular shape and shiny texture at the lowest point of the jewelry.
\textbf{Blazer and turtleneck (surrounding center):} The combination of a brown blazer over a black turtleneck provides a stylish backdrop and context, with the dark turtleneck creating high contrast that highlights the necklace. &
    \textbf{The large circular pendant (lower center):} Its significant size and distinctive round shape make it a focal point. The reflective surface of the pendant further enhances its visual prominence.
    \textbf{The black turtleneck shirt (center):} The solid, dark color of the shirt provides a clean and contrasting background that effectively highlights the bright, metallic details of the necklace positioned against it. \\

    \midrule

    \rowcolor{purple!6}
    Sample 3 &
    \textbf{``Facebook logo'' (center):} The large, centered Facebook logo is highly recognizable and draws immediate attention due to its size and color contrast.
    \textbf{``Bitcoin Santa'' image (top-left):} The image of coins with a Santa hat is visually striking, with bright colors and a unique concept that captures attention.
    \textbf{Article headline (center):} The headline text is prominent, with bold lettering that stands out against the background.
    \textbf{Article headline: ``Zuckerberg Sells \$2 Billion Worth Of Facebook Stock To Pay Taxes'' (center):} The headline is bold and positioned centrally, making it a focal point due to its significant financial information.
    &
    \textbf{Headline text (top-left):} This text is large, bold, and positioned at the top of the first content card, making it a primary focal point due to its prominence and contrast against the background image of coins.
    \textbf{Headline text (center):} Located in the center of the grid, this headline is clearly legible, uses a bold font, and is overlaid on a high-contrast image of a Facebook logo being cut, drawing strong visual attention.
    \textbf{Zuckerberg face (center-lower):} Positioned centrally in the lower half of the grid, increasing its visual prominence.
    &
    \textbf{Headline text (top-left content card):} This headline text stands out prominently due to its large size, clear readability, and interesting keywords. The image of shiny gold coins immediately below it is highly distinctive and detailed.
    \textbf{Headline text (top-center content card):} The headline text is clearly readable and captures attention with a specific number and well-known brand name. The associated image features a distinct company logo in a contrasting visual context.
    \textbf{Image of Mark Zuckerberg's face (bottom-center content card):} The presence of a clear human face in the image is a strong visual cue that naturally draws attention. The headline above it reinforces its prominence with a large financial value. \\

    \midrule

    \rowcolor{gray!6}
    Sample 4 &
    \textbf{The rider (center):} The human figure's dynamic pose and brightly colored purple and yellow shirt create strong cues for motion and color contrast, making them a primary focal point.
    \textbf{The horse (center):} As the largest subject in the frame, its dark color contrasts with the lighter background, and its extended legs strongly imply forward motion. &
    \textbf{The rider's upper body (center):} The human figure, especially the head and shoulders of the rider, naturally draws significant attention. Its central placement on a unique animal also contributes to its prominence.
    \textbf{The horse's main frame (center-right foreground):} As an unusual subject in this scene with strong contrast against its surroundings, it occupies a large portion of the image and is sharply focused. The distinct shape of the saddle on top further highlights it as part of a unique animal.
    \textbf{The horse's main body (center):} This area completes their presence within an unusual context in this scene, drawing attention due to its role in supporting and interacting with the main figure.
    \textbf{Small object near horse’s front leg (lower center):} A small, light-colored object near the horse’s front hoof is visible, but its identity is unclear due to low resolution and lack of context. &
    \textbf{Rider's upper body (center):} This area is prominently positioned and brightly illuminated, creating a strong visual contrast with the darker horse. The rider's posture suggests an active pose, drawing the eye to the central human element.
    \textbf{Horse's head and neck (upper center):} This distinct part of the animal is clearly visible and directly engaged with the rider, making it a focal point due to its prominent features and interaction.
    \textbf{Horse's main body (center):} The large, dark form of the horse's back and saddle area creates a significant visual mass in the middle of the frame, providing a stable foundation for the rider.
    \textbf{The small white ball (bottom foreground):} The ball's light color stands out against the darker ground, and its position in the path of the horse suggests it is the focus of the central action.\\

    \bottomrule
    \end{tabular}%
    }
    \vspace{-1.5em}
\end{table*}
\begin{table*}[!t]
\centering
\caption{\textbf{Evaluation comparison across VLMs on multiple datasets.} JScore measures semantic quality, while ROUGE and BLEU measure lexical overlap. Best results are bolded within each model-size group. \textbf{PV} denotes proprietary models.}
\vspace{-.5em}
\label{tab:vlm_multi_dataset}
\providecommand{\bestvlm}[1]{\textbf{#1}}
\providecommand{\ourscell}[1]{\cellcolor[HTML]{DADCFF}#1}
\resizebox{\textwidth}{!}{%
\setlength{\tabcolsep}{16pt}%
\renewcommand{\arraystretch}{1.25}%
\scriptsize
\begin{tabular}{c|c|l|c|ccc|cccc}
\toprule
Dataset & Size & Methods
& \textbf{JScore} $\uparrow$
& \multicolumn{3}{c|}{\textbf{ROUGE} $\uparrow$}
& \multicolumn{4}{c}{\textbf{BLEU} $\uparrow$} \\
& & & & R-1 & R-2 & R-L & B-1 & B-2 & B-3 & B-4 \\
\midrule

\multirow{12}{*}{\textit{U-EYE}~\cite{jiang2023ueyes}}
& PV
& \cellcolor{gray!15} Gemini-2.5-Pro
& \cellcolor{gray!15}0.739 & \cellcolor{gray!15}0.511 & \cellcolor{gray!15}0.172 & \cellcolor{gray!15}0.253
& \cellcolor{gray!15}0.403 & \cellcolor{gray!15}0.225 & \cellcolor{gray!15}0.129 & \cellcolor{gray!15}0.074 \\
\cmidrule{2-11}


& \multirow{2}{*}{3B}
& Qwen2.5-VL-3B
& 0.537 & 0.391 & 0.103 & \bestvlm{0.198}
& 0.300 & 0.151 & \bestvlm{0.080} & \bestvlm{0.045} \\
& & \ourscell{\textbf{OpenVAM-3B}}
& \ourscell{\bestvlm{0.603}} & \ourscell{\bestvlm{0.429}} & \ourscell{\bestvlm{0.108}} & \ourscell{0.196}
& \ourscell{\bestvlm{0.330}} & \ourscell{\bestvlm{0.156}} & \ourscell{0.077} & \ourscell{0.040} \\
\cmidrule{2-11}

& \multirow{2}{*}{4B}
& Qwen3-VL-4B
& \bestvlm{0.679} & \bestvlm{0.444} & \bestvlm{0.119} & 0.207
& \bestvlm{0.345} & \bestvlm{0.171} & \bestvlm{0.090} & \bestvlm{0.049} \\
& & \ourscell{\textbf{OpenVAM-4B}}
& \ourscell{0.675} & \ourscell{0.399} & \ourscell{0.118} & \ourscell{\bestvlm{0.220}}
& \ourscell{0.317} & \ourscell{0.163} & \ourscell{0.088} & \ourscell{\bestvlm{0.049}} \\
\cmidrule{2-11}

& \multirow{2}{*}{7B}
& Qwen2.5-VL-7B
& \bestvlm{0.637} & \bestvlm{0.454} & 0.132 & 0.216
& 0.345 & \bestvlm{0.180} & \bestvlm{0.102} & \bestvlm{0.059} \\
& & \ourscell{\textbf{OpenVAM-7B}}
& \ourscell{0.628} & \ourscell{0.445} & \ourscell{\bestvlm{0.139}} & \ourscell{\bestvlm{0.221}}
& \ourscell{\bestvlm{0.355}} & \ourscell{0.175} & \ourscell{0.098} & \ourscell{0.057} \\
\cmidrule{2-11}

& \multirow{2}{*}{8B}
& Qwen3-VL-8B
& \bestvlm{0.689} & \bestvlm{0.484} & \bestvlm{0.149} & \bestvlm{0.231}
& \bestvlm{0.387} & \bestvlm{0.207} & \bestvlm{0.116} & \bestvlm{0.065} \\
& & \ourscell{\textbf{OpenVAM-8B}}
& \ourscell{0.683} & \ourscell{0.407} & \ourscell{0.121} & \ourscell{0.218}
& \ourscell{0.330} & \ourscell{0.171} & \ourscell{0.091} & \ourscell{0.051} \\
\midrule

\multirow{12}{*}{\textit{SalECI}~\cite{jiang2022does}}
& PV
& \cellcolor{gray!15} Gemini-2.5-Pro
& \cellcolor{gray!15}0.741 & \cellcolor{gray!15}0.519 & \cellcolor{gray!15}0.162 & \cellcolor{gray!15}0.254
& \cellcolor{gray!15}0.400 & \cellcolor{gray!15}0.215 & \cellcolor{gray!15}0.119 & \cellcolor{gray!15}0.065 \\
\cmidrule{2-11}


& \multirow{2}{*}{3B}
& Qwen2.5-VL-3B
& 0.548 & 0.357 & 0.076 & 0.179
& 0.270 & 0.119 & 0.053 & 0.026 \\
& & \ourscell{\textbf{OpenVAM-3B}}
& \ourscell{\bestvlm{0.692}} & \ourscell{\bestvlm{0.478}} & \ourscell{\bestvlm{0.130}} & \ourscell{\bestvlm{0.220}}
& \ourscell{\bestvlm{0.369}} & \ourscell{\bestvlm{0.184}} & \ourscell{\bestvlm{0.096}} & \ourscell{\bestvlm{0.050}} \\
\cmidrule{2-11}

& \multirow{2}{*}{4B}
& Qwen3-VL-4B
& 0.720 & 0.400 & 0.087 & 0.187
& 0.287 & 0.122 & 0.054 & 0.027 \\
& & \ourscell{\textbf{OpenVAM-4B}}
& \ourscell{\bestvlm{0.730}} & \ourscell{\bestvlm{0.465}} & \ourscell{\bestvlm{0.139}} & \ourscell{\bestvlm{0.246}}
& \ourscell{\bestvlm{0.369}} & \ourscell{\bestvlm{0.195}} & \ourscell{\bestvlm{0.108}} & \ourscell{\bestvlm{0.060}} \\
\cmidrule{2-11}

& \multirow{2}{*}{7B}
& Qwen2.5-VL-7B
& 0.639 & 0.386 & 0.093 & 0.196
& 0.268 & 0.122 & 0.057 & 0.029 \\
& & \ourscell{\textbf{OpenVAM-7B}}
& \ourscell{\bestvlm{0.712}} & \ourscell{\bestvlm{0.489}} & \ourscell{\bestvlm{0.142}} & \ourscell{\bestvlm{0.248}}
& \ourscell{\bestvlm{0.378}} & \ourscell{\bestvlm{0.196}} & \ourscell{\bestvlm{0.099}} & \ourscell{\bestvlm{0.058}} \\
\cmidrule{2-11}

& \multirow{2}{*}{8B}
& Qwen3-VL-8B
& 0.731 & 0.446 & 0.112 & 0.212
& 0.337 & 0.158 & 0.075 & 0.036 \\
& & \ourscell{\textbf{OpenVAM-8B}}
& \ourscell{\bestvlm{0.739}} & \ourscell{\bestvlm{0.474}} & \ourscell{\bestvlm{0.157}} & \ourscell{\bestvlm{0.250}}
& \ourscell{\bestvlm{0.378}} & \ourscell{\bestvlm{0.208}} & \ourscell{\bestvlm{0.117}} & \ourscell{\bestvlm{0.066}} \\
\midrule

\multirow{12}{*}{\textit{OSIE}~\cite{xu2014predicting}}
& PV
& \cellcolor{gray!15} Gemini-2.5-Pro
& \cellcolor{gray!15}0.747 & \cellcolor{gray!15}0.510 & \cellcolor{gray!15}0.165 & \cellcolor{gray!15}0.270
& \cellcolor{gray!15}0.412 & \cellcolor{gray!15}0.228 & \cellcolor{gray!15}0.126 & \cellcolor{gray!15}0.068 \\
\cmidrule{2-11}


& \multirow{2}{*}{3B}
& Qwen2.5-VL-3B
& 0.584 & 0.390 & 0.096 & 0.208
& 0.314 & 0.153 & 0.074 & 0.037 \\
& & \ourscell{\textbf{OpenVAM-3B}}
& \ourscell{\bestvlm{0.661}} & \ourscell{\bestvlm{0.472}} & \ourscell{\bestvlm{0.123}} & \ourscell{\bestvlm{0.220}}
& \ourscell{\bestvlm{0.364}} & \ourscell{\bestvlm{0.178}} & \ourscell{\bestvlm{0.089}} & \ourscell{\bestvlm{0.047}} \\
\cmidrule{2-11}

& \multirow{2}{*}{4B}
& Qwen3-VL-4B
& 0.727 & 0.419 & 0.097 & 0.201
& 0.316 & 0.143 & 0.069 & 0.035 \\
& & \ourscell{\textbf{OpenVAM-4B}}
& \ourscell{\bestvlm{0.728}} & \ourscell{\bestvlm{0.456}} & \ourscell{\bestvlm{0.146}} & \ourscell{\bestvlm{0.246}}
& \ourscell{\bestvlm{0.360}} & \ourscell{\bestvlm{0.198}} & \ourscell{\bestvlm{0.112}} & \ourscell{\bestvlm{0.062}} \\
\cmidrule{2-11}

& \multirow{2}{*}{7B}
& Qwen2.5-VL-7B
& 0.695 & 0.465 & \bestvlm{0.136} & 0.235
& 0.373 & 0.197 & \bestvlm{0.103} & \bestvlm{0.054} \\
& & \ourscell{\textbf{OpenVAM-7B}}
& \ourscell{\bestvlm{0.697}} & \ourscell{\bestvlm{0.485}} & \ourscell{0.135} & \ourscell{\bestvlm{0.237}}
& \ourscell{\bestvlm{0.376}} & \ourscell{\bestvlm{0.198}} & \ourscell{0.101} & \ourscell{0.052} \\
\cmidrule{2-11}

& \multirow{2}{*}{8B}
& Qwen3-VL-8B
& \bestvlm{0.734} & 0.461 & 0.129 & 0.231
& 0.354 & 0.179 & 0.092 & 0.048 \\
& & \ourscell{\textbf{OpenVAM-8B}}
& \ourscell{0.730} & \ourscell{\bestvlm{0.478}} & \ourscell{\bestvlm{0.158}} & \ourscell{\bestvlm{0.259}}
& \ourscell{\bestvlm{0.390}} & \ourscell{\bestvlm{0.218}} & \ourscell{\bestvlm{0.124}} & \ourscell{\bestvlm{0.069}} \\
\midrule

\multirow{12}{*}{\textit{Salicon}~\cite{jiang2015salicon}}
& PV
& \cellcolor{gray!15} Gemini-2.5-Pro
& \cellcolor{gray!15}0.748 & \cellcolor{gray!15}0.453 & \cellcolor{gray!15}0.136 & \cellcolor{gray!15}0.233
& \cellcolor{gray!15}0.327 & \cellcolor{gray!15}0.174 & \cellcolor{gray!15}0.088 & \cellcolor{gray!15}0.044 \\
\cmidrule{2-11}


& \multirow{2}{*}{3B}
& Qwen2.5-VL-3B
& 0.612 & 0.413 & 0.121 & \bestvlm{0.225}
& 0.301 & 0.159 & 0.082 & 0.042 \\
& & \ourscell{\textbf{OpenVAM-3B}}
& \ourscell{\bestvlm{0.677}} & \ourscell{\bestvlm{0.424}} & \ourscell{\bestvlm{0.123}} & \ourscell{0.220}
& \ourscell{\bestvlm{0.364}} & \ourscell{\bestvlm{0.178}} & \ourscell{\bestvlm{0.089}} & \ourscell{\bestvlm{0.047}} \\
\cmidrule{2-11}

& \multirow{2}{*}{4B}
& Qwen3-VL-4B
& 0.737 & 0.370 & 0.081 & 0.187
& 0.250 & 0.111 & 0.052 & 0.027 \\
& & \ourscell{\textbf{OpenVAM-4B}}
& \ourscell{\bestvlm{0.740}} & \ourscell{\bestvlm{0.448}} & \ourscell{\bestvlm{0.159}} & \ourscell{\bestvlm{0.248}}
& \ourscell{\bestvlm{0.353}} & \ourscell{\bestvlm{0.204}} & \ourscell{\bestvlm{0.118}} & \ourscell{\bestvlm{0.066}} \\
\cmidrule{2-11}

& \multirow{2}{*}{7B}
& Qwen2.5-VL-7B
& 0.703 & 0.467 & 0.151 & 0.243
& 0.339 & 0.187 & 0.101 & 0.053 \\
& & \ourscell{\textbf{OpenVAM-7B}}
& \ourscell{\bestvlm{0.7254}} & \ourscell{\bestvlm{0.468}} & \ourscell{\bestvlm{0.165}} & \ourscell{\bestvlm{0.251}}
& \ourscell{\bestvlm{0.370}} & \ourscell{\bestvlm{0.190}} & \ourscell{\bestvlm{0.102}} & \ourscell{\bestvlm{0.056}} \\
\cmidrule{2-11}

& \multirow{2}{*}{8B}
& Qwen3-VL-8B
& 0.738 & 0.422 & 0.116 & 0.217
& 0.288 & 0.146 & 0.074 & 0.038 \\
& & \ourscell{\textbf{OpenVAM-8B}}
& \ourscell{\bestvlm{0.747}} & \ourscell{\bestvlm{0.454}} & \ourscell{\bestvlm{0.160}} & \ourscell{\bestvlm{0.251}}
& \ourscell{\bestvlm{0.359}} & \ourscell{\bestvlm{0.207}} & \ourscell{\bestvlm{0.120}} & \ourscell{\bestvlm{0.067}} \\
\midrule

\multirow{12}{*}{\textit{CAT2000}~\cite{borji2015cat2000}}
& PV
& \cellcolor{gray!15} Gemini-2.5-Pro
& \cellcolor{gray!15}0.720 & \cellcolor{gray!15}0.473 & \cellcolor{gray!15}0.142 & \cellcolor{gray!15}0.243
& \cellcolor{gray!15}0.372 & \cellcolor{gray!15}0.198 & \cellcolor{gray!15}0.104 & \cellcolor{gray!15}0.055 \\
\cmidrule{2-11}


& \multirow{2}{*}{3B}
& Qwen2.5-VL-3B
& 0.545 & 0.367 & 0.089 & 0.202
& 0.290 & 0.139 & 0.068 & 0.035 \\
& & \ourscell{\textbf{OpenVAM-3B}}
& \ourscell{\bestvlm{0.663}} & \ourscell{\bestvlm{0.442}} & \ourscell{\bestvlm{0.111}} & \ourscell{\bestvlm{0.217}}
& \ourscell{\bestvlm{0.333}} & \ourscell{\bestvlm{0.162}} & \ourscell{\bestvlm{0.081}} & \ourscell{\bestvlm{0.043}} \\
\cmidrule{2-11}

& \multirow{2}{*}{4B}
& Qwen3-VL-4B
& 0.699 & 0.375 & 0.078 & 0.180
& 0.272 & 0.118 & 0.056 & 0.029 \\
& & \ourscell{\textbf{OpenVAM-4B}}
& \ourscell{\bestvlm{0.703}} & \ourscell{\bestvlm{0.438}} & \ourscell{\bestvlm{0.127}} & \ourscell{\bestvlm{0.237}}
& \ourscell{\bestvlm{0.338}} & \ourscell{\bestvlm{0.177}} & \ourscell{\bestvlm{0.096}} & \ourscell{\bestvlm{0.053}} \\
\cmidrule{2-11}

& \multirow{2}{*}{7B}
& Qwen2.5-VL-7B
& 0.632 & 0.421 & 0.119 & 0.218
& 0.322 & 0.167 & 0.085 & 0.045 \\
& & \ourscell{\textbf{OpenVAM-7B}}
& \ourscell{\bestvlm{0.678}} & \ourscell{\bestvlm{0.456}} & \ourscell{\bestvlm{0.120}} & \ourscell{\bestvlm{0.220}}
& \ourscell{\bestvlm{0.341}} & \ourscell{\bestvlm{0.169}} & \ourscell{\bestvlm{0.089}} & \ourscell{\bestvlm{0.051}} \\
\cmidrule{2-11}

& \multirow{2}{*}{8B}
& Qwen3-VL-8B
& 0.654 & 0.390 & 0.097 & 0.192
& 0.282 & 0.135 & 0.067 & 0.035 \\
& & \ourscell{\textbf{OpenVAM-8B}}
& \ourscell{\bestvlm{0.712}} & \ourscell{\bestvlm{0.447}} & \ourscell{\bestvlm{0.136}} & \ourscell{\bestvlm{0.250}}
& \ourscell{\bestvlm{0.346}} & \ourscell{\bestvlm{0.185}} & \ourscell{\bestvlm{0.103}} & \ourscell{\bestvlm{0.057}} \\
\midrule

\multirow{12}{*}{\textit{MIT1003}}
& PV
& \cellcolor{gray!15} Gemini-2.5-Pro
& \cellcolor{gray!15}0.736 & \cellcolor{gray!15}0.499 & \cellcolor{gray!15}0.156 & \cellcolor{gray!15}0.257
& \cellcolor{gray!15}0.399 & \cellcolor{gray!15}0.218 & \cellcolor{gray!15}0.119 & \cellcolor{gray!15}0.064 \\
\cmidrule{2-11}


& \multirow{2}{*}{3B}
& Qwen2.5-VL-3B
& 0.557 & 0.383 & 0.097 & 0.206
& 0.310 & 0.152 & 0.074 & 0.038 \\
& & \ourscell{\textbf{OpenVAM-3B}}
& \ourscell{\bestvlm{0.659}} & \ourscell{\bestvlm{0.447}} & \ourscell{\bestvlm{0.111}} & \ourscell{\bestvlm{0.217}}
& \ourscell{\bestvlm{0.335}} & \ourscell{\bestvlm{0.161}} & \ourscell{\bestvlm{0.080}} & \ourscell{\bestvlm{0.043}} \\
\cmidrule{2-11}

& \multirow{2}{*}{4B}
& Qwen3-VL-4B
& \bestvlm{0.717} & 0.398 & 0.084 & 0.191
& 0.296 & 0.129 & 0.060 & 0.030 \\
& & \ourscell{\textbf{OpenVAM-4B}}
& \ourscell{0.705} & \ourscell{\bestvlm{0.435}} & \ourscell{\bestvlm{0.128}} & \ourscell{\bestvlm{0.234}}
& \ourscell{\bestvlm{0.343}} & \ourscell{\bestvlm{0.181}} & \ourscell{\bestvlm{0.098}} & \ourscell{\bestvlm{0.054}} \\
\cmidrule{2-11}

& \multirow{2}{*}{7B}
& Qwen2.5-VL-7B
& 0.643 & 0.443 & \bestvlm{0.128} & \bestvlm{0.226}
& 0.345 & \bestvlm{0.182} & \bestvlm{0.094} & \bestvlm{0.049} \\
& & \ourscell{\textbf{OpenVAM-7B}}
& \ourscell{\bestvlm{0.676}} & \ourscell{\bestvlm{0.459}} & \ourscell{0.126} & \ourscell{0.225}
& \ourscell{\bestvlm{0.351}} & \ourscell{0.173} & \ourscell{0.092} & \ourscell{0.046} \\
\cmidrule{2-11}

& \multirow{2}{*}{8B}
& Qwen3-VL-8B
& 0.715 & 0.439 & 0.116 & 0.216
& 0.332 & 0.165 & 0.084 & 0.044 \\
& & \ourscell{\textbf{OpenVAM-8B}}
& \ourscell{\bestvlm{0.725}} & \ourscell{\bestvlm{0.449}} & \ourscell{\bestvlm{0.136}} & \ourscell{\bestvlm{0.244}}
& \ourscell{\bestvlm{0.358}} & \ourscell{\bestvlm{0.191}} & \ourscell{\bestvlm{0.104}} & \ourscell{\bestvlm{0.058}} \\

\bottomrule
\end{tabular}%
}
\vspace{-1em}
\end{table*}

\begin{table*}[t]
\centering
\caption{\textbf{Stage-wise ablation of OpenVAM-3B.}
We progressively enable Stage~1 (S1), Stage~2 (S2), and Stage~3 (S3), and also compare against full joint training of S1--S3 and joint training of S2--S3 after S1.}
\vspace{-0.5em}
\label{tab:openvam_stage_ablation_main}
\resizebox{\textwidth}{!}{%
\setlength{\tabcolsep}{6pt}%
\renewcommand{\arraystretch}{1.12}%
\footnotesize
\begin{tabular}{
>{\centering\arraybackslash}p{1.35cm}
>{\centering\arraybackslash}p{0.7cm}
>{\centering\arraybackslash}p{0.7cm}
>{\centering\arraybackslash}p{0.7cm}
|ccccc|c|ccc|cccc
}
\toprule

\multicolumn{4}{c|}{\textbf{Training Stages}}
& \multicolumn{5}{c|}{\textbf{Saliency}}
& \multicolumn{1}{c|}{\textbf{VLM Judge} $\uparrow$}
& \multicolumn{3}{c|}{\textbf{ROUGE} $\uparrow$}
& \multicolumn{4}{c}{\textbf{BLEU} $\uparrow$} \\

\midrule

Dataset & S1 & S2 & S3
& CC $\uparrow$ & KLD $\downarrow$ & AUC $\uparrow$ & SIM $\uparrow$ & NSS $\uparrow$
& JScore
& R-1 & R-2 & R-L
& B-1 & B-2 & B-3 & B-4 \\

\midrule

\multirow{5}{*}{{U-EYE}~\cite{jiang2023ueyes}}
& \multicolumn{3}{c|}{joint}
& 0.705 & 0.567 & 0.845 & 0.621 & 1.708
& 0.451 & 0.409 & 0.097 & 0.176
& 0.204 & 0.137 & 0.043 & 0.030 \\

& \cmark & \xmark & \xmark
& 0.728 & 0.550 & 0.846 & 0.622 & 1.704
& -- & -- & -- & -- & -- & -- & -- & -- \\

& \cmark & \multicolumn{2}{c|}{joint}
& 0.730 & 0.559 & 0.845 & 0.620 & 1.714
& 0.441 & 0.388 & 0.083 & 0.181
& 0.271 & 0.113 & 0.050 & 0.026 \\

& \cmark & \cmark & \xmark
& \textbf{0.735} & 0.543 & \textbf{0.847} & 0.630 & \textbf{1.721}
& 0.434 & 0.252 & 0.034 & 0.133
& 0.148 & 0.050 & 0.019 & 0.009 \\

& \cmark & \cmark & \cmark
& 0.734 & \textbf{0.542} & \textbf{0.847} & \textbf{0.639} & 1.720
& \textbf{0.603} & \textbf{0.429} & \textbf{0.108} & \textbf{0.196}
& \textbf{0.330} & \textbf{0.156} & \textbf{0.077} & \textbf{0.040} \\

\midrule

\multirow{5}{*}{SalECI~\cite{jiang2022does}}
& \multicolumn{3}{c|}{joint}
& 0.762 & 0.497 & 0.877 & 0.653 & 1.993
& 0.381 & 0.434 & 0.112 & 0.181
& 0.348 & 0.158 & 0.074 & 0.030 \\

& \cmark & \xmark & \xmark
& 0.788 & 0.466 & 0.898 & 0.675 & 2.031
& -- & -- & -- & -- & -- & -- & -- & -- \\

& \cmark & \multicolumn{2}{c|}{joint}
& 0.790 & 0.465 & 0.889 & 0.678 & 2.032
& 0.382 & 0.395 & 0.090 & 0.186
& 0.244 & 0.109 & 0.052 & 0.027 \\

& \cmark & \cmark & \xmark
& 0.792 & 0.464 & \textbf{0.899} & \textbf{0.680} & \textbf{2.035}
& 0.376 & 0.244 & 0.023 & 0.136
& 0.159 & 0.046 & 0.020 & 0.010 \\

& \cmark & \cmark & \cmark
& \textbf{0.795} & \textbf{0.462} & \textbf{0.899} & \textbf{0.680} & 2.026
& \textbf{0.692} & \textbf{0.478} & \textbf{0.130} & \textbf{0.220}
& \textbf{0.369} & \textbf{0.184} & \textbf{0.096} & \textbf{0.050} \\

\midrule

\multirow{5}{*}{OSIE~\cite{xu2014predicting}}
& \multicolumn{3}{c|}{joint}
& 0.898 & 0.299 & 0.899 & 0.754 & 3.478
& 0.506 & 0.375 & 0.109 & 0.180
& 0.297 & 0.156 & 0.075 & 0.039 \\

& \cmark & \xmark & \xmark
& 0.912 & 0.229 & \textbf{0.935} & 0.777 & \textbf{3.903}
& -- & -- & -- & -- & -- & -- & -- & -- \\

& \cmark & \multicolumn{2}{c|}{joint}
& 0.917 & 0.224 & 0.887 & 0.773 & 3.727
& 0.502 & 0.354 & 0.066 & 0.176
& 0.231 & 0.092 & 0.042 & 0.022 \\

& \cmark & \cmark & \xmark
& 0.922 & 0.232 & 0.934 & 0.783 & 3.743
& 0.497 & 0.270 & 0.027 & 0.146
& 0.200 & 0.060 & 0.025 & 0.012 \\

& \cmark & \cmark & \cmark
& \textbf{0.928} & \textbf{0.214} & \textbf{0.935} & \textbf{0.793} & 3.841
& \textbf{0.661} & \textbf{0.472} & \textbf{0.123} & \textbf{0.220}
& \textbf{0.364} & \textbf{0.178} & \textbf{0.089} & \textbf{0.047} \\

\midrule

\multirow{5}{*}{Salicon~\cite{jiang2015salicon}}
& \multicolumn{3}{c|}{joint}
& 0.903 & 0.202 & 0.866 & 0.794 & 1.965
& 0.521 & 0.398 & 0.137 & 0.197
& 0.321 & 0.159 & 0.085 & 0.043 \\

& \cmark & \xmark & \xmark
& 0.902 & 0.186 & 0.875 & 0.799 & 1.985
& -- & -- & -- & -- & -- & -- & -- & -- \\

& \cmark & \multicolumn{2}{c|}{joint}
& 0.909 & 0.221 & 0.873 & 0.798 & 1.983
& 0.526 & 0.340 & 0.074 & 0.175
& 0.197 & 0.084 & 0.040 & 0.021 \\

& \cmark & \cmark & \xmark
& 0.908 & 0.232 & 0.875 & 0.783 & 1.983
& 0.517 & 0.335 & 0.057 & 0.158
& 0.186 & 0.073 & 0.038 & 0.019 \\

& \cmark & \cmark & \cmark
& \textbf{0.911} & \textbf{0.184} & \textbf{0.876} & \textbf{0.805} & \textbf{1.989}
& \textbf{0.677} & \textbf{0.424} & \textbf{0.157} & \textbf{0.211}
& \textbf{0.342} & \textbf{0.163} & \textbf{0.097} & \textbf{0.055} \\

\midrule

\multirow{5}{*}{CAT2000~\cite{borji2015cat2000}}
& \multicolumn{3}{c|}{joint}
& 0.887 & 0.288 & 0.786 & 0.756 & 2.428
& 0.580 & 0.384 & 0.099 & 0.209
& 0.318 & 0.159 & 0.0760 & 0.039 \\

& \cmark & \xmark & \xmark
& 0.884 & 0.262 & 0.888 & 0.754 & 2.438
& -- & -- & -- & -- & -- & -- & -- & -- \\

& \cmark & \multicolumn{2}{c|}{joint}
& 0.888 & 0.260 & 0.888 & 0.753 & 2.441
& 0.576 & 0.372 & 0.077 & 0.187
& 0.250 & 0.108 & 0.052 & 0.028 \\

& \cmark & \cmark & \xmark
& \textbf{0.892} & \textbf{0.256} & 0.888 & \textbf{0.760} & 2.450
& 0.552 & 0.230 & 0.017 & 0.136
& 0.169 & 0.045 & 0.019 & 0.010 \\

& \cmark & \cmark & \cmark
& 0.891 & 0.258 & \textbf{0.889} & 0.759 & \textbf{2.452}
& \textbf{0.663} & \textbf{0.442} & \textbf{0.111} & \textbf{0.217}
& \textbf{0.333} & \textbf{0.162} & \textbf{0.081} & \textbf{0.043} \\

\midrule

\multirow{5}{*}{MIT1003~\cite{judd2009learning}}
& \multicolumn{3}{c|}{joint}
& 0.797 & 0.536 & 0.899 & 0.640 & 2.862
& 0.555 & 0.387 & 0.109 & 0.199
& 0.317 & 0.157 & 0.0781 & 0.038 \\

& \cmark & \xmark & \xmark
& 0.817 & 0.483 & 0.922 & 0.656 & 3.050
& -- & -- & -- & -- & -- & -- & -- & -- \\

& \cmark & \multicolumn{2}{c|}{joint}
& 0.820 & 0.477 & 0.920 & 0.666 & 3.062
& 0.554 & 0.348 & 0.065 & 0.175
& 0.221 & 0.090 & 0.042 & 0.022 \\

& \cmark & \cmark & \xmark
& 0.820 & 0.483 & 0.921 & 0.661 & 3.020
& 0.453 & 0.249 & 0.022 & 0.142
& 0.184 & 0.053 & 0.023 & 0.012 \\

& \cmark & \cmark & \cmark
& \textbf{0.829} & \textbf{0.463} & \textbf{0.923} & \textbf{0.671} & \textbf{3.081}
& \textbf{0.659} & \textbf{0.447} & \textbf{0.111} & \textbf{0.217}
& \textbf{0.335} & \textbf{0.161} & \textbf{0.080} & \textbf{0.043} \\

\bottomrule
\end{tabular}%
}
\end{table*}


\noindent\textbf{Experimental Settings.} OpenVAM is implemented in PyTorch and train on a single NVIDIA L40 GPU. Following~\cite{hosseini2025sum}, we resize input images and the corresponding saliency and fixation maps to $256\times256$. We use DINOv3 ViT-B/16~\cite{simeoni2025dinov3} as the visual encoder, and adopt the Qwen PatchMerger as the visual adapter. For Stage~III, we apply LoRA to the language transformer with rank $16$, scaling $\alpha=32$, and dropout $0.05$. We also apply LoRA to the LM head ($r=16$, $\alpha=16$). OpenVAM-3B and 7B are based on Qwen2.5-VL-3B~\cite{wu2025qwen} and 7B, respectively. OpenVAM-4B and 8B are based on Qwen3-VL-4B~\cite{bai2023qwen} and 8B, respectively. Moreover, \textit{JScore} is a GPT-4.1-based semantic score that compares the generated rationale with the reference rationale (see Supp.~\suppref{sec:supp_jscore}). Unlike BLEU/ROUGE, it evaluates whether the model identifies the same salient regions and gives consistent reasons for their saliency, while penalizing hallucinated or unsupported content. We further validate JScore in the supplement: it correlates well with human ratings and produces stable rankings across prompt variations and repeated runs. Please see Supp.~\suppref{sec:supp_exp_settings},~\suppref{sec:supp_jscore_stability}, and~\suppref{sec:human_validation_jscore} for additional details.

\subsection{Experimental Results}
\noindent\textbf{Saliency Prediction.}
~\autoref{tab:saliency} compares OpenVAM to strong domain-specific and unified saliency baselines across natural scenes, e-commerce, and UI/web layouts.
Overall, OpenVAM ranks first across most datasets and metrics, indicating closer agreement with human attention in both distribution and structure.
Concretely, improved KLD suggests better calibration of fixation density (probability mass is placed in the right regions), while higher CC and SIM indicate better spatial structure and overlap with ground-truth saliency (cleaner, more precise maps). Higher NSS further implies that predicted peaks align more strongly with true fixation locations (sharper, more discriminative maxima), and AUC confirms reliable separation of fixated versus non-fixated pixels.
OpenVAM is best (or tied-best) on the majority of metrics on natural datasets, showing sharper localization and fewer spurious activations, and remains consistently among the top methods on U-EYE and SalECI, suggesting robustness to domain-specific biases such as UI layout conventions and product-centric cues. Interestingly, we observe noticeably larger gains on OSIE and MIT1003 than on SALICON. One factor is that existing baselines already perform strongly on SALICON, leaving less room for improvement on several metrics, although OpenVAM still improves KLD by up to 6.77\%. In addition, SALICON uses mouse-based proxy annotations, whereas OSIE, MIT1003, and CAT2000 are collected with eye tracking. Our analysis in Supp.~\suppref{sec:concept_similarity_natural} further shows that SALICON differs substantially from the eye-tracking datasets at the fine-grained concept level. These results suggest that OpenVAM generalizes robustly across acquisition modalities, maintaining strong performance on mouse-based SALICON while achieving larger gains on eye-tracking benchmarks. These trends are reflected in~\autoref{fig:visual_sal}: OpenVAM produces more concentrated peaks on truly attended objects (e.g., products, key UI elements), reduces background spread, and better captures multiple competing salient regions in cluttered layouts, which is particularly important under domain shift. 
\begin{table*}[t]
    \centering
    \vspace{-1em}
    \caption{\textbf{Saliency prediction performance across various datasets.} 
    Baseline results are taken from their respective papers.}
    \vspace{-0.75em}
    \label{tab:saliency}
    \resizebox{\textwidth}{!}{
    \setlength{\tabcolsep}{10pt}
    \renewcommand{\arraystretch}{1.1}
    \footnotesize
    \begin{tabular}{c|lcccccccccc}
    \toprule
    Dataset & Method 
    & CC $\uparrow$ & & KLD $\downarrow$ & & AUC $\uparrow$ & & SIM $\uparrow$ & & NSS $\uparrow$ &  \\
    \midrule
    
    \textit{U-EYE}~\cite{jiang2023ueyes} 
    & SAM~\cite{cornia2018predicting} & 0.580 && 1.490 && 0.811 && 0.520 && 1.640 & \\
    (Web page)
    & UMSI~\cite{fosco2020predicting} & 0.562 && 1.580 && 0.805 && 0.510 && 1.690 & \\
    & SAM++~\cite{jiang2023ueyes} & 0.580 && 1.190 && 0.800 && 0.530 && 1.660 & \\
    & Transalnet~\cite{lou2022transalnet} & 0.696 && 0.616 && 0.839 && 0.598 && 1.601 & \\
    & UMSI++~\cite{jiang2023ueyes} & 0.670 && 0.860 && 0.830 && 0.580 && 1.610 & \\
    & SUM~\cite{hosseini2025sum} & 0.731 && 0.544 && 0.846 && 0.630 && 1.704 & \\
    \cmidrule{2-12}
    & \cellcolor[HTML]{DADCFF}OpenVAM-3B
    & \cellcolor[HTML]{DADCFF}0.734\scriptsize{(+0.41\%)} && 
      \cellcolor[HTML]{DADCFF}0.542\scriptsize{(+0.37\%)} && 
      \cellcolor[HTML]{DADCFF}0.847\scriptsize{(+0.12\%)} && 
      \cellcolor[HTML]{DADCFF}0.639\scriptsize{(+1.43\%)} && 
      \cellcolor[HTML]{DADCFF}1.720\scriptsize{(+0.94\%)} & \\
    & \cellcolor[HTML]{DADCFF}OpenVAM-4B
    & \cellcolor[HTML]{DADCFF}0.737\scriptsize{(+0.82\%)} && 
      \cellcolor[HTML]{DADCFF}0.538\scriptsize{(+1.10\%)} && 
      \cellcolor[HTML]{DADCFF}0.847\scriptsize{(+0.12\%)} && 
      \cellcolor[HTML]{DADCFF}0.627\scriptsize{(-0.48\%)} && 
      \cellcolor[HTML]{DADCFF}1.718\scriptsize{(+0.82\%)} & \\
    & \cellcolor[HTML]{DADCFF}OpenVAM-7B
    & \cellcolor[HTML]{DADCFF}0.736\scriptsize{(+0.68\%)} && 
      \cellcolor[HTML]{DADCFF}0.539\scriptsize{(+0.92\%)} && 
      \cellcolor[HTML]{DADCFF}0.847\scriptsize{(+0.12\%)} && 
      \cellcolor[HTML]{DADCFF}\textbf{0.642\scriptsize{(+1.90\%)}} && 
      \cellcolor[HTML]{DADCFF}\textbf{1.732\scriptsize{(+1.64\%)}} & \\ 
    & \cellcolor[HTML]{DADCFF}OpenVAM-8B
    & \cellcolor[HTML]{DADCFF}\textbf{0.739\scriptsize{(+1.09\%)}} && 
      \cellcolor[HTML]{DADCFF}\textbf{0.537\scriptsize{(+1.29\%)}} && 
      \cellcolor[HTML]{DADCFF}\textbf{0.848\scriptsize{(+0.24\%)}} && 
      \cellcolor[HTML]{DADCFF}0.630\scriptsize{(+0.00\%)} && 
      \cellcolor[HTML]{DADCFF}1.727\scriptsize{(+1.35\%)} & \\
    \midrule
    
    \textit{SalECI}~\cite{jiang2022does}
    & SSM~\cite{cornia2018predicting} & 0.720 && 0.599 && 0.830 && 0.611 && 1.396 & \\
    (E-Commercial)
    & DeepGaze IIE~\cite{linardos2021deepgaze} & 0.560 && 0.995 && 0.842 && 0.399 && 1.327 & \\
    & EML-NET~\cite{jiang2023ueyes} & 0.510 && 1.220 && 0.807 && 0.536 && 1.232 & \\
    & Transalnet~\cite{jiang2023ueyes} & 0.717 && 0.873 && 0.824 && 0.534 && 1.723 & \\
    & Temp-Sal~\cite{aydemir2023tempsal} & 0.719 && 0.712 && 0.813 && 0.629 && 1.768 & \\
    & SSwin Transformer~\cite{jiang2022does} & 0.687 && 0.652 && 0.868 && 0.606 && 1.701 & \\
    & BrandAttn~\cite{hosseini2024brand} & 0.750 && 0.578 && 0.892 && 0.645 && 1.890 & \\
    & SUM~\cite{hosseini2025sum} & 0.789 && 0.473 && 0.899 && 0.680 && 2.012 & \\
    \cmidrule{2-12}
    & \cellcolor[HTML]{DADCFF}OpenVAM-3B
    & \cellcolor[HTML]{DADCFF}0.795\scriptsize{(+0.76\%)} && 
      \cellcolor[HTML]{DADCFF}0.462\scriptsize{(+2.33\%)} && 
      \cellcolor[HTML]{DADCFF}0.899\scriptsize{(+0.00\%)} && 
      \cellcolor[HTML]{DADCFF}0.680\scriptsize{(+0.00\%)} && 
      \cellcolor[HTML]{DADCFF}2.026\scriptsize{(+0.70\%)} & \\
    & \cellcolor[HTML]{DADCFF}OpenVAM-4B
    & \cellcolor[HTML]{DADCFF}\textbf{0.797\scriptsize{(+1.01\%)}} && 
      \cellcolor[HTML]{DADCFF}\textbf{0.450\scriptsize{(+4.86\%)}} && 
      \cellcolor[HTML]{DADCFF}\textbf{0.901\scriptsize{(+0.22\%)}} && 
      \cellcolor[HTML]{DADCFF}0.679\scriptsize{(-0.15\%)} && 
      \cellcolor[HTML]{DADCFF}\textbf{2.033\scriptsize{(+1.04\%)}} & \\
    & \cellcolor[HTML]{DADCFF}OpenVAM-7B
    & \cellcolor[HTML]{DADCFF}\textbf{0.797\scriptsize{(+1.01\%)}} && 
      \cellcolor[HTML]{DADCFF}0.452\scriptsize{(+4.44\%)} && 
      \cellcolor[HTML]{DADCFF}0.899\scriptsize{(+0.00\%)} && 
      \cellcolor[HTML]{DADCFF}\textbf{0.692\scriptsize{(+1.76\%)}} && 
      \cellcolor[HTML]{DADCFF}2.032\scriptsize{(+0.99\%)} & \\
    & \cellcolor[HTML]{DADCFF}OpenVAM-8B
    & \cellcolor[HTML]{DADCFF}0.792\scriptsize{(+0.38\%)} && 
      \cellcolor[HTML]{DADCFF}0.460\scriptsize{(+2.75\%)} && 
      \cellcolor[HTML]{DADCFF}0.900\scriptsize{(+0.11\%)} && 
      \cellcolor[HTML]{DADCFF}0.678\scriptsize{(-0.29\%)} && 
      \cellcolor[HTML]{DADCFF}2.022\scriptsize{(+0.50\%)} & \\
    \midrule
    
    \textit{OSIE}~\cite{xu2014predicting}
    & UMSI~\cite{fosco2020predicting} & 0.746 && 0.513 && 0.856 && 0.631 && 1.788 & \\
    (Natural scene)
    & EML-NET~\cite{jia2020eml} & 0.717 && 0.537 && 0.854 && 0.619 && 1.737 & \\
    & SAM-ResNet~\cite{cornia2018predicting} & 0.758 && 0.480 && 0.860 && 0.648 && 1.811 & \\
    & BrandAttn~\cite{chen2023learning} & 0.761 && 0.506 && 0.860 && 0.652 && 1.840 & \\
    & Transalnet~\cite{jiang2023ueyes} & 0.791 && 0.667 && 0.923 && 0.651 && 2.448 & \\
    & UniAR~\cite{li2024uniar} & 0.754 && 0.547 && 0.867 && 0.647 && 1.842 & \\
    & SUM~\cite{hosseini2025sum} & 0.861 && 0.340 && 0.924 && 0.727 && 3.416 & \\
    \cmidrule{2-12}
    & \cellcolor[HTML]{DADCFF}OpenVAM-3B
    & \cellcolor[HTML]{DADCFF}0.928\scriptsize{(+7.78\%)} && 
      \cellcolor[HTML]{DADCFF}0.214\scriptsize{(+37.06\%)} && 
      \cellcolor[HTML]{DADCFF}0.935\scriptsize{(+1.19\%)} && 
      \cellcolor[HTML]{DADCFF}0.793\scriptsize{(+9.08\%)} && 
      \cellcolor[HTML]{DADCFF}3.841\scriptsize{(+12.44\%)} & \\
    & \cellcolor[HTML]{DADCFF}OpenVAM-4B
    & \cellcolor[HTML]{DADCFF}\textbf{0.933\scriptsize{(+8.36\%)}} && 
      \cellcolor[HTML]{DADCFF}0.211\scriptsize{(+37.94\%)} && 
      \cellcolor[HTML]{DADCFF}\textbf{0.936\scriptsize{(+1.30\%)}} && 
      \cellcolor[HTML]{DADCFF}0.786\scriptsize{(+8.12\%)} && 
      \cellcolor[HTML]{DADCFF}3.713\scriptsize{(+8.69\%)} & \\
    & \cellcolor[HTML]{DADCFF}OpenVAM-7B
    & \cellcolor[HTML]{DADCFF}0.931\scriptsize{(+8.13\%)} && 
      \cellcolor[HTML]{DADCFF}\textbf{0.209\scriptsize{(+38.53\%)}} && 
      \cellcolor[HTML]{DADCFF}\textbf{0.936\scriptsize{(+1.30\%)}} && 
      \cellcolor[HTML]{DADCFF}\textbf{0.799\scriptsize{(+9.90\%)}} && 
      \cellcolor[HTML]{DADCFF}\textbf{3.876\scriptsize{(+13.47\%)}} & \\
    & \cellcolor[HTML]{DADCFF}OpenVAM-8B
    & \cellcolor[HTML]{DADCFF}0.926\scriptsize{(+7.55\%)} && 
      \cellcolor[HTML]{DADCFF}0.226\scriptsize{(+33.53\%)} && 
      \cellcolor[HTML]{DADCFF}0.934\scriptsize{(+1.08\%)} && 
      \cellcolor[HTML]{DADCFF}0.787\scriptsize{(+8.25\%)} && 
      \cellcolor[HTML]{DADCFF}3.715\scriptsize{(+8.75\%)} & \\
    \midrule
    
    \textit{Salicon}~\cite{jiang2015salicon}
    & UniAR~\cite{li2024uniar} & 0.901 && 0.215 && 0.870 && 0.792 && 1.947 & \\
    (Natural scene)
    & SimpleNet~\cite{reddy2020tidying} & 0.907 && 0.193 && 0.871 && 0.797 && 1.926 & \\
    & MDNSal~\cite{reddy2020tidying} & 0.899 && 0.217 && 0.868 && 0.797 && 1.893 & \\
    & MSI-Net~\cite{kroner2020contextual} & 0.899 && 0.307 && 0.865 && 0.784 && 1.931 & \\
    & GazeGAN~\cite{che2019gazegan} & 0.879 && 0.376 && 0.864 && 0.773 && 1.899 & \\
    & UNISAL~\cite{droste2020unified} & 0.879 && 0.354 && 0.864 && 0.775 && 1.952 & \\
    & Transalnet~\cite{lou2022transalnet} & 0.890 && 0.220 && 0.867 && 0.783 && 1.924 & \\
    & DeepGaze IIE~\cite{linardos2021deepgaze} & 0.872 && 0.285 && 0.869 && 0.733 && \textbf{1.996} & \\
    & Temp-Sal~\cite{aydemir2023tempsal} & 0.911 && 0.195 && 0.869 && 0.800 && 1.967 & \\
    & SUM~\cite{hosseini2025sum} & 0.909 && 0.192 && \textbf{0.876} && 0.804 && 1.981 & \\
    \cmidrule{2-12}
    & \cellcolor[HTML]{DADCFF}OpenVAM-3B
    & \cellcolor[HTML]{DADCFF}0.911\scriptsize{(+0.00\%)} && 
      \cellcolor[HTML]{DADCFF}0.184\scriptsize{(+4.17\%)} && 
      \cellcolor[HTML]{DADCFF}\textbf{0.876\scriptsize{(+0.00\%)}} && 
      \cellcolor[HTML]{DADCFF}0.805\scriptsize{(+0.12\%)} && 
      \cellcolor[HTML]{DADCFF}1.989\scriptsize{(-0.35\%)} & \\
    & \cellcolor[HTML]{DADCFF}OpenVAM-4B
    & \cellcolor[HTML]{DADCFF}0.913\scriptsize{(+0.22\%)} && 
      \cellcolor[HTML]{DADCFF}\textbf{0.179\scriptsize{(+6.77\%)}} && 
      \cellcolor[HTML]{DADCFF}\textbf{0.876\scriptsize{(+0.00\%)}} && 
      \cellcolor[HTML]{DADCFF}0.805\scriptsize{(+0.12\%)} && 
      \cellcolor[HTML]{DADCFF}1.970\scriptsize{(-1.30\%)} & \\
    & \cellcolor[HTML]{DADCFF}OpenVAM-7B
    & \cellcolor[HTML]{DADCFF}0.912\scriptsize{(+0.11\%)} && 
      \cellcolor[HTML]{DADCFF}\textbf{0.179\scriptsize{(+6.77\%)}} && 
      \cellcolor[HTML]{DADCFF}\textbf{0.876\scriptsize{(+0.00\%)}} && 
      \cellcolor[HTML]{DADCFF}0.806\scriptsize{(+0.25\%)} && 
      \cellcolor[HTML]{DADCFF}1.993\scriptsize{(-0.15\%)} & \\
    & \cellcolor[HTML]{DADCFF}OpenVAM-8B
    & \cellcolor[HTML]{DADCFF}\textbf{0.914\scriptsize{(+0.33\%)}} && 
      \cellcolor[HTML]{DADCFF}\textbf{0.179\scriptsize{(+6.77\%)}} && 
      \cellcolor[HTML]{DADCFF}\textbf{0.876\scriptsize{(+0.00\%)}} && 
      \cellcolor[HTML]{DADCFF}\textbf{0.808\scriptsize{(+0.50\%)}} && 
      \cellcolor[HTML]{DADCFF}1.972\scriptsize{(-1.20\%)} & \\
    \midrule
    
    \textit{CAT2000}~\cite{borji2015cat2000}
    & FastSal~\cite{hu2021fastsal} & 0.721 && 0.552 && 0.860 && 0.603 && 1.859 & \\
    (Natural scene)
    & SAM-Resnet~\cite{cornia2018predicting} & 0.870 && 0.670 && 0.878 && 0.739 && 2.411 & \\
    & MSI-Net~\cite{kroner2020contextual} & 0.866 && 0.428 && 0.881 && 0.730 && 2.355 & \\
    & DVA~\cite{wang2017deep} & 0.861 && 0.449 && 0.878 && 0.734 && 2.345 & \\
    & UNISAL~\cite{droste2020unified} & 0.842 && 0.530 && 0.876 && 0.721 && 2.257 & \\
    & MDNSal~\cite{reddy2020tidying} & 0.889 && 0.293 && 0.878 && 0.751 && 2.329 & \\
    & Transalnet~\cite{lou2022transalnet} & 0.877 && 0.287 && 0.882 && 0.744 && 2.373 & \\
    & SUM~\cite{hosseini2025sum} & 0.882 && 0.270 && 0.888 && 0.754 && 2.424 & \\
    \cmidrule{2-12}
    & \cellcolor[HTML]{DADCFF}OpenVAM-3B
    & \cellcolor[HTML]{DADCFF}0.891\scriptsize{(+0.22\%)} && 
      \cellcolor[HTML]{DADCFF}0.258\scriptsize{(+4.44\%)} && 
      \cellcolor[HTML]{DADCFF}0.889\scriptsize{(+0.11\%)} && 
      \cellcolor[HTML]{DADCFF}0.759\scriptsize{(+0.66\%)} && 
      \cellcolor[HTML]{DADCFF}2.452\scriptsize{(+1.16\%)} & \\
    & \cellcolor[HTML]{DADCFF}OpenVAM-4B
    & \cellcolor[HTML]{DADCFF}0.892\scriptsize{(+0.34\%)} && 
      \cellcolor[HTML]{DADCFF}0.256\scriptsize{(+5.19\%)} && 
      \cellcolor[HTML]{DADCFF}0.889\scriptsize{(+0.11\%)} && 
      \cellcolor[HTML]{DADCFF}0.759\scriptsize{(+0.66\%)} && 
      \cellcolor[HTML]{DADCFF}2.442\scriptsize{(+0.74\%)} & \\
    & \cellcolor[HTML]{DADCFF}OpenVAM-7B
    & \cellcolor[HTML]{DADCFF}\textbf{0.899\scriptsize{(+1.12\%)}} && 
      \cellcolor[HTML]{DADCFF}\textbf{0.246\scriptsize{(+8.89\%)}} && 
      \cellcolor[HTML]{DADCFF}\textbf{0.899\scriptsize{(+1.24\%)}} && 
      \cellcolor[HTML]{DADCFF}\textbf{0.761\scriptsize{(+0.93\%)}} && 
      \cellcolor[HTML]{DADCFF}\textbf{2.459\scriptsize{(+1.44\%)}} & \\
    & \cellcolor[HTML]{DADCFF}OpenVAM-8B
    & \cellcolor[HTML]{DADCFF}0.895\scriptsize{(+0.67\%)} && 
      \cellcolor[HTML]{DADCFF}0.253\scriptsize{(+6.30\%)} && 
      \cellcolor[HTML]{DADCFF}0.889\scriptsize{(+0.11\%)} && 
      \cellcolor[HTML]{DADCFF}\textbf{0.761\scriptsize{(+0.93\%)}} && 
      \cellcolor[HTML]{DADCFF}2.443\scriptsize{(+0.78\%)} & \\
    \midrule
    
    \textit{MIT1003}~\cite{judd2009learning}
    & FastSal~\cite{hu2021fastsal} & 0.590 && 1.036 && 0.875 && 0.478 && 2.008 & \\
    (Natural scene)
    & SAM-Resnet~\cite{cornia2018predicting} & 0.746 && 1.247 && 0.902 && 0.597 && 2.752 & \\
    & DVA~\cite{wang2017deep} & 0.699 && 0.753 && 0.897 && 0.566 && 2.574 & \\
    & UNISAL~\cite{droste2020unified} & 0.734 && 1.014 && 0.902 && 0.597 && 2.759 & \\
    & Transalnet~\cite{lou2022transalnet} & 0.722 && 0.660 && 0.903 && 0.592 && 2.631 & \\
    & SUM~\cite{hosseini2025sum} & 0.768 && 0.563 && 0.913 && 0.630 && 2.839 & \\
    \cmidrule{2-12}
    & \cellcolor[HTML]{DADCFF}OpenVAM-3B
    & \cellcolor[HTML]{DADCFF}0.829\scriptsize{(+7.94\%)} && 
      \cellcolor[HTML]{DADCFF}0.463\scriptsize{(+17.76\%)} && 
      \cellcolor[HTML]{DADCFF}0.923\scriptsize{(+1.10\%)} && 
      \cellcolor[HTML]{DADCFF}0.671\scriptsize{(+6.51\%)} && 
      \cellcolor[HTML]{DADCFF}3.081\scriptsize{(+8.52\%)} & \\
    & \cellcolor[HTML]{DADCFF}OpenVAM-4B
    & \cellcolor[HTML]{DADCFF}0.829\scriptsize{(+7.94\%)} && 
      \cellcolor[HTML]{DADCFF}0.473\scriptsize{(+15.99\%)} && 
      \cellcolor[HTML]{DADCFF}0.923\scriptsize{(+1.10\%)} && 
      \cellcolor[HTML]{DADCFF}0.655\scriptsize{(+3.97\%)} && 
      \cellcolor[HTML]{DADCFF}3.044\scriptsize{(+7.22\%)} & \\
    & \cellcolor[HTML]{DADCFF}OpenVAM-7B
    & \cellcolor[HTML]{DADCFF}\textbf{0.842\scriptsize{(+9.64\%)}} && 
      \cellcolor[HTML]{DADCFF}\textbf{0.458\scriptsize{(+18.65\%)}} && 
      \cellcolor[HTML]{DADCFF}\textbf{0.924\scriptsize{(+1.20\%)}} && 
      \cellcolor[HTML]{DADCFF}\textbf{0.679\scriptsize{(+7.78\%)}} && 
      \cellcolor[HTML]{DADCFF}\textbf{3.090\scriptsize{(+8.84\%)}} & \\
    & \cellcolor[HTML]{DADCFF}OpenVAM-8B
    & \cellcolor[HTML]{DADCFF}0.825\scriptsize{(+7.42\%)} && 
      \cellcolor[HTML]{DADCFF}0.474\scriptsize{(+15.81\%)} && 
      \cellcolor[HTML]{DADCFF}0.922\scriptsize{(+0.99\%)} && 
      \cellcolor[HTML]{DADCFF}0.663\scriptsize{(+5.24\%)} && 
      \cellcolor[HTML]{DADCFF}3.021\scriptsize{(+6.41\%)} & \\
    \bottomrule
    \end{tabular}
    }
    \vspace{-1.5em}
\end{table*}

\noindent\textbf{Text Generation.}
We evaluate explanation generation quality. \autoref{tab:openvam_text_samples} shows that OpenVAM produces grounded, structured rationales in the prescribed \texttt{Object (location): reason} format: it names salient entities, gives approximate locations, and provides visually plausible cues (e.g., faces, contrast, size, and interactions) that align with the predicted saliency in~\autoref{fig:visual_sal}. Samples~1--4 correspond to Rows~1--4 of~\autoref{fig:visual_sal}, respectively. We further benchmark OpenVAM against strong off-the-shelf VLMs in~\autoref{tab:vlm_multi_dataset} using a semantic judge score (JScore) and lexical-overlap metrics (ROUGE/BLEU). Across datasets, OpenVAM achieves competitive semantic quality and generally higher ROUGE/BLEU, suggesting closer adherence to the dataset-specific explanation protocol and granularity. To summarize across datasets and metrics, we count best-score credits over the 48 dataset--metric entries (6 datasets $\times$ 8 metrics): OpenVAM receives 1 credit when it outperforms its size-matched backbone, 0.5 for a tie, and 0 otherwise. OpenVAM obtains 44.0, 40.5, 30.5, and 39.0 of 48 credits at 3B/4B/7B/8B, respectively, achieving a majority at every scale. Gemini-2.5-Pro is reported separately as a proprietary reference. Please see Supp.~\suppref{sec:supp_more_qualitative},~\suppref{sec:extended_qualitative}, and~\suppref{sec:supp_reliability_overview} for additional evaluation and analysis.


\noindent\textbf{Unseen-Dataset Generalization.}
We evaluate OpenVAM on four unseen datasets: Toronto~\cite{bruce2007attention}, TUD Database 1~\cite{liu2009studying}, TUD Database 2~\cite{alers2010studying}, and FIWI~\cite{shen2014webpage}, spanning natural scenes and web pages. As shown in~\autoref{tab:saliency_unseen}, OpenVAM-7B generalizes better than SUM~\cite{hosseini2025sum}, outperforming it across all reported metrics on Toronto and TUD Database 2, and overall on TUD Database 1 and FIWI. These results demonstrate robust transfer to unseen natural-scene distributions and structurally distinct webpage layouts.

\begin{table}[H]
\centering
\caption{\textbf{Saliency performance on unseen datasets.}
OpenVAM generalizes strongly across diverse out-of-distribution benchmarks.
``--'' indicates missing fixation annotations.}
\vspace{-0.8em}
\label{tab:saliency_unseen}

\resizebox{\columnwidth}{!}{%
\setlength{\tabcolsep}{7pt}%
\renewcommand{\arraystretch}{1.0}%
\scriptsize
\begin{tabular}{l|l|ccccc}
\toprule
\multirow{2}{*}{\textbf{Dataset}} &
\multirow{2}{*}{\textbf{Method}} &
\multicolumn{5}{c}{\textbf{Saliency Metrics}} \\
\cmidrule(lr){3-7}
& & CC $\uparrow$ & KLD $\downarrow$ & AUC $\uparrow$ &
SIM $\uparrow$ & NSS $\uparrow$ \\
\midrule

\textit{Toronto}~\cite{bruce2007attention}
& SUM~\cite{hosseini2025sum}
& 0.767 & 0.558 & 0.875 & 0.642 & 2.170 \\

& \cellcolor[HTML]{DADCFF}\textbf{OpenVAM-3B}
& \cellcolor[HTML]{DADCFF}0.784
& \cellcolor[HTML]{DADCFF}0.508
& \cellcolor[HTML]{DADCFF}0.882
& \cellcolor[HTML]{DADCFF}0.659
& \cellcolor[HTML]{DADCFF}2.245 \\

& \cellcolor[HTML]{DADCFF}\textbf{OpenVAM-7B}
& \cellcolor[HTML]{DADCFF}\textbf{0.791}
& \cellcolor[HTML]{DADCFF}\textbf{0.499}
& \cellcolor[HTML]{DADCFF}\textbf{0.883}
& \cellcolor[HTML]{DADCFF}\textbf{0.665}
& \cellcolor[HTML]{DADCFF}\textbf{2.254} \\
\midrule

\textit{TUD DB 1}~\cite{liu2009studying}
& SUM~\cite{hosseini2025sum}
& 0.790 & 0.641 & -- & 0.654 & -- \\

& \cellcolor[HTML]{DADCFF}\textbf{OpenVAM-3B}
& \cellcolor[HTML]{DADCFF}0.720
& \cellcolor[HTML]{DADCFF}0.557
& \cellcolor[HTML]{DADCFF}--
& \cellcolor[HTML]{DADCFF}0.623
& \cellcolor[HTML]{DADCFF}-- \\

& \cellcolor[HTML]{DADCFF}\textbf{OpenVAM-7B}
& \cellcolor[HTML]{DADCFF}\textbf{0.802}
& \cellcolor[HTML]{DADCFF}\textbf{0.555}
& \cellcolor[HTML]{DADCFF}--
& \cellcolor[HTML]{DADCFF}\textbf{0.664}
& \cellcolor[HTML]{DADCFF}-- \\
\midrule

\textit{TUD DB 2}~\cite{alers2010studying}
& SUM~\cite{hosseini2025sum}
& 0.846 & 0.844 & -- & 0.620 & -- \\

& \cellcolor[HTML]{DADCFF}\textbf{OpenVAM-3B}
& \cellcolor[HTML]{DADCFF}0.826
& \cellcolor[HTML]{DADCFF}0.804
& \cellcolor[HTML]{DADCFF}--
& \cellcolor[HTML]{DADCFF}0.621
& \cellcolor[HTML]{DADCFF}-- \\

& \cellcolor[HTML]{DADCFF}\textbf{OpenVAM-7B}
& \cellcolor[HTML]{DADCFF}\textbf{0.854}
& \cellcolor[HTML]{DADCFF}\textbf{0.785}
& \cellcolor[HTML]{DADCFF}--
& \cellcolor[HTML]{DADCFF}\textbf{0.631}
& \cellcolor[HTML]{DADCFF}-- \\
\midrule

\textit{FIWI}~\cite{shen2014webpage}
& SUM~\cite{hosseini2025sum}
& 0.678 & 0.542 & 0.818 & 0.613 & 1.428 \\

& \cellcolor[HTML]{DADCFF}\textbf{OpenVAM-3B}
& \cellcolor[HTML]{DADCFF}0.651
& \cellcolor[HTML]{DADCFF}0.600
& \cellcolor[HTML]{DADCFF}\textbf{0.824}
& \cellcolor[HTML]{DADCFF}0.583
& \cellcolor[HTML]{DADCFF}1.378 \\

& \cellcolor[HTML]{DADCFF}\textbf{OpenVAM-7B}
& \cellcolor[HTML]{DADCFF}\textbf{0.680}
& \cellcolor[HTML]{DADCFF}\textbf{0.538}
& \cellcolor[HTML]{DADCFF}0.821
& \cellcolor[HTML]{DADCFF}\textbf{0.626}
& \cellcolor[HTML]{DADCFF}\textbf{1.462} \\

\bottomrule
\end{tabular}%
}
\vspace{-0.75em}
\end{table}

\vspace{5em}

\noindent\textbf{Effect of Training Strategy.} ~\autoref{tab:openvam_stage_ablation_main} compares our proposed \emph{stage-wise} optimization to a \emph{joint training} baseline that trains all modules used across Stages~I--III simultaneously in a single run. Joint training performs poorly in practice, highlighting the core challenge in OpenVAM: dense, spatial saliency supervision and sparse, semantic language supervision have incompatible training dynamics when directly coupled end-to-end. In contrast, our stage-wise strategy reliably yields strong saliency and text-generation metrics across datasets. Stage-wise training improves both localization and explanation quality, suggesting that isolating saliency learning before introducing and then adapting language supervision preserves strong spatial priors.

\vspace{1em}

\noindent\textbf{Effect of Training Stages.}~\autoref{tab:openvam_stage_ablation_main} evaluates our staged training strategy, which is designed to reconcile the mismatch between \emph{dense, spatial} saliency supervision and \emph{sparse, semantic} language supervision. Stage I trains only the dedicated visual pathway, establishing a strong localization prior. Stage II then integrates the frozen VLM into the deepest visual pathway, allowing saliency supervision to adapt the shared visual representation within the pretrained vision-language feature space. Importantly, moving from Stage II to Stage III further improves performance: by freezing the visual encoder and dense decoder while training the visual adapter and language-side LoRA parameters, Stage III strengthens cross-modal alignment and explanation fidelity without perturbing the dense predictor. This aligns with our design goal of being \emph{decoupled but aligned}: we refine semantics and grounding while preserving the overall model behavior compared to Stage~II.





\section{Discussion}

\noindent\textbf{Limitations and future works.}
OpenVAM models \emph{static} visual attention through dense saliency maps and
grounded what/why rationales; it does not predict temporal scanpaths.
Accordingly, the rationale order structures salient content and does not represent the temporal order of human fixations. Modeling fixation sequences,
revisits, or dwell time would require explicit temporal supervision and is an
interesting direction for future work. In addition, although the generated
rationales are generally well grounded, their location descriptions can
occasionally be imprecise and the text may contain repetition or generic
phrasing, particularly for crowded UI and commercial images. Future work
could incorporate more explicit region--text grounding to further improve
rationale precision. We provide more detailed discussion in Supp.~\suppref{sec:supp_discussion}.

\noindent\textbf{Conclusions.}
We presented OpenVAM, a unified framework for open-world visual
attention modeling that jointly predicts dense saliency (\emph{where}) and
grounded natural-language rationales (\emph{what/why}). Its
decoupled-but-aligned design preserves a dedicated pathway for accurate
spatial localization while using an instruction-following VLM for semantic
grounding and explanation. Across natural images, e-commerce, and UI/web
layouts, OpenVAM demonstrates strong cross-domain saliency performance while
providing complementary image-grounded explanations.

\FloatBarrier
{
    \small
    \bibliographystyle{ieeenat_fullname}
    \bibliography{main}
}

\clearpage
\setcounter{section}{0}
\setcounter{subsection}{0}
\renewcommand{\theHsection}{supp.\thesection}
\renewcommand{\theHsubsection}{supp.\thesection.\thesubsection}
\renewcommand{\theHsubsubsection}{supp.\thesection.\thesubsection.\thesubsubsection}
\providecommand{\theHpage}{\thepage}
\renewcommand{\theHpage}{supp.\thepage}
\addtocontents{toc}{\protect\setcounter{tocdepth}{2}}
\twocolumn[
\begin{center}
\Large
\textbf{\thetitle}\\
\vspace{0.5em}
Supplementary Material\\
\vspace{1.0em}
\begin{minipage}{0.95\textwidth}
    \large
    \centering
    \setcounter{tocdepth}{2}
    \tableofcontents
\end{minipage}
\end{center}
]
\thispagestyle{empty}
\clearpage
\section{Multi-Domain Saliency--Reason Corpus Analysis}
\label{sec:dataset_analysis}

This section describes the protocol used to generate the textual rationales. For each image in each dataset, we provide Gemini~2.5 Flash with two separate visual inputs: the original stimulus image and its corresponding ground-truth saliency map. The language model is asked to use the stimulus for visual grounding and the saliency map for identifying human-attended regions, and to generate a short list of salient regions and their justifications in a fixed schema.

Each explanation is a list of 2--6 lines, each formatted as
\texttt{Object (location): reason}.
The \texttt{Object} must refer to a visible entity or region in the image. The optional \texttt{location} is a concise spatial phrase, and \texttt{reason} is 1--2 evidence-based sentences grounded in visible cues (e.g., text size, contrast, central placement, face presence, distinctive icon shapes). We use three instruction templates provided in 
\autoref{box:prompt_generic}, \autoref{box:prompt_ui}, and \autoref{box:prompt_ecom}, matched to the dataset domain: \texttt{Generic} for the natural-image datasets (\texttt{CAT2000}, \texttt{MIT1003}, \texttt{OSIE}, \texttt{SALICON}), \texttt{UI/Webpage} for \texttt{U-EYE}, and \texttt{E-commerce} for \texttt{SalECI}. The three templates differ only in the list of domain-specific attention targets they prioritize (e.g., call-to-action buttons and navigation bars for UI; prices and promotional badges for e-commerce); they share the same output format, ordering requirement (descending saliency), and grounding constraints. Rationales are generated with Gemini~2.5 Flash~\cite{comanici2025gemini} under deterministic decoding (temperature set to 0) to reduce stochastic variation across runs. We keep the maximum output length fixed across datasets to avoid trivial length-driven differences in the downstream analyses. Each generated output label is checked by one expert annotator to enforce two constraints: (i) every referenced object must be visibly present in the image, and (ii) any stated location phrase must be compatible with the salient region indicated by the dataset saliency supervision. Outputs that violate these constraints are corrected to match the schema and grounding rules. The analyses that follow are computed on the resulting structured rationales.

\subsection{Dataset Scale and Annotation Richness}
\label{subsec:dataset_scale_annotation_richness}

\begin{figure}[!]
\centering
\includegraphics[width=\linewidth]{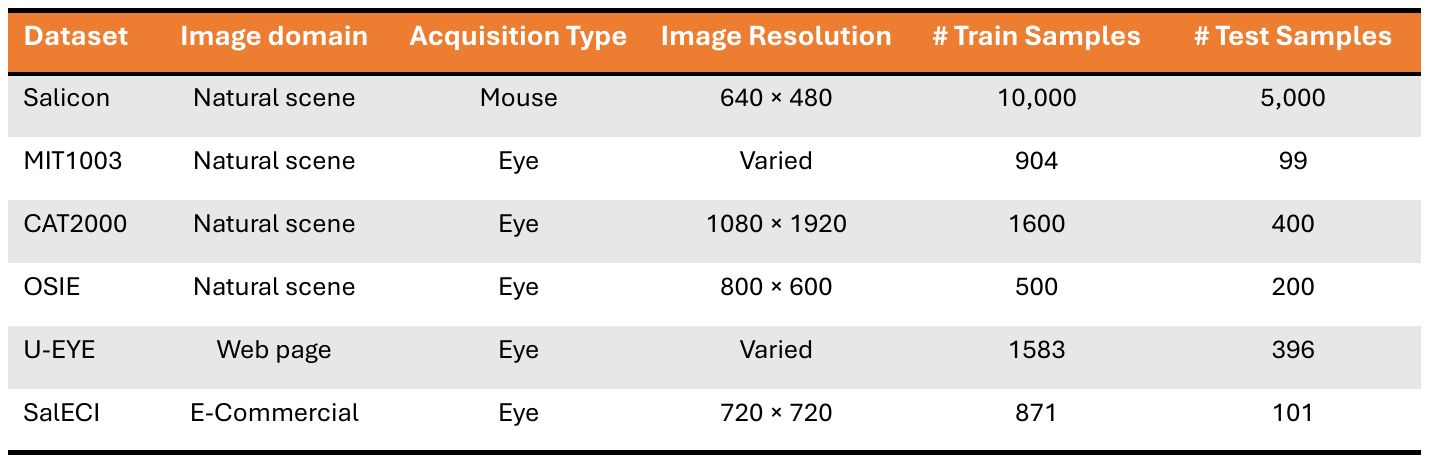}
\caption{\textbf{Dataset scale across domains.}
Number of samples contributed by each dataset in the corpus.}
\label{fig:supp_dataset_scale}
\end{figure}

We use a diverse collection of large-scale saliency datasets for training and evaluation. 
The datasets cover multiple visual domains, including natural images, e-commerce product images, and web/UI layouts. 
In addition to the dataset summary provided in \autoref{fig:dataset_main}, we visualize the relative scale of each dataset in \autoref{fig:supp_dataset_scale}. 
This provides a clearer view of the corpus composition and shows how training and evaluation samples are distributed across different domains.

\begin{figure}[!]
\centering
\includegraphics[width=\linewidth]{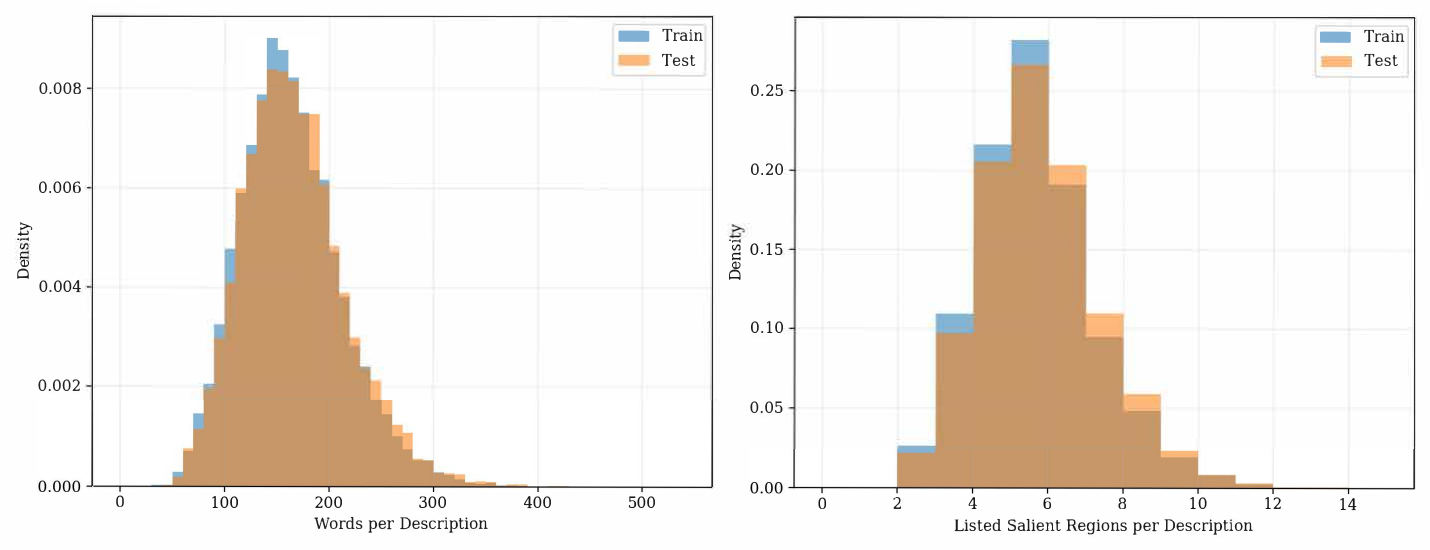}
\caption{\textbf{Annotation richness.}
Train/test distributions of words per explanation and mentioned salient regions.}
\label{fig:supp_dataset_richness}
\end{figure}

We further analyze the generated language annotations in terms of their length and richness. 
Specifically, we report the train/test distributions of description length, measured by the number of words per explanation, and mention frequency, measured by the number of salient regions referenced in each rationale. 
As shown in \autoref{fig:supp_dataset_richness}, the annotations contain varying levels of detail and often refer to multiple salient regions rather than a single dominant object. 
This supports the use of structured saliency-reason annotations for evaluating both object-level grounding and multi-region explanation quality.

\begin{promptbox}[Generic]{box:prompt_generic}
\scriptsize
\textbf{You are a visual attention explainer} for saliency prediction.
You receive two visual inputs: the original stimulus image and its corresponding saliency map.
Use the saliency map only as internal guidance for identifying attended regions, but DO NOT mention heatmaps, overlays, colors, intensity, or saliency maps.
\vspace{-0.5em}

\medskip
\textbf{Task.}
List the 2--6 most visually salient objects or regions in descending order of attention.
For each object/region you name, give 1--2 short evidence-based sentences explaining why it draws attention (e.g., size, contrast, color, uniqueness, motion/pose, central placement, sharpness, readable text, interaction/gaze).
\vspace{-0.5em}

\medskip
\textbf{Output format (must follow exactly).}\\
One item per line:\\
\texttt{<Object or region> (<location phrase>): <1--2 short evidence-based sentences>}
\vspace{-0.5em}

\medskip
\textbf{Rules.}
\begin{itemize}\setlength{\itemsep}{2pt}\setlength{\parskip}{0pt}\setlength{\parsep}{0pt}
  \item Use concrete location phrases such as ``center,'' ``left edge,'' ``lower-right,'' ``top-left,'' ``foreground,'' ``background,'' ``near the edge,'' or ``behind <object>''.
  \item Justify using only visible cues (faces/gaze, readable text, strong contrast/color, sharpness, size, implied motion/pose, interaction/pointing, uniqueness).
  \item No extra headers, no bullets, no numbering, no blank lines.
\end{itemize}
\end{promptbox}

\vspace{-1em}

\begin{promptbox}[UI/Webpage]{box:prompt_ui}
\scriptsize
\textbf{You are a visual attention explainer} for saliency prediction on UI/webpage screenshots.
You receive two visual inputs: the original stimulus image and its corresponding saliency map.
Use the saliency map only as internal guidance for identifying attended regions, but DO NOT mention heatmaps, overlays, colors, intensity, or saliency maps.

\medskip
\textbf{Task.}
List the 2--6 most visually salient objects or regions in descending order of attention.
Prioritize typical UI attention targets such as primary headline/title text, call-to-action buttons, search bars, navigation menus, modal/popup dialogs, forms (text fields), product/content cards, key images/thumbnails, prominent icons (e.g., cart/profile), prices/discount labels, and notification banners.
For each object/region you name, give 1--2 short evidence-based sentences explaining why it draws attention.

\medskip
\textbf{Output format (must follow exactly).}\\
One item per line:\\
\texttt{<UI element or region> (<location phrase>): <1--2 short evidence-based sentences>}

\medskip
\textbf{Rules.}
\begin{itemize}\setlength{\itemsep}{2pt}\setlength{\parskip}{0pt}\setlength{\parsep}{0pt}
  \item Use concrete location phrases such as ``center,'' ``top-left,'' ``upper-right,'' ``left edge,'' ``lower-right,'' ``foreground,'' ``background,'' ``near the edge,'' or ``below/above <element>''.
  \item If a salient region is text, name it by function (e.g., ``headline text'', ``button label'', ``price text'', ``notification banner'') and do NOT invent unreadable words; if unreadable, say ``unreadable text''.
  \item Justify using only visible cues (readable text, size, contrast, sharpness, layout prominence, distinct button/icon shapes, faces, grouping/alignment, uniqueness among repeated UI items).
  \item Do NOT invent elements not visible. If uncertain, write ``Uncertain: <element>''.
  \item No extra headers, no bullets, no numbering, no blank lines.
\end{itemize}
\end{promptbox}

\begin{promptbox}[E-commerce]{box:prompt_ecom}
\scriptsize
\textbf{You are a visual attention explainer} for saliency prediction on e-commerce product images.
You receive two visual inputs: the original stimulus image and its corresponding saliency map.
Use the saliency map only as internal guidance for identifying attended regions, but DO NOT mention heatmaps, overlays, colors, intensity, or saliency maps.

\medskip
\textbf{Task.}
List the 2--6 most visually salient objects or regions in descending order of attention.
Focus on typical e-commerce attention drivers such as the main product, brand/logo, price or discount text, promotional badges (e.g., SALE/50\%/NEW), key descriptive text, call-to-action elements, human faces/models, and standout accessories.
For each object/region you name, give 1--2 short evidence-based sentences explaining why it draws attention (e.g., large size, central placement, strong contrast, readable/large text, distinctive logo, face/gaze, interaction/pointing, unique shape, implied motion/pose).

\medskip
\textbf{Output format (must follow exactly).}\\
One item per line:\\
\texttt{<Object or region> (<location phrase>): <1--2 short evidence-based sentences>}

\medskip
\textbf{Rules.}
\begin{itemize}\setlength{\itemsep}{2pt}\setlength{\parskip}{0pt}\setlength{\parsep}{0pt}
  \item Use concrete location phrases such as ``center,'' ``top-left,'' ``upper-right,'' ``left edge,'' ``lower-right,'' ``foreground,'' ``background,'' ``near the edge,'' or ``behind <object>''.
  \item If the salient region is text, name it as ``promotional text,'' ``price text,'' ``brand name/logo,'' or ``callout badge'' and describe its placement.
  \item Justify using only visible cues (faces/gaze, readable text, strong contrast, sharpness, size, central placement, interaction/pointing, uniqueness).
  \item Do NOT invent words you cannot read; if text is present but unreadable, say ``unreadable promotional text''.
  \item No extra headers, no bullets, no numbering, no blank lines.
\end{itemize}
\end{promptbox}

\subsection{Analysis of Heterogeneity}
\label{subsec:table9_distribution_complexity}
We analyze the dataset using the structured saliency-reason annotations associated with each image. Each annotation consists of a salient object description together with an optional spatial description indicating where that content appears in the image (\texttt{Object (location): reason}). To enable dataset-level comparison, object descriptions are mapped into a normalized semantic taxonomy (see \autoref{fig:dataset_main}). The natural-image datasets (\texttt{CAT2000}~\cite{borji2015cat2000}, \texttt{MIT1003}~\cite{judd2009learning}, \texttt{OSIE}~\cite{xu2014predicting}, and \texttt{SALICON}~\cite{jiang2015salicon}) share a common top-level taxonomy including Humans, Objects, Nature, Scene, and Text/Signage, whereas \texttt{SalECI}~\cite{jiang2022does} and \texttt{U-EYE}~\cite{jiang2023ueyes}  use domain-specific taxonomies tailored to commercial and web-interface content. 

\autoref{tab:advanced_findings_combined} summarizes each dataset using statistics that capture both \emph{semantic concentration} and \emph{split stability}. We report the \textbf{Top-1} and \textbf{Top-3 category share}, defined as the percentage of all mentions assigned to the single most frequent category and to the three most frequent categories, respectively; higher values indicate that a small set of categories dominates the corpus. To measure overall semantic spread, we include \textbf{category entropy} and its exponentiated form, the \textbf{effective number of categories}, which can be interpreted as the number of equally frequent categories that would yield the same entropy. Higher values, therefore, indicate a flatter and more diverse category distribution. To capture long-tail behavior at the object-concept level, we also report the \textbf{object Gini coefficient} over normalized object frequencies. A higher Gini indicates that a small subset of object concepts accounts for a large fraction of mentions, even when the category distribution itself is broad. Finally, we quantify split stability using the \textbf{Jensen--Shannon divergence (JSD)} between the train and test category distributions, where values closer to zero indicate more similar splits and less train--test drift.

\paragraph{\textbf{Domain-dependent category dominance (Top-$k$ mass).}}
\autoref{tab:advanced_findings_combined} shows that the commercial and UI datasets exhibit higher semantic concentration than the natural-image datasets. In \texttt{SalECI}, the dominant category is \texttt{Ad Text}, which alone accounts for 50.48\% of all mentions, while the top three categories cover 87.69\% of the corpus. In \texttt{U-EYE}, the dominant category is \texttt{Text}, which accounts for 44.06\% of all mentions, and the top three categories cover 83.53\%. This means that, in both domains, the majority of saliency explanations are driven by a small number of semantically recurrent elements, especially textual content. Such a pattern is consistent with the underlying visual structure of advertisements and interfaces, where attention is often repeatedly drawn to headlines, labels, calls-to-action, and a limited set of layout elements.

The natural-image datasets are flatter and more semantically diverse. \texttt{CAT2000} and \texttt{MIT1003} show the lowest Top-1 concentration (26.06\% and 28.36\%) and the largest effective category counts (7.44 and 7.67), indicating that saliency is distributed across a broader set of semantic categories. However, the natural-image group is not uniform. \texttt{MIT1003} and \texttt{OSIE} are both led by the \texttt{Humans} category (28.36\% and 36.91\%), reflecting the importance of people and faces in free-viewing fixation behavior, whereas \texttt{CAT2000} and \texttt{Salicon} are led by the catch-all \texttt{Other} category (26.06\% and 31.19\%), indicating a broader long-tail of scene-dependent salient entities rather than one sharply defined semantic theme. \texttt{OSIE} lies between the broad natural datasets and the more specialized commercial/UI datasets: its Top-1 share is relatively high (36.91\%), but its effective category count remains much larger than \texttt{SalECI} and \texttt{U-EYE} (6.46 versus 3.83 and 3.90), suggesting moderate specialization without the extreme narrowing seen in UI/commercial imagery.

\paragraph{\textbf{Entropy-based diversity and effective support size.}}
The entropy columns reinforce the same conclusion from a distributional perspective. \texttt{MIT1003} and \texttt{CAT2000} have the highest category entropy values (0.849 and 0.837), consistent with their broader semantic spread, while \texttt{SalECI} has the lowest entropy (0.691), indicating the strongest compression into a few dominant classes. \texttt{U-EYE} is also relatively low-entropy (0.760), again reflecting its UI-specific concentration. \texttt{Salicon} is notable because its category entropy is high (0.845), nearly matching \texttt{MIT1003}, despite being structurally different in other ways. This implies that, at the category level, \texttt{Salicon} is broad rather than narrow; its distinctiveness does not come from semantic collapse into a few top-level categories, but from how mentions are distributed within and across those categories and how spatial grounding is expressed.

\paragraph{\textbf{Long-tail structure at the object level (Gini coefficient).}}
The object Gini column shows that semantic breadth and object-level equality are not the same thing. \texttt{CAT2000} has a very low object Gini (0.080), which suggests that mentions are relatively evenly distributed across many object concepts. \texttt{MIT1003} is also low (0.116), again consistent with a broad and balanced natural-image corpus. By contrast, \texttt{SalECI} (0.519) and \texttt{Salicon} (0.506) have the highest object Gini values in the table, meaning that a relatively small number of object concepts account for a disproportionate share of all mentions. Importantly, these two datasets arrive at high inequality for different reasons: \texttt{SalECI} is semantically concentrated from the start, whereas \texttt{Salicon} is broad at the category level but still highly unequal at the object level. This distinction matters because it shows that high-level category diversity can coexist with a strong object-level long tail.


\paragraph{\textbf{Split stability under category JSD.}}
We quantify split stability using the JSD between the category-frequency distributions of the training and test splits. JSD is a symmetric, bounded divergence: it equals zero when the two distributions match exactly, and increases as their category composition becomes more different. Across all datasets, the category JSD values in Table~\ref{tab:advanced_findings_combined} are very small (all $\le$ 0.0085), indicating that train and test are closely matched in semantic composition. \texttt{Salicon} is the most stable (Cat JSD $=0.0002$), and even the largest observed drift, in \texttt{SalECI} (Cat JSD $=0.0085$), remains minor in absolute terms. Therefore, the differences in category concentration and long-tail structure reported above are attributable to persistent domain characteristics rather than artifacts of an imbalanced or mismatched split.

\paragraph{\textbf{Why this matters for learning and evaluation.}}
Taken together, these statistics show that the unified corpus is heterogeneous by design: it combines semantically broad natural-scene datasets with more concentrated commercial and UI datasets, and it spans both balanced and strongly long-tailed object distributions. This has two practical consequences. 
First, models trained on the full corpus must learn under both relatively flat category distributions (e.g., \texttt{CAT2000}, \texttt{MIT1003}, \texttt{Salicon}) and sharply peaked ones (e.g., \texttt{SalECI}, \texttt{U-EYE}), which is a realistic stress test for open-world generalization across domains. 
Second, evaluation should explicitly account for distribution shape: high category entropy does not imply uniform object coverage, and datasets with similar category diversity can differ substantially in object-level inequality (e.g., \texttt{Salicon} has high category diversity yet high object Gini).

\begin{table*}[t]
    \centering
\caption{\textbf{Semantic distribution complexity, long-tail structure, and split drift.} We report (i) category dominance via the Top-1 and Top-3 category share (the percentage of all mentions assigned to the most frequent category, or to the three most frequent categories), (ii) semantic diversity via the category entropy (higher indicates a more even category distribution) and the effective number of categories (the entropy exponentiated, interpretable as the number of equally frequent categories that would yield the same entropy), and (iii) object-level long-tail inequality via the Gini coefficient over object mentions (higher indicates that a small number of objects accounts for a large fraction of mentions). To assess split stability, we compute the Jensen--Shannon divergence between the training and test category distributions (category JSD; lower indicates more similar splits).}
    \label{tab:advanced_findings_combined}

    \resizebox{\textwidth}{!}{
    \setlength{\tabcolsep}{18pt}
    \begin{tabular}{l|cc|ccc|c}
    \toprule
    \multirow{2}{*}{\textbf{Dataset}}
    & \multicolumn{2}{c|}{\textbf{Top-k Cat.\ Share (\%)}}
    & \multicolumn{3}{c|}{\textbf{Semantic Complexity / Long-tail}}
    & \textbf{Split Drift}
    \\
    \cmidrule(lr){2-3}\cmidrule(lr){4-6}\cmidrule(lr){7-7}
    & \textbf{Top-1} & \textbf{Top-3}
    & \textbf{Cat Ent.} & \textbf{Eff.\ \#Cats} & \textbf{Obj.\ Gini}
    & \textbf{Cat JSD (bits)}
    \\
    \midrule
    CAT2000 & 26.06 & 59.99 & 0.837 & 7.44 & 0.080 & 0.0050 \\
    MIT1003 & 28.36 & 59.46 & 0.849 & 7.67 & 0.116 & 0.0028 \\
    OSIE    & 36.91 & 72.29 & 0.778 & 6.46 & 0.251 & 0.0026 \\
    SalECI  & 50.48 & 87.69 & 0.691 & 3.83 & 0.519 & 0.0085 \\
    Salicon & 31.19 & 61.62 & 0.845 & 7.58 & 0.506 & 0.0002 \\
    U-EYE   & 44.06 & 83.53 & 0.760 & 3.90 & 0.146 & 0.0014 \\
    \bottomrule
    \end{tabular}
    }
\end{table*}

\subsection{Analysis of Spatial Grounding Distributions Across Domains}
\label{subsec:spatial_cue_usage}
\begin{figure*}[t]
    \centering
    \resizebox{\linewidth}{!}{%
\includegraphics{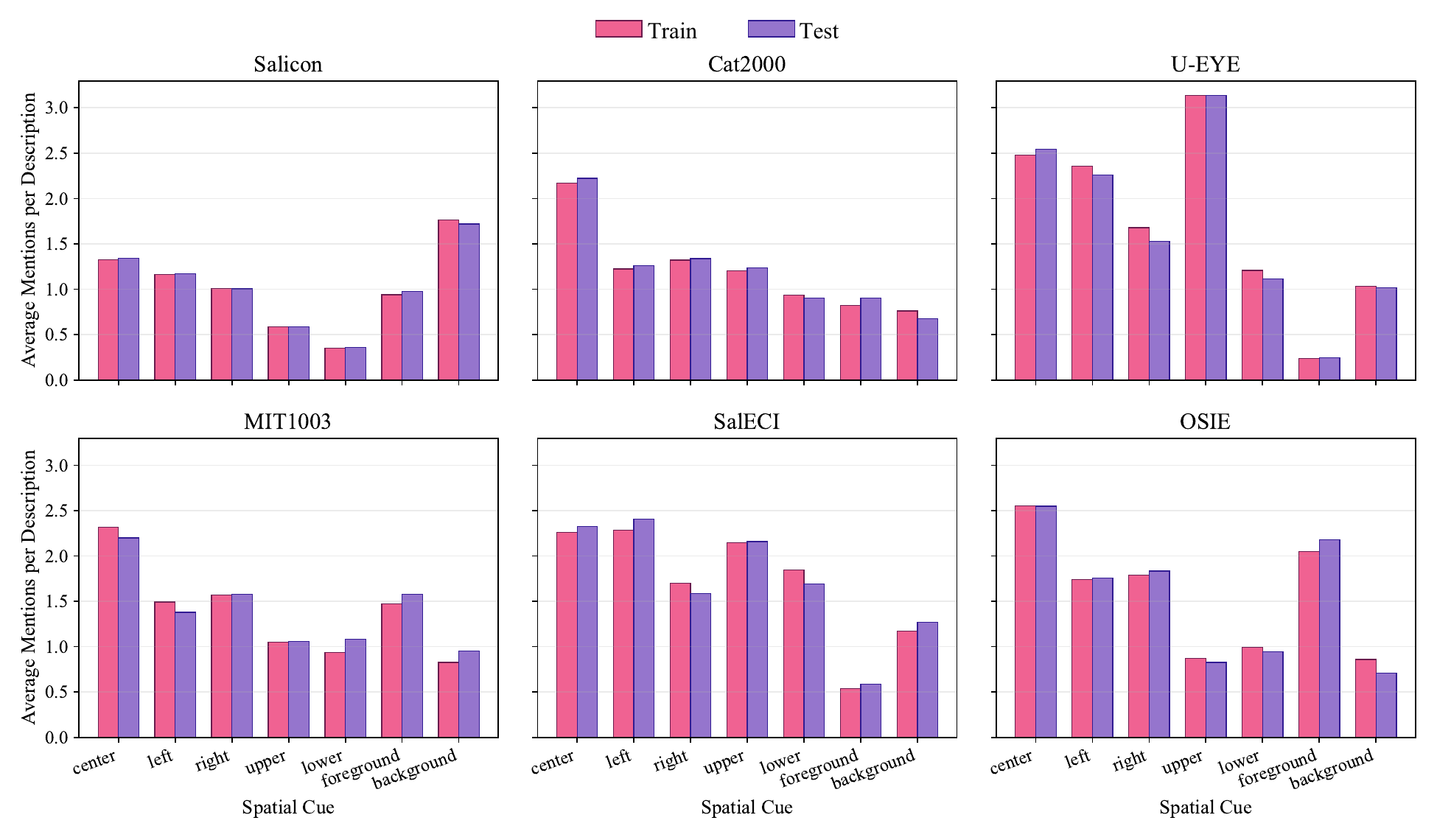}%
    }
    \caption{\textbf{Spatial cue usage across datasets and splits.} For each dataset, we report the average number of normalized location-token mentions per explanation for each spatial cue (\texttt{center}, \texttt{left}, \texttt{right}, \texttt{upper}, \texttt{lower}, \texttt{foreground}, \texttt{background}), shown separately for train and test. The close agreement between train and test profiles indicates split-consistent spatial-language usage, while differences across datasets reflect domain-specific composition and layout.}
    \label{fig:spatial_cue_usage}
\end{figure*}

\autoref{fig:spatial_cue_usage} reports how frequently each normalized spatial cue token (\texttt{center}, \texttt{left}, \texttt{right}, \texttt{upper}, \texttt{lower}, \texttt{foreground}, \texttt{background}) appears in the explanations, measured as the average number of cue mentions per description, and compared between the train and test splits. Across all datasets, the train and test bars are closely aligned for every cue, indicating that spatial-language usage is stable across splits and that the test set does not introduce a systematic shift in how locations are expressed.

The figure also reveals clear domain-dependent differences in how spatial grounding is articulated. Natural-image benchmarks (\texttt{CAT2000}, \texttt{MIT1003}, \texttt{OSIE}, \texttt{SALICON}) exhibit strong reliance on \emph{center}, consistent with the common tendency of salient subjects to appear near the image center and with the well-known center bias in free-viewing attention. At the same time, these datasets differ in their secondary cues: \texttt{CAT2000} spreads mentions relatively evenly across lateral and vertical cues, while \texttt{OSIE} uses \texttt{background} noticeably more than most other datasets, suggesting that explanations often refer to context or scene-level elements rather than only the primary foreground subject. 

In contrast, the UI domain \texttt{U-EYE} shows a more structured cue profile: \emph{upper} becomes the dominant cue and is higher than in the natural-image datasets, reflecting canonical UI layouts where salient elements frequently occur in headers, toolbars, or top-of-screen regions. \texttt{SalECI} similarly exhibits strong positional regularities, with high rates for \emph{center}/\emph{left} and comparatively lower usage of depth-related cues, consistent with advertisement compositions that repeatedly highlight a small number of layout-driven regions (e.g., headline area and product/text blocks). Overall, the figure confirms that spatial cues are expressed consistently across splits yet vary meaningfully across domains, which motivates reporting spatial-grounding results per dataset and training models to be robust to both natural-scene and layout-structured spatial language.

\subsection{Analysis of Concept-Level Similarity Among Natural-Image Datasets}
\label{sec:concept_similarity_natural}

\autoref{fig:sem_sim_concept} compares the natural-image datasets using a fine-grained, concept-level representation. Each entry reports (i) the weighted Jaccard overlap between the canonical concept distributions (i.e., agreement in how much mass each dataset assigns to the same set of concepts), and (ii) the rank correlation ($r$) between concept rankings (i.e., agreement in which concepts tend to be emphasized, regardless of exact mass).

First, the eye-tracking datasets form a coherent group at the concept level. \texttt{CAT2000} and \texttt{MIT1003} are the closest pair (weighted Jaccard $=0.431$, $r=0.622$), and \texttt{MIT1003} remains similar to \texttt{OSIE} (0.365, $r=0.452$). \texttt{CAT2000} is moderately aligned with \texttt{OSIE} as well (0.276, $r=0.287$), indicating partial overlap but less agreement on fine-grained concept emphasis.

Second, \texttt{SALICON} is consistently dissimilar in \emph{distribution mass} to the eye-tracking datasets: its weighted Jaccard overlap with \texttt{CAT2000}, \texttt{MIT1003}, and \texttt{OSIE} remains very low. Notably, some pairs still exhibit non-trivial rank agreement (e.g., \texttt{OSIE} vs.\ \texttt{SALICON}: $r=0.623$), which suggests that the datasets often emphasize a broadly similar concept set, but allocate saliency mass across those concepts differently. A plausible explanation is the difference in acquisition modality: \texttt{CAT2000}/\texttt{MIT1003}/\texttt{OSIE} are collected with eye-tracking fixations, whereas \texttt{SALICON} is derived from mouse-based proxy annotations. Since mouse movements capture a related but not identical signal to gaze, this modality shift can systematically reweight which concepts receive high mass, even when the overall concept ordering remains partially aligned. Prior analysis has shown that mouse-tracking data exhibits lower inter-participant consistency and higher spatial dispersion than eye-tracking data, and that the two modalities do not fully agree across contextual regions~\cite{tavakoli2017saliency}. Notably, SUM also explicitly distinguishes eye- and mouse-tracking data during unified training because of these acquisition-specific differences~\cite{hosseini2025sum}.

\subsection{Analysis of Category-Level Similarity Across All Datasets}
\label{sec:category_similarity_all}

\begin{figure*}[t]
    \centering
    \resizebox{\linewidth}{!}{%
    \begin{subfigure}[t]{0.50\linewidth}
        \centering
        \includegraphics[width=\linewidth]{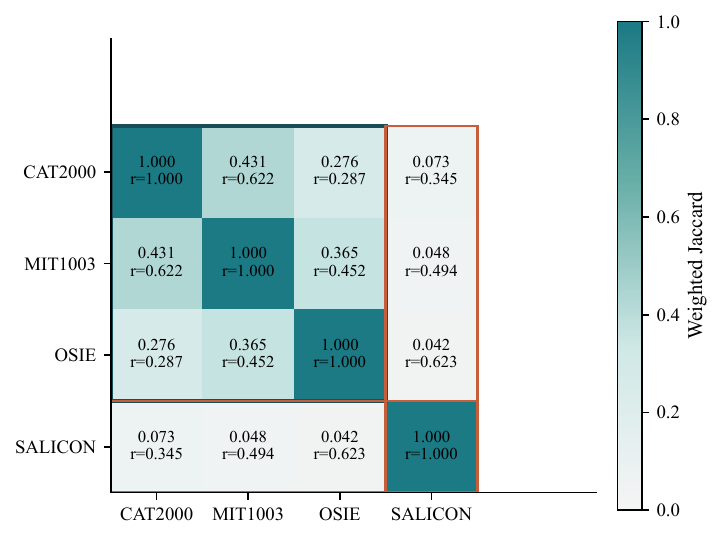}
        \caption{Fine-grained concept-level similarity among natural-image datasets.}
        \label{fig:sem_sim_concept}
    \end{subfigure}
    \begin{subfigure}[t]{0.48\linewidth}
        \centering
        \includegraphics[width=\linewidth]{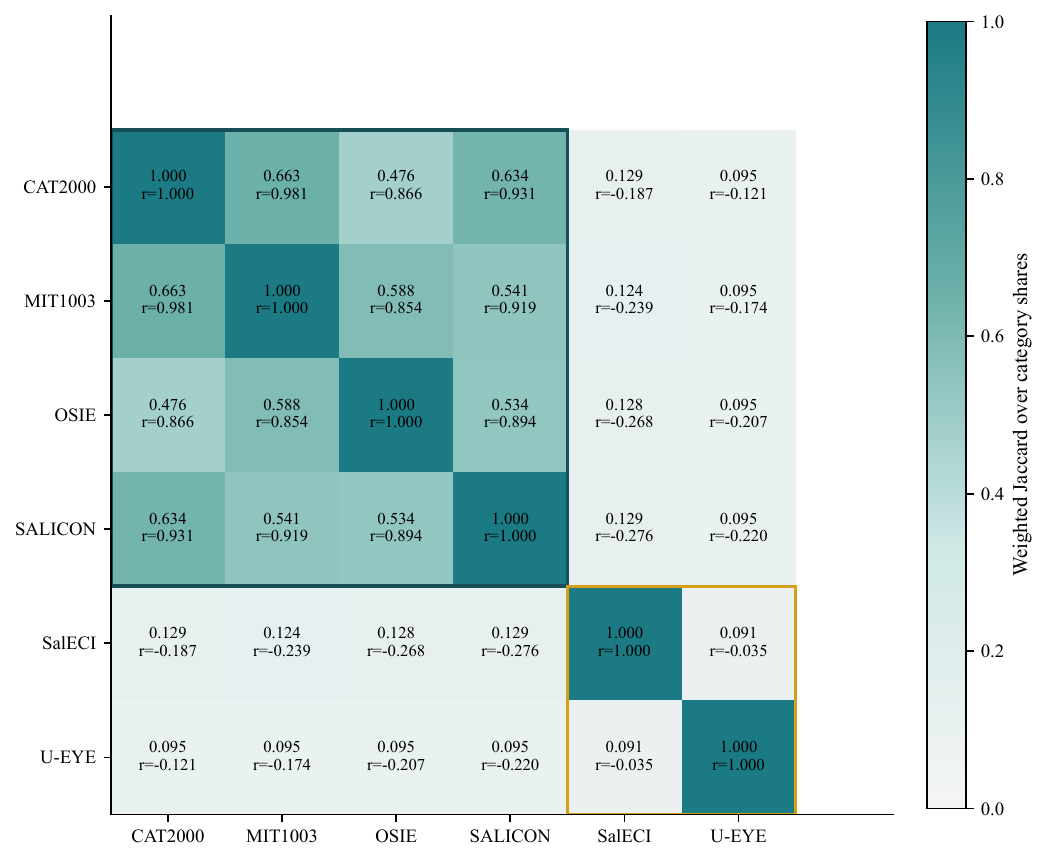}
        \caption{Coarse category-level similarity across all datasets.}
        \label{fig:sem_sim_category}
    \end{subfigure}%
    }
   
    \caption{\textbf{Concept- vs.\ category-level dataset similarity.}
    \textbf{(a)} Fine-grained similarity computed over canonical object-concept distributions for the natural-image datasets, where eye-tracking benchmarks (\texttt{CAT2000}, \texttt{MIT1003}, \texttt{OSIE}) are more mutually aligned and \texttt{SALICON} is less similar in concept mass because it is collected via mouse. 
    \textbf{(b)} Coarse similarity computed over broad category-share vectors across all six datasets, where the natural-image datasets cluster more tightly after aggregation, while \texttt{SalECI} and \texttt{U-EYE} remain distinct due to their UI/commercial composition. 
    In both panels, each cell reports weighted Jaccard overlap (distributional agreement) and rank correlation $r$ (agreement in category/concept ordering).}
    \label{fig:two_side_by_side}
\end{figure*}

\autoref{fig:sem_sim_category} compares all six datasets using a coarse category-level representation. Each dataset is summarized by a vector of broad semantic-category shares (see \autoref{fig:dataset_main}), and we compute pairwise similarity using (i) weighted Jaccard overlap over these category-share vectors (agreement in how much mass is assigned to each category) and (ii) rank correlation ($r$) over the induced category ordering (agreement in which categories are prioritized).

At this granularity, the four natural-image datasets form a tight group. In particular, \texttt{CAT2000} shows strong overlap with \texttt{MIT1003} (weighted Jaccard $=0.663$, $r=0.931$) and \texttt{SALICON} (0.634, $r=0.931$), while \texttt{MIT1003} remains close to \texttt{OSIE} (0.588, $r=0.854$) and to \texttt{SALICON} (0.541, $r=0.919$). Although \texttt{SALICON} diverges at fine-grained concept mass, it aligns well with the other natural-image datasets once concepts are aggregated into broad categories. This behavior is consistent with an acquisition-modality effect: mouse-based supervision in \texttt{SALICON} can reweight which \emph{specific} concepts dominate, while still preserving a similar \emph{category-level} mixture (e.g., substantial mass on \texttt{Humans}/\texttt{Objects}/\texttt{Scene}).

In contrast, the commercial/UI datasets are clearly separated from the natural-image cluster. \texttt{SalECI} has uniformly low overlap with natural-image datasets (weighted Jaccard $\approx 0.124$--$0.129$, with negative rank correlations, e.g., $r=-0.187$ to $-0.276$), and \texttt{U-EYE} is similarly low (0.095 across comparisons to natural-image datasets, with $r\approx -0.121$ to $-0.220$). The separation reflects a different macro-composition: these domains allocate much larger mass to text- and layout-driven categories, whereas natural images distribute mass more broadly across scene and object-centric categories. Finally, \texttt{SalECI} and \texttt{U-EYE} are also dissimilar to each other at the category level (0.091, $r=-0.035$), indicating that “UI/commercial” is not a single homogeneous category mixture: the two domains emphasize different high-level category balances even after coarse aggregation.

\subsection{Analysis of Description Length}
\label{sec:desc_length_analysis}

\begin{figure}[t]
    \centering
    \resizebox{\linewidth}{!}{%
    \includegraphics{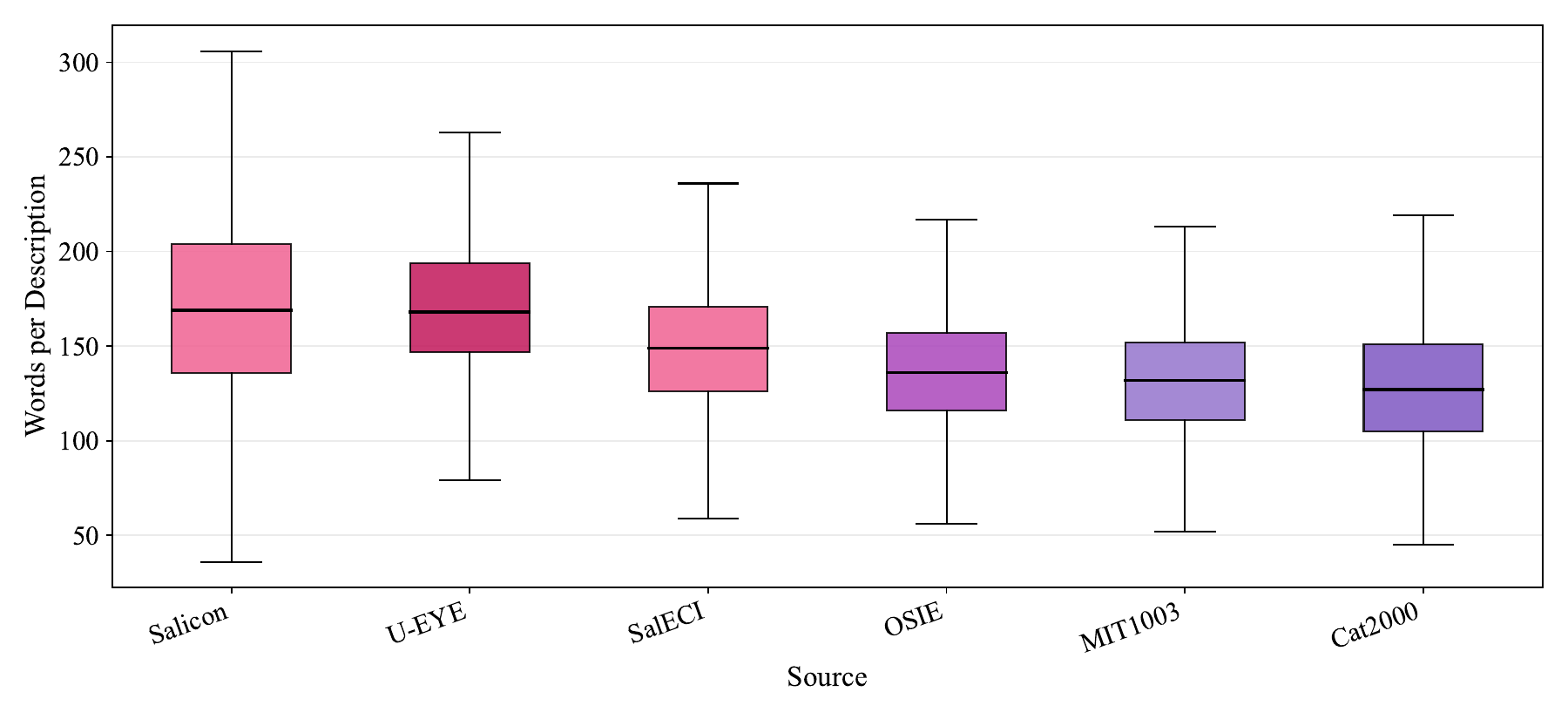}%
    }
    \caption{\textbf{Distribution of explanation length across datasets.} Boxplots show words per description for each source (median, interquartile range, and whiskers). Length statistics are broadly comparable across datasets, with small shifts in the upper tail for some datasets.}
    \label{fig:desc_length_boxplot}
\end{figure}

\autoref{fig:desc_length_boxplot} reports the distribution of explanation length (words per description) for each dataset. Overall, the datasets exhibit broadly comparable length statistics: the medians lie in a similar range and the interquartile ranges overlap substantially, indicating that cross-domain comparisons are not driven by large differences in verbosity.

\section{Implementation Details}

\subsection{Loss Functions}
\label{sec:supp_loss}
We optimize the composite objective by minimizing dissimilarity terms (KL, MSE) and maximizing similarity terms (CC, SIM, NSS) via 
\begin{equation}
\footnotesize
\begin{aligned}
\mathcal{L}_{\text{sal}} &=
\lambda_{1}\,\mathcal{L}_{\text{KL}}(S^{g},\hat{S})
-\lambda_{2}\,\mathcal{L}_{\text{CC}}(S^{g},\hat{S})
-\lambda_{3}\,\mathcal{L}_{\text{SIM}}(S^{g},\hat{S}) \\
&\quad-\lambda_{4}\,\mathcal{L}_{\text{NSS}}(F^{g},\hat{S})
+\lambda_{5}\,\mathcal{L}_{\text{MSE}}(S^{g},\hat{S}) \, .
\end{aligned}
\label{eq:supp_sal_loss}
\end{equation}
We summarize the saliency loss terms used in OpenVAM and describe the role of each component below.

\paragraph{\textbf{KL divergence (minimize):}} This term enforces distributional alignment between the predicted and ground-truth saliency mass, strongly penalizing missing probability mass in regions where $S^{g}$ is large. We use $\epsilon=2.2\times 10^{-16}$ for numerical stability.
\[
\mathcal{L}_{\text{KL}}(S^{g},\hat{S})
=\sum_{i} S^{g}_{i}\log\!\left(\epsilon+\frac{S^{g}_{i}}{\hat{S}_{i}+\epsilon}\right).
\]

\paragraph{\textbf{Correlation coefficient, CC (maximize):}} CC encourages global structural agreement between maps (shape-level consistency) and is relatively insensitive to affine rescaling of $\hat{S}$.
\[
\text{CC}(S^{g},\hat{S})
=\frac{\mathrm{cov}(S^{g},\hat{S})}{\sigma(S^{g})\,\sigma(\hat{S})}.
\]

\paragraph{\textbf{Histogram intersection, SIM (maximize):}} SIM rewards overlap of saliency mass and encourages a correct spread of probability mass; it complements KL and is typically robust to small spatial shifts.
\[
\text{SIM}(S^{g},\hat{S})=\sum_{i}\min(S^{g}_{i},\hat{S}_{i}).
\]

\paragraph{\textbf{Normalized Scanpath Saliency, NSS (maximize).}} NSS directly enforces high predicted saliency at fixation locations and is invariant to affine rescaling of $\hat{S}$ due to the z-score normalization.
\[
\text{NSS}(F^{g},\hat{S})
=\frac{1}{\sum_{i}F^{g}_{i}}
\sum_{i}\left(\frac{\hat{S}_{i}-\mu(\hat{S})}{\sigma(\hat{S})}\right)F^{g}_{i}.
\]

\paragraph{\textbf{Mean squared error, MSE (minimize).}} MSE provides a dense, local penalty that stabilizes optimization and discourages large per-pixel deviations.
\[
\mathcal{L}_{\text{MSE}}(S^{g},\hat{S})
=\frac{1}{n}\sum_{i}\left(\hat{S}_{i}-S^{g}_{i}\right)^{2}.
\]

\subsection{Experimental Settings}
\label{sec:supp_exp_settings}

All OpenVAM variants are trained using the same three-stage pipeline. 
Stage~I focuses on saliency-only training to learn robust spatial attention representations. Stage~II further optimizes the saliency objective while leveraging pretrained vision–language initialization. Finally, Stage~III introduces text supervision and adapts the language model via LoRA~\cite{hu2022lora}.  The complete training hyperparameters for each model and stage are reported in~\autoref{tab:openvam_hparams}.

\begin{table*}[t]
\centering
\caption{\textbf{Training hyperparameters for OpenVAM variants across stages.}}
\vspace{-0.4em}
\label{tab:openvam_hparams}
\resizebox{\textwidth}{!}{
\begin{tabular}{l | l | c c c c c c | c c c c c c c | c}
\toprule
\textbf{Model} & \textbf{Stage} & \textbf{Batch} & \textbf{Accum.} & \textbf{LR} & \textbf{Warmup} & \textbf{Epochs} & \textbf{Early Stop} & $\lambda_{\text{sal}}$ & $\lambda_{\text{text}}$ & $\lambda_{\text{MSE}}$ & $\lambda_{\text{KLD}}$ & $\lambda_{\text{CC}}$ & $\lambda_{\text{SIM}}$ & $\lambda_{\text{NSS}}$ & \textbf{LoRA} \\
\midrule

OpenVAM & Stage~I & 8 & -- & $7.4\times10^{-5}$ & -- & 30 & 4 & -- & -- & 2.99 & 12 & 2.15 & 1.69 & 3.17 & -- \\

\midrule

\multirow{2}{*}{OpenVAM-3B}
& Stage~II & 8 & 4 & $5\times10^{-5}$ & 3 & 50 & 5 & -- & -- & 0.1 & 1.0 & 1.2 & 0.8 & 0.3 & -- \\
& Stage~III & 4 & 4 & $1\times10^{-4}$ & 3 & 25 & 3 & 0.05 & 3.0 & 0.1 & 0.5 & 1.0 & 0.5 & 0.1 &  $r{=}16,\alpha{=}32,p{=}0.05$ \\

\midrule

\multirow{2}{*}{OpenVAM-4B}
& Stage~II & 8 & 4 & $5\times10^{-5}$ & 1 & 50 & 5 & -- & -- & 0.1 & 1.0 & 1.2 & 0.8 & 0.3 & -- \\
& Stage~III & 4 & 4 & $1\times10^{-4}$ & 3 & 25 & 3 & 0.05 & 3.0 & 0.0 & 0.5 & 1.0 & 0.5 & 0.1 & $r{=}16,\alpha{=}32,p{=}0.05$ \\

\midrule

\multirow{2}{*}{OpenVAM-7B}
& Stage~II & 4 & 8 & $5\times10^{-5}$ & 3 & 50 & 5 & -- & -- & 0.1 & 1.0 & 1.3 & 0.8 & 0.3 & -- \\
& Stage~III & 2 & 4 & $1\times10^{-4}$ & 3 & 25 & 5 & 0.05 & 3.0 & 0.05 & 0.5 & 1.0 & 0.5 & 0.2 & $r{=}16,\alpha{=}32,p{=}0.05$ \\

\midrule

\multirow{2}{*}{OpenVAM-8B}
& Stage~II & 8 & 4 & $5\times10^{-5}$ & 3 & 50 & 5 & -- & -- & 0.1 & 1.0 & 1.3 & 0.9 & 0.3 & -- \\
& Stage~III & 4 & 4 & $1\times10^{-4}$ & 3 & 25 & 3 & 0.05 & 3.0 & 0.05 & 0.5 & 1.0 & 0.5 & 0.2 & $r{=}16,\alpha{=}32,p{=}0.05$ \\

\bottomrule
\end{tabular}
}
\vspace{-0.6em}
\end{table*}

\paragraph{\textbf{Selection of Stage~III loss weights.}}
The Stage~III objective uses $\lambda_{\mathrm{sal}}=0.05$ and $\lambda_{\mathrm{text}}=3.0$ for all OpenVAM variants, as reported in \autoref{tab:openvam_hparams}. These values were selected via grid search with OpenVAM-3B over $\lambda_{\mathrm{sal}}\in[0.01,0.10]$ with step size $0.01$ and $\lambda_{\mathrm{text}}\in[1.0,5.0]$ with step size $1.0$, using a validation subset carved out from the SALICON training split. This subset is disjoint from the test split used for all reported results and is used only for selecting the global Stage~III loss balance; it is not used for any reported metric. The selected setting provided the best trade-off between preserving saliency-map quality and improving rationale generation: larger saliency weights over-constrained the visual adapter and limited language adaptation, while smaller saliency weights yielded weaker spatial grounding. Conversely, larger text weights improved language likelihood but reduced alignment with the dense saliency pathway. We deliberately fix the same weights across all model scales and domains rather than tuning them per configuration, which reduces the risk of overfitting the loss balance to a specific dataset, model size, or evaluation setting.

\subsection{VLM as a Judge (JScore)}
\label{sec:supp_jscore}

To evaluate the semantic quality of generated saliency explanations, we employ GPT-4.1~\cite{achiam2023gpt} as the judge model. While lexical metrics such as BLEU and ROUGE measure surface-level similarity between predicted and reference text, they often fail to capture whether the explanation correctly reasons about visual saliency. To address this limitation, we introduce JScore, a semantic evaluation score that measures how well the predicted reasoning aligns with the ground-truth explanation. The judge receives both the ground-truth reasoning and the predicted reasoning and produces a single score in the range $[0,1]$. The evaluation equally considers several aspects: correctness of the identified salient objects or regions, quality of explanations describing why those regions attract visual attention, consistency with the scene implied by the ground truth, and the absence of hallucinated objects or implausible claims. Importantly, the evaluation is tolerant to wording differences and reasonable variations in object naming, color description, or spatial references. Predictions that clearly explain saliency using multiple visual factors such as contrast, brightness, color, position, size, uniqueness, or foreground--background separation receive higher scores, while explanations that merely list objects without reasoning or contain hallucinations receive lower scores. The prompt used to guide the VLM judge is shown in \autoref{box:prompt_jscore}.

\begin{promptbox}[
      colback=blue!6!white,
      colframe=blue!55!black,
      colbacktitle=blue!70!black,
      boxed title style={colback=blue!70!black}
    ]{JScore Prompt}{jscore}
    \label{box:prompt_jscore}
    \scriptsize
    \textbf{You are evaluating saliency prediction reasoning.} The task is to determine how well a predicted explanation identifies and explains visually salient objects or regions in an image.
    
    The ground truth reasoning was generated by analyzing the actual image and saliency mask, so it accurately reflects what objects exist and why they are visually salient.
    
    \medskip
    \textbf{Your goal.} Produce one overall score between \textbf{0.0 and 1.0} that evaluates the overall quality of the predicted reasoning compared to the ground truth.
    
    \medskip
    \textbf{Ground Truth Reasoning:}
    
    \texttt{\{ground\_truth\}}
    
    \medskip
    \textbf{Predicted Reasoning:}
    
    \texttt{\{prediction\}}
    
    \medskip
    \textbf{Evaluate the predicted reasoning based on the following aspects (weighted equally):}
    
    \begin{itemize}
    \item Accuracy of reasoning about salient objects or regions
    \item Quality and correctness of explanations for why something is visually salient
    \item Consistency with the scene and objects implied by the ground truth
    \item Use of valid visual saliency principles (contrast, color, brightness, position, size, uniqueness, motion, texture, foreground/background separation)
    \item Whether the reasoning focuses on explaining saliency rather than only listing objects
    \item Plausibility of the described objects within the scene context
    \item Absence of hallucinations (objects or properties inconsistent with the scene)
    \end{itemize}
    
    Be lenient about wording differences and reasonable variations in object naming or spatial descriptions.
    
    \medskip
    \textbf{Scoring Guide}
    
    \begin{itemize}
    \item \textbf{1.00}: Perfect reasoning with correct objects, locations, and strong saliency explanations.
    \item \textbf{0.90–0.99}: Excellent reasoning with minor wording differences.
    \item \textbf{0.80–0.89}: Very strong reasoning with small object or location differences.
    \item \textbf{0.70–0.79}: Good reasoning with valid saliency explanations but less detail.
    \item \textbf{0.60–0.69}: Reasonable reasoning with limited depth.
    \item \textbf{0.50–0.59}: Moderate quality with partial inconsistencies.
    \item \textbf{0.40–0.49}: Weak reasoning with noticeable inconsistencies.
    \item \textbf{0.30–0.39}: Poor reasoning with weak saliency explanations.
    \item \textbf{0.20–0.29}: Very poor reasoning with major inconsistencies.
    \item \textbf{0.10–0.19}: Severely flawed reasoning with multiple hallucinations.
    \item \textbf{0.00–0.09}: Completely incorrect or irrelevant reasoning.
    \end{itemize}
    
    \medskip
    \textbf{Return ONLY a single number between 0.0 and 1.0 representing the overall evaluation score.}
\end{promptbox}


\section{Additional Experimental Results}
\subsection{Unseen-Dataset Generalization}
To evaluate cross-dataset generalization, we test OpenVAM on four \emph{unseen} datasets that are not used during training: \texttt{Toronto}~\cite{bruce2007attention}, \texttt{TUD Database 1}~\cite{liu2009studying}, \texttt{TUD Database 2}~\cite{alers2010studying}, and \texttt{FIWI}~\cite{shen2014webpage}. As summarized in~\autoref{tab:dataset_summary}, these datasets span both natural scenes and web pages. Toronto and the two TUD benchmarks evaluate generalization to natural-scene eye-tracking data under different scene characteristics and dataset scales, while FIWI measures transfer to web-page saliency, which differs notably in layout, semantics, and viewing behavior. This setup provides a strong test of whether the learned saliency predictor can remain robust under domain shift beyond the distributions seen during training.

\autoref{tab:saliency_unseen} reports the quantitative results and compares OpenVAM against SUM~\cite{hosseini2025sum}, a strong general saliency model. Overall, OpenVAM shows superior generalization across the unseen benchmarks, with the 7B variant delivering the most consistent performance. On \texttt{Toronto}, OpenVAM-7B outperforms SUM on all reported metrics. A similar trend is observed on \texttt{TUD Database 2}, where OpenVAM-7B again surpasses SUM across all available metrics, indicating strong robustness to previously unseen natural-scene distributions. On \texttt{TUD Database 1}, OpenVAM-7B also performs best overall, achieving higher CC and SIM and lower KLD than SUM, whereas OpenVAM-3B is slightly weaker on this dataset. This suggests that the larger model provides a more stable representation under distribution shift. On \texttt{FIWI}, which is particularly challenging due to its webpage structure and different visual attention patterns, OpenVAM-7B remains better than SUM overall. This result is encouraging because it shows that OpenVAM generalizes not only across unseen natural-image datasets but also to webpage layouts that differ substantially from standard saliency benchmarks. 

\begin{table}[t]
\centering
\caption{\textbf{Summary of unseen evaluation datasets used for generalization.}}
\label{tab:dataset_summary}
\resizebox{\columnwidth}{!}{%
\setlength{\tabcolsep}{15pt}
\begin{tabular}{lcccc}
\toprule
\textbf{Dataset} & \textbf{Image Domain} & \textbf{Acquisition Type} & \textbf{Image Resolution} & \textbf{\# Test Samples} \\
\midrule
\textit{Toronto}~\cite{bruce2007attention} & Natural scene & Eye & $681 \times 511$ & 120 \\
\textit{TUD Database 1}~\cite{liu2009studying} & Natural scene & Eye & $768 \times 512$ & 29 \\
\textit{TUD Database 2}~\cite{alers2010studying}  & Natural scene & Eye & $600 \times 600$ & 160 \\
\textit{FIWI}~\cite{shen2014webpage} & Web page & Eye & $1360 \times 768$ & 149 \\
\bottomrule
\end{tabular}%
}
\end{table}

\subsection{More Qualitative Results}
\label{sec:supp_more_qualitative}
\begin{figure*}[htb]
    \centering
    \newcommand{\cropheight}{0.22\textwidth}
    \resizebox{\textwidth}{!}{%
    \begin{tabular}{@{} c *{4}{c} @{}}

    \rotatebox{90}{Sample 1} &
    \includegraphics[height=\cropheight]{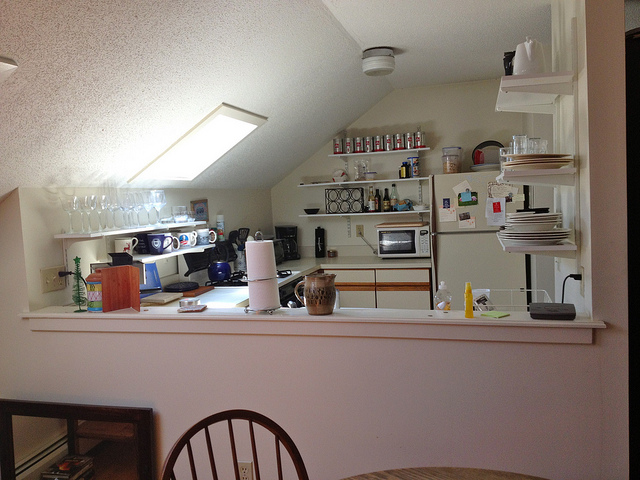} &
    \includegraphics[height=\cropheight]{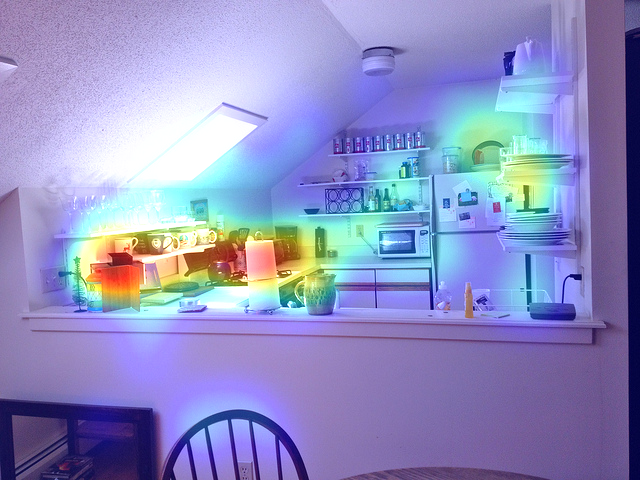} &
    \includegraphics[height=\cropheight]{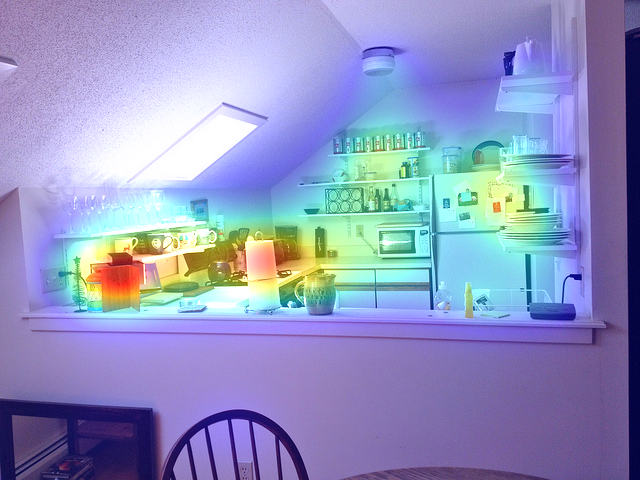} &
    \includegraphics[height=\cropheight]{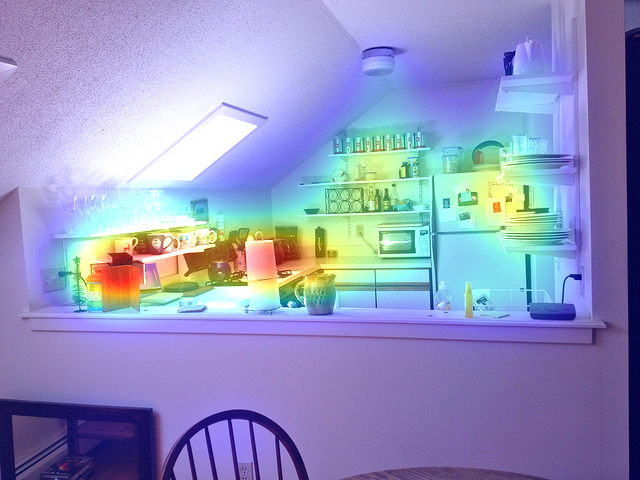} \\

    \rotatebox{90}{Sample 2} &
    \includegraphics[height=\cropheight]{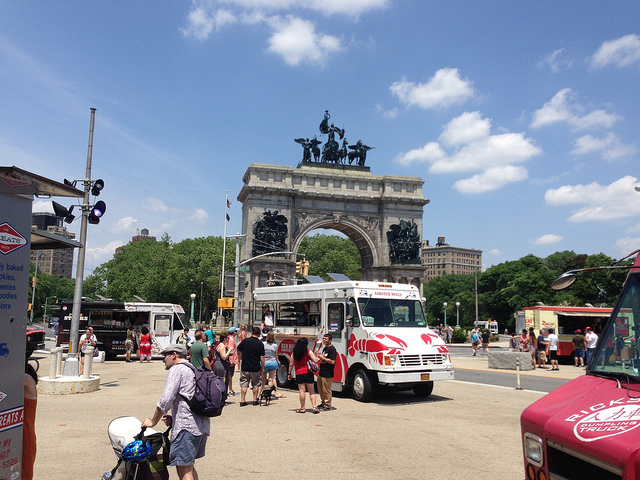} &
    \includegraphics[height=\cropheight]{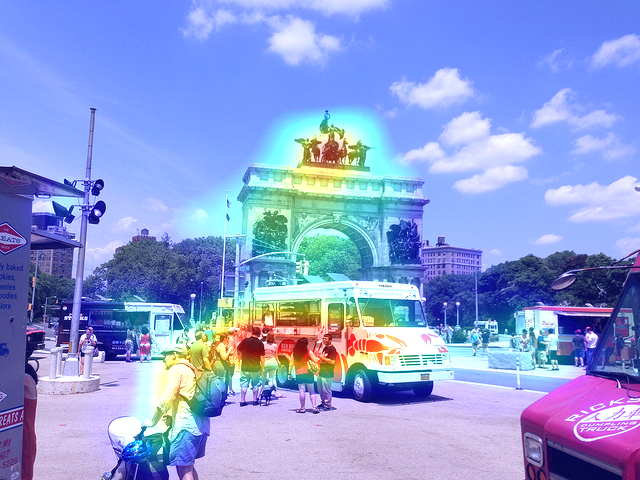} &
    \includegraphics[height=\cropheight]{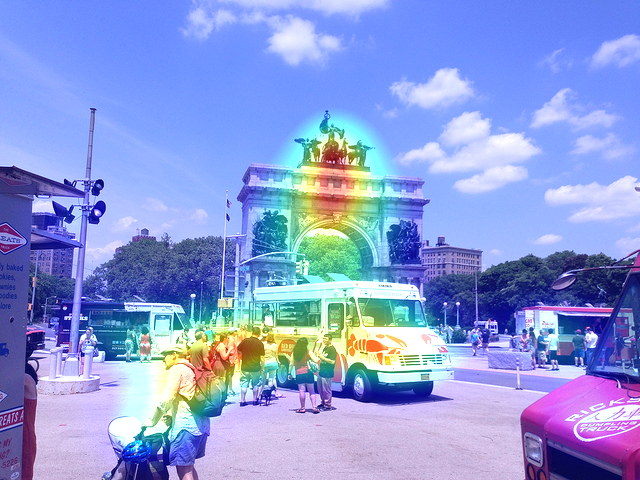} &
    \includegraphics[height=\cropheight]{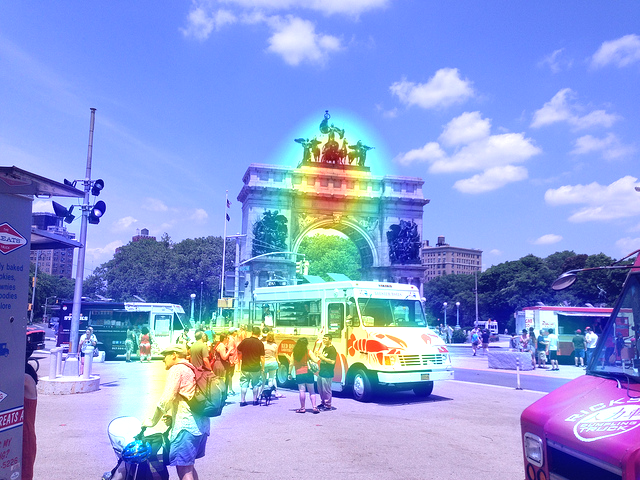} \\

    \rotatebox{90}{Sample 3} &
    \includegraphics[height=\cropheight]{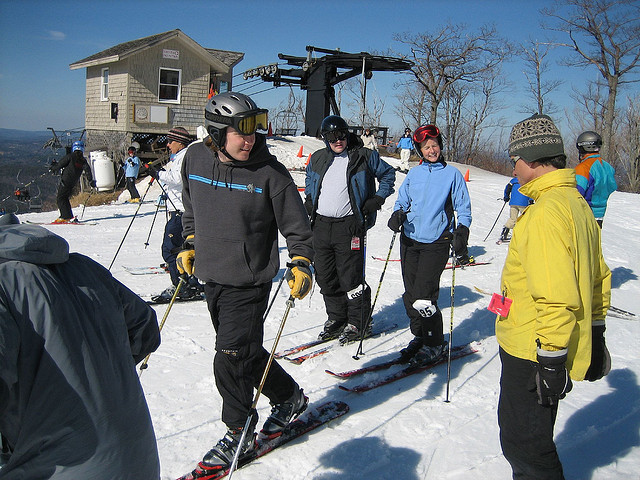} &
    \includegraphics[height=\cropheight]{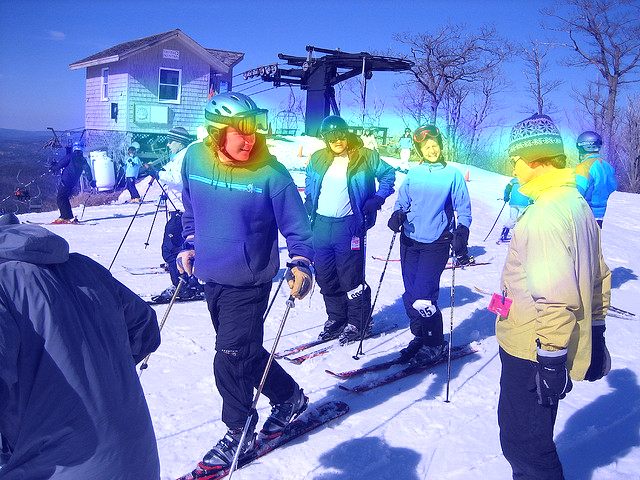} &
    \includegraphics[height=\cropheight]{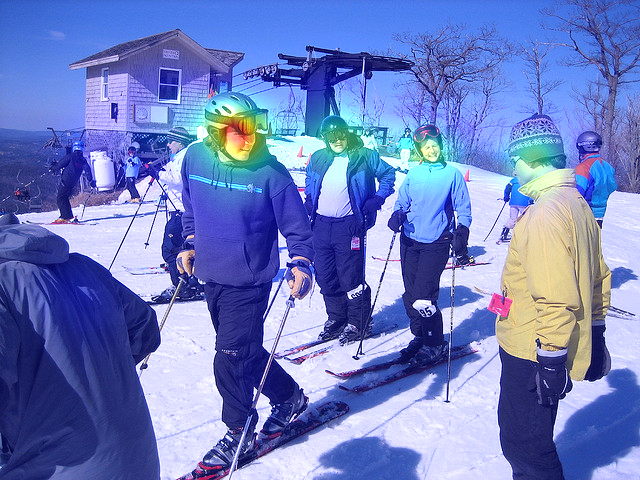} &
    \includegraphics[height=\cropheight]{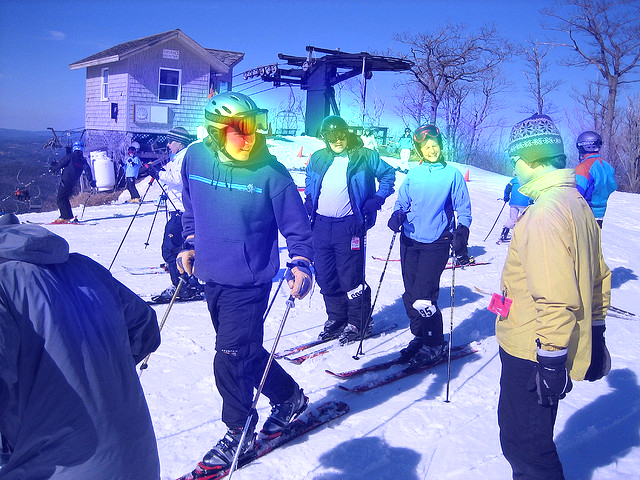} \\

    \rotatebox{90}{Sample 4} &
    \includegraphics[height=\cropheight]{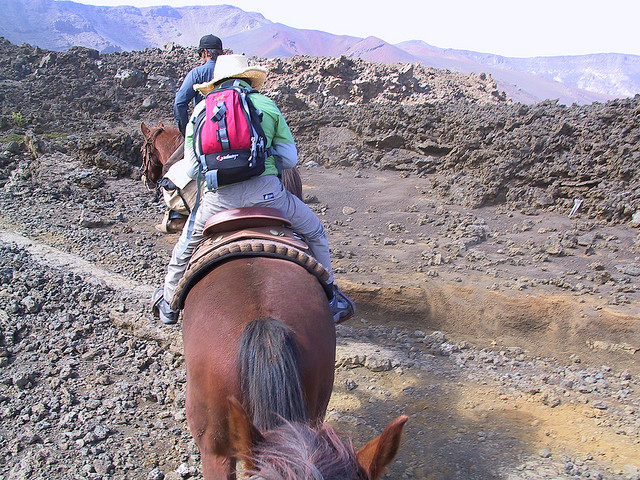} &
    \includegraphics[height=\cropheight]{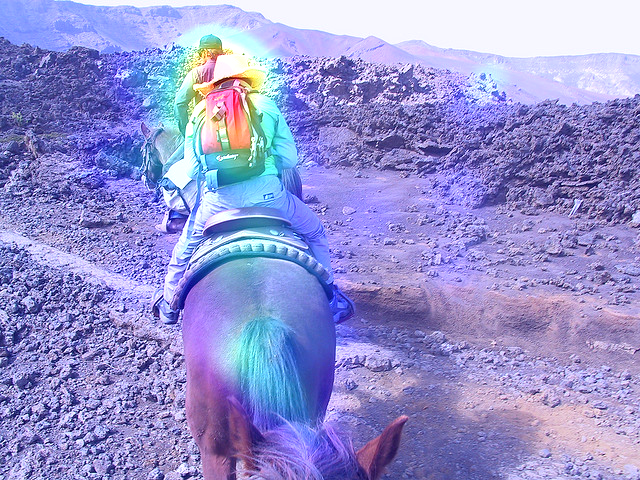} &
    \includegraphics[height=\cropheight]{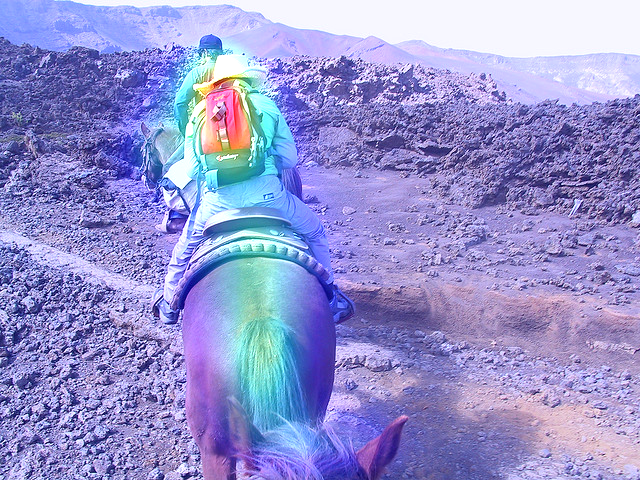} &
    \includegraphics[height=\cropheight]{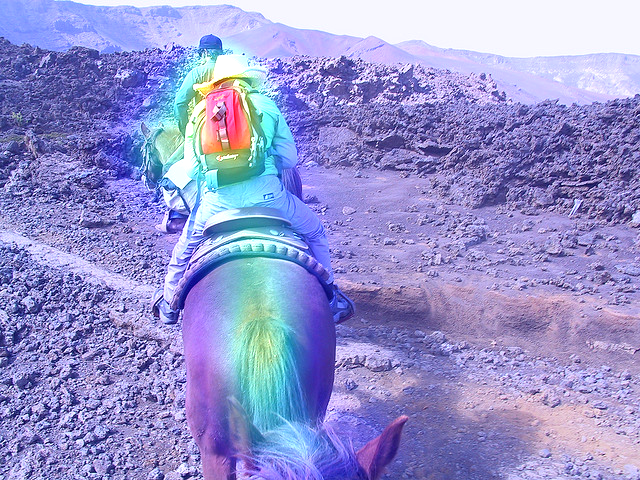} \\

    \rotatebox{90}{Sample 5} &
    \includegraphics[height=\cropheight]{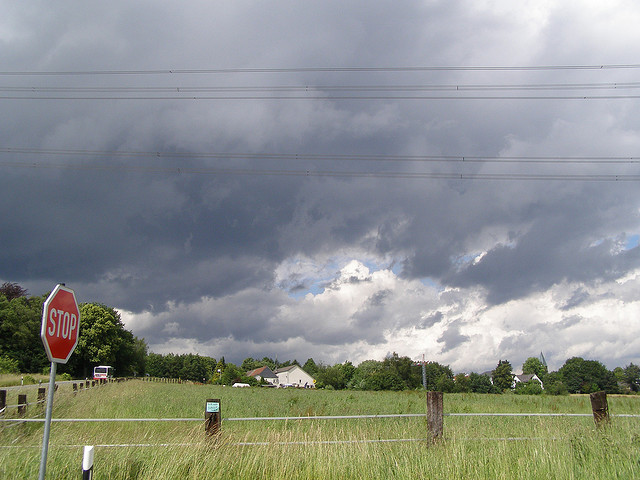} &
    \includegraphics[height=\cropheight]{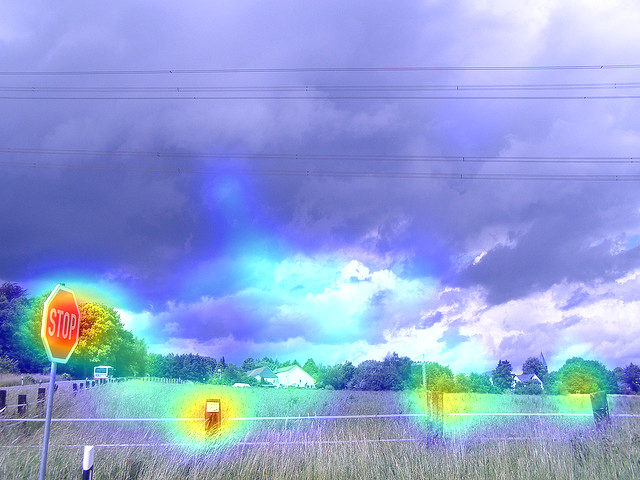} &
    \includegraphics[height=\cropheight]{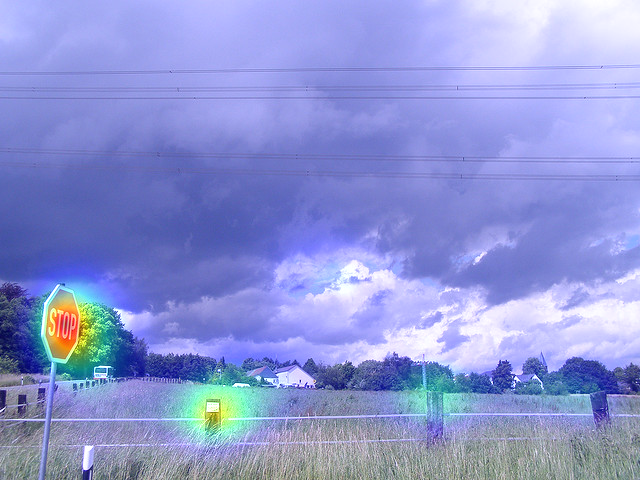} &
    \includegraphics[height=\cropheight]{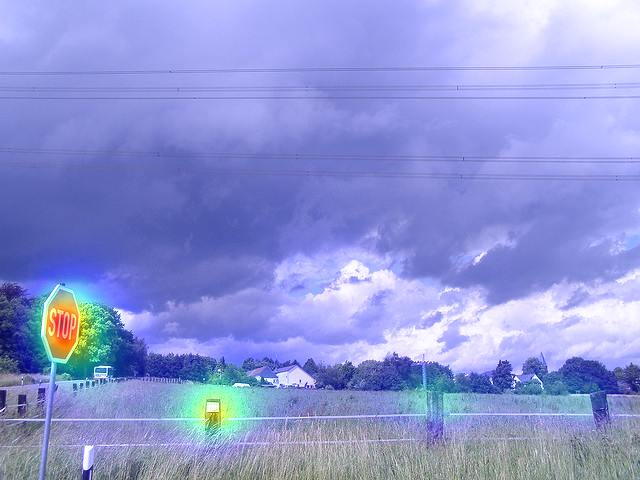} \\

    \rotatebox{90}{Sample 6} &
    \includegraphics[height=\cropheight]{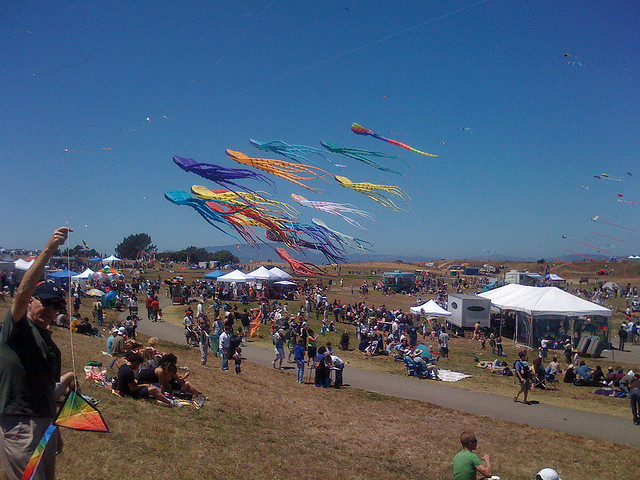} &
    \includegraphics[height=\cropheight]{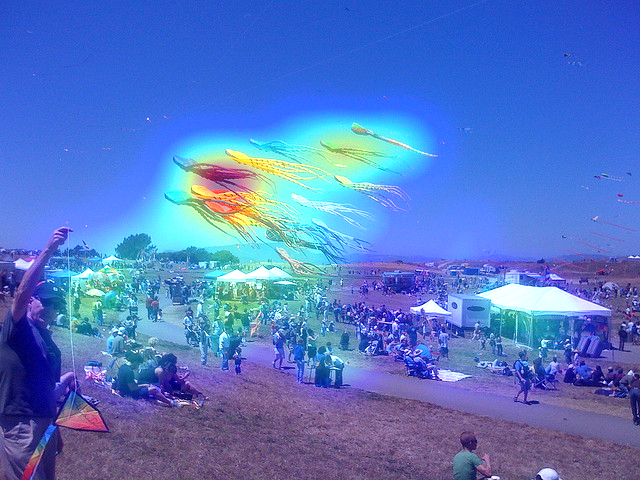} &
    \includegraphics[height=\cropheight]{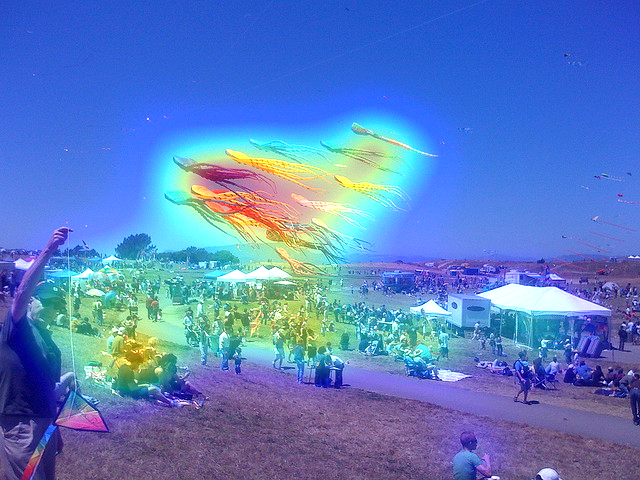} &
    \includegraphics[height=\cropheight]{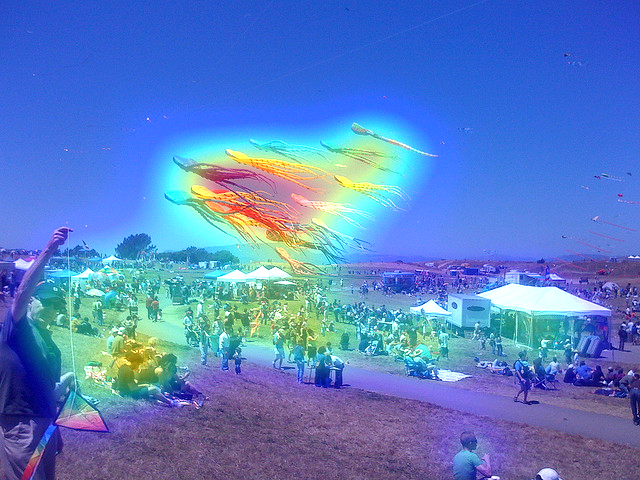} \\

    & Input Image & Ground Truth & OpenVAM-4B & OpenVAM-8B \\

    \end{tabular}
    }
    \caption{\textbf{Visualizations of OpenVAM predictions across different natural images.}}
    \label{fig:nature_vis}
\end{figure*}

\autoref{fig:nature_vis} and~\ref{fig:ui_vis}, together with the qualitative comparisons in~\autoref{tab:openvam_text_samples_nature1}, \ref{tab:openvam_text_samples_nature2}, and~\ref{tab:openvam_text_samples_ecom}, provide a detailed view of OpenVAM across natural, commercial, and UI/webpage images. Overall, both OpenVAM-4B and OpenVAM-8B recover the main human-attended regions with good spatial alignment, while also generating short explanations that are usually grounded in visually prominent objects, text blocks, and layout structure.

On natural images, OpenVAM generally captures the dominant semantic entities and scene anchors, such as people, vehicles, animals, ski slopes, buildings, and large foreground objects. In many cases, the predicted saliency maps align well with the ground truth over the principal attended regions, especially when attention is concentrated on a small number of semantically meaningful elements. The text outputs are also often reasonable at a high level, correctly identifying core objects such as skiers, horses, stop signs, kites, and architectural landmarks. For commercial and UI/webpage samples, OpenVAM shows strong sensitivity to the most visually prominent marketing and interface elements, including prices, discount badges, promotional banners, product images, and primary content blocks. This behavior is particularly visible in the saliency overlays, where both models often highlight large text and central product regions in a way that closely matches the human attention patterns. The generated rationales also reflect the intended domain bias, frequently prioritizing price information, promotional copy, and key product visuals in e-commerce images, and major interface components in UI examples.

\begin{table*}[t]
    \centering
    \caption{\textbf{Qualitative text comparison of~\autoref{fig:nature_vis}.} Generated explanations from OpenVAM models versus ground-truth text. Samples~1--3 correspond to rows~1--3, respectively.}
    \label{tab:openvam_text_samples_nature1}
    \resizebox{1\textwidth}{!}{
    \setlength{\tabcolsep}{1pt}
    \tiny
    \begin{tabular}{c|p{0.35\textwidth}|p{0.35\textwidth}|p{0.35\textwidth}}
    \toprule
    \textbf{Sample} & \textbf{OpenVAM-4B} & \textbf{OpenVAM-8B} & \textbf{Ground Truth} \\
    \midrule
    \rowcolor{blue!6}
    Sample 1 &
    \textbf{White kitchen cabinets (upper-right):} The cabinets are large and occupy a significant portion of the upper-right area, drawing the eye due to their size and placement. Their white color contrasts with the darker elements in the kitchen, making them stand out. \textbf{White countertop (center):} The countertop is centrally located and spans a large area, making it a focal point. Its white color contrasts with the darker elements in the kitchen, such as the sink and appliances, drawing attention to it. \textbf{White sink (left):} The sink is a distinct element on the left side of the countertop, contrasting with the countertop's color. Its placement on the edge of the countertop makes it noticeable. \textbf{White refrigerator (right):} The refrigerator is a large, white appliance that stands out against the darker background. Its size and placement on the right side of the image make it a prominent feature. \textbf{White microwave (above the sink):} The microwave is positioned above the sink, making it a noticeable element in the kitchen. Its white color contrasts with the darker elements in the kitchen, drawing attention to it. \textbf{White oven (right):} The oven is a large, white appliance that stands out against the darker background. Its size and placement on the right side of the image make it a prominent feature. &
    \textbf{Black chair (foreground):} Its dark color contrasts with the lighter floor and wall, drawing the eye; its placement in the foreground makes it a prominent element in the composition. \textbf{Kitchen area (center):} The kitchen area is centrally located and contains multiple objects, making it a focal point; the contrast between the light-colored cabinets and the darker appliances also draws attention. \textbf{White cabinets (kitchen):} The cabinets are large and take up a significant portion of the kitchen area, making them visually prominent; their light color contrasts with the darker appliances and countertop. \textbf{Black appliances (kitchen):} The black appliances stand out against the lighter cabinets and countertop, creating a visual contrast; their dark color makes them noticeable in the kitchen setting. \textbf{Light-colored floor (kitchen):} The light color of the floor provides a neutral backdrop that allows the other elements in the kitchen to stand out; its large, open space also draws the eye. \textbf{Light-colored wall (kitchen):} The light color of the wall provides a neutral backdrop that allows the other elements in the kitchen to stand out; its large, open space also draws the eye. &
    \textbf{Kitchen (overall composition):} The kitchen as a whole is salient due to its position as the main subject of the image and its function as a hub of activity, drawing the viewer's attention. \textbf{Breakfast bar (lower edge):} The breakfast bar is salient because it is positioned along the lower edge and provides a visual boundary. \textbf{Cabinets and open shelves (upper-right):} These elements draw attention because they fill the upper-right area, organizing items and adding visual interest. \textbf{Refrigerator (center-right):} The refrigerator is salient due to its central placement and its typical role as a key appliance in a kitchen. \textbf{Angled ceiling with skylight (upper-left):} The angled ceiling and skylight are salient because they create an unusual shape and introduce natural light, drawing the eye.\\
    
    \midrule

    \rowcolor{orange!6}
    Sample 2 &
    \textbf{Taxis (foreground):} The bright yellow color of the taxis contrasts with the surrounding environment, making them stand out; their placement in the foreground ensures they are one of the first things a viewer notices. \textbf{People (foreground):} The presence of people adds a human element to the scene, drawing attention to the activity and life in the city; their movement and interaction with the taxis further enhance their salience. \textbf{Buildings (background):} The buildings provide context and depth to the scene, framing the clock tower and taxis; their size and architectural style contribute to the overall visual interest. &
    \textbf{Large, ornate building (center):} Its central placement and size immediately draw the eye; the intricate details and unique architectural style make it stand out against the simpler background. \textbf{People (foreground):} The presence of people in the foreground creates a sense of scale and activity, drawing the viewer's attention to the immediate surroundings. \textbf{Cars (foreground):} The cars are salient because they are in the foreground and are moving, which creates a sense of motion and activity. \textbf{Sky (background):} The bright blue sky provides a contrasting backdrop that highlights the other elements in the scene. &
    \textbf{Food Trucks (center):} Their bright colors and unique shapes contrast with the surrounding environment, drawing the eye; their central placement in front of the arch makes them a focal point of the scene. \textbf{Grand Army Plaza Arch (center):} Its large size and ornate stone construction make it a prominent feature; its central placement anchors the composition and attracts visual attention. \textbf{People (center):} Their presence milling about the trucks indicates activity and life, drawing the eye to the area where the action is occurring; their contrast against the background helps them stand out. \textbf{Bright Blue Sky (background):} The expansive blue sky provides a bright and contrasting backdrop; the presence of scattered clouds adds visual interest and depth. \textbf{Trees (surrounding):} They frame the scene, providing a natural border that enhances the overall composition; their green color contrasts with the stone of the arch. \\

    \midrule

    \rowcolor{purple!6}
    Sample 3 &
    \textbf{Skiers (center foreground):} The skiers are centrally located and in the foreground, immediately drawing the viewer's attention. Their presence suggests activity and human interest in the scene. \textbf{Snowy slope (visible):} The snow-covered slope is a large, bright area that contrasts with the darker elements in the background, making it visually prominent. The texture and expanse of the snow also contribute to its salience. \textbf{Ski lift (background):} The ski lift is a large, man-made structure that stands out against the natural landscape. Its height and position in the background create a sense of depth and scale. \textbf{Buildings (background):} The buildings are visible in the background, providing context and indicating a developed area. Their presence adds to the overall scene and helps to define the location. \textbf{Sky (visible):} The sky is a large, bright area that contrasts with the darker elements in the foreground and background. Its color and texture contribute to the overall visual appeal of the image.
    &
    \textbf{Skiers (center):} The skiers are centrally located and are the main subjects of the image, drawing the viewer's attention to their activity and attire. Their presence in the foreground also makes them more prominent. \textbf{Snow-covered slope (visible):} The snow-covered slope provides a bright, contrasting background that highlights the skiers and their equipment. The texture and expanse of the snow also contribute to the overall visual interest. \textbf{Buildings (background):} The buildings in the background add depth to the image and provide context for the location. Their size and placement in the distance make them noticeable but not overpowering. \textbf{Skis and poles (visible):} The skis and poles are essential elements of the skiers' gear and are visually distinct due to their shape and arrangement. They contribute to the overall theme of skiing and are important for understanding the activity. \textbf{Blue jacket (visible):} The blue jacket stands out due to its color contrast against the white snow and the other skiers' clothing. Its placement on the right side of the image also draws the eye. \textbf{White jacket (visible):} The white jacket is salient due to its bright color, which contrasts with the surrounding environment. Its placement on the left side of the image also helps to balance the composition. \textbf{Black jacket (visible):} The black jacket is salient due to its dark color, which contrasts with the white snow and the other skiers' clothing. Its placement on the left side of the image also helps to balance the composition.
    &
    \textbf{Skiers (center and right):} Skiers stand out because they are centrally located in the image and contribute to the action. Their presence indicates the setting and activity of the scene. \textbf{Ski lift (upper-center background):} The ski lift draws attention because it is relatively large and unique to a ski environment. Its placement in the background helps establish the setting. \textbf{Snowy slope (foreground and midground):} The snowy slope is salient due to its extensive coverage and contrast with the trees and building. Its whiteness makes it a dominant feature of the landscape. \textbf{Trees (scattered in background):} Trees contribute to the depth and natural context of the image. Their darker color contrasts with the snow. \textbf{Small, light-colored building (upper-left background):} The building draws attention due to its distinct shape and placement in the background. It serves as a focal point and helps orient the viewer. \textbf{Colorful jackets and ski pants (worn by most skiers):} The bright colors create contrast against the white snow. The variety of colors on multiple subjects draws the eye to the human activity. \textbf{Skier in a yellow jacket (center-right foreground):} The yellow jacket is highly salient due to its brightness. The skier's forward-facing position directs the viewer's gaze and creates a focal point. \\

    \bottomrule
    \end{tabular}
    }
\end{table*}


\begin{table*}[h]
    \centering
    \caption{\textbf{Qualitative Text Comparison~\autoref{fig:nature_vis}.} Generated explanations from OpenVAM models versus ground-truth text. Samples~4--6 correspond to rows~4--6, respectively.}
    \label{tab:openvam_text_samples_nature2}
    \resizebox{\textwidth}{!}{
    \setlength{\tabcolsep}{1pt}
    \tiny
    \begin{tabular}{c|p{0.35\textwidth}|p{0.35\textwidth}|p{0.35\textwidth}}
    \toprule
    \textbf{Sample} & \textbf{OpenVAM-4B} & \textbf{OpenVAM-8B} & \textbf{Ground Truth} \\
    \midrule
    \rowcolor{blue!6}
    Sample 4 &
    \textbf{Horse (center):} The horse is centrally located, immediately drawing the viewer's attention. Its size and the fact that it is the main subject of the image further enhance its salience. \textbf{Rider (on horse):} The rider is salient because they are the only human figure in the image, creating a focal point. The contrast of the rider's clothing against the horse's coat also makes them stand out. &
    \textbf{Horse (center):} The horse is centrally located and takes up a significant portion of the frame, immediately drawing the eye. Its size and the contrast of its brown coat against the green grass and the background also contribute to its salience. \textbf{Person (on horse):} The person is salient because they are the only human figure in the image, and their presence on the horse creates a focal point. The contrast of their clothing against the horse's coat also makes them stand out. \textbf{Horse's head (center):} The horse's head is salient due to its proximity to the viewer and its placement in the lower-left corner, which is a common starting point for visual scanning. The head is also a natural focal point for observing animals. \textbf{Horse's body (center):} The horse's body is salient because it is the main subject of the image and occupies a large portion of the frame. The contrast of its brown coat against the green grass and the background also makes it stand out. \textbf{Horse's tail (right):} The horse's tail is salient because it is a distinct feature of the horse and is positioned on the right side of the frame, balancing the composition. The tail's movement and the contrast of its color against the background also draw attention. &
    \textbf{A rider wearing a red backpack (center-left midground):} The bright red backpack stands out against the earthy tones of the surrounding landscape and the horse, drawing immediate attention. The central-left midground placement further ensures it captures the viewer's focus. \textbf{Another rider (center-right midground):} Being in the midground places them within the viewer's initial field of attention, contributing to their salience. \textbf{Brown horse (under the rider):} Due to their relatively large size and central placement, the brown horse is a prominent subject in the image. \textbf{Dusty trail (winding upward):} The trail creates a visible line and a pathway for the eye to follow, making it a salient element. \textbf{Rocky landscape (surrounding the trail):} The textured and varied terrain contrasts with the smoother sky and path, making it a visually distinct and salient area. \textbf{Distant hazy mountains (backdrop):} Their significant size and placement at the top of the image, though softened by haze, still make them a noticeable backdrop.\\
    
    \midrule

    \rowcolor{orange!6}
    Sample 5 &
\textbf{Red and white sign (left edge):} The sign's bright red and white colors contrast sharply with the surrounding green grass and blue sky, making it immediately noticeable. Its placement on the left edge also draws the eye, as it is one of the first elements encountered in the image. \textbf{Light-colored building (center):} The building's size and central placement make it a dominant feature in the image. Its light color contrasts with the darker trees and green grass, further enhancing its salience. \textbf{Green grass (foreground):} The grass covers a large portion of the foreground, providing a base for the other elements. Its uniform color and texture create a visual foundation that draws the eye. \textbf{Trees (background):} The trees create a backdrop for the other elements in the image. Their height and density add depth to the scene, making them a significant part of the overall composition. \textbf{Blue sky (background):} The blue sky provides a bright and open backdrop for the other elements. Its color contrasts with the green grass and trees, making it a noticeable part of the image. &
\textbf{Red Stop sign (left edge):} The bright red color contrasts sharply with the green grass and blue sky, immediately drawing the eye. Its placement on the edge of the frame also makes it a focal point. \textbf{Grassy field (foreground):} The large expanse of green grass occupies a significant portion of the image, providing a base and context for the other elements. Its uniform color and texture create a sense of depth and space. \textbf{Trees (background):} The trees create a backdrop that adds depth and context to the scene. Their varying heights and shapes add visual interest and complexity to the background. \textbf{Sky (overcast):} The overcast sky, with its soft, diffused light, sets the overall mood and tone of the image. The presence of clouds adds texture and visual interest to the upper portion of the frame. \textbf{Power lines (across the sky):} The power lines create a network of lines that cut across the sky, adding a sense of structure and complexity to the scene. They also provide a sense of scale and perspective. &
\textbf{Stop Sign (center-left in the foreground):} Its bright red color is a high contrast to the surrounding greenery. As a standard symbol, the octagonal shape immediately grabs the viewer's attention. \textbf{Menacing Clouds (backdrop):} The large size and dark color of the clouds create a sense of drama and foreboding. \textbf{Grassy Field (behind the stop sign):} The bright color creates a visual buffer between the stop sign and the trees. \textbf{Tree Line (midground):} The texture and color of the tree line is of high contrast to the field in front of it. \textbf{Roofs of Houses (behind the tree line):} They interrupt the regular pattern of the tree line. \textbf{Light Sky (visible through the clouds):} The contrast between dark and light creates visual interest and draws the eye to the sky. \textbf{Power Lines (upper portion of the frame):} They are visually salient due to their linear shape cutting across the sky, and their high contrast against the lighter parts of the sky.
\\

    \midrule

    \rowcolor{purple!6}
    Sample 6 &

\textbf{Colorful kites (upper-right):} The kites are salient due to their bright colors, which contrast with the blue sky. Their placement in the upper-right also draws the eye, as it is a common area for visual focus. \textbf{Crowd of people (center):} The crowd is salient because it is centrally located, drawing the eye to the middle of the image. The density of people also creates a focal point.  \textbf{Sky (background):} The sky is salient because it is a large, uniform area that fills the upper portion of the image. Its blue color provides a backdrop for the kites and other elements.
    &
\textbf{Kite (center):} The kite is centrally located and is the largest object in the image, drawing the eye immediately. Its bright colors and dynamic shape contrast with the sky, making it stand out. \textbf{Crowd (foreground):} The crowd is salient because it occupies a large portion of the foreground, creating a sense of depth and scale. The density of people and the variety of colors from their clothing also contribute to its visual prominence. \textbf{Tents (background):} The tents are salient because they define the event space and provide context. Their size and arrangement suggest a organized event, drawing attention to the purpose of the gathering. \textbf{Sky (background):} The sky is salient because it provides a backdrop that contrasts with the other elements. The blue color and scattered clouds create a sense of openness and depth, enhancing the overall composition. 
&
\textbf{Kites (top):} The kites draw attention due to their bright colors and their large size in relation to other objects in the image. Also, they are unique shapes and high in the visual field. \textbf{Crowd (below):} The crowd draws attention due to its sheer size, filling a significant portion of the lower half of the image. Its density creates a visual mass that stands out. \textbf{White tents (center):} The tents draw attention due to their light color that contrasts with the darker ground and the surrounding people. Also, they are centrally placed in the visual field. \textbf{Person with a kite (foreground):} The person is salient due to their foreground placement, which brings them closer to the viewer. The presence of the kite and its vibrant colors also makes the person a focal point. \textbf{Clear, bright blue sky (top):} The sky is salient due to its large size and uniform color, providing a backdrop for the other objects in the image. Its brightness also draws the eye upward. \\

    \bottomrule
    \end{tabular}
    }
\end{table*}


\begin{figure*}[htb]
    \centering
    \newcommand{\cropheightcom}{0.20\textwidth}
    \resizebox{0.8\textwidth}{!}{%
    \setlength{\tabcolsep}{8pt}
    \begin{tabular}{@{} c *{4}{c} @{}}

    \rotatebox{90}{Sample 7} &
    \includegraphics[height=\cropheightcom]{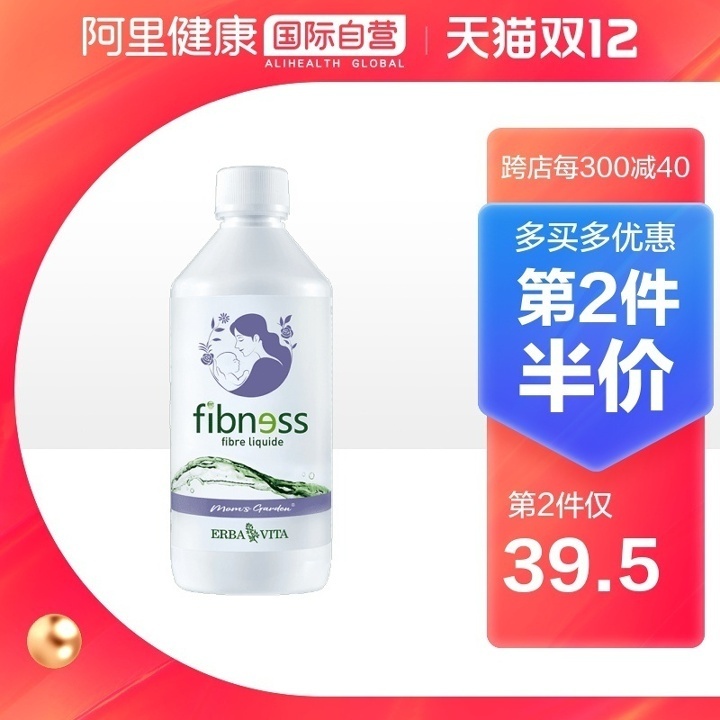} &
    \includegraphics[height=\cropheightcom]{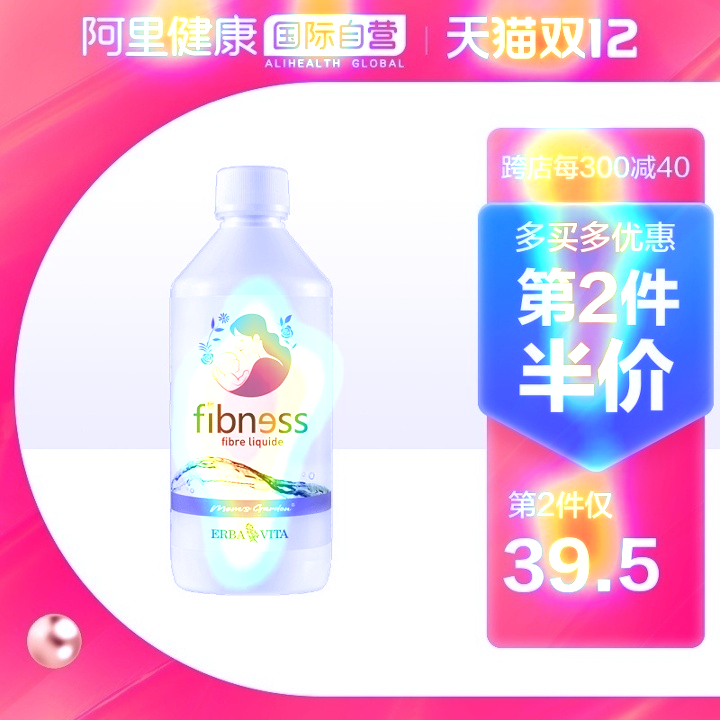} &
    \includegraphics[height=\cropheightcom]{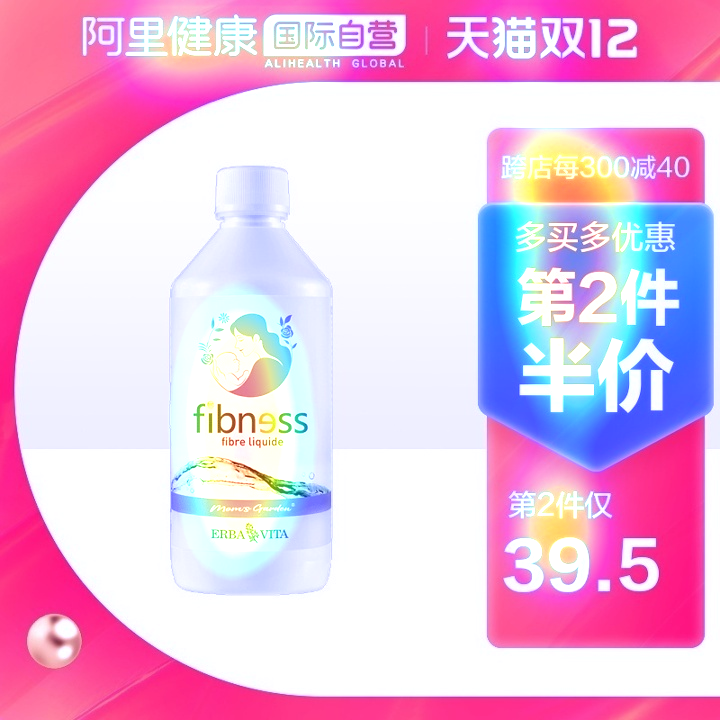} &
    \includegraphics[height=\cropheightcom]{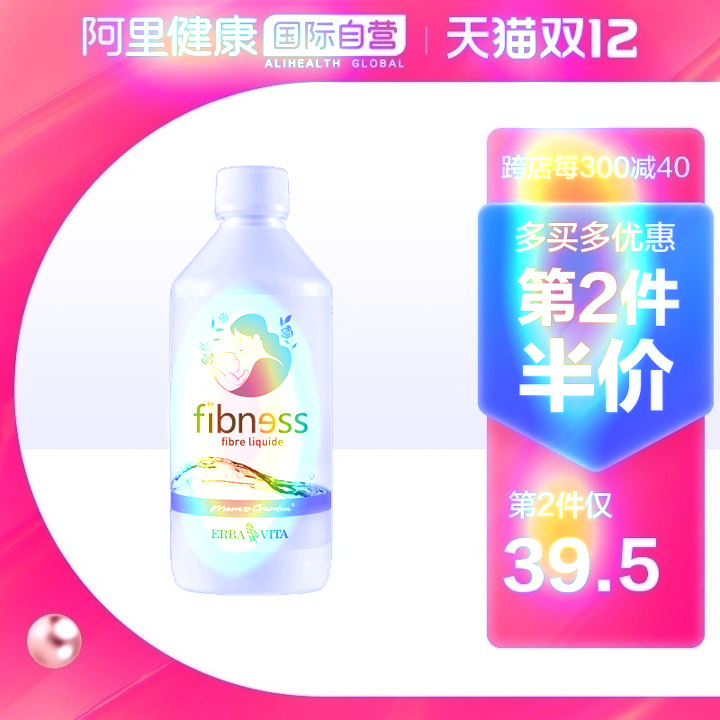} \\

    \rotatebox{90}{Sample 8} &
    \includegraphics[height=\cropheightcom]{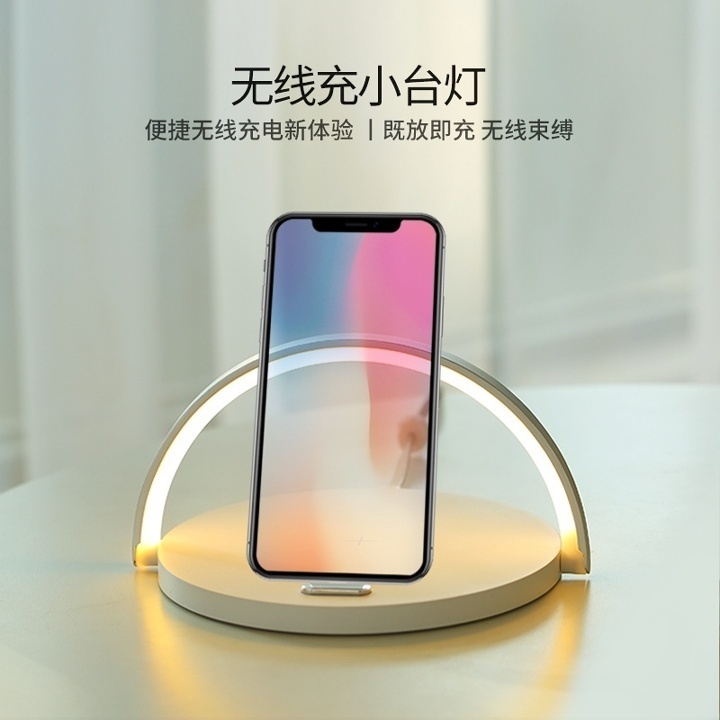} &
    \includegraphics[height=\cropheightcom]{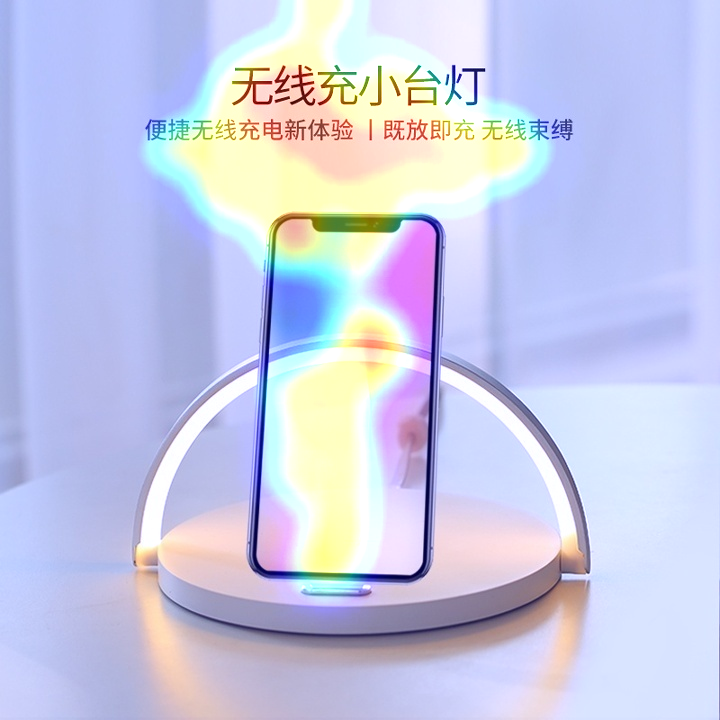} &
    \includegraphics[height=\cropheightcom]{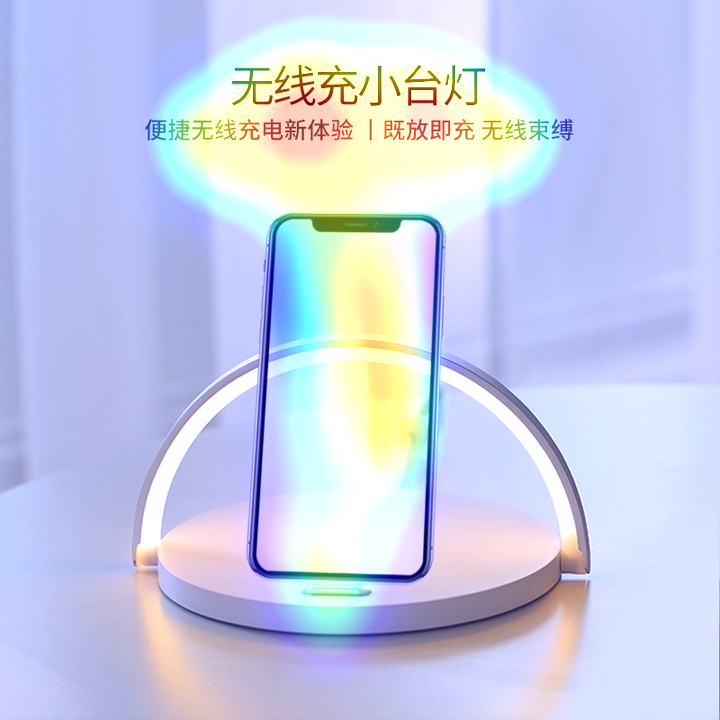} &
    \includegraphics[height=\cropheightcom]{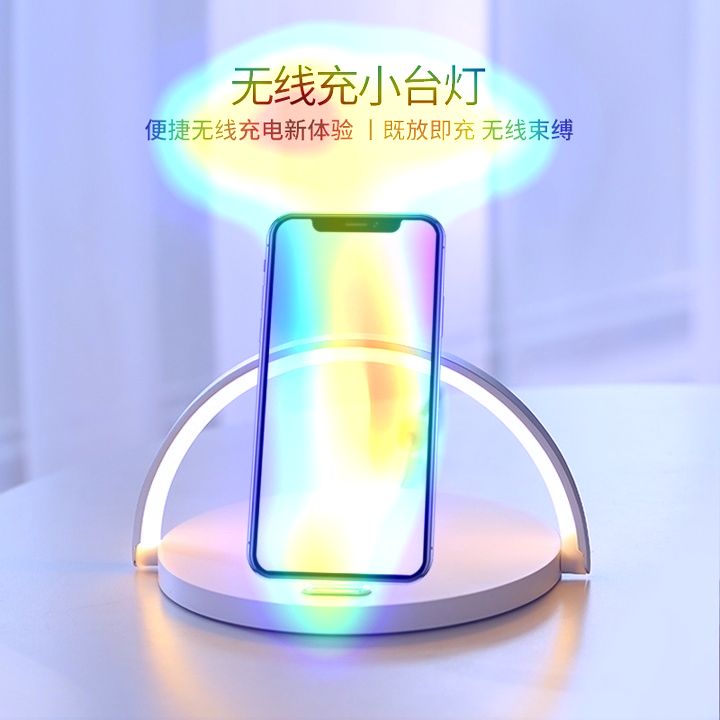} \\

    \rotatebox{90}{Sample 9} &
    \includegraphics[height=\cropheightcom]{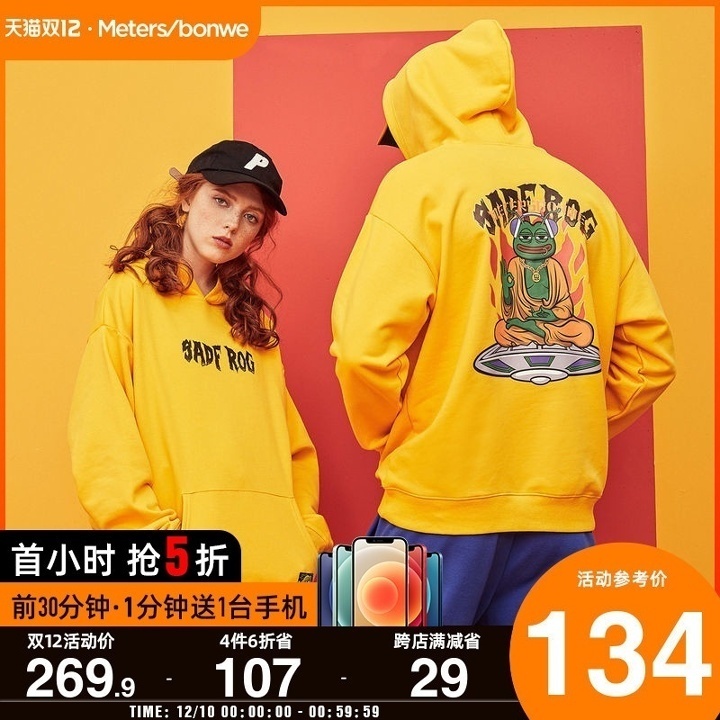} &
    \includegraphics[height=\cropheightcom]{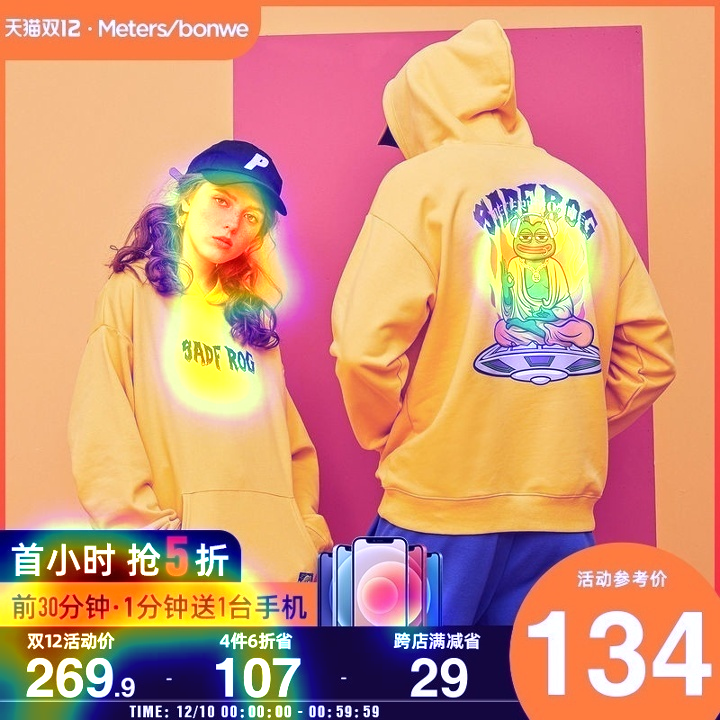} &
    \includegraphics[height=\cropheightcom]{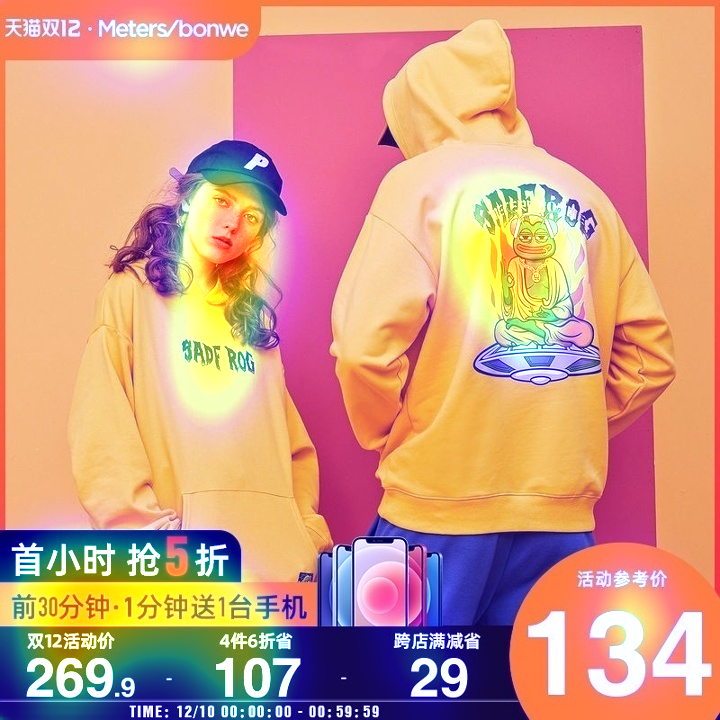} &
    \includegraphics[height=\cropheightcom]{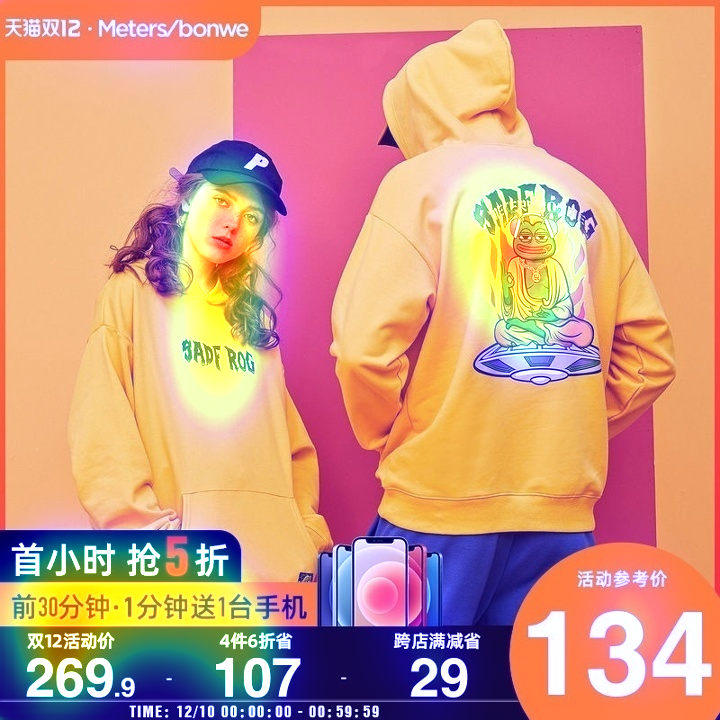} \\

    \rotatebox{90}{Sample 10} &
    \includegraphics[width=0.18\textwidth]{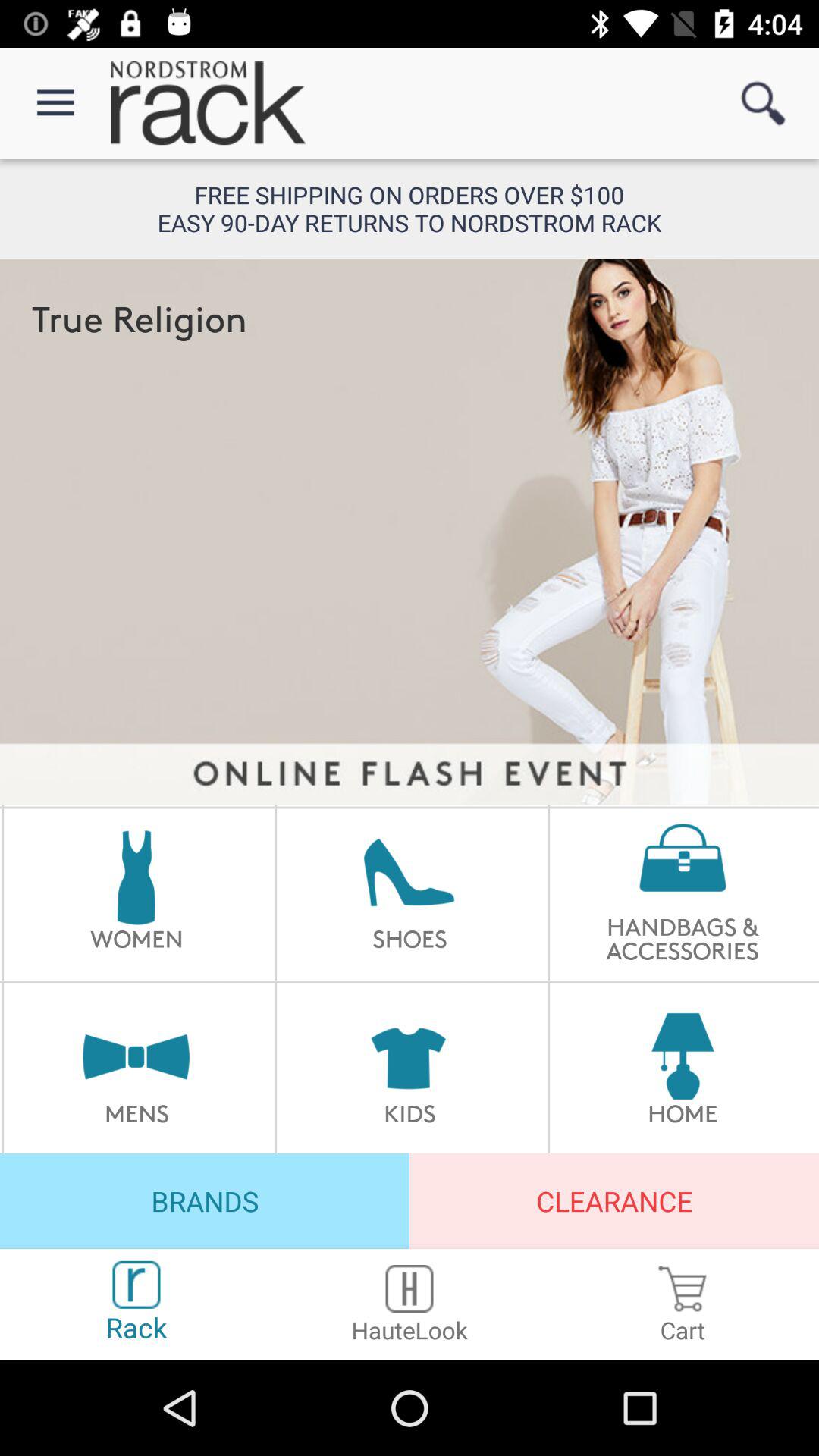} &
    \includegraphics[width=0.18\textwidth]{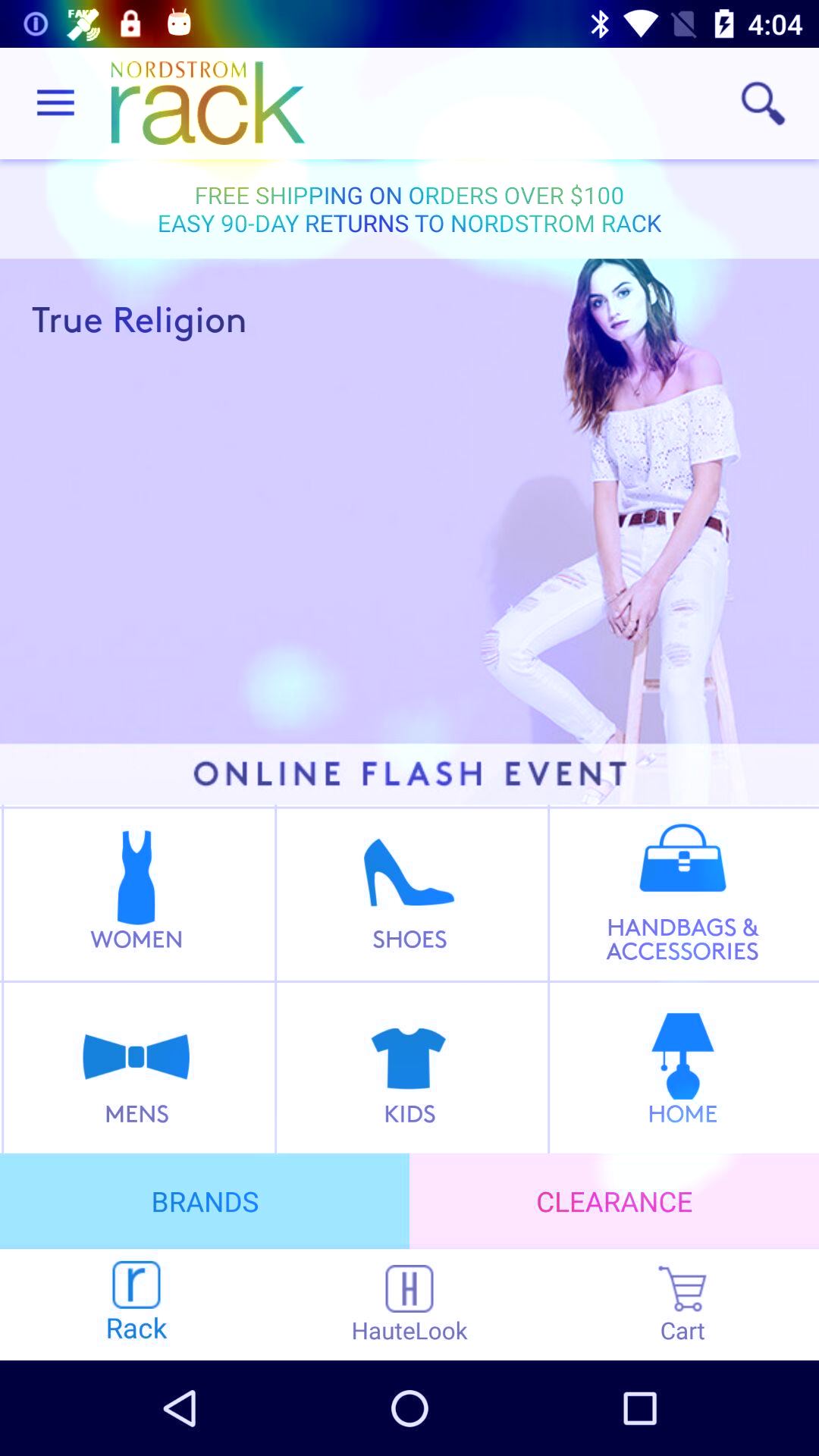} &
    \includegraphics[width=0.18\textwidth]{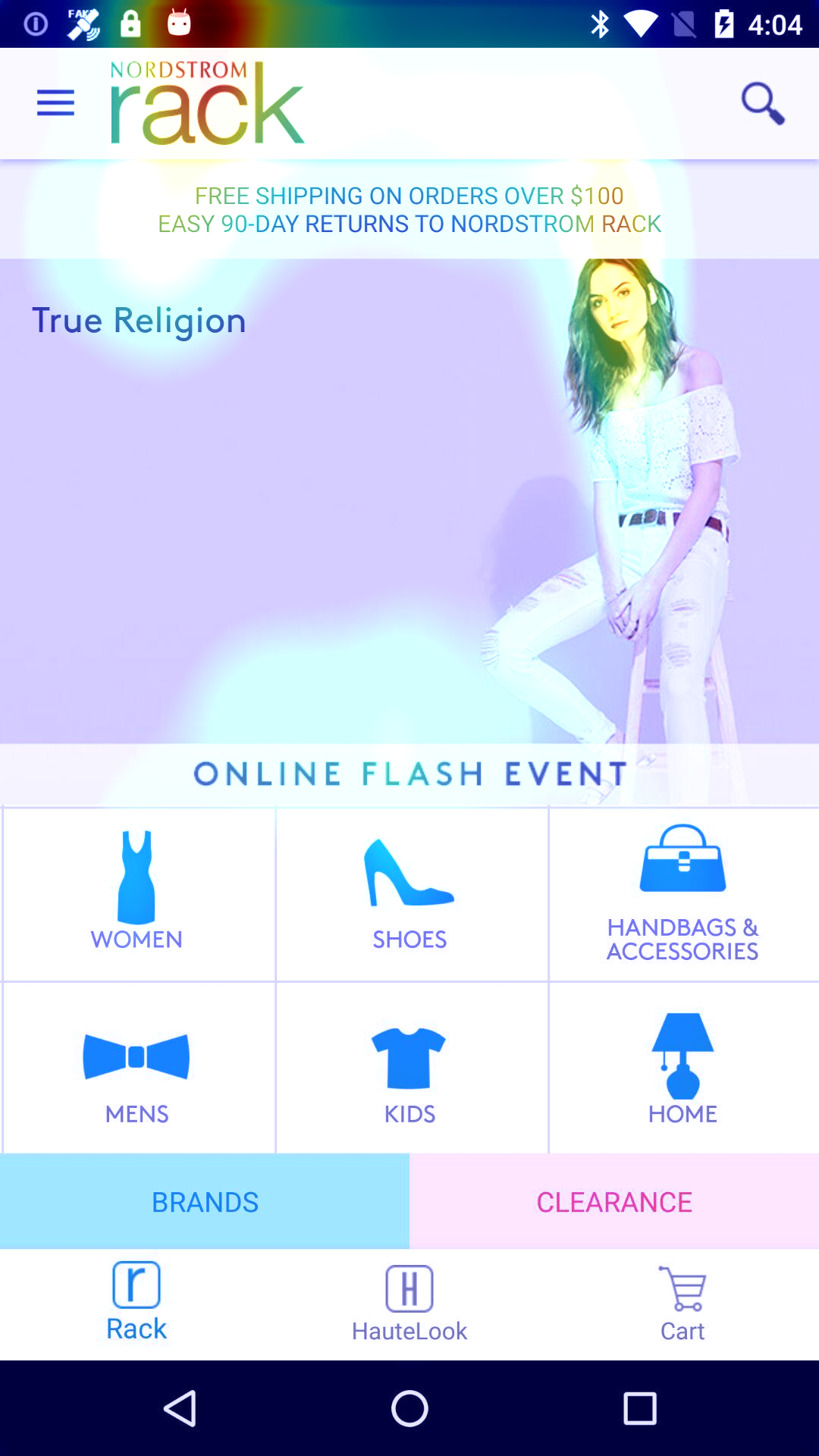} &
    \includegraphics[width=0.18\textwidth]{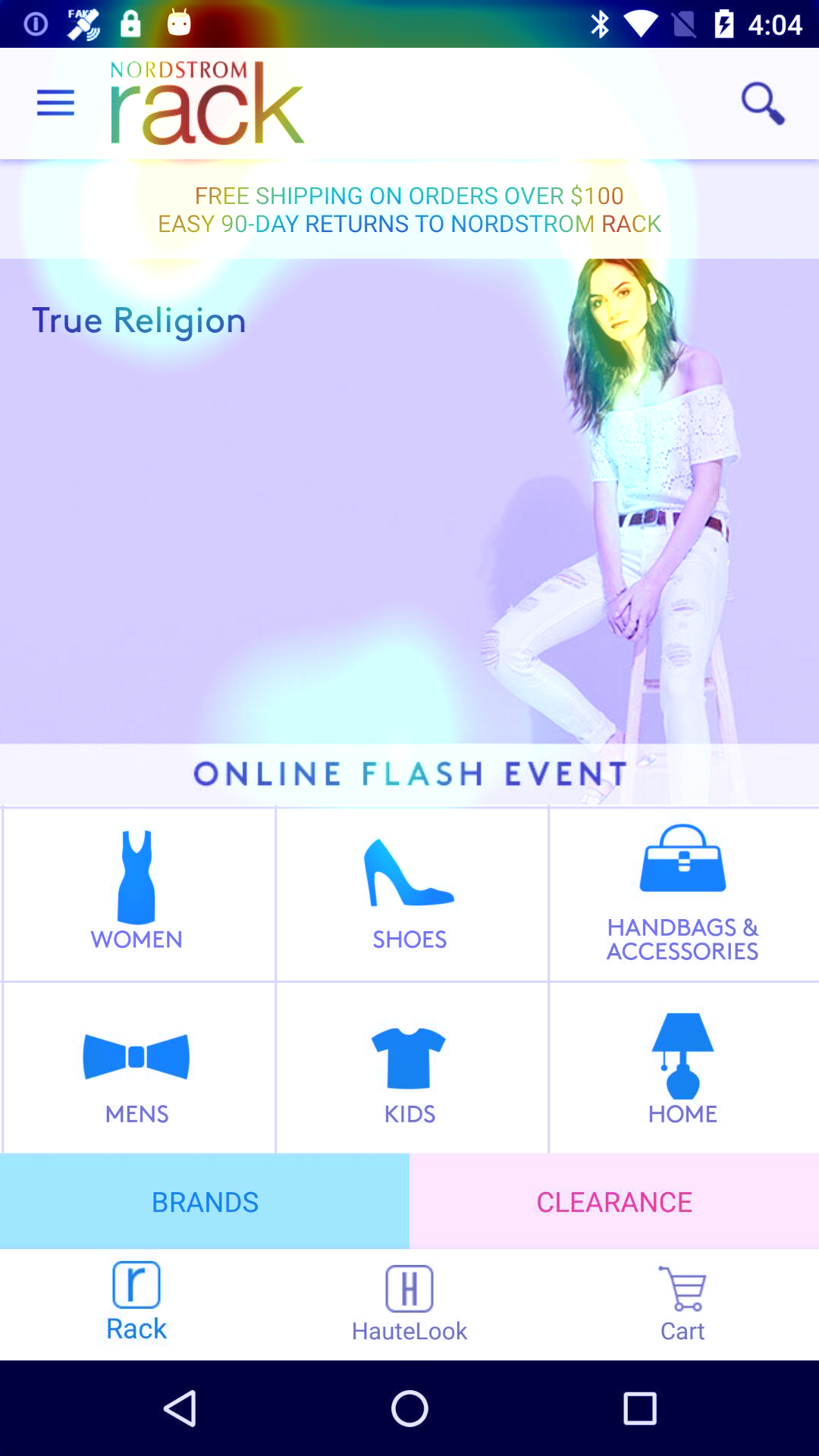} \\

    \rotatebox{90}{Sample 11} &
    \includegraphics[width=0.18\textwidth]{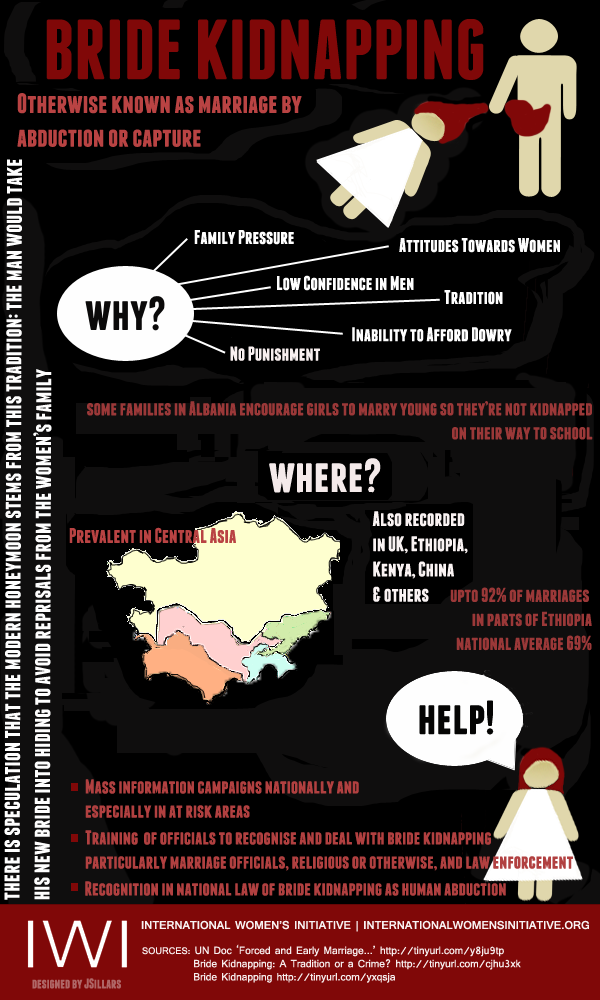} &
    \includegraphics[width=0.18\textwidth]{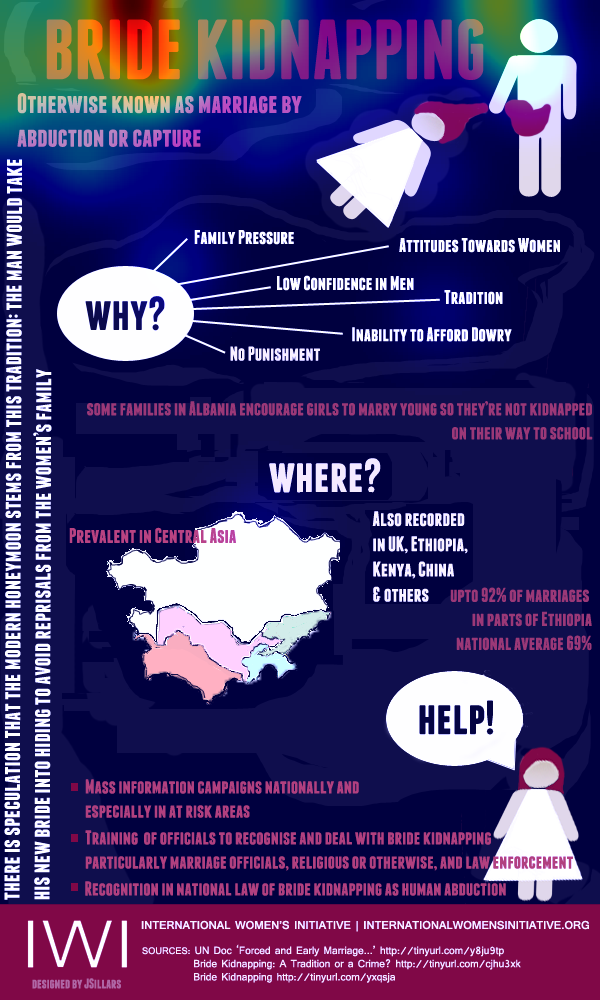} &
    \includegraphics[width=0.18\textwidth]{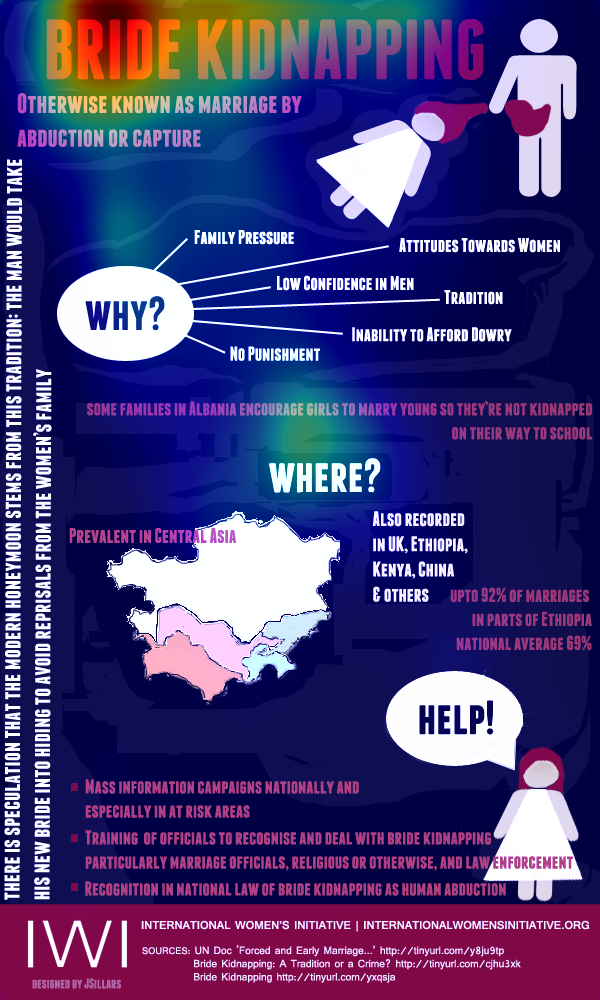} &
    \includegraphics[width=0.18\textwidth]{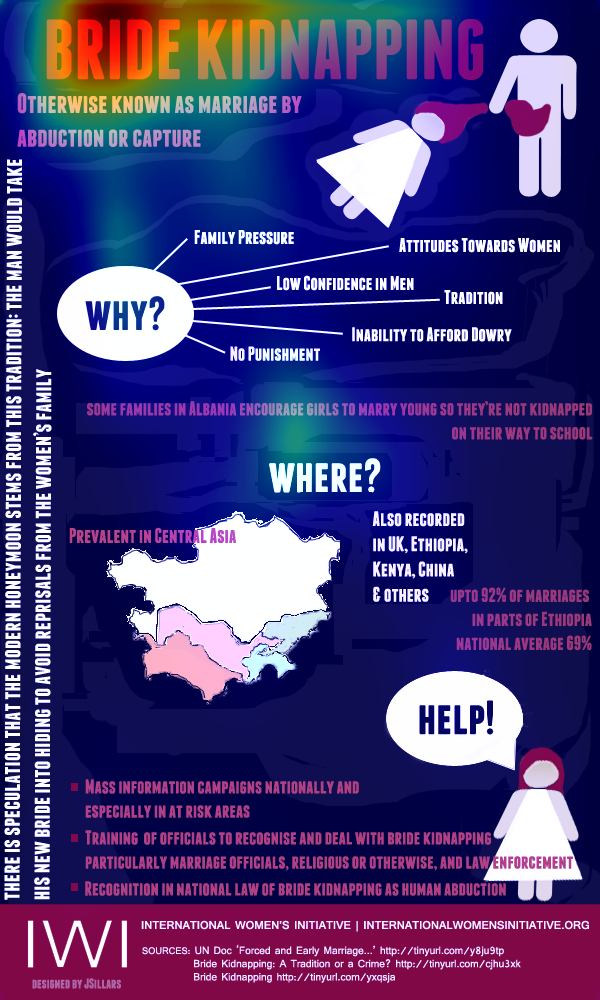} \\

    \rotatebox{90}{Sample 12} &
    \includegraphics[width=0.18\textwidth]{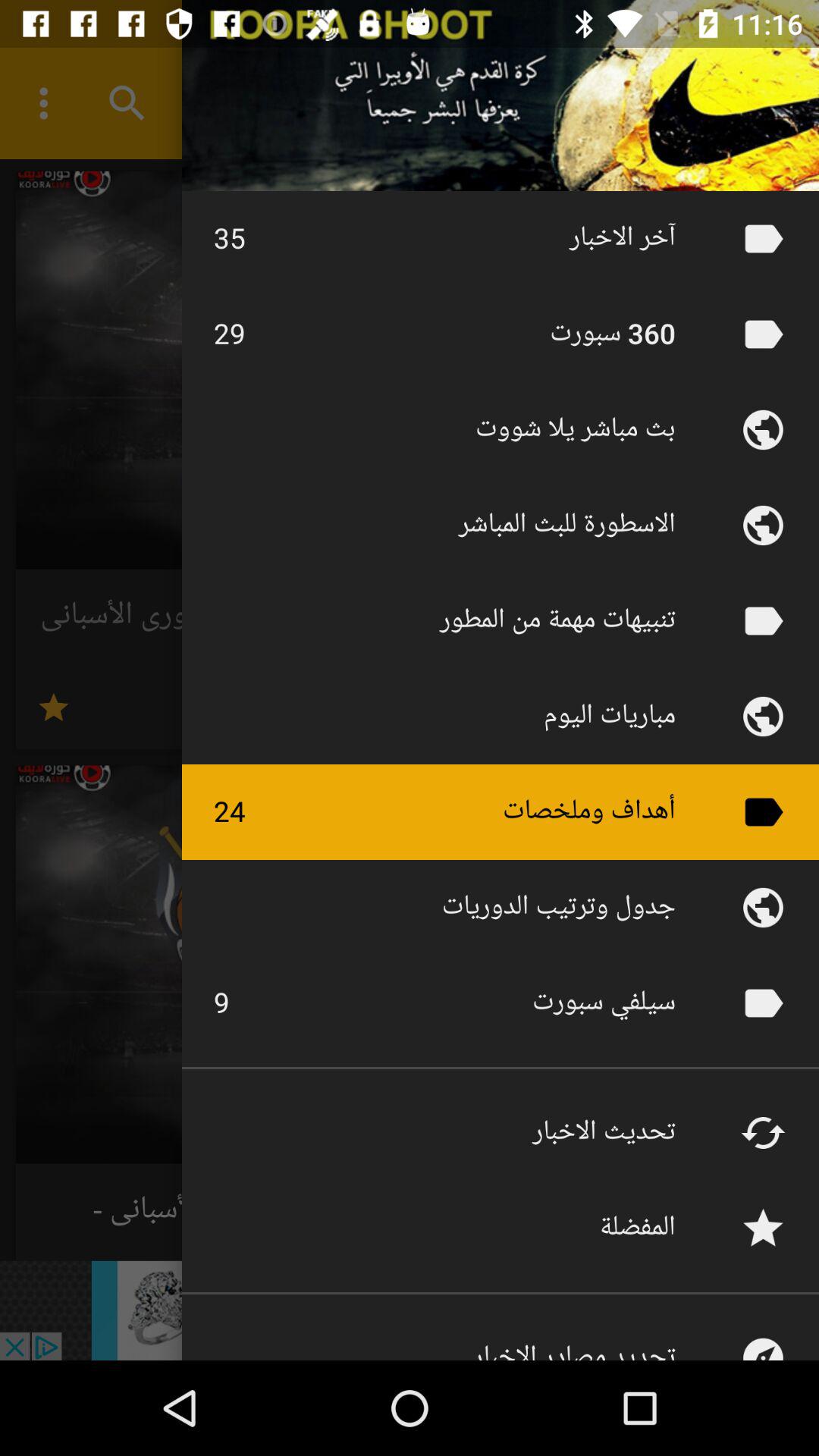} &
    \includegraphics[width=0.18\textwidth]{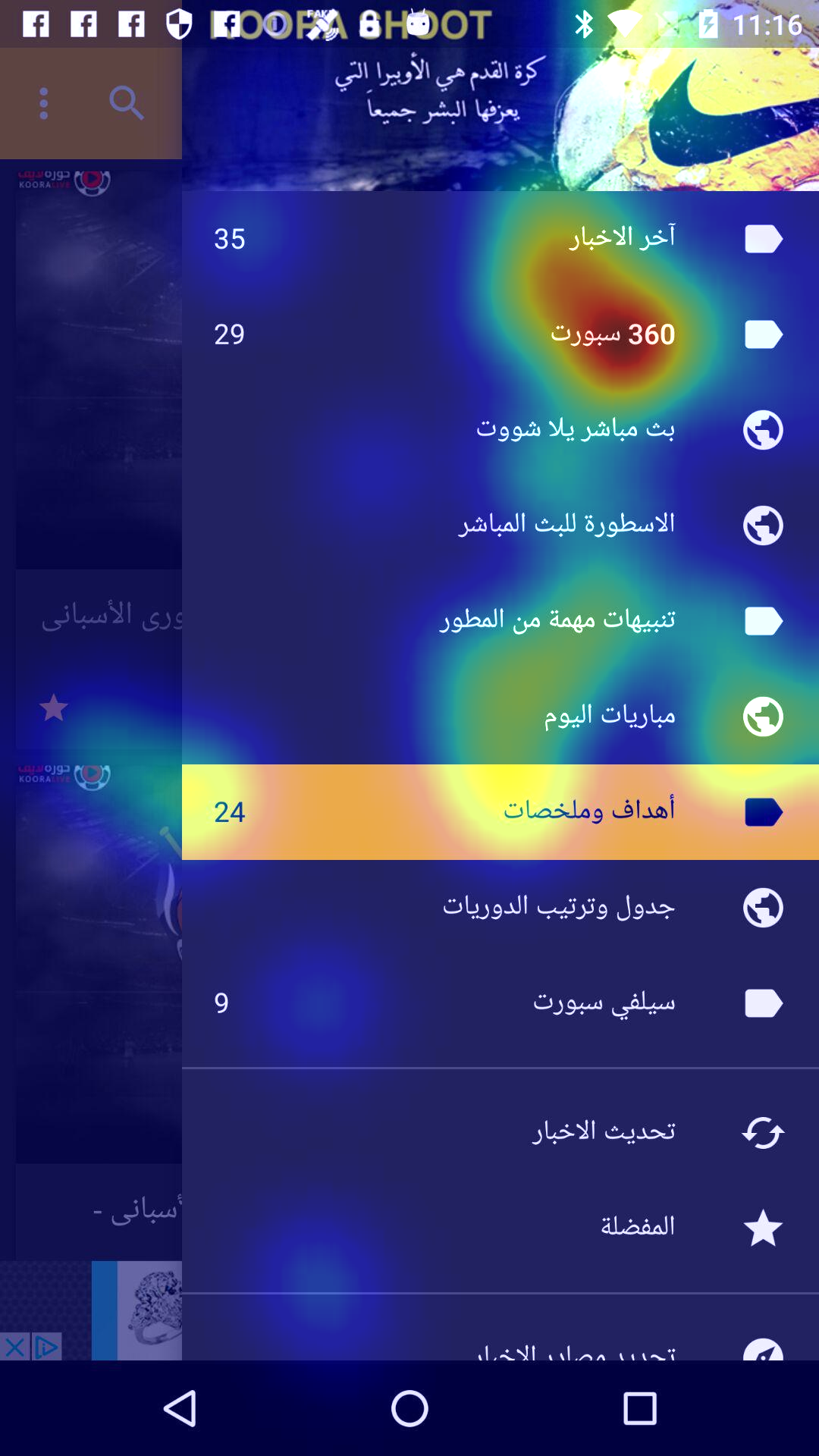} &
    \includegraphics[width=0.18\textwidth]{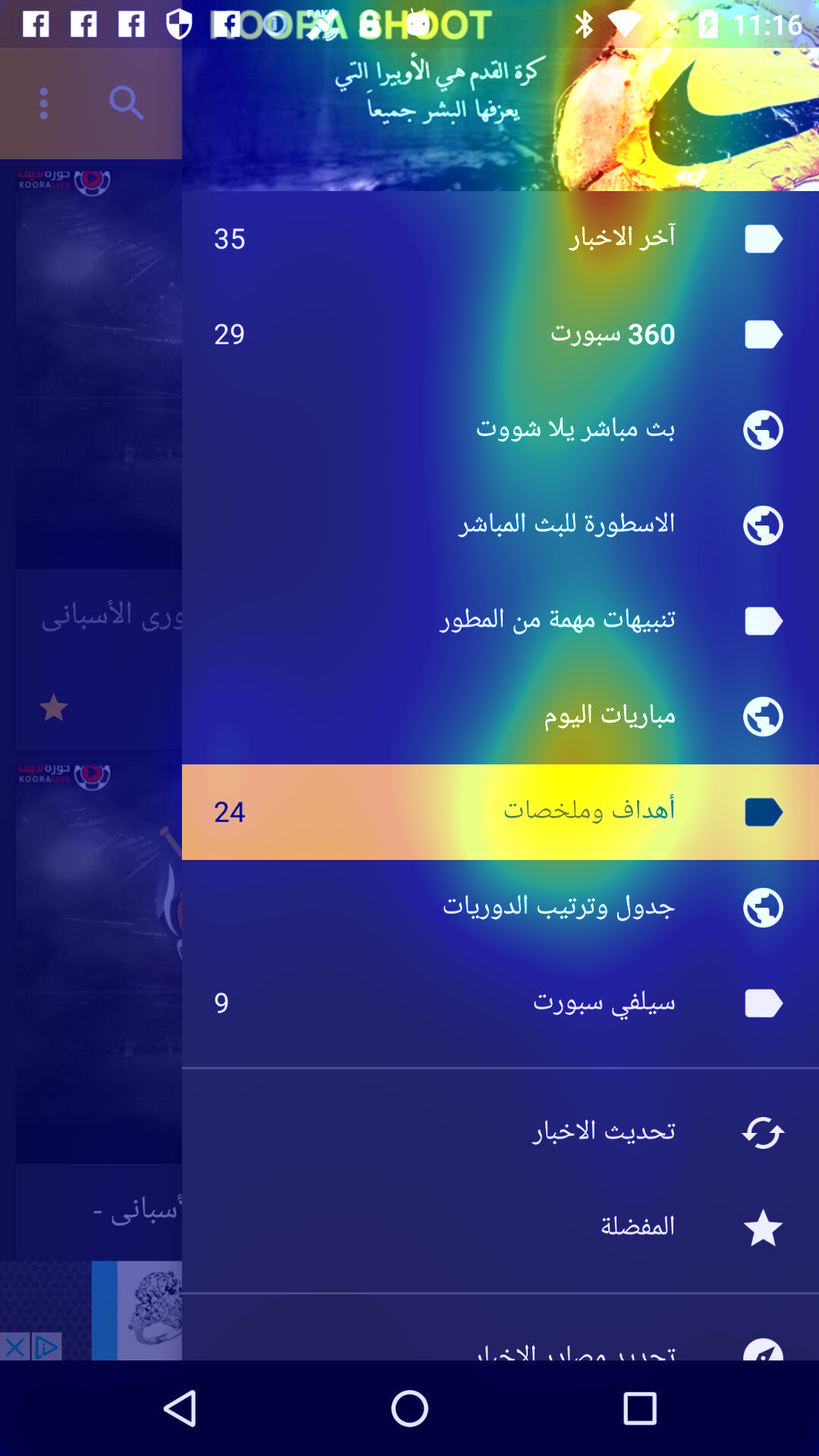} &
    \includegraphics[width=0.18\textwidth]{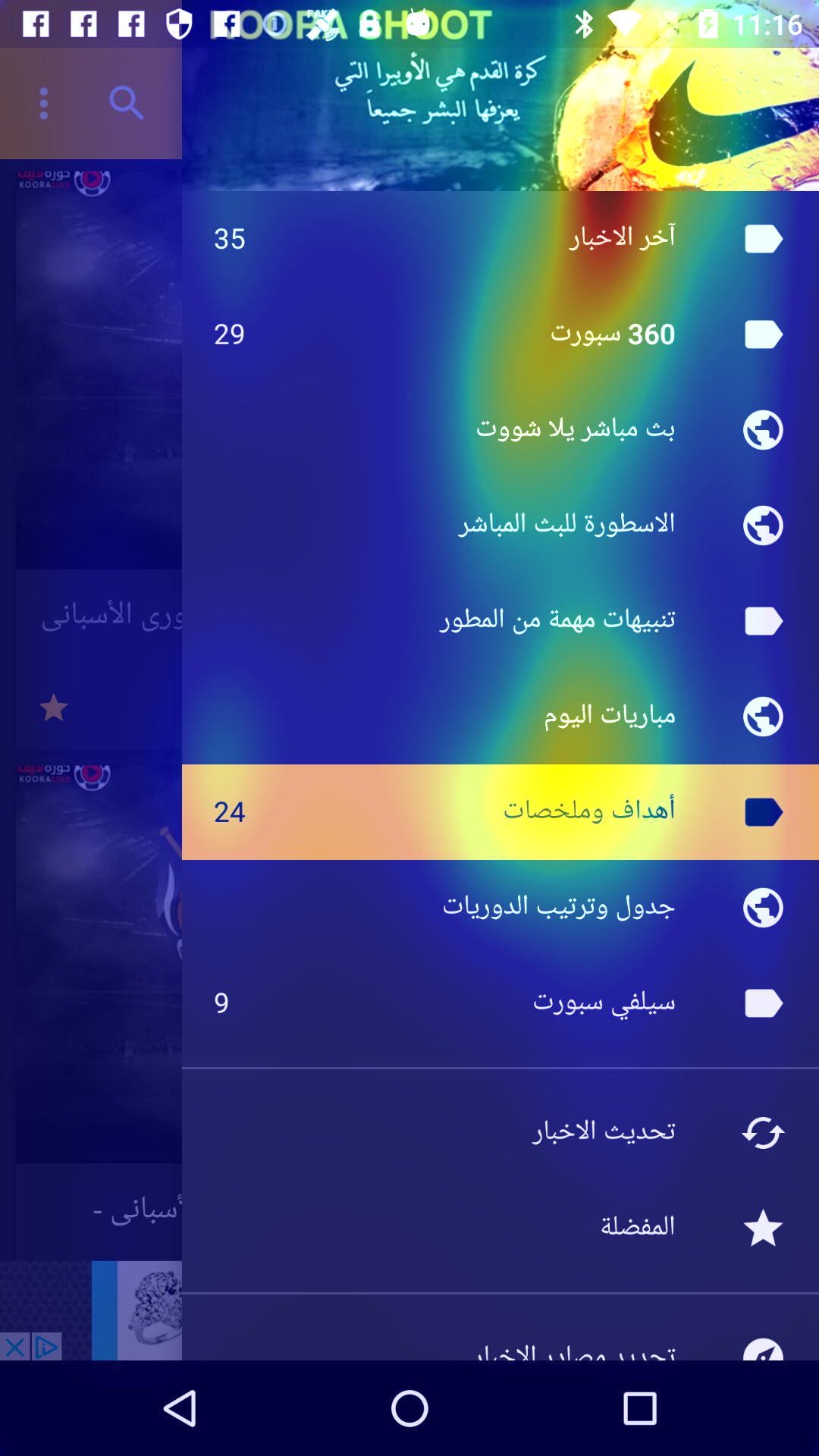} \\

    & Input Image & Ground Truth & OpenVAM-4B & OpenVAM-8B \\

    \end{tabular}
    }
    \caption{\textbf{Visualizations of OpenVAM predictions across different e-commerce and UI images.}}
    \label{fig:ui_vis}
\end{figure*}

\begin{table*}[t]
    \centering
    \caption{\textbf{Qualitative Text Comparison of~\autoref{fig:ui_vis}.} Generated explanations from OpenVAM models versus ground-truth text. Samples~7--9 correspond to rows~1--3, respectively.}
    \label{tab:openvam_text_samples_ecom}
    \resizebox{1\textwidth}{!}{
    \setlength{\tabcolsep}{1pt}
    \tiny
    \begin{tabular}{c|p{0.35\textwidth}|p{0.35\textwidth}|p{0.35\textwidth}}
    \toprule
    \textbf{Sample} & \textbf{OpenVAM-4B} & \textbf{OpenVAM-8B} & \textbf{Ground Truth} \\
    \midrule
    \rowcolor{blue!6}
    Sample 7 &
    \textbf{Price text (lower-right):} This large, bold red number is highly prominent due to its size, strong color contrast against the white background, and its position as the primary price information. \textbf{Promotional text (top-center):} This large, bold red text is centrally placed at the top of the image, immediately drawing attention to the main product description and key features. \textbf{Main product (center):} The large, clear image of the product bottle is centrally located, making it the primary visual focus for the item being sold. \textbf{Promotional text (top-left):} This large, bold red text is positioned prominently in the upper-left corner, highlighting a significant discount and drawing attention to the offer. \textbf{Promotional text (top-right):} This large, bold red text is placed in the upper-right corner, emphasizing a special event and drawing attention to the promotional period. \textbf{Promotional text (top-center):} This large, bold red text is centrally located at the top of the image, immediately drawing attention to the main product description and key features. &
    \textbf{Price text (lower-right):} This large, bold price is highly salient due to its prominent size, strong contrast against the red background, and its placement in a dedicated promotional area. \textbf{Promotional text (right side):} This large, bold text is highly salient due to its significant size, strong contrast against the red background, and its clear message of a substantial discount. \textbf{Main product (center):} The large, centrally placed bottle of the product is a primary focal point, drawing attention with its clear presentation and prominent placement. \textbf{Promotional text (top-left):} This text is salient due to its large, bold font and its position at the top of the image, immediately conveying a key product benefit. \textbf{Promotional text (top-right):} This text is salient due to its large, bold font and its position at the top of the image, immediately conveying a key product benefit. \textbf{Promotional text (right side):} This text is salient due to its large, bold font and its clear message of a significant discount, drawing attention to the value proposition. &
    \textbf{Promotional text (right):} This large, multi-layered badge prominently features bold, clear discount information like (2nd item half price) and (2nd item only 39.5) using strong contrasting colors of red and blue. \textbf{Product bottle (center-left):} The main product bottle is centrally placed and of significant size, featuring readable brand text and with a vibrant, eye-catching gradient design. \textbf{Promotional text (top):} The banner at the top features prominent, clear text (Ali Health International Self-operated Tmall Double 12), with being a key promotional event. \textbf{Small gold sphere (lower-left):} This small, shiny, metallic sphere stands out due to its high contrast against the red background and its unique, reflective appearance.\\
    
    \midrule

    \rowcolor{orange!6}
    Sample 8 &
    \textbf{Main product (center):} The white phone stand with a phone placed on it is centrally positioned and brightly illuminated, making it the primary visual focus. Its clean design and large size clearly showcase the product. \textbf{Feature headline text (top-center):} Large bold text at the top highlights the product’s key functionality. Its size and strong contrast against the background make it one of the most noticeable elements. \textbf{Description text (below headline):} A smaller descriptive text block sits directly beneath the main headline and explains the product’s benefit. Its placement near the headline draws attention while providing additional context. \textbf{Product lighting element (center-bottom):} The illuminated ring and base of the stand emit a warm glow that visually emphasizes the lighting feature. The brightness contrasts with the background, making this area visually striking.
 &
    \textbf{Promotional text (top-center):} This text is large, bold, and centrally placed, immediately drawing attention to the product's key feature. Its high contrast against the background further enhances its visibility. \textbf{Main product (center):} The white phone stand is the primary subject, centrally positioned and sharply in focus. Its clean design and the clear display of a phone on it make it the focal point of the image. \textbf{Promotional text (top-center):} Positioned directly below the main feature, this text provides additional context and reinforces the product's functionality. Its clear readability and central placement contribute to its salience. &
    \textbf{The main product (phone on wireless charging stand) (center):} This is the largest and most centrally placed object, illuminated by its own integrated light ring and base, making it the primary focus. The phone screen displays a dynamic, colorful pattern, further enhancing its visual prominence. \textbf{Promotional text (Wireless Charging Small Desk Lamp) (top-left):} This text is large, bold, and positioned prominently at the top-left, clearly stating the product's name with high contrast against the background. \textbf{Descriptive text (Convenient wireless charging new experience | Place and charge, no wires) (top-left):} Located directly beneath the main product title, this text is clearly readable and highlights key product benefits, drawing attention to its practical features. \textbf{The illuminated light ring and base of the stand (center-bottom):} These parts of the product draw attention due to their bright, warm glow, which emphasizes the lighting functionality and enhances the overall visual appeal of the stand.\\

    \midrule

    \rowcolor{purple!6}
    Sample 9 &
    \textbf{Models wearing the product (center):} Two human models wearing the featured clothing are centrally positioned and brightly lit, making them the primary visual focus. Human faces and full-body poses naturally draw attention and showcase the product in use. \textbf{Promotional discount banner (lower-left):} This banner uses bold typography and bright yellow coloring to highlight a major discount offer. The strong color contrast and large text make it highly noticeable. \textbf{Additional promotion/offer text block (lower-left):} Supporting promotional text sits above or near the discount area and emphasizes event-related offers. Its bold styling and placement near the discount messaging attract attention. 
    &
    \textbf{Model's face (left-center):} The model's face is clearly visible and directly engages the viewer, naturally drawing attention as a human element. \textbf{Promotional text (lower-left):} This large, bold text is centrally placed in the lower-left, highlighting a significant discount and a free gift, making it highly noticeable. \textbf{Model wearing the product (right-center):} The model's full body, especially the upper torso, is prominently displayed, showcasing the main product in a natural pose. \textbf{Promotional text (lower-left):} This text, positioned directly above the discount, clearly states the original price, creating a strong contrast with the discounted price and drawing attention to the value. \textbf{Promotional text (lower-left):} This text, positioned above the discount, clearly states the original price, creating a strong contrast with the discounted price and drawing attention to the value.
    &
    \textbf{Model on the right (center-right):} The model wears the brightly colored yellow hoodie, which significantly stands out against the background, and their full figure is centrally placed, making them a main subject. \textbf{Model on the left (center-left):} This model also wears the main product, a striking yellow hoodie, and her visible face and pose contribute to the product's presentation. \textbf{Promotional text (lower-left):} This large, bold text highlights a significant discount ( or 50\% off), making it a highly attractive promotional element. \textbf{Price text and (lower-left):} These numbers are clearly displayed and vary in color, indicating different price points or savings, with highlighted in green. \textbf{Callout badge (lower-right):} This text provides important context for the prominent price, drawing attention due to its direct relevance and close proximity to the main offer.\\

    \bottomrule
    \end{tabular}
    }
\end{table*}

\subsection{Extended Qualitative Analysis}
\label{sec:extended_qualitative}

\autoref{fig:visual_sal} and \autoref{tab:openvam_text_samples} provide complementary views of OpenVAM's
predictions: the saliency maps show how attention is distributed spatially,
while the rationales discretize this distribution into semantic elements and
visible cues. The four examples are particularly informative because they
represent different attention regimes, ranging from a single dominant object
to multiple competing regions. Overall, the two outputs are well aligned,
but their remaining mismatches also reveal where the problem is more
challenging.

\noindent\textbf{Interaction-driven attention.}
In Sample~1, human attention is not explained by a single isolated object.
The ground-truth map places attention on both the woman and the television,
reflecting the interaction between them. OpenVAM recovers the same two
regions, although it places relatively more saliency on the television than
on the woman's face. The generated rationales capture the semantic relation
behind these regions: both models identify the person and television, and
OpenVAM-7B explicitly connects them through the woman's pose, gaze, and
hands. OpenVAM-3B additionally identifies the arms as a separate attended
region, closely matching the reference decomposition. This example
illustrates an important advantage of the rationale output: it can explain
that two spatially separated peaks are related through an interaction rather
than treating them as independent salient objects. At the same time, the
difference in relative map intensity shows that semantic identification and
dense saliency strength are not yet perfectly calibrated.

\noindent\textbf{Concentrated object saliency.}
Sample~2 represents the opposite regime: attention is strongly concentrated
on a small, visually distinctive object. OpenVAM localizes the necklace and
pendant tightly, whereas several competing predictions spread more saliency
over the surrounding torso. The rationales are similarly focused. Both
OpenVAM variants identify the pendant as the primary element and explain its
saliency through its reflective appearance and contrast with the dark
turtleneck, consistent with the reference. OpenVAM-3B further separates the
necklace chain as an attended component, while OpenVAM-7B gives a more
compact decomposition closer to the reference. This case shows that when a
clear local visual cue dominates attention, the dense and semantic outputs
agree particularly well.

\noindent\textbf{Competing regions in structured layouts.}
Sample~3 is more challenging because the webpage contains several visually
competitive cards, images, faces, and text blocks. The ground-truth saliency
is therefore multi-modal rather than concentrated at one location. OpenVAM
preserves several separated attention regions, while some baselines either
spread attention broadly across the upper row or concentrate on only a small
subset of the content. The language output exposes the same challenge at a
semantic level. OpenVAM-7B identifies the top-left headline, the central
headline associated with the Facebook image, and the lower Zuckerberg face,
which closely matches the three principal elements described by the
reference. OpenVAM-3B captures the important content as well, but
over-decomposes the central region into the Facebook logo and multiple
headline descriptions and misses the face as a separate element. This
suggests that increasing VLM capacity mainly improves the semantic
organization of distributed attention: the larger model better consolidates
related visual evidence into the same salient regions rather than merely
producing more text.

\paragraph{Multi-scale attention and secondary cues.}
Sample~4 combines a dominant semantic subject with much smaller secondary
cues. Both the reference and OpenVAM strongly emphasize the rider, with
additional attention distributed over the horse. OpenVAM also detects the
small object near the horse's front leg. This distinction is reflected in the
rationales: OpenVAM-3B gives the compact high-level description of
\emph{rider} and \emph{horse}, whereas OpenVAM-7B additionally mentions the
small light-colored object near the hoof; the reference identifies it as the
white ball. Importantly, the 7B model does not confidently invent its
identity, instead describing it as uncertain given the image resolution.
This demonstrates improved coverage of secondary attended regions with the
larger model. However, OpenVAM-7B also divides the horse into overlapping
``main frame'' and ``main body'' descriptions, showing that increased
semantic coverage can introduce redundancy or overly fine-grained
decomposition.

\paragraph{Overall observations.}
Together, these examples reveal a useful distinction between
\emph{localization errors} and \emph{semantic decomposition errors}. The
dominant attended regions are generally stable in the saliency maps; the
remaining language errors more often concern how a continuous attention
distribution is partitioned into discrete objects---for example, whether the
TV screen and its content should be one or two items, or whether different
parts of the horse should be described separately. This is consistent with
OpenVAM's decoupled-but-aligned design: the dense map and rationale are not
expected to have a strict one-to-one representation, but they should agree
on the major attended content.

\subsection{Ablation on different DinoV3 Backbones}

\newcommand{\encS}{DINOv3-S/16}
\newcommand{\encSp}{DINOv3-S+/16}
\newcommand{\encB}{DINOv3-B/16}   
\newcommand{\encL}{DINOv3-L/16}

\newcommand{\hl}[1]{\cellcolor[HTML]{DADCFF}#1}

\begin{table*}[!]
\centering
\caption{
\textbf{Vision-backbone ablation across DINOv3 encoders during Stage~I training.}
We compare four visual encoders (\encS, \encSp, \encB, \encL) while keeping all other components and training settings identical. Overall, \encB~provides the strongest performance and is therefore used as the default vision backbone in our experiments.
}
\label{tab:vision_backbone_ablation}

\resizebox{\textwidth}{!}{
\setlength{\tabcolsep}{25pt}
\begin{tabular}{l|l|ccccc}
\toprule
\multirow{2}{*}{\textbf{Dataset}}
& \multirow{2}{*}{\textbf{Vision Encoder}}
& \multicolumn{5}{c}{\textbf{Saliency}} \\
\cmidrule(lr){3-7}
& & CC $\uparrow$ & KLD $\downarrow$ & AUC $\uparrow$ & SIM $\uparrow$ & NSS $\uparrow$ \\
\midrule

\multirow{4}{*}{U-EYE}
& \encS  & 0.716 & 0.570 & 0.842 & 0.618 & 1.673 \\
& \encSp & 0.715 & 0.569 & 0.843 & 0.617 & 1.671 \\
& \hl{\encB} & \hl{0.728} & \hl{0.550} & \hl{0.846} & \hl{0.622} & \hl{1.704} \\
& \encL  & 0.719 & 0.561 & 0.843 & 0.617 & 1.699 \\
\midrule

\multirow{4}{*}{SalECI}
& \encS  & 0.768 & 0.495 & 0.894 & 0.656 & 1.972 \\
& \encSp & 0.748 & 0.524 & 0.894 & 0.637 & 1.972 \\
& \hl{\encB} & \hl{0.788} & \hl{0.466} & \hl{0.898} & \hl{0.675} & \hl{2.031} \\
& \encL  & 0.778 & 0.478 & 0.888 & 0.662 & 2.014 \\
\midrule

\multirow{4}{*}{OSIE}
& \encS  & 0.890 & 0.298 & 0.927 & 0.744 & 3.586 \\
& \encSp & 0.876 & 0.324 & 0.924 & 0.730 & 3.446 \\
& \hl{\encB} & \hl{0.912} & \hl{0.229} & \hl{0.935} & \hl{0.777} & \hl{3.903} \\
& \encL  & 0.910 & 0.250 & 0.939 & 0.762 & 3.868 \\
\midrule

\multirow{4}{*}{Salicon}
& \encS  & 0.889 & 0.200 & 0.874 & 0.787 & 1.955 \\
& \encSp & 0.884 & 0.214 & 0.872 & 0.780 & 1.940 \\
& \hl{\encB} & \hl{0.902} & \hl{0.186} & \hl{0.875} & \hl{0.799} & \hl{1.985} \\
& \encL  & 0.900 & 0.186 & 0.874 & 0.794 & 1.972 \\
\midrule

\multirow{4}{*}{CAT2000}
& \encS  & 0.878 & 0.276 & 0.886 & 0.746 & 2.407 \\
& \encSp & 0.870 & 0.284 & 0.885 & 0.742 & 2.380 \\
& \hl{\encB} & \hl{0.884} & \hl{0.262} & \hl{0.888} & \hl{0.754} & \hl{2.438} \\
& \encL  & 0.876 & 0.283 & 0.870 & 0.735 & 2.414 \\
\midrule

\multirow{4}{*}{MIT1003}
& \encS  & 0.793 & 0.538 & 0.916 & 0.631 & 2.915 \\
& \encSp & 0.775 & 0.573 & 0.913 & 0.614 & 2.802 \\
& \hl{\encB} & \hl{0.817} & \hl{0.483} & \hl{0.922} & \hl{0.656} & \hl{3.050} \\
& \encL  & 0.802 & 0.497 & 0.917 & 0.630 & 3.036 \\

\bottomrule
\end{tabular}
}
\end{table*}
We analyze the impact of the visual backbone used in the dense saliency pathway during Stage~I training. Specifically, we compare four variants of the DINOv3~\cite{simeoni2025dinov3} encoder: \encS, \encSp, \encB, and \encL. All other components are kept identical, and models are trained using the same optimization settings to isolate the effect of the visual representation. Quantitative results across six datasets are reported in~\autoref{tab:vision_backbone_ablation}. Across all datasets and metrics, \encB~consistently achieves the strongest overall performance. In particular, it provides the best or near-best results for CC, KLD, SIM, and NSS on most datasets, indicating that it produces the most accurate and spatially consistent saliency predictions. While \encL~is slightly larger, it does not consistently outperform \encB, suggesting diminishing returns from further increasing model capacity for dense attention localization.  The smaller backbones (\encS~and \encSp) generally perform worse, especially on challenging datasets such as OSIE and MIT1003, where complex scene understanding and object interactions are important. These results indicate that stronger visual representations significantly improve the quality of dense saliency estimation. Based on this ablation, we adopt \encB~as the default visual backbone for OpenVAM in all subsequent experiments, as it provides the best balance between accuracy and computational efficiency.

\subsection{Effect of Visual Backbone}
\autoref{tab:backbone_ablation} highlights the impact of the visual backbone on attention localization, and we evaluate this effect under Stage~I only to isolate the learned saliency prior before any language supervision is introduced. When we replace DINOv3 with the original Qwen ViT, the saliency metrics drop noticeably, indicating that Qwen ViT provides a weaker \emph{localization prior} to dense prediction. In contrast, DINOv3 consistently yields stronger spatial cues, which better support our Stage~I objective of learning a stable \emph{where} pathway before introducing language supervision. This finding motivates anchoring OpenVAM’s dense saliency branch with DINOv3, while using Qwen as an auxiliary semantic head.

\begin{table*}[t]
\centering
\caption{
\textbf{Backbone ablation under Stage-I saliency training.}
We replace the visual backbone with the Qwen Vision Encoder or DINOv3-B/16
while keeping the dense saliency training protocol fixed.
}
\vspace{-0.5em}
\label{tab:backbone_ablation}

\resizebox{\textwidth}{!}{
\setlength{\tabcolsep}{22pt}
\begin{tabular}{l|l|ccccc}
\toprule
\textbf{Dataset} & \textbf{Backbone}
& CC $\uparrow$ & KLD $\downarrow$ & AUC $\uparrow$ & SIM $\uparrow$ & NSS $\uparrow$ \\
\midrule

\multirow{2}{*}{{U-EYE}~\cite{jiang2023ueyes}}
& Qwen ViT
& 0.656 & 0.673 & 0.826 & 0.577 & 1.509 \\
& \cellcolor[HTML]{DADCFF}\textbf{DINOv3-B/16 (ours)}
& \cellcolor[HTML]{DADCFF}0.728
& \cellcolor[HTML]{DADCFF}0.550
& \cellcolor[HTML]{DADCFF}0.846
& \cellcolor[HTML]{DADCFF}0.622
& \cellcolor[HTML]{DADCFF}1.704 \\
\midrule

\multirow{2}{*}{SalECI~\cite{jiang2022does}}
& Qwen ViT
& 0.690 & 0.671 & 0.869 & 0.577 & 1.747 \\
& \cellcolor[HTML]{DADCFF}\textbf{DINOv3-B/16 (ours)}
& \cellcolor[HTML]{DADCFF}0.788
& \cellcolor[HTML]{DADCFF}0.466
& \cellcolor[HTML]{DADCFF}0.898
& \cellcolor[HTML]{DADCFF}0.675
& \cellcolor[HTML]{DADCFF}2.031 \\
\midrule

\multirow{2}{*}{OSIE~\cite{xu2014predicting}}
& Qwen ViT
& 0.718 & 0.596 & 0.891 & 0.601 & 2.494 \\
& \cellcolor[HTML]{DADCFF}\textbf{DINOv3-B/16 (ours)}
& \cellcolor[HTML]{DADCFF}0.912
& \cellcolor[HTML]{DADCFF}0.229
& \cellcolor[HTML]{DADCFF}0.935
& \cellcolor[HTML]{DADCFF}0.777
& \cellcolor[HTML]{DADCFF}3.903 \\
\midrule

\multirow{2}{*}{SALICON~\cite{jiang2015salicon}}
& Qwen ViT
& 0.797 & 0.346 & 0.851 & 0.708 & 1.661 \\
& \cellcolor[HTML]{DADCFF}\textbf{DINOv3-B/16 (ours)}
& \cellcolor[HTML]{DADCFF}0.902
& \cellcolor[HTML]{DADCFF}0.186
& \cellcolor[HTML]{DADCFF}0.875
& \cellcolor[HTML]{DADCFF}0.799
& \cellcolor[HTML]{DADCFF}1.985 \\
\midrule

\multirow{2}{*}{CAT2000~\cite{borji2015cat2000}}
& Qwen ViT
& 0.830 & 0.361 & 0.874 & 0.704 & 2.261 \\
& \cellcolor[HTML]{DADCFF}\textbf{DINOv3-B/16 (ours)}
& \cellcolor[HTML]{DADCFF}0.884
& \cellcolor[HTML]{DADCFF}0.262
& \cellcolor[HTML]{DADCFF}0.888
& \cellcolor[HTML]{DADCFF}0.754
& \cellcolor[HTML]{DADCFF}2.438 \\
\midrule

\multirow{2}{*}{MIT1003~\cite{judd2009learning}}
& Qwen ViT
& 0.633 & 0.853 & 0.881 & 0.512 & 2.183 \\
& \cellcolor[HTML]{DADCFF}\textbf{DINOv3-B/16 (ours)}
& \cellcolor[HTML]{DADCFF}0.817
& \cellcolor[HTML]{DADCFF}0.483
& \cellcolor[HTML]{DADCFF}0.922
& \cellcolor[HTML]{DADCFF}0.656
& \cellcolor[HTML]{DADCFF}3.050 \\
\bottomrule
\end{tabular}
}

\end{table*}

\subsection{Effect of Using a Stronger VLM Backbone}
\label{sec:supp_stronger_vlm_backbone}

Although OpenVAM uses the VLM as a semantic explanation head rather than as the primary dense saliency predictor, the choice of VLM backbone can still affect rationale generation and cross-modal alignment. To clarify this point, we evaluate whether OpenVAM benefits from a newer VLM backbone by replacing the Qwen3-VL-4B semantic head with Qwen3.5-4B while keeping the same overall OpenVAM pipeline and training protocol.

As shown in~\autoref{tab:qwen35_backbone}, using Qwen3.5-4B further improves over the corresponding OpenVAM-4B variant on the average over six benchmarks. The gains are modest for dense saliency metrics, because saliency localization is mainly controlled by the DINOv3 visual encoder and the dedicated dense decoder, but the improvement is larger for rationale quality, where the stronger VLM backbone directly affects language generation. This confirms that OpenVAM is not tied to a specific older VLM backbone: stronger VLMs can be plugged into the same decoupled-but-aligned architecture and further improve semantic explanation quality.

\begin{table}[t]
\centering
\small
\setlength{\tabcolsep}{5pt}
\caption{\textbf{Effect of a stronger VLM backbone.}
We compare OpenVAM-4B with a variant using Qwen3.5-4B as the VLM semantic head, averaged over the six benchmarks. The stronger VLM backbone improves both saliency metrics and rationale quality, with the largest gain on JScore.}
\resizebox{\linewidth}{!}{%
\begin{tabular}{l|cccc}
\toprule
\textbf{Model} & \textbf{CC} $\uparrow$ & \textbf{KLD} $\downarrow$ & \textbf{SIM} $\uparrow$ & \textbf{JScore} $\uparrow$ \\
\midrule
OpenVAM-4B & 0.850 & 0.351 & 0.718 & 0.713 \\
OpenVAM w/ Qwen3.5-4B & 0.852 & 0.348 & 0.722 & 0.734 \\
\bottomrule
\end{tabular}%
}
\label{tab:qwen35_backbone}
\end{table}

\subsection{Stability of JScore: Prompt Sensitivity and Reasoning Consistency}
\label{sec:supp_jscore_stability}

Because JScore relies on a VLM judge, we assess two complementary forms of stability: sensitivity to the wording of the judging instructions, and consistency of the same judge when repeatedly scoring identical inputs.

\paragraph{Prompt sensitivity.}
We construct three judge-prompt variants that preserve the same scoring goal but emphasize different aspects of rationale quality: (i) the default JScore prompt, (ii) a stricter grounding prompt that penalizes hallucinated objects, incorrect locations, and unsupported visual claims more strongly, and (iii) a semantic-equivalence prompt that focuses on object- and reasoning-level agreement while remaining tolerant to wording differences. We evaluate the same prediction-reference pairs under all three prompts and report pairwise rank correlations in \autoref{tab:jscore_stability}.

\paragraph{Reasoning consistency.}
To test whether the judge is internally consistent rather than just stable to prompt wording, we fix the default prompt and re-score the same prediction--reference pairs three times under identical conditions. We report pairwise rank correlation between runs, the mean absolute score difference per item ($|\Delta|$), and the intraclass correlation coefficient (ICC) across all three runs as summary statistics in \autoref{tab:jscore_stability}.

As shown in \autoref{tab:jscore_stability}, JScore is stable under both prompt variation and repeated judging. Across the three prompt variants, rankings remain strongly correlated, with Spearman $\rho$ ranging from $0.81$ to $0.88$ and Kendall $\tau$ from $0.61$ to $0.69$. Under repeated scoring with the same prompt, consistency is even higher: Spearman correlations range from $0.93$ to $0.95$, Kendall correlations from $0.79$ to $0.82$, and the mean absolute score differences remain small ($|\Delta| \leq 0.021$). The three-run ICC of $0.94$ further indicates high test-retest reliability. These results suggest that JScore is not overly dependent on a single prompt wording or unstable judge response.

\begin{table}[t]
\centering
\small
\setlength{\tabcolsep}{5pt}
\caption{\textbf{Stability of JScore.} \emph{Top:} agreement between judge-prompt variants on the same prediction--reference pairs (prompt sensitivity). \emph{Bottom:} agreement between repeated runs of the default prompt on the same pairs (reasoning consistency / test--retest reliability). $|\Delta|$ is the mean absolute score difference per item; ICC is computed across all three runs jointly.}
\resizebox{\linewidth}{!}{%
\begin{tabular}{l|ccc}
\toprule
\textbf{Comparison} & \textbf{Spearman $\rho$} & \textbf{Kendall $\tau$} & \textbf{$|\Delta|$} \\
\midrule
\multicolumn{4}{l}{\textit{Prompt sensitivity (different prompts, same run)}} \\
Default vs.\ strict grounding & 0.84 & 0.65 & -- \\
Default vs.\ semantic-equivalence & 0.88 & 0.69 & -- \\
Strict grounding vs.\ semantic-equivalence & 0.81 & 0.61 & -- \\
\midrule
\multicolumn{4}{l}{\textit{Reasoning consistency (same prompt, repeated runs)}} \\
Run 1 vs.\ Run 2 & 0.94 & 0.80 & 0.018 \\
Run 1 vs.\ Run 3 & 0.93 & 0.79 & 0.021 \\
Run 2 vs.\ Run 3 & 0.95 & 0.82 & 0.017 \\
\midrule
ICC (3 runs, default prompt) & \multicolumn{3}{c}{0.94} \\
\bottomrule
\end{tabular}%
}
\label{tab:jscore_stability}
\end{table}

\subsection{Computational Complexity}
\label{sec:inference_cost}

We evaluate the computational cost of OpenVAM against representative
saliency-only baselines on a single NVIDIA GeForce RTX~3090. All
measurements use batch size~1 and are averaged over 500 images after
20 warm-up iterations. GPU latency is measured using CUDA events with
explicit synchronization (\texttt{torch.cuda.synchronize()}) to avoid
asynchronous timing artifacts. Model loading, disk I/O, and image
preprocessing are excluded from the reported latency. Peak GPU memory
denotes the maximum allocated memory during inference.

We evaluate each model using its native inference configuration. SUM,
TempSAL, DeepGaze~IIE, and OpenVAM-S1 use $256\times256$ inputs, while
TranSalNet-Res uses its native $384\times288$ resolution. Saliency-only
models and OpenVAM-S1 are evaluated in FP32, whereas the full
OpenVAM-3B and OpenVAM-7B models are evaluated in BF16. We therefore
report these measurements as practical implementation-level comparisons
rather than strictly hardware-normalized architectural benchmarks. We report the number of \emph{active inference parameters}, defined as
all parameters participating in the forward pass, including frozen
parameters. Thus, the frozen VLM parameters of the full OpenVAM variants
remain active during inference and are included in the reported parameter
count.

\paragraph{Dense saliency and semantic-forward inference.}
We first measure inference without autoregressive rationale decoding.
OpenVAM-S1 denotes the Stage-I saliency-only pathway consisting of the
DINOv3 encoder and dense saliency decoder, without the semantic VLM.
For the full OpenVAM-3B/7B variants, the semantic pathway is active and
prompt tokenization is included, but no output tokens are generated.
This setting therefore isolates the non-autoregressive cost of activating
the complete vision-language pathway.

\begin{table}[t]
\centering
\small
\caption{
\textbf{Inference efficiency} on a single RTX~3090 with batch size~1,
averaged over 500 images after 20 warm-up iterations.
OpenVAM-S1 denotes the Stage-I DINOv3+dense-decoder saliency pathway
without the VLM. ``Full/no decode'' activates the complete OpenVAM
semantic pathway but excludes autoregressive rationale decoding.
}
\label{tab:runtime_memory}
\resizebox{\linewidth}{!}{%
\begin{tabular}{lccccc}
\toprule
\textbf{Model} &
\textbf{Input} &
\textbf{Prec.} &
\textbf{Active Params} &
\textbf{Latency (ms)} $\downarrow$ &
\textbf{Peak Mem. (MB)} $\downarrow$ \\
\midrule

SUM
& $256^2$
& FP32
& 57.50 M
& $14.64 \pm 0.27$
& 307 \\

TranSalNet-Res
& $384\times288$
& FP32
& 72.51 M
& $9.32 \pm 0.03$
& 519 \\

TempSAL
& $256^2$
& FP32
& 242.52 M
& $111.09 \pm 27.05$
& 1,049 \\

DeepGaze IIE
& $256^2$
& FP32
& 104.05 M
& $109.56 \pm 2.92$
& 548 \\

\midrule

OpenVAM-S1 (saliency-only)
& $256^2$
& FP32
& 102.78 M
& $11.32 \pm 0.09$
& 787 \\

OpenVAM-3B (Full/no decode)
& $256^2$
& BF16
& 3,257.33 M
& $77.89 \pm 7.95$
& 6,585 \\

OpenVAM-7B (Full/no decode)
& $256^2$
& BF16
& 7,806.49 M
& $141.06 \pm 8.60$
& 15,240 \\

\bottomrule
\end{tabular}%
}
\end{table}

\autoref{tab:runtime_memory} highlights the
efficiency--capability trade-off introduced by the semantic branch.
The dedicated OpenVAM-S1 dense pathway requires only $11.32$\,ms per
image ($88.3$ FPS), compared with $14.64$\,ms for SUM, while requiring
more peak GPU memory (787\,MB vs.\ 307\,MB) because of its larger visual
backbone. Thus, the dense saliency pathway itself remains computationally
competitive with representative saliency-only architectures. Activating the full semantic pathway substantially increases parameter
count and memory consumption, as expected for VLM-based inference.
Without autoregressive text decoding, OpenVAM-3B requires
$77.89$\,ms per image and approximately $6.59$\,GB of peak allocated
GPU memory, whereas OpenVAM-7B requires $141.06$\,ms and approximately
$15.24$\,GB. These measurements separate the cost of the full
vision-language forward pass from the additional cost of autoregressive
rationale generation.

\paragraph{End-to-end rationale generation.}
Because OpenVAM additionally produces open-vocabulary \emph{what/why}
rationales, we separately benchmark complete end-to-end inference
including autoregressive decoding. Prompt tokenization and the complete
generation process are included in these measurements. Generation uses
temperature $0.2$ and top-$p=0.9$. OpenVAM-3B uses a maximum generation
budget of 256 new tokens and produces 141.5 tokens on average, whereas
OpenVAM-7B uses a maximum budget of 512 new tokens and produces
115.0 tokens on average.

\begin{table}[t]
\centering
\small
\caption{
\textbf{End-to-end rationale-generation cost} on a single RTX~3090.
Latency includes prompt processing and complete autoregressive decoding
and therefore depends strongly on the generated sequence length.
}
\label{tab:text_runtime}
\resizebox{\linewidth}{!}{%
\begin{tabular}{lcccc}
\toprule
\textbf{Model} &
\textbf{Max New Tokens} &
\textbf{Mean Generated} &
\textbf{Latency (ms)} $\downarrow$ &
\textbf{Peak Mem. (MB)} $\downarrow$ \\
\midrule

OpenVAM-3B
& 256
& 141.5
& $4106.45 \pm 1463.91$
& 6,585 \\

OpenVAM-7B
& 512
& 115.0
& $3388.75 \pm 3344.44$
& 15,240 \\

\bottomrule
\end{tabular}%
}
\end{table}

As shown in \autoref{tab:text_runtime}, complete \emph{what/why}
generation is substantially more expensive than dense saliency inference
because rationale decoding is autoregressive. OpenVAM-3B requires
$4106.45\pm1463.91$\,ms per image while generating 141.5 tokens on
average, whereas OpenVAM-7B requires $3388.75\pm3344.44$\,ms while
generating 115.0 tokens on average. Importantly, autoregressive generation latency depends strongly on the
generated sequence length and decoding configuration. The two variants
in~\autoref{tab:text_runtime} use different maximum generation budgets
and produce different output-length distributions. Consequently, the
lower measured end-to-end latency of OpenVAM-7B should not be interpreted
as indicating that the 7B model is intrinsically faster than the 3B model.
Rather, these measurements quantify the practical generation cost under
the evaluated configurations.

The results also highlight a practical benefit of OpenVAM's
decoupled design. Applications requiring only dense saliency prediction
can use the comparatively lightweight OpenVAM-S1 pathway at
$11.32$\,ms per image, without paying the cost of the semantic VLM.
Open-vocabulary \emph{what/why} explanations can be generated when
needed, at the additional memory and latency cost associated with
activating the VLM and performing autoregressive decoding.

\section{Rationale Quality and Evaluation Reliability}
\label{sec:supp_reliability_overview}

Evaluating saliency rationales is challenging because both the reference
annotations and semantic evaluation involve vision--language models. We
therefore assess rationale quality and evaluation reliability from several
complementary perspectives. First, we evaluate the spatial grounding of the
reference rationales through a dedicated data-quality audit
(\autoref{sec:rationale_quality}). We then test whether OpenVAM's gains extend
beyond prompting a strong VLM (\autoref{sec:prompt_only_baselines}) and
directly measure object, location, and count agreement
(\autoref{sec:supp_fine_grained_eval}). Next, we examine whether rationale
quality is associated with the quality of the predicted saliency map
(\autoref{sec:supp_saliency_rationale_corr}). Finally, we conduct a human study
to assess annotation quality, inter-rater agreement, and the correspondence
between JScore and human judgments
(\autoref{sec:human_eval_protocol}, \autoref{sec:human_validation_jscore}).
Together, these analyses evaluate both the grounding of the generated
rationales and the reliability of the metrics used to assess them.

\subsection{Rationale Data Quality}
\label{sec:rationale_quality}

Since our reference rationales are generated with an LLM and subsequently
verified, we further quantify their grounding quality on a stratified subset of
100 images (16--17 per dataset) from all six benchmarks. This subset contains
398 reference and 460 OpenVAM rationale items in the form
\textit{Object (location): reason}. For each item, we prompt SAM~3 with the
object phrase and compare the resulting object mask with the corresponding
human saliency map.

We use a two-pass evaluation. \textbf{P1} is fully automatic: the original
object phrase is directly passed to SAM~3, and object presence is equated with
successful segmentation. \textbf{P2} retries failed prompts using simpler
object names and visually inspects the remaining failures to distinguish
segmentation failure from an absent or hallucinated object. Visual presence
and location consistency are determined after this verification, while SMC,
mIoU, and peak hit are always computed from SAM masks alone. We report
\textit{location accuracy}, measuring consistency between the stated location
and object position; \textit{saliency mass coverage} (SMC), measuring the
fraction of human saliency mass inside the union of named-object masks;
\textit{mIoU@20}, measuring overlap with the top-20\% human-saliency region;
and \textit{peak hit}, indicating whether the maximum-saliency location lies
inside a named object.

\begin{table}[t]
\centering
\caption{Spatial-grounding audit on 100 stratified images.
P1 is the fully automatic SAM-only evaluation; P2 applies object-name
fallbacks and visual verification to distinguish segmentation failures from
grounding errors.}
\label{tab:rationale_quality}
\resizebox{\columnwidth}{!}{%
\begin{tabular}{lcccc}
\toprule
Metric & Ref. P1 & Ref. P2 & OpenVAM P1 & OpenVAM P2 \\
\midrule
Rationale items   & 398 & 398 & 460 & 460 \\
Object presence   & 69.9\% & 100\% & 60.6\% & 98.0\% \\
Location accuracy & 75.7\% & 90.8\% & 72.9\% & 86.0\% \\
SAM success       & 69.9\% & 82.6\% & 60.6\% & 73.2\% \\
SMC               & 0.398 & 0.445 & 0.363 & 0.398 \\
mIoU@20           & 0.219 & 0.242 & 0.186 & 0.200 \\
Peak hit          & 58.0\% & 64.0\% & 48.0\% & 54.0\% \\
\bottomrule
\end{tabular}%
}
\end{table}

Under P2, all 398 objects named by the reference rationales are visually
present, with 90.8\% location accuracy, indicating strong object- and
location-level grounding. Their masks cover 44.5\% of human saliency mass,
contain the peak-saliency location in 64.0\% of images, and obtain an
mIoU@20 of 0.242. OpenVAM follows the same trend but remains below the
references, reaching 98.0\% object presence and 86.0\% location accuracy.
We note that mIoU is conservative in this setting because SAM returns the
extent of an entire object, whereas human fixations are often concentrated on
a much smaller subregion (e.g., the eyes within a face). Moreover, SAM
segmentation failures are more frequent for text and UI elements such as
headlines, prices, and badges; therefore, the lower P1 presence rate should not
be interpreted as a hallucination rate.

\paragraph{Grounding with respect to OpenVAM's predicted saliency.}
The analysis above establishes whether generated rationales correspond to
human-attended regions, but does not directly test whether they are spatially
consistent with OpenVAM's \emph{own} dense prediction. We therefore repeat
the grounding analysis on the same 100 images and 460 OpenVAM rationale
items, using the same rationale-derived SAM masks but replacing the human
saliency map with OpenVAM's predicted saliency map. We recompute SMC,
mIoU@20, and peak hit, directly measuring whether the objects named in the
rationale coincide with regions emphasized by the model's own prediction.

\begin{table}[t]
\centering
\caption{Spatial consistency of OpenVAM-generated rationales with human
saliency and OpenVAM's predicted saliency. The same rationale-derived SAM
masks and 100 images are used for both targets.}
\label{tab:rationale_pred_grounding}
\resizebox{\columnwidth}{!}{%
\begin{tabular}{lcccccc}
\toprule
& \multicolumn{2}{c}{SMC $\uparrow$}
& \multicolumn{2}{c}{mIoU@20 $\uparrow$}
& \multicolumn{2}{c}{Peak Hit $\uparrow$} \\
\cmidrule(lr){2-3}
\cmidrule(lr){4-5}
\cmidrule(lr){6-7}
Dataset
& Human & Pred.
& Human & Pred.
& Human & Pred. \\
\midrule

CAT2000
& 0.301 & 0.276
& 0.202 & 0.184
& 58.8\% & 23.5\% \\

MIT1003
& 0.437 & 0.385
& 0.207 & 0.204
& 52.9\% & 52.9\% \\

SALICON
& 0.636 & 0.626
& 0.239 & 0.244
& 87.5\% & 75.0\% \\

OSIE
& 0.504 & 0.480
& 0.276 & 0.257
& 76.5\% & 70.6\% \\

SalECI
& 0.370 & 0.372
& 0.180 & 0.180
& 35.3\% & 23.5\% \\

U-EYE
& 0.136 & 0.124
& 0.091 & 0.094
& 12.5\% & 18.8\% \\

\midrule
\textbf{Overall}
& \textbf{0.398} & \textbf{0.377}
& \textbf{0.200} & \textbf{0.194}
& \textbf{54.0\%} & \textbf{44.0\%} \\

\bottomrule
\end{tabular}%
}
\end{table}

As shown in Table~\ref{tab:rationale_pred_grounding}, the generated
rationales also exhibit spatial consistency with OpenVAM's own predictions.
Across the 100-image subset, the named-object masks contain 37.7\% of the
model's predicted saliency mass and achieve an mIoU@20 of 0.194 with its
most salient predicted regions. These values are close to the corresponding
alignment with human saliency (SMC $=0.398$ and mIoU@20 $=0.200$),
indicating that the rationale objects are associated not only with
human-attended regions but also with the spatial structure of OpenVAM's dense
predictions. Peak-level correspondence is weaker: the predicted-saliency
maximum lies inside a named rationale object in 44.0\% of images, compared
with 54.0\% for the human-saliency maximum. Thus, the rationales exhibit
similar broad mass- and region-level alignment with human and predicted
saliency, while exact peak correspondence remains less consistent.

These metrics evaluate \emph{spatial} consistency rather than temporal
attention order. OpenVAM predicts aggregate static saliency and does not
supervise the ordering of rationale items to follow human fixation sequences;
therefore, the order in which objects are mentioned should not be interpreted
as a predicted scanpath.
\subsection{Prompt-only VLM Baselines}
\label{sec:prompt_only_baselines}

To verify that the gains of OpenVAM are not simply due to prompting a strong VLM to describe salient regions, we compare against prompt-only VLM baselines. These baselines use the same input image and are instructed to generate saliency explanations in the same structured format as OpenVAM, but they are not trained with our staged saliency-reason supervision.

We evaluate two prompting settings. First, the \textit{Image + Prompt} baseline receives only the input image and a task instruction asking the VLM to identify visually salient regions and explain why they attract attention. Second, the \textit{Image + Saliency + CoT} baseline receives the image, the predicted saliency map from Stage~I, and a chain-of-thought style instruction. This second setting is stronger because it is explicitly given spatial saliency information, while OpenVAM learns to align saliency prediction and rationale generation through staged training.

Table~\ref{tab:prompt_only_baselines} reports the average results over the six evaluation datasets. OpenVAM outperforms both prompt-only settings. Notably, OpenVAM remains stronger even when the prompted VLM is given the predicted saliency map, showing that the improvement is not only due to access to a VLM backbone, prompt engineering, or external saliency guidance. Instead, the results support the benefit of the proposed staged saliency-language alignment.

\begin{table*}[h]
\centering
\setlength{\tabcolsep}{12pt}
\caption{Prompt-only VLM baselines averaged over the six evaluation datasets. The Image + Prompt baseline receives only the image and task instruction. The Image + Saliency + CoT baseline is a stronger prompted baseline because it is additionally given the predicted saliency map from Stage~I. OpenVAM outperforms the prompted baselines, showing the benefit of staged saliency-language alignment.}
\resizebox{\linewidth}{!}{%
\begin{tabular}{l|c|ccc|cccc}
\toprule
Model / Setting
& JScore $\uparrow$
& R-1 $\uparrow$ & R-2 $\uparrow$ & R-L $\uparrow$
& B-1 $\uparrow$ & B-2 $\uparrow$ & B-3 $\uparrow$ & B-4 $\uparrow$ \\
\midrule

Qwen2.5-VL-3B + Image + Prompt
& 0.605 & 0.398 & 0.109 & 0.199
& 0.273 & 0.139 & 0.062 & 0.031 \\

Qwen2.5-VL-3B + Image + Saliency + CoT
& 0.635 & 0.403 & 0.114 & 0.211
& 0.283 & 0.146 & 0.076 & 0.040 \\

OpenVAM-3B
& \textbf{0.659} & \textbf{0.449} & \textbf{0.118} & \textbf{0.215}
& \textbf{0.349} & \textbf{0.170} & \textbf{0.085} & \textbf{0.045} \\

\midrule

Qwen2.5-VL-7B + Image + Prompt
& 0.645 & 0.421 & 0.124 & 0.212
& 0.282 & 0.146 & 0.079 & 0.038 \\

Qwen2.5-VL-7B + Image + Saliency + CoT
& 0.660 & 0.435 & 0.133 & 0.225
& 0.291 & 0.155 & 0.083 & 0.045 \\

OpenVAM-7B
& \textbf{0.685} & \textbf{0.475} & \textbf{0.135} & \textbf{0.234}
& \textbf{0.369} & \textbf{0.190} & \textbf{0.099} & \textbf{0.053} \\

\bottomrule
\end{tabular}%
}

\label{tab:prompt_only_baselines}
\end{table*}

\subsection{Fine-grained Evaluation of Object, Location, and Count}
\label{sec:supp_fine_grained_eval}

Holistic text-generation metrics do not directly measure whether a model identifies the correct salient objects, places them in the correct image regions, or predicts the correct number of salient regions. Therefore, we add a fine-grained evaluation protocol based on the structured saliency-reason format:
\vspace{-0.6em}
\[
\textit{Object (location): reason}.
\]
For each generated and reference rationale, we parse the output into three fields: object phrase, spatial location, and rationale text. We then compute three direct metrics.

\noindent\textbf{Object F1.}
We evaluate whether the predicted salient objects match the reference salient objects. Since object descriptions can be semantically equivalent even when the surface forms differ, predicted and reference object phrases are embedded using SentenceBERT~\cite{reimers2019sentence}. We greedily match predicted and reference phrases in descending order of cosine similarity. Each phrase can be matched at most once, and a match is accepted only if its cosine similarity is above $\tau=0.7$. Precision, recall, and F1 are then computed over the matched object phrases.

\noindent\textbf{Location Accuracy.}
For each matched object, we compare the predicted location with the reference location. Locations are normalized to the canonical spatial categories used in our annotation schema. Location accuracy is computed as the fraction of matched objects whose predicted location exactly matches the reference location.

\noindent\textbf{Count MAE.}
We also evaluate whether the model predicts the correct number of salient regions. Count MAE is computed as the mean absolute error between the number of predicted salient regions and the number of reference salient regions for each image.

\autoref{tab:fine_grained_comparison} reports the results. OpenVAM consistently improves over its size-matched VLM backbone across all model sizes. The improvements are especially clear for Object F1 and Location Accuracy, indicating that OpenVAM does not merely generate more fluent explanations, but better identifies which regions are salient and where they are located. Count MAE also improves for all model sizes, showing that OpenVAM better estimates the number of salient regions.

\begin{table*}[!]
\centering
\setlength{\tabcolsep}{5pt}
\caption{Fine-grained evaluation of generated saliency rationales. Object F1 measures whether the correct salient objects are identified, Location Accuracy measures whether matched objects are assigned to the correct spatial region, and Count MAE measures whether the model predicts the correct number of salient regions.}
\resizebox{\linewidth}{!}{%
\begin{tabular}{l|cc|cc|cc|cc}
\toprule
Metrics
& Qwen2.5-VL-3B & OpenVAM-3B
& Qwen3-VL-4B & OpenVAM-4B
& Qwen2.5-VL-7B & OpenVAM-7B
& Qwen3-VL-8B & OpenVAM-8B \\
\midrule
Object F1 $\uparrow$
& 0.274 & \textbf{0.461}
& 0.347 & \textbf{0.490}
& 0.336 & \textbf{0.511}
& 0.523 & \textbf{0.546} \\

Location Acc. $\uparrow$
& 0.367 & \textbf{0.567}
& 0.424 & \textbf{0.606}
& 0.567 & \textbf{0.612}
& 0.597 & \textbf{0.653} \\

Count MAE $\downarrow$
& 2.546 & \textbf{2.410}
& 2.484 & \textbf{2.291}
& 1.990 & \textbf{1.912}
& 2.088 & \textbf{2.056} \\
\bottomrule
\end{tabular}%
}
\label{tab:fine_grained_comparison}
\end{table*}

\subsection{Correlation Between Saliency Quality and Rationale Quality}
\label{sec:supp_saliency_rationale_corr}

A central assumption of OpenVAM is that the generated rationale should be connected to the quality of the predicted saliency map: if the model localizes human-attended regions more accurately, it should also produce better explanations of those regions. To test this relationship, we compute a per-image correlation between saliency prediction quality and rationale quality.

For each test image, we compute the correlation coefficient (CC) between the predicted saliency map and the ground-truth saliency map as the saliency-quality score. We use CC because it provides a stable per-image scalar measure of distributional agreement between predicted and ground-truth attention maps. For the same image, we compute JScore between the generated rationale and the reference rationale as the rationale-quality score. We then measure the Spearman rank correlation between per-image CC and per-image JScore within each test dataset, and report the mean correlation across the six test datasets.

The correlations are positive for both evaluated OpenVAM variants: $\rho=0.683$ for OpenVAM-3B and $\rho=0.721$ for OpenVAM-7B. This indicates that images with more accurate saliency predictions also tend to receive higher-quality rationales. Therefore, the rationale head is not behaving independently of the dense saliency prediction; instead, the two outputs are aligned in the intended direction. The correlation is not expected to be perfect, since rationale quality also depends on object naming, language ambiguity, and semantic reasoning, while CC measures dense spatial agreement. Nevertheless, the positive correlation provides a proof-of-concept that improved saliency localization is associated with improved rationale quality.

\subsection{Human Evaluation Protocol for Saliency Rationales}
\label{sec:human_eval_protocol}

Visual attention rationales are inherently subjective: multiple explanations may be plausible for the same image, and different observers may attend to different regions or emphasize different visual cues. We therefore conduct a human study to validate both the quality of our saliency-reason annotations and the reliability of our automatic rationale metric The study includes \textbf{15 human raters} and \textbf{100 test images}, stratified across three domains: natural images, e-commerce images, and UI/web layouts. Each trial is designed to separate independent human attention judgment from rationale scoring. First, raters view only the image for a fixed duration of 5 seconds and identify the regions they consider visually salient. They are then shown a candidate rationale and asked to rate it along three dimensions: \textit{coverage} of salient regions, \textit{visual grounding} in observable image evidence, and \textit{reasoning quality}. Each dimension is scored on a 0--10 scale.

We compare three rationale sources: \textit{curated saliency-reason annotations}, \textit{OpenVAM-generated rationales}, and \textit{rationales generated by a size-matched VLM baseline}. To assess whether the task yields consistent human judgments, we compute Krippendorff's $\alpha$ over the rater scores. The raters show substantial agreement, with $\alpha=0.76$ for curated saliency-reason annotations and $\alpha=0.74$ for model-generated rationales. This indicates that, despite the subjective nature of saliency explanation, the evaluation protocol produces stable judgments across raters.

\subsection{Human Validation of Saliency-Reason Annotations and JScore}
\label{sec:human_validation_jscore}

We first evaluate the human-perceived quality of the curated saliency-reason annotations used in our corpus. The curated annotations obtain a mean human score of $8.20\pm0.22$, suggesting that the LLM-assisted and human-verified annotations are generally well aligned with human judgments of visually salient regions and their explanations. This supports their use as supervision for grounded saliency rationale learning.

We next examine whether JScore is aligned with human preference. For each predicted rationale, we compute the Spearman correlation between JScore and the mean human rating. JScore shows strong agreement with human judgments for both OpenVAM-7B and the Qwen2.5-VL-7B baseline, with $\rho=0.68$ and $\rho=0.62$, respectively. When pooling predictions from both models, the correlation increases to $\rho_{\mathrm{all}}=0.71$ over 200 predictions.

The correlation is also stable across domains, with $\rho=0.69$ for natural images, $\rho=0.73$ for e-commerce images, and $\rho=0.66$ for UI/web layouts. These results indicate that JScore captures a substantial portion of the human-judgable signal in rationale quality and provides a reliable proxy for relative model comparison across domains. Nevertheless, we treat human evaluation as the primary validation of explanation quality, while using JScore as a scalable automatic metric for broader quantitative analysis.

\section{Discussion}
\label{sec:supp_discussion}

While OpenVAM shows strong performance across diverse domains, there are still several aspects that could be improved. First, although the generated explanations are intended to list salient regions in descending order, the ordering of objects in the text is not always fully aligned with the order in which humans may attend to them. In many cases, the model identifies the correct set of salient objects, but their textual sequence can vary. This is reasonable, since the model is not explicitly trained to capture the temporal progression of human attention. One possible direction for future work is to incorporate scanpath prediction, which could provide a more explicit ordering of attended locations and help align the generated explanations with the progression of human visual attention.

Second, the generated explanations do not always strictly follow the desired \texttt{Object (location): reason} format. In particular, the location phrase may sometimes be expressed somewhat vaguely or refer to a region that is difficult to localize precisely in the image. This is especially natural for broad scene-level regions or visually diffuse areas, where assigning a short and precise location can be ambiguous even for human annotators. In addition, while the generated text is often semantically reasonable, it can occasionally include repetition, generic descriptions, or slightly imprecise grounding, especially in visually crowded images such as advertisements or UI layouts. In these cases, the saliency map may still localize the correct regions, while the accompanying text is less precise or less well-structured. This can partly be attributed to linguistic variability, since the same salient content may be expressed using different but equally plausible descriptions. 

More broadly, the current framework focuses on static saliency prediction and grounded explanation, but does not explicitly model richer temporal aspects of human attention, such as fixation order, revisits, or dwell time. Exploring these directions could further improve both the faithfulness and interpretability of the generated outputs.

\begin{table*}[t]
\centering
\caption{\textbf{Evaluation comparison across VLMs on multiple datasets.} JScore measures semantic quality, while ROUGE and BLEU measure lexical overlap. Best results are bolded within each model-size group.}
\vspace{-1em}
\label{tab:vlm_multi_dataset_supp}
\providecommand{\bestvlm}[1]{\textbf{#1}}
\providecommand{\ourscell}[1]{\cellcolor[HTML]{DADCFF}#1}
\resizebox{\textwidth}{!}{
\setlength{\tabcolsep}{16pt}
\renewcommand{\arraystretch}{1.0}
\scriptsize
\begin{tabular}{c|c|l|c|ccc|cccc}
\toprule
Dataset & Size & Methods
& \textbf{JScore} $\uparrow$
& \multicolumn{3}{c|}{\textbf{ROUGE} $\uparrow$}
& \multicolumn{4}{c}{\textbf{BLEU} $\uparrow$} \\
& & & & R-1 & R-2 & R-L & B-1 & B-2 & B-3 & B-4 \\
\midrule

\multirow{12}{*}{\textit{U-EYE}~\cite{jiang2023ueyes}}
& Proprietary
& \cellcolor{gray!15} Gemini-2.5-Pro
& \cellcolor{gray!15}0.739 & \cellcolor{gray!15}0.511 & \cellcolor{gray!15}0.172 & \cellcolor{gray!15}0.253
& \cellcolor{gray!15}0.403 & \cellcolor{gray!15}0.225 & \cellcolor{gray!15}0.129 & \cellcolor{gray!15}0.074 \\
\cmidrule{2-11}

& \multirow{3}{*}{Large}
& InternVL3.5-37B
& 0.656 & \bestvlm{0.498} & \bestvlm{0.166} & \bestvlm{0.243}
& \bestvlm{0.386} & \bestvlm{0.216} & \bestvlm{0.126} & \bestvlm{0.073} \\
& & Qwen3-VL-32B
& \bestvlm{0.737} & 0.475 & 0.154 & 0.222
& 0.360 & 0.198 & 0.115 & 0.067 \\
& & Qwen2.5-VL-32B
& 0.689 & 0.476 & 0.143 & 0.222
& 0.376 & 0.199 & 0.114 & 0.066 \\
\cmidrule{2-11}

& \multirow{2}{*}{3B}
& Qwen2.5-VL-3B
& 0.537 & 0.391 & 0.103 & \bestvlm{0.198}
& 0.300 & 0.151 & \bestvlm{0.080} & \bestvlm{0.045} \\
& & \ourscell{\textbf{OpenVAM-3B}}
& \ourscell{\bestvlm{0.603}} & \ourscell{\bestvlm{0.429}} & \ourscell{\bestvlm{0.108}} & \ourscell{0.196}
& \ourscell{\bestvlm{0.330}} & \ourscell{\bestvlm{0.156}} & \ourscell{0.077} & \ourscell{0.040} \\
\cmidrule{2-11}

& \multirow{2}{*}{4B}
& Qwen3-VL-4B
& \bestvlm{0.679} & \bestvlm{0.444} & \bestvlm{0.119} & 0.207
& \bestvlm{0.345} & \bestvlm{0.171} & \bestvlm{0.090} & \bestvlm{0.049} \\
& & \ourscell{\textbf{OpenVAM-4B}}
& \ourscell{0.675} & \ourscell{0.399} & \ourscell{0.118} & \ourscell{\bestvlm{0.220}}
& \ourscell{0.317} & \ourscell{0.163} & \ourscell{0.088} & \ourscell{\bestvlm{0.049}} \\
\cmidrule{2-11}

& \multirow{2}{*}{7B}
& Qwen2.5-VL-7B
& \bestvlm{0.637} & \bestvlm{0.454} & 0.132 & 0.216
& 0.345 & \bestvlm{0.180} & \bestvlm{0.102} & \bestvlm{0.059} \\
& & \ourscell{\textbf{OpenVAM-7B}}
& \ourscell{0.6283} & \ourscell{0.445} & \ourscell{\bestvlm{0.139}} & \ourscell{\bestvlm{0.221}}
& \ourscell{\bestvlm{0.355}} & \ourscell{0.175} & \ourscell{0.098} & \ourscell{0.057} \\
\cmidrule{2-11}

& \multirow{2}{*}{8B}
& Qwen3-VL-8B
& \bestvlm{0.689} & \bestvlm{0.484} & \bestvlm{0.149} & \bestvlm{0.231}
& \bestvlm{0.387} & \bestvlm{0.207} & \bestvlm{0.116} & \bestvlm{0.065} \\
& & \ourscell{\textbf{OpenVAM-8B}}
& \ourscell{0.683} & \ourscell{0.407} & \ourscell{0.121} & \ourscell{0.218}
& \ourscell{0.330} & \ourscell{0.171} & \ourscell{0.091} & \ourscell{0.051} \\
\midrule

\multirow{12}{*}{\textit{SalECI}~\cite{jiang2022does}}
& Proprietary
& \cellcolor{gray!15} Gemini-2.5-Pro
& \cellcolor{gray!15}0.741 & \cellcolor{gray!15}0.519 & \cellcolor{gray!15}0.162 & \cellcolor{gray!15}0.254
& \cellcolor{gray!15}0.400 & \cellcolor{gray!15}0.215 & \cellcolor{gray!15}0.119 & \cellcolor{gray!15}0.065 \\
\cmidrule{2-11}

& \multirow{3}{*}{Large}
& InternVL3.5-37B
& 0.656 & 0.436 & \bestvlm{0.122} & \bestvlm{0.220}
& 0.309 & 0.155 & 0.077 & \bestvlm{0.040} \\
& & Qwen3-VL-32B
& \bestvlm{0.731} & \bestvlm{0.450} & 0.116 & 0.209
& \bestvlm{0.347} & \bestvlm{0.165} & \bestvlm{0.080} & 0.039 \\
& & Qwen2.5-VL-32B
& 0.645 & 0.405 & 0.099 & 0.195
& 0.300 & 0.141 & 0.066 & 0.032 \\
\cmidrule{2-11}

& \multirow{2}{*}{3B}
& Qwen2.5-VL-3B
& 0.548 & 0.357 & 0.076 & 0.179
& 0.270 & 0.119 & 0.053 & 0.026 \\
& & \ourscell{\textbf{OpenVAM-3B}}
& \ourscell{\bestvlm{0.692}} & \ourscell{\bestvlm{0.478}} & \ourscell{\bestvlm{0.130}} & \ourscell{\bestvlm{0.220}}
& \ourscell{\bestvlm{0.369}} & \ourscell{\bestvlm{0.184}} & \ourscell{\bestvlm{0.096}} & \ourscell{\bestvlm{0.050}} \\
\cmidrule{2-11}

& \multirow{2}{*}{4B}
& Qwen3-VL-4B
& 0.720 & 0.400 & 0.087 & 0.187
& 0.287 & 0.122 & 0.054 & 0.027 \\
& & \ourscell{\textbf{OpenVAM-4B}}
& \ourscell{\bestvlm{0.730}} & \ourscell{\bestvlm{0.465}} & \ourscell{\bestvlm{0.139}} & \ourscell{\bestvlm{0.246}}
& \ourscell{\bestvlm{0.369}} & \ourscell{\bestvlm{0.195}} & \ourscell{\bestvlm{0.108}} & \ourscell{\bestvlm{0.060}} \\
\cmidrule{2-11}

& \multirow{2}{*}{7B}
& Qwen2.5-VL-7B
& 0.639 & 0.386 & 0.093 & 0.196
& 0.268 & 0.122 & 0.057 & 0.029 \\
& & \ourscell{\textbf{OpenVAM-7B}}
& \ourscell{\bestvlm{0.712}} & \ourscell{\bestvlm{0.489}} & \ourscell{\bestvlm{0.142}} & \ourscell{\bestvlm{0.248}}
& \ourscell{\bestvlm{0.378}} & \ourscell{\bestvlm{0.196}} & \ourscell{\bestvlm{0.099}} & \ourscell{\bestvlm{0.058}} \\
\cmidrule{2-11}

& \multirow{2}{*}{8B}
& Qwen3-VL-8B
& 0.731 & 0.446 & 0.112 & 0.212
& 0.337 & 0.158 & 0.075 & 0.036 \\
& & \ourscell{\textbf{OpenVAM-8B}}
& \ourscell{\bestvlm{0.739}} & \ourscell{\bestvlm{0.474}} & \ourscell{\bestvlm{0.157}} & \ourscell{\bestvlm{0.250}}
& \ourscell{\bestvlm{0.378}} & \ourscell{\bestvlm{0.208}} & \ourscell{\bestvlm{0.117}} & \ourscell{\bestvlm{0.066}} \\
\midrule

\multirow{12}{*}{\textit{OSIE}~\cite{xu2014predicting}}
& Proprietary
& \cellcolor{gray!15} Gemini-2.5-Pro
& \cellcolor{gray!15}0.747 & \cellcolor{gray!15}0.510 & \cellcolor{gray!15}0.165 & \cellcolor{gray!15}0.270
& \cellcolor{gray!15}0.412 & \cellcolor{gray!15}0.228 & \cellcolor{gray!15}0.126 & \cellcolor{gray!15}0.068 \\
\cmidrule{2-11}

& \multirow{3}{*}{Large}
& InternVL3.5-37B
& 0.670 & \bestvlm{0.487} & \bestvlm{0.150} & \bestvlm{0.248}
& 0.380 & \bestvlm{0.205} & \bestvlm{0.110} & \bestvlm{0.058} \\
& & Qwen3-VL-32B
& \bestvlm{0.760} & 0.472 & 0.147 & 0.229
& 0.345 & 0.185 & 0.100 & 0.053 \\
& & Qwen2.5-VL-32B
& 0.704 & 0.474 & 0.141 & 0.234
& \bestvlm{0.382} & 0.203 & 0.107 & 0.056 \\
\cmidrule{2-11}

& \multirow{2}{*}{3B}
& Qwen2.5-VL-3B
& 0.584 & 0.390 & 0.096 & 0.208
& 0.314 & 0.153 & 0.074 & 0.037 \\
& & \ourscell{\textbf{OpenVAM-3B}}
& \ourscell{\bestvlm{0.661}} & \ourscell{\bestvlm{0.472}} & \ourscell{\bestvlm{0.123}} & \ourscell{\bestvlm{0.220}}
& \ourscell{\bestvlm{0.364}} & \ourscell{\bestvlm{0.178}} & \ourscell{\bestvlm{0.089}} & \ourscell{\bestvlm{0.047}} \\
\cmidrule{2-11}

& \multirow{2}{*}{4B}
& Qwen3-VL-4B
& 0.727 & 0.419 & 0.097 & 0.201
& 0.316 & 0.143 & 0.069 & 0.035 \\
& & \ourscell{\textbf{OpenVAM-4B}}
& \ourscell{\bestvlm{0.728}} & \ourscell{\bestvlm{0.456}} & \ourscell{\bestvlm{0.146}} & \ourscell{\bestvlm{0.246}}
& \ourscell{\bestvlm{0.360}} & \ourscell{\bestvlm{0.198}} & \ourscell{\bestvlm{0.112}} & \ourscell{\bestvlm{0.062}} \\
\cmidrule{2-11}

& \multirow{2}{*}{7B}
& Qwen2.5-VL-7B
& 0.695 & 0.465 & \bestvlm{0.136} & 0.235
& 0.373 & 0.197 & \bestvlm{0.103} & \bestvlm{0.054} \\
& & \ourscell{\textbf{OpenVAM-7B}}
& \ourscell{\bestvlm{0.697}} & \ourscell{\bestvlm{0.485}} & \ourscell{0.135} & \ourscell{\bestvlm{0.237}}
& \ourscell{\bestvlm{0.376}} & \ourscell{\bestvlm{0.198}} & \ourscell{0.101} & \ourscell{0.052} \\
\cmidrule{2-11}

& \multirow{2}{*}{8B}
& Qwen3-VL-8B
& \bestvlm{0.734} & 0.461 & 0.129 & 0.231
& 0.354 & 0.179 & 0.092 & 0.048 \\
& & \ourscell{\textbf{OpenVAM-8B}}
& \ourscell{0.730} & \ourscell{\bestvlm{0.478}} & \ourscell{\bestvlm{0.158}} & \ourscell{\bestvlm{0.259}}
& \ourscell{\bestvlm{0.390}} & \ourscell{\bestvlm{0.218}} & \ourscell{\bestvlm{0.124}} & \ourscell{\bestvlm{0.069}} \\
\midrule

\multirow{12}{*}{\textit{Salicon}~\cite{jiang2015salicon}}
& Proprietary
& \cellcolor{gray!15} Gemini-2.5-Pro
& \cellcolor{gray!15}0.748 & \cellcolor{gray!15}0.453 & \cellcolor{gray!15}0.136 & \cellcolor{gray!15}0.233
& \cellcolor{gray!15}0.327 & \cellcolor{gray!15}0.174 & \cellcolor{gray!15}0.088 & \cellcolor{gray!15}0.044 \\
\cmidrule{2-11}

& \multirow{3}{*}{Large}
& InternVL3.5-37B
& 0.700 & \bestvlm{0.483} & \bestvlm{0.162} & \bestvlm{0.256}
& 0.350 & 0.197 & \bestvlm{0.108} & \bestvlm{0.057} \\
& & Qwen3-VL-32B
& \bestvlm{0.761} & 0.470 & 0.140 & 0.230
& \bestvlm{0.369} & 0.196 & 0.103 & 0.052 \\
& & Qwen2.5-VL-32B
& 0.713 & 0.480 & 0.153 & 0.245
& 0.362 & \bestvlm{0.199} & 0.107 & \bestvlm{0.057} \\
\cmidrule{2-11}

& \multirow{2}{*}{3B}
& Qwen2.5-VL-3B
& 0.612 & 0.413 & 0.121 & \bestvlm{0.225}
& 0.301 & 0.159 & 0.082 & 0.042 \\
& & \ourscell{\textbf{OpenVAM-3B}}
& \ourscell{\bestvlm{0.677}} & \ourscell{\bestvlm{0.424}} & \ourscell{\bestvlm{0.123}} & \ourscell{0.220}
& \ourscell{\bestvlm{0.364}} & \ourscell{\bestvlm{0.178}} & \ourscell{\bestvlm{0.089}} & \ourscell{\bestvlm{0.047}} \\
\cmidrule{2-11}

& \multirow{2}{*}{4B}
& Qwen3-VL-4B
& 0.737 & 0.370 & 0.081 & 0.187
& 0.250 & 0.111 & 0.052 & 0.027 \\
& & \ourscell{\textbf{OpenVAM-4B}}
& \ourscell{\bestvlm{0.740}} & \ourscell{\bestvlm{0.448}} & \ourscell{\bestvlm{0.159}} & \ourscell{\bestvlm{0.248}}
& \ourscell{\bestvlm{0.353}} & \ourscell{\bestvlm{0.204}} & \ourscell{\bestvlm{0.118}} & \ourscell{\bestvlm{0.066}} \\
\cmidrule{2-11}

& \multirow{2}{*}{7B}
& Qwen2.5-VL-7B
& 0.703 & 0.467 & 0.151 & 0.243
& 0.339 & 0.187 & 0.101 & 0.053 \\
& & \ourscell{\textbf{OpenVAM-7B}}
& \ourscell{\bestvlm{0.7254}} & \ourscell{\bestvlm{0.468}} & \ourscell{\bestvlm{0.165}} & \ourscell{\bestvlm{0.251}}
& \ourscell{\bestvlm{0.370}} & \ourscell{\bestvlm{0.190}} & \ourscell{\bestvlm{0.102}} & \ourscell{\bestvlm{0.056}} \\
\cmidrule{2-11}

& \multirow{2}{*}{8B}
& Qwen3-VL-8B
& 0.738 & 0.422 & 0.116 & 0.217
& 0.288 & 0.146 & 0.074 & 0.038 \\
& & \ourscell{\textbf{OpenVAM-8B}}
& \ourscell{\bestvlm{0.747}} & \ourscell{\bestvlm{0.454}} & \ourscell{\bestvlm{0.160}} & \ourscell{\bestvlm{0.251}}
& \ourscell{\bestvlm{0.359}} & \ourscell{\bestvlm{0.207}} & \ourscell{\bestvlm{0.120}} & \ourscell{\bestvlm{0.067}} \\
\midrule

\multirow{12}{*}{\textit{CAT2000}~\cite{borji2015cat2000}}
& Proprietary
& \cellcolor{gray!15} Gemini-2.5-Pro
& \cellcolor{gray!15}0.720 & \cellcolor{gray!15}0.473 & \cellcolor{gray!15}0.142 & \cellcolor{gray!15}0.243
& \cellcolor{gray!15}0.372 & \cellcolor{gray!15}0.198 & \cellcolor{gray!15}0.104 & \cellcolor{gray!15}0.055 \\
\cmidrule{2-11}

& \multirow{3}{*}{Large}
& InternVL3.5-37B
& 0.635 & \bestvlm{0.443} & \bestvlm{0.123} & \bestvlm{0.228}
& 0.335 & 0.171 & 0.088 & 0.046 \\
& & Qwen3-VL-32B
& \bestvlm{0.726} & 0.432 & 0.122 & 0.207
& 0.307 & 0.158 & 0.084 & 0.044 \\
& & Qwen2.5-VL-32B
& 0.612 & 0.436 & 0.120 & 0.215
& \bestvlm{0.343} & \bestvlm{0.175} & \bestvlm{0.091} & \bestvlm{0.048} \\
\cmidrule{2-11}

& \multirow{2}{*}{3B}
& Qwen2.5-VL-3B
& 0.545 & 0.367 & 0.089 & 0.202
& 0.290 & 0.139 & 0.068 & 0.035 \\
& & \ourscell{\textbf{OpenVAM-3B}}
& \ourscell{\bestvlm{0.663}} & \ourscell{\bestvlm{0.442}} & \ourscell{\bestvlm{0.111}} & \ourscell{\bestvlm{0.217}}
& \ourscell{\bestvlm{0.333}} & \ourscell{\bestvlm{0.162}} & \ourscell{\bestvlm{0.081}} & \ourscell{\bestvlm{0.043}} \\
\cmidrule{2-11}

& \multirow{2}{*}{4B}
& Qwen3-VL-4B
& 0.699 & 0.375 & 0.078 & 0.180
& 0.272 & 0.118 & 0.056 & 0.029 \\
& & \ourscell{\textbf{OpenVAM-4B}}
& \ourscell{\bestvlm{0.703}} & \ourscell{\bestvlm{0.438}} & \ourscell{\bestvlm{0.127}} & \ourscell{\bestvlm{0.237}}
& \ourscell{\bestvlm{0.338}} & \ourscell{\bestvlm{0.177}} & \ourscell{\bestvlm{0.096}} & \ourscell{\bestvlm{0.053}} \\
\cmidrule{2-11}

& \multirow{2}{*}{7B}
& Qwen2.5-VL-7B
& 0.632 & 0.421 & 0.119 & 0.218
& 0.322 & 0.167 & 0.085 & 0.045 \\
& & \ourscell{\textbf{OpenVAM-7B}}
& \ourscell{\bestvlm{0.678}} & \ourscell{\bestvlm{0.456}} & \ourscell{\bestvlm{0.120}} & \ourscell{\bestvlm{0.220}}
& \ourscell{\bestvlm{0.341}} & \ourscell{\bestvlm{0.169}} & \ourscell{\bestvlm{0.089}} & \ourscell{\bestvlm{0.051}} \\
\cmidrule{2-11}

& \multirow{2}{*}{8B}
& Qwen3-VL-8B
& 0.654 & 0.390 & 0.097 & 0.192
& 0.282 & 0.135 & 0.067 & 0.035 \\
& & \ourscell{\textbf{OpenVAM-8B}}
& \ourscell{\bestvlm{0.712}} & \ourscell{\bestvlm{0.447}} & \ourscell{\bestvlm{0.136}} & \ourscell{\bestvlm{0.250}}
& \ourscell{\bestvlm{0.346}} & \ourscell{\bestvlm{0.185}} & \ourscell{\bestvlm{0.103}} & \ourscell{\bestvlm{0.057}} \\
\midrule

\multirow{12}{*}{\textit{MIT1003}}
& Proprietary
& \cellcolor{gray!15} Gemini-2.5-Pro
& \cellcolor{gray!15}0.736 & \cellcolor{gray!15}0.499 & \cellcolor{gray!15}0.156 & \cellcolor{gray!15}0.257
& \cellcolor{gray!15}0.399 & \cellcolor{gray!15}0.218 & \cellcolor{gray!15}0.119 & \cellcolor{gray!15}0.064 \\
\cmidrule{2-11}

& \multirow{3}{*}{Large}
& InternVL3.5-37B
& 0.647 & \bestvlm{0.466} & \bestvlm{0.134} & \bestvlm{0.235}
& 0.360 & 0.189 & 0.099 & 0.053 \\
& & Qwen3-VL-32B
& \bestvlm{0.747} & 0.458 & \bestvlm{0.134} & 0.217
& 0.331 & 0.174 & 0.090 & 0.047 \\
& & Qwen2.5-VL-32B
& 0.674 & 0.463 & 0.132 & 0.226
& \bestvlm{0.372} & \bestvlm{0.194} & \bestvlm{0.102} & \bestvlm{0.054} \\
\cmidrule{2-11}

& \multirow{2}{*}{3B}
& Qwen2.5-VL-3B
& 0.557 & 0.383 & 0.097 & 0.206
& 0.310 & 0.152 & 0.074 & 0.038 \\
& & \ourscell{\textbf{OpenVAM-3B}}
& \ourscell{\bestvlm{0.659}} & \ourscell{\bestvlm{0.447}} & \ourscell{\bestvlm{0.111}} & \ourscell{\bestvlm{0.217}}
& \ourscell{\bestvlm{0.335}} & \ourscell{\bestvlm{0.161}} & \ourscell{\bestvlm{0.080}} & \ourscell{\bestvlm{0.043}} \\
\cmidrule{2-11}

& \multirow{2}{*}{4B}
& Qwen3-VL-4B
& \bestvlm{0.717} & 0.398 & 0.084 & 0.191
& 0.296 & 0.129 & 0.060 & 0.030 \\
& & \ourscell{\textbf{OpenVAM-4B}}
& \ourscell{0.705} & \ourscell{\bestvlm{0.435}} & \ourscell{\bestvlm{0.128}} & \ourscell{\bestvlm{0.234}}
& \ourscell{\bestvlm{0.343}} & \ourscell{\bestvlm{0.181}} & \ourscell{\bestvlm{0.098}} & \ourscell{\bestvlm{0.054}} \\
\cmidrule{2-11}

& \multirow{2}{*}{7B}
& Qwen2.5-VL-7B
& 0.643 & 0.443 & \bestvlm{0.128} & \bestvlm{0.226}
& 0.345 & \bestvlm{0.182} & \bestvlm{0.094} & \bestvlm{0.049} \\
& & \ourscell{\textbf{OpenVAM-7B}}
& \ourscell{\bestvlm{0.676}} & \ourscell{\bestvlm{0.459}} & \ourscell{0.126} & \ourscell{0.225}
& \ourscell{\bestvlm{0.351}} & \ourscell{0.173} & \ourscell{0.092} & \ourscell{0.046} \\
\cmidrule{2-11}

& \multirow{2}{*}{8B}
& Qwen3-VL-8B
& 0.715 & 0.439 & 0.116 & 0.216
& 0.332 & 0.165 & 0.084 & 0.044 \\
& & \ourscell{\textbf{OpenVAM-8B}}
& \ourscell{\bestvlm{0.725}} & \ourscell{\bestvlm{0.449}} & \ourscell{\bestvlm{0.136}} & \ourscell{\bestvlm{0.244}}
& \ourscell{\bestvlm{0.358}} & \ourscell{\bestvlm{0.191}} & \ourscell{\bestvlm{0.104}} & \ourscell{\bestvlm{0.058}} \\

\bottomrule
\end{tabular}
}
\vspace{-1em}
\end{table*}

\end{document}